\documentclass[nohyperref]{article}

\usepackage{microtype}
\usepackage{graphicx}
\usepackage{tabularx}
\usepackage{subcaption}
\usepackage{booktabs} 

\usepackage{hyperref}

\usepackage[accepted]{icml2022}

\usepackage{amsmath}
\usepackage{amssymb}
\usepackage{mathtools}
\usepackage{amsthm}

\usepackage[capitalize,noabbrev]{cleveref}

\theoremstyle{plain}

\theoremstyle{definition}

\theoremstyle{remark}

\usepackage[textsize=tiny]{todonotes}

\icmltitlerunning{Showing Your Offline Reinforcement Learning Work}

\begin{document}

\twocolumn[
\icmltitle{Showing Your Offline Reinforcement Learning Work: \\ Online Evaluation Budget Matters}



\icmlsetsymbol{equal}{*}

\begin{icmlauthorlist}
\icmlauthor{Vladislav Kurenkov}{comp}
\icmlauthor{Sergey Kolesnikov}{comp}
\end{icmlauthorlist}

\icmlaffiliation{comp}{Tinkoff, Moscow, Russia}

\icmlcorrespondingauthor{Vladislav Kurenkov}{v.kurenkov@tinkoff.ai}

\icmlkeywords{Machine Learning, ICML}

\vskip 0.3in
]



\printAffiliationsAndNotice{}  

\begin{abstract}

In this work, we argue for the importance of an online evaluation budget for a reliable comparison of deep offline RL algorithms. First, we delineate that the online evaluation budget is problem-dependent, where some problems allow for less but others for more. And second, we demonstrate that the preference between algorithms is budget-dependent across a diverse range of decision-making domains such as Robotics, Finance, and Energy Management. Following the points above, we suggest reporting the performance of deep offline RL algorithms under varying online evaluation budgets. To facilitate this, we propose to use a reporting tool from the NLP field, Expected Validation Performance. This technique makes it possible to reliably estimate expected maximum performance under different budgets while not requiring any additional computation beyond hyperparameter search. By employing this tool, we also show that Behavioral Cloning is often more favorable to offline RL algorithms when working within a limited budget.
\end{abstract}

\section{Introduction}

In recent years, significant success has been achieved in applying Reinforcement Learning (RL) to different real-world scenarios \cite{chenTopKOffPolicyCorrection2019, tangDeepValuenetworkBased2019, gauciHorizonFacebookOpen2019}. 
Offline Reinforcement Learning (ORL) \cite{levineOfflineReinforcementLearning2020} is pioneering a real-world adaptation of RL, focusing on algorithms that can learn a policy from a fixed, previously recorded dataset, without having to interact with the environment during training.
Many recent community efforts were focused on advanced benchmarks \cite{fuD4RLDatasetsDeep2021,gulcehreRLUnpluggedSuite2021} and nuanced algorithm comparisons \cite{brandfonbrenerOfflineRLOffPolicy2021, qinNeoRLRealWorldBenchmark2021}. 

However, it is widely recognized that even deep online RL is plagued with evaluation issues, such as insufficient account for statistical variability \cite{agarwalDeepReinforcementLearning2022}, or sensitivity to the choice of hyperparameters \cite{hendersonDeepReinforcementLearning2018}. The latter problem is one of the reasons why the results of many deep RL methods are usually reported for a narrow range of values.

When it comes to deep ORL algorithms, the choice of hyperparameters also plays a major role in the final performance \cite{wuBehaviorRegularizedOffline2019}. While many in the community note that the whole training and evaluation pipeline needs improvement \cite{fuBenchmarksDeepOffPolicy2021}, the comparisons of new deep ORL algorithms are still mostly done through online performance reports on the best set of hyperparameters \cite{kostrikovOfflineReinforcementLearning2021, fujimotoMinimalistApproachOffline2021}.

In this paper, we argue that the hyperparameter search should not be ignored in the deep offline RL setting, demonstrating that the conclusions about the algorithms change when we control for the number of trained policies deployed online\footnote{Code is available at \href{https://tinkoff-ai.github.io/eop/}{tinkoff-ai.github.io/eop}}. To this end, we introduce the notion of an online budget, i.e. the number of policies deployed online, and suggest to use an entire pipeline similar to the one in \cite{paineHyperparameterSelectionOffline2020} for reporting results: training, offline selection, and online evaluation, but one where the selection is done by uniform sampling. As we will demonstrate, this decision allows us to re-use the Expected Validations Performance (EVP) \cite{dodgeShowYourWork2019} technique from the NLP field to get reliable estimates of expected maximum performance under different online budgets from just one round of hyperparameter search.

\begin{figure*}[ht]
\vskip 0.2in
\begin{center}
\centerline{\includegraphics[width=\textwidth]{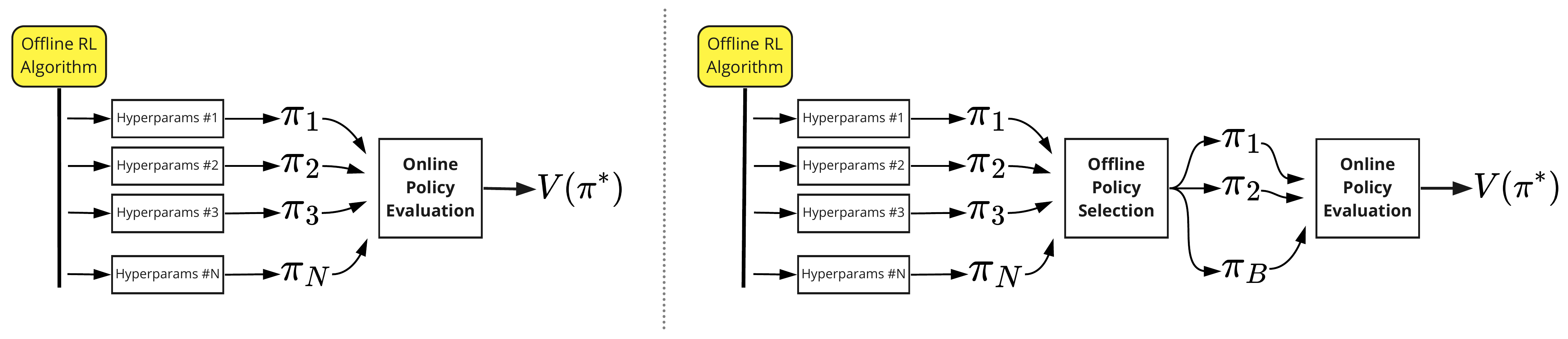}}
\caption{\textbf{Left}: A widespreaded approach for reporting deep offline RL results, commonly known as online policy selection. \textbf{Right}: Full deep offline RL evaluation pipeline. We argue for reporting results under the second pipeline with varying sizes of the online evaluation budget $B$. Note that selecting hyperparameters that perform best overall on the online policy selection tasks from the same domain \cite{fuD4RLDatasetsDeep2021, gulcehreRLUnpluggedSuite2021} can also be put to the right pipeline as a special case.}
\label{fig:eval_pipelines}
\end{center}
\vskip -0.2in
\end{figure*}

\textbf{Our Contributions} Here, we list the main contributions of our work:
\begin{itemize}
    \item We demonstrate that the preference between deep offline RL algorithms is budget-dependent. We stress that this is more critical for offline settings than for online ones, and that current evaluation methodology does not account for such dependence.
    \item We propose to use Expected Validation Performance \cite{dodgeShowYourWork2019}, a technique actively employed in NLP, for reliable comparison of deep offline RL algorithms under varying online evaluation budgets. To stress the online nature of comparison (in opposition to validation), we refer to it as Expected Online Performance (EOP). This tool can take the both major components of deep offline RL into account: the offline policy selection (OPS) method as well as online evaluation budget. Furthermore, it can be applied without additional computational expenses.
    \item Using the proposed tool, we also demonstrate that Behavioral Cloning \cite{pomerleauEfficientTrainingArtificial1991} is often more favorable under a limited evaluation budget.
    \item In addition, EOP can be applied to comparisons of OPS methods. Using EOP, we illustrate that their preference is also budget-dependent.
\end{itemize}

In the end, we also discuss how the proposed solution relates to the recently introduced Active-OPS \cite{konyushkovaActiveOfflinePolicy2021} and deployment-constrainted RL setup \cite{matsushimaDeploymentEfficientReinforcementLearning2020}.

\section{Background}
\subsection{Offline RL}

Reinforcement learning (\textit{RL}) is a framework for solving sequential decision-making problems. It is typically formulated as a Markov Decision Process (MDP) over a 5-tuple $(S, A, P, r, \gamma)$, with action space $A$, state space $S$, transition dynamics $P$, reward function $r$, and discount factor $\gamma$. The goal of the learning agent is to obtain a policy $\pi(s, a)$ that maximizes the expected discounted return  $E_{\pi}[\sum_{t=0}^{\infty}\gamma^{t}r_{t+1}]$ through interaction with the MDP.

In \textit{Offline RL}, also known as Batch RL, instead of obtaining data and learning via environment interactions, the agent solely relies on a static dataset $D$ that was collected under some unknown behavioral policy (or a mixture of policies) $\pi_{\mu}$. This setting is considered more challenging, as the agent loses its ability for exploration \cite{levineOfflineReinforcementLearning2020} and is faced with the problem of extrapolation error -- being unable to correct its estimation inaccuracies when selected actions are not present in the training dataset \cite{fujimotoBenchmarkingBatchDeep2019}.

\subsection{Offline RL Evaluation}

\begin{table*}[h]
\caption{\textbf{Best final performance of deep offline RL algorithms if they were evaluated under a different number of policies deployed online} for Hopper-v3 environment. This table highlights that the usage of different online evaluation budgets ($B$) may lead to different conclusions on preference between algorithms. $N$ is the total number of hyperparameters evaluated for a specific algorithm.}
\vspace{0.15in}
\centering
\begin{tabular}{lccccccc|cr}
\toprule
Algorithm & $B = 1$ & $B = 2$ & $B = 3$ & $B = 4$ & $B = 8$ & $B = 15$ & $B = 30$ & Final & $N$
\\
\midrule
{BC}  &   \textbf{1794} & \textbf{2057} & \textbf{2179} & - & - & - & - & 2179 & 3 \\
{CQL} &  1773 & 1954 & 2072 & \textbf{2161} & \textbf{2391} & \textbf{2603} & \textbf{2832} & \textbf{2832} & 30 \\
{PLAS} & 1475 & 1833 & 1996 & 2096 & 2316 & 2507 & - & 2507 & 15 \\
{BCQ} &  1325 & 1605 & 1742 & 1826 & 1986 & - & - & 2062 & 12\\
{CRR}      & 1013 & 1339 & 1477 & 1545 & 1636 & - & - & 1668 & 12 \\
{BREMEN}      & 883 & 1148 & 1318 & 1439 & 1691 & - & - & 1795 & 12 \\
{MOPO}      & 11 & 18 & 24 & 30 & 46 & 63 & 78 & 78 & 30 \\
\bottomrule
\end{tabular}\\
\label{table:online-budget-dependence-hopper}
\end{table*}

Training and evaluation of deep offline RL algorithms is still in active development, and various authors approach it in different ways by simplifying the genuine offline setting \cite{gulcehreRLUnpluggedSuite2021}. At the core of the simplification are two primary issues: (1) unlimited amount of online evaluations available, and therefore (2) sidestepping offline policy selection. For example, it is common to report the maximum performance for the best set of hyperparameters (Figure \ref{fig:eval_pipelines}, Left). Moreover, in many cases, the number of search trials is not made explicit \cite{kumarConservativeQLearningOffline2020}.

To eliminate these simplifications, we adhere to a more general setup for training and evaluating offline RL algorithms similar to \citet{paineHyperparameterSelectionOffline2020} in order to satisfy hard offline constraints (Figure \ref{fig:eval_pipelines}, Right).

First, the dataset $D$ is randomly split trajectory-wise into training $D_{T}$ and validation $D_{V}$ subsets accordingly. Then a sequence of hyperparameter assignments $(h_{1}, h_{2}, ..., h_{N})$ is sampled for running an algorithm of interest, resulting in a sequence of policies $(\pi_{1}, \pi_{2}, ..., \pi_{N})$. Note that at this stage, we do not know how good these policies are.

Then, $B \leq N$ of policies are arbitrarily chosen for online evaluation, which we refer to as an \textit{online evaluation budget}. In the most restricted offline RL setting, $B = 1$. However, the generalization to $B > 1$ is justified by the online evaluation budget being conditioned on the relevant decision-making problem and the available resources.

To choose policies for online evaluation, offline policy selection (OPS) methods can be used. In specific domains, like recommender systems, policies can be picked based on established offline metrics, e.g. Recall, computed on the validation $D_{V}$ dataset \cite{xinSelfSupervisedReinforcementLearning2020a}. However, such metrics do not always exist, and it is often necessary to rely on general methods \cite{voloshinEmpiricalStudyOffPolicy2020, fuBenchmarksDeepOffPolicy2021} or proxy tasks \cite{fuD4RLDatasetsDeep2021, gulcehreRLUnpluggedSuite2021}.

\section{Online Evaluation Budget Matters}
\label{sec:online-eval-matters}

As can be seen in Figure \ref{fig:eval_pipelines}, Left, using online policy selection makes the evaluation budget $B$ equivalent to the number of hyperparameter search trials $N$. On the other hand, \citet{fuD4RLDatasetsDeep2021, gulcehreRLUnpluggedSuite2021} search for the best set of hyperparameters using proxy tasks, but the online evaluation budget on the target task is $B = 1$.

Meanwhile, there is a whole spectrum of values in-between that could be relevant not only for a specific problem, but for a specific context. By context we mean a certain space of resources (computational resources, robotics hardware, time constraints, online testing capacity). Here, a practitioner may work on the same problem, but have a lower or bigger amount of resources available for online evaluation. 

Therefore, a natural question to ask when analysing results of deep ORL algorithms is “Will the conclusions about the algorithms change, if I have a lower or higher online evaluation budget than the one reported in the paper?”. Unfortunately, current evaluation and report methodologies do not provide an answer, and the dependence between varied online evaluation budget and the resulting performance of the algorithms is left unreported. 

To address this issue independently, it is necessary to access detailed experimental results showing which hyperparameters resulted in which performance. While some authors open-source such data \cite{qinNeoRLRealWorldBenchmark2021}, it is not a common practice to do so.

To demonstrate that the conclusions about algorithm preference are dependent on the available online evaluation budget, we rely on open-sourced\footnote{Note that there is a discrepancy in open-sourced and reported results. There is additional data on CRR and BREMEN, but data on MB-PPO is not provided.} results by \citet{qinNeoRLRealWorldBenchmark2021}. For each algorithm in Table \ref{table:online-budget-dependence-hopper}, we compute the expected maximum performance under uniform policy selection (i.e. the policies are chosen at random) given a specific online evaluation budget $B$. The final column is what would be reported in the paper, demonstrating that CQL \cite{kumarConservativeQLearningOffline2020} significantly outperforms its competitors. However, in budgets up to 4, Behavioral Cloning performs the best. Also, note that the preference between CRR \cite{wangCriticRegularizedRegression2020} and BREMEN \cite{matsushimaDeploymentEfficientReinforcementLearning2020} is reversed starting from the budget of 8.

\section{Accounting for the Budget \raisebox{-0.32ex}{\includegraphics[scale=0.08]{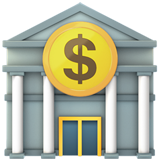}}}

\begin{figure*}[!h]
    \begin{subfigure}[b]{0.32\textwidth}
        \centering
        \centerline{\includegraphics[width=\columnwidth]{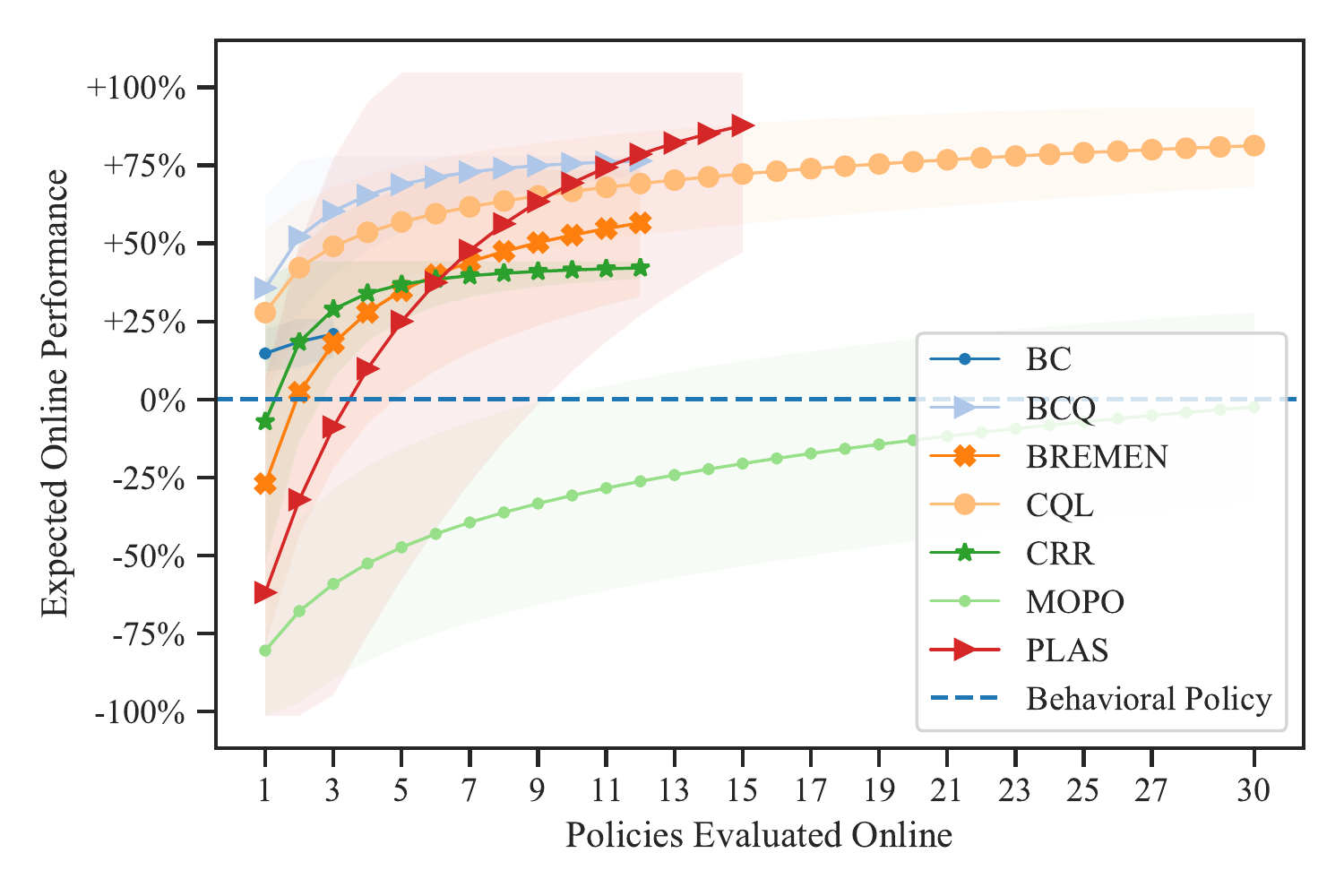}}
        \caption{Walker2d, Low-1000}
    \end{subfigure}
    \begin{subfigure}[b]{0.32\textwidth}
        \centering
        \centerline{\includegraphics[width=\columnwidth]{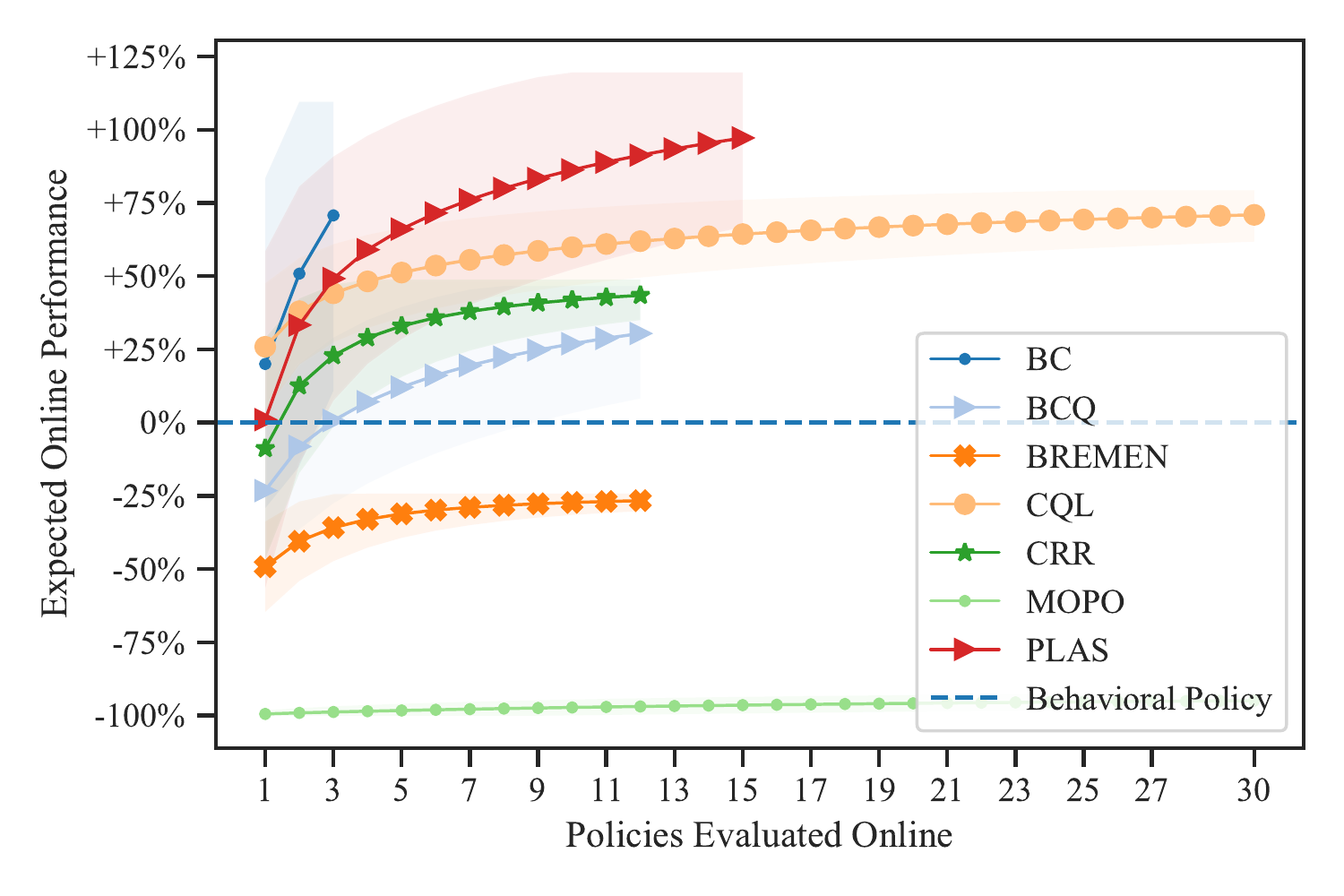}}
        \caption{Hopper, Medium-1000}
    \end{subfigure}
    \begin{subfigure}[b]{0.32\textwidth}
        \centering
        \centerline{\includegraphics[width=\columnwidth]{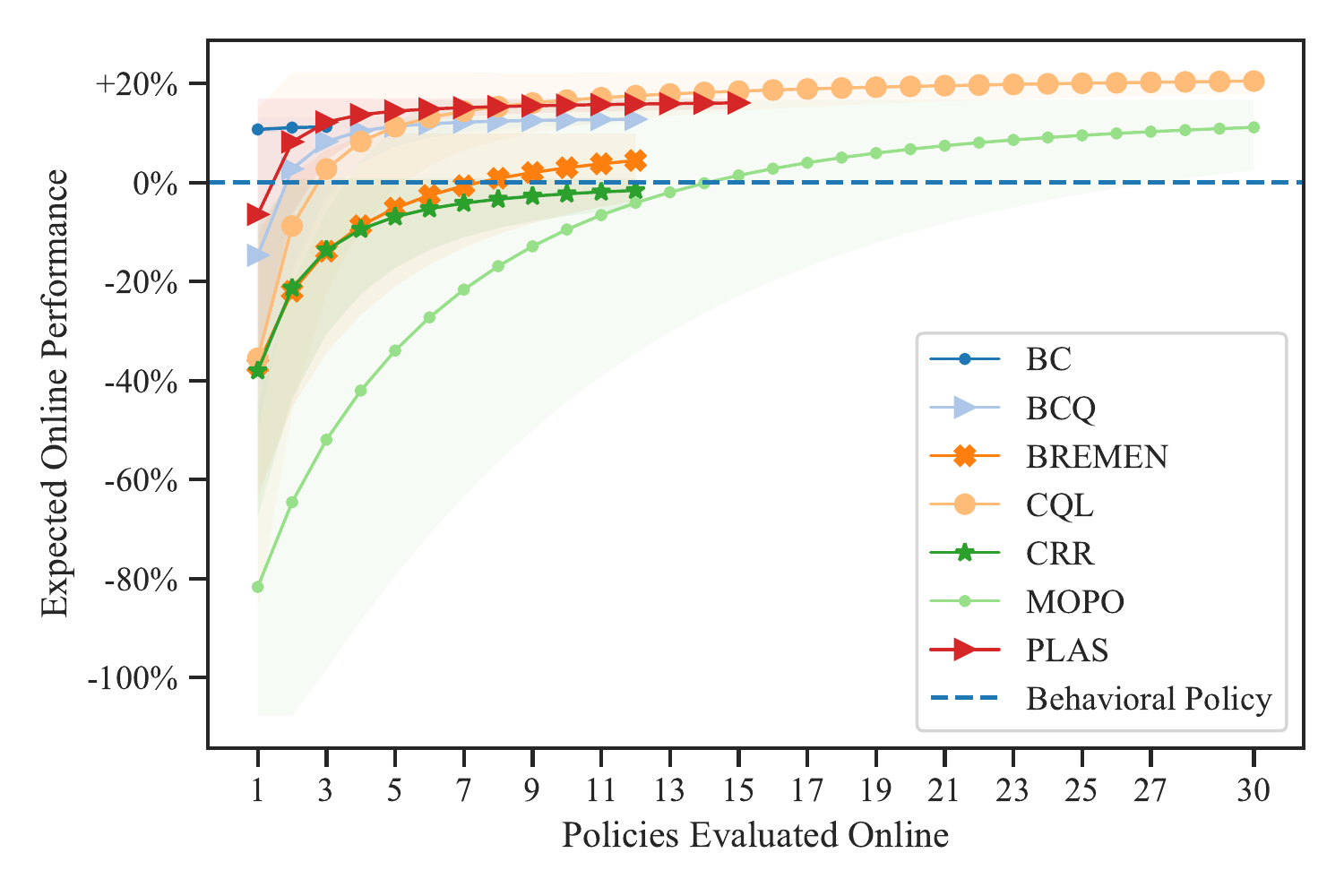}}
        \caption{HalfCheetah, High-1000}
    \end{subfigure}
    \caption{\textbf{Expected Online Performance graphs under uniform offline policy selection} on \citet{qinNeoRLRealWorldBenchmark2021} data. The proposed EOP graph clearly demonstrates that preference between algorithms is budget-dependent. Furthermore, it highlights that BC is often more favorable to offline RL algorithms under limited online evaluation budgets. The X-axis denotes the number of policies deployed online, and Y-axis refers to the normalized performance. Note that the number of estimates for a concrete algorithm is upper-bounded by the total number of hyperparameter assignments ($N$) evaluated for this algorithm. Shadowed area represents one standard deviation.}
    \label{fig:eop_neorl}
\end{figure*}

In the previous section, we demonstrated multiple model comparisons where authors would have reached a different conclusion if they had used a smaller (or bigger) online evaluation budget. To resolve this issue, we use a tool from the NLP field, Expected Validation Performance (EVP) \cite{dodgeShowYourWork2019}, that can be adapted for enhancing the quality of experimental reports in a deep ORL setting.

\subsection{Expected Online Performance}
\label{sec:eop_math}

Here, we give a detailed description for EVP, reframed for an offline RL setting, and with the computational budget replaced by an online evaluation budget (typically, $B \ll N$). We refer to this approach as Expected Online Performance (EOP).

Having all $N$ policies evaluated online after hyperparameter search, we want an estimate of the expected maximum performance, given that we could deploy only $1 \leq B \leq N$ policies out of $N$. 
\newpage
The parameters of interest to us are $\theta_{1}$,...,$\theta_{B}$, where 
\begin{equation} \label{eq:expected_online_performace}
    \theta_{b} := E[max(V(\pi_{1}),...,V(\pi_{b}))] = E[V_{(b:b)}]
\end{equation}
for $1 \leq b \leq B$ and $V$ is an random variable (RV) representing the result of online evaluation.
In other words, $\theta_{b}$ is the expected value of the $b^{th}$ order statistic for a sample of size $b$. The $i^{th}$ order statistic $V_{(i:b)}$ is an RV representing the $i^{th}$ smallest value if the RVs were sorted.

Originally, EVP operates over one stage -- hyperparameter value selection. But in an ORL setting, there is also a second stage -- policy selection (see Figure \ref{fig:eval_pipelines}, Right). To account for this discrepancy, we note that uniformly sampled hyperparameter values and then uniformly sampled policies result in the probability of a policy being selected for online evaluation proportional to the probability of its hyperparameters being used. Virtually, that makes $\theta_{b}$ be based on a sample size $b$ drawn independent and identically distributed. 

In this case, the estimator proposed in \citet{dodgeShowYourWork2019} can be readily applied. The derivation is similar to \citet{tangShowingYourWork2020}:
\begin{equation}
\begin{aligned}
    Pr[V_{(b:b)} < v] & = Pr[V(\pi_{1}) \leq v \wedge ... \wedge V(\pi_{b}) \leq v] \\
    & = \prod_{i=1}^{b} Pr[V(\pi_{i}) \leq v],
\end{aligned}
\end{equation}
which we denote as $F^{b}(v)$. Then
\begin{equation} \label{eq:expected_online_performace_cdf}
    \theta_{b} = E[V_{(b:b)}] = \int_{-\inf}^{\inf} vdF^{b}(v).
\end{equation}
Without loss of generality, assume $V(\pi_{1}) \leq ... \leq V(\pi_{N})$. To approximate the Cumulative Distribution Function (CDF), use Empirical Cumulative Distribution Function (ECDF)
\begin{equation}
    \hat{F}^{b}_{N}(v) = (\frac{1}{N}\sum_{i=1}^{N}I[V(\pi_{i}) \leq v])^{b}
\end{equation}
To arrive at the final estimator, replace CDF with an ECDF in Equation \ref{eq:expected_online_performace_cdf}
\begin{equation}
    \hat{\theta}_{b} = \int_{-\inf}^{\inf} vd\hat{F}^{b}_{N}(v)
\end{equation}
which, by definition, evaluates to
\begin{equation} \label{eq:expected_online_performace_uniform_selection}
    \hat{\theta}_{b} = \frac{1}{N}\sum_{i=1}^{N}V(\pi_{i})(\hat{F}^{b}_{N}(V(\pi_{i})) - \hat{F}^{b}_{N}(V(\pi_{i-1}))
\end{equation}

To summarize, $\hat{\theta_{b}}$ corresponds to the estimated expected maximum performance given that (1) hyperparameters were randomly sampled from a pre-defined grid, (2) we could deploy $1 \leq b \leq B$ policies out of $N$ for online evaluation, and (3) these $b$ policies were picked by uniform policy selection.

The major advantage of this estimator is that, if our evaluation methodology satisfies all three conditions described above, then the computation within a single round of hyperparameter search is sufficient to construct a reliable estimate of
expected online performance for different values of $b$, without requiring any further experimentation \cite{dodgeShowYourWork2019}.
Moreover, for a compact presentation, we can plot a graph over the entire range of values for $b$, demonstrating the dependence between the final performance and online evaluation budget (Figure \ref{fig:eop_neorl}).

Note, that there are alternative estimators for the quantity of interest \cite{tangShowingYourWork2020}. However, \citet{dodgeExpectedValidationPerformance2021} compared different approaches for estimating the expected maximum and found that the employed estimator \cite{dodgeShowYourWork2019} is favored amongst existing approaches in terms of both MSE criterion and a percent of incorrect conclusions.


\subsection{Target Metric}

\begin{figure*}[!h]
    \begin{subfigure}[b]{0.32\textwidth}
        \centering
        \centerline{\includegraphics[width=\columnwidth]{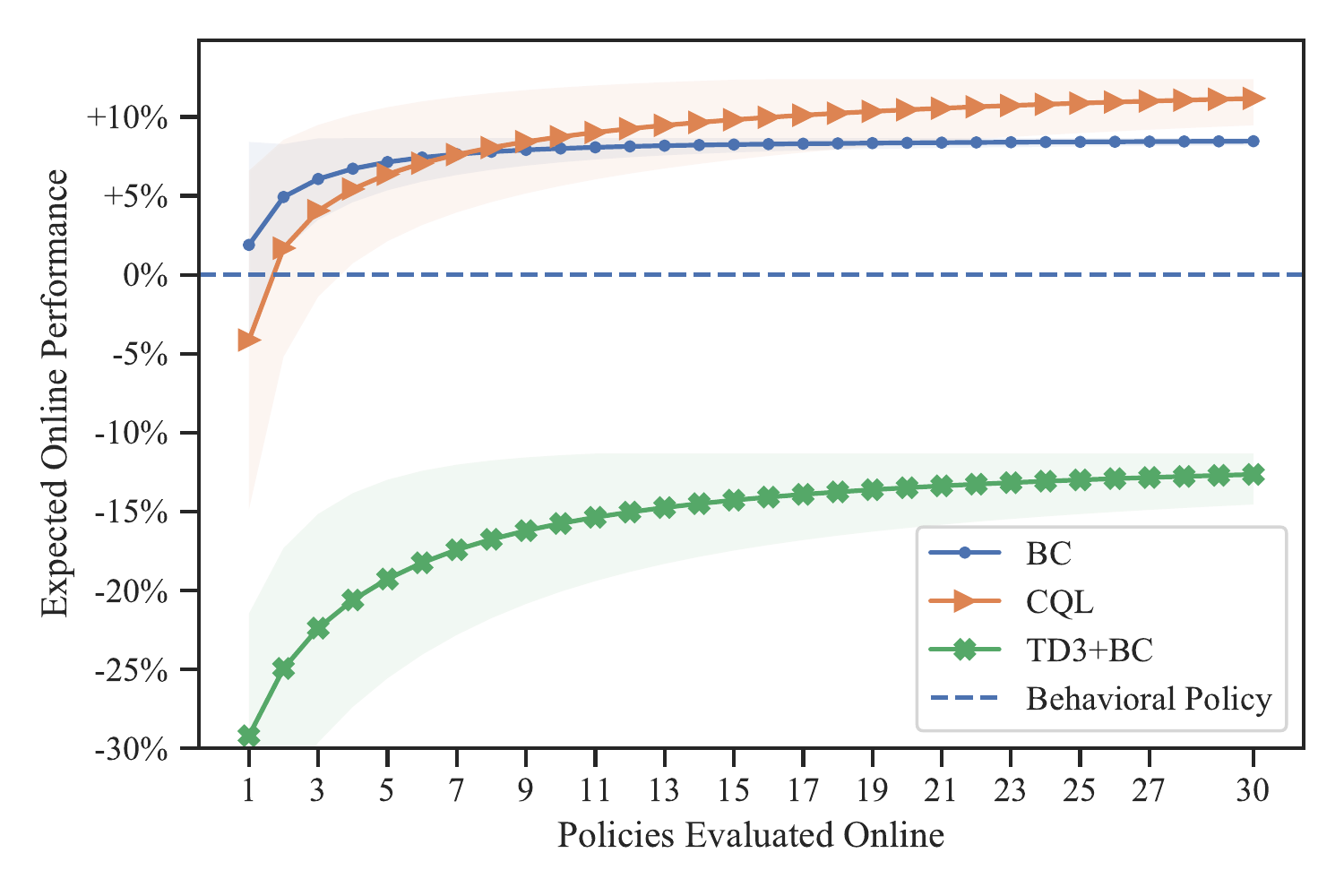}}
        \caption{CityLearn, Medium-1000}
    \end{subfigure}
    \begin{subfigure}[b]{0.32\textwidth}
        \centering
        \centerline{\includegraphics[width=\columnwidth]{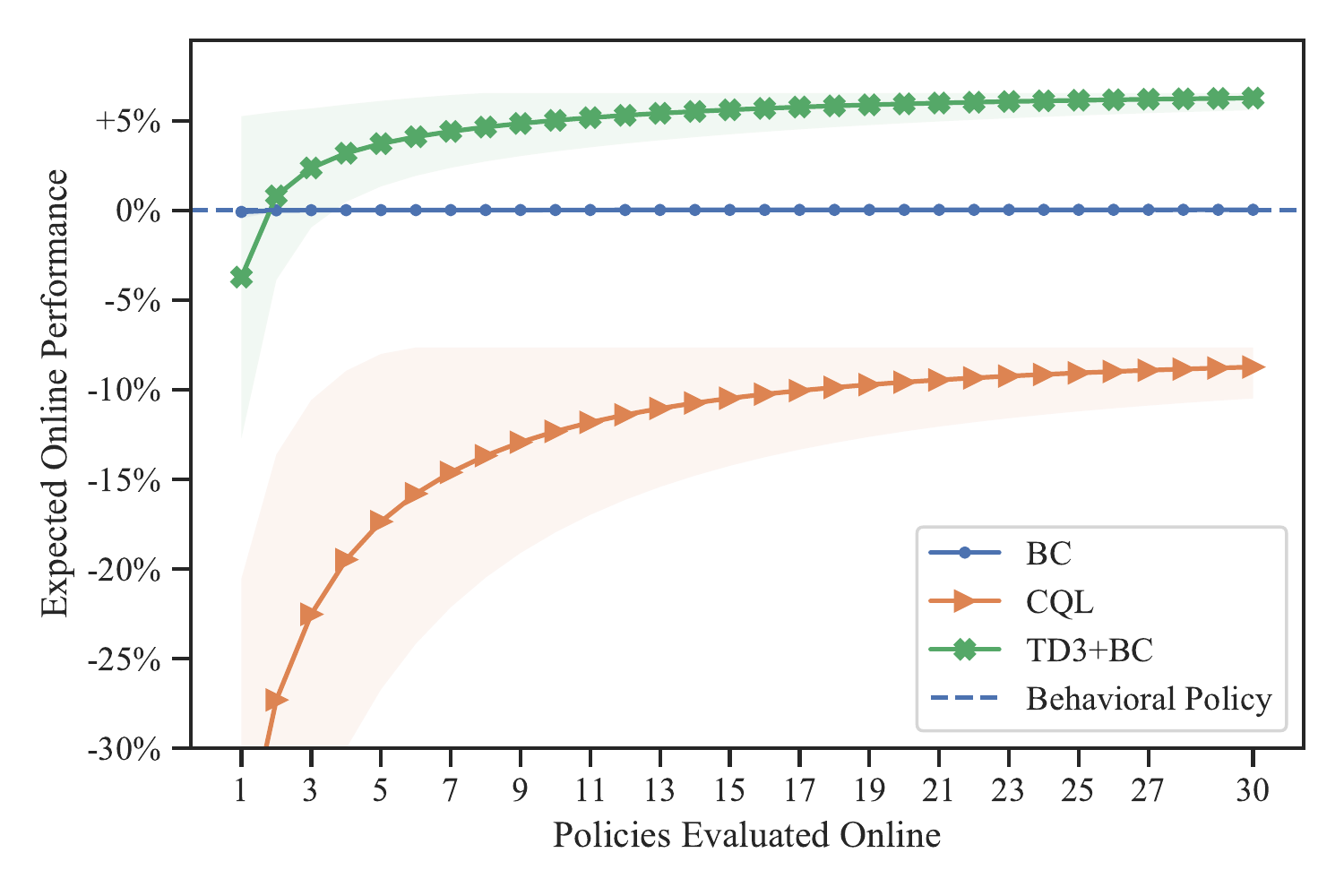}}
        \caption{FinRL, High-1000}
    \end{subfigure}
    \begin{subfigure}[b]{0.32\textwidth}
        \centering
        \centerline{\includegraphics[width=\columnwidth]{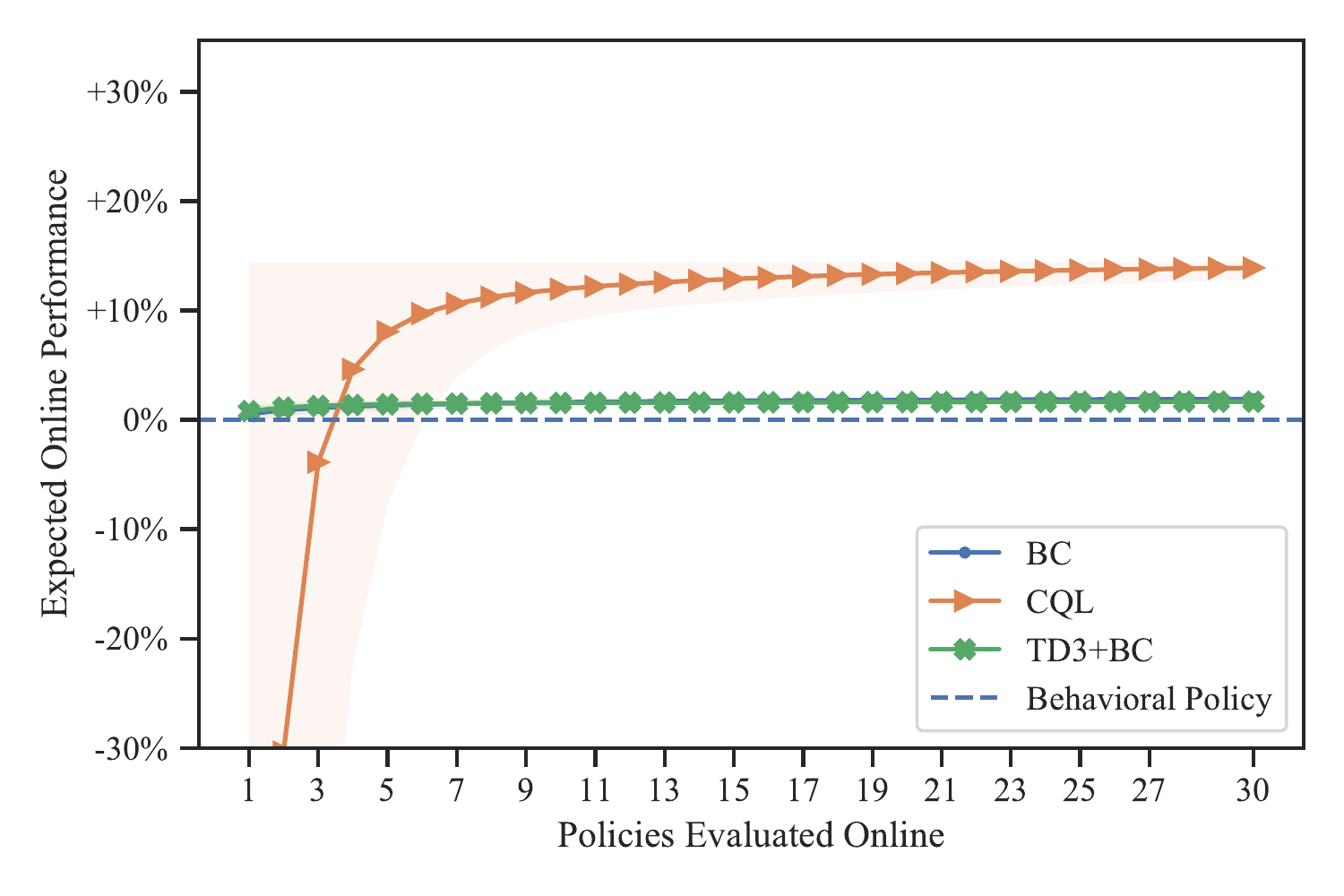}}
        \caption{Industrial Benchmark, Medium-1000}
    \end{subfigure}
    \caption{\textbf{Expected Online Performance graphs under uniform offline policy selection.} The proposed EOP graph clearly demonstrates that preference between algorithms is budget dependent. Furthermore, it highlights that BC is often more favorable to offline RL algorithms under limited online evaluation budgets. Shadowed area represents one standard deviation.}
    \label{fig:eop_other_domains}
\end{figure*}

The target metric can be represented by any convenient measure used in literature, e.g., absolute policy performance or policy performance normalized by an expert \cite{fuD4RLDatasetsDeep2021}. In our case studies (Section \ref{sec:case_studies}), we rely on a modified version of the latter. The main motivation behind this modification is that the original metric is normalized by the value provided by a domain-specific expert \cite{fuD4RLDatasetsDeep2021}. However, the final results of offline RL algorithms are highly dependent on training data, and expecting to achieve expert performance while training on data from weak policies can be too optimistic. Therefore, we normalize by the performance of the best policy (as there can be multiple) that collected the training data.

\subsection{Online Evaluation Budget}

The original EVP makes it possible to use various quantities as an argument to the target metric, e.g. number of hyperparameters enumerated or training time. Similarly, for EOP in the deep ORL setting, several options can be used, such as the number of trajectories, number of timesteps, or number of policies. We suggest using the latter option, and equate the online evaluation budget $B$ with the number of policies deployed for online evaluation. This choice provides researchers with the flexibility of defining their own amount of computation for getting reliable estimates for policy values.






\subsection{Beyond Uniform Policy Selection}

\begin{figure*}[!h]
    \begin{subfigure}[b]{0.32\textwidth}
        \centering
        \centerline{\includegraphics[width=\columnwidth]{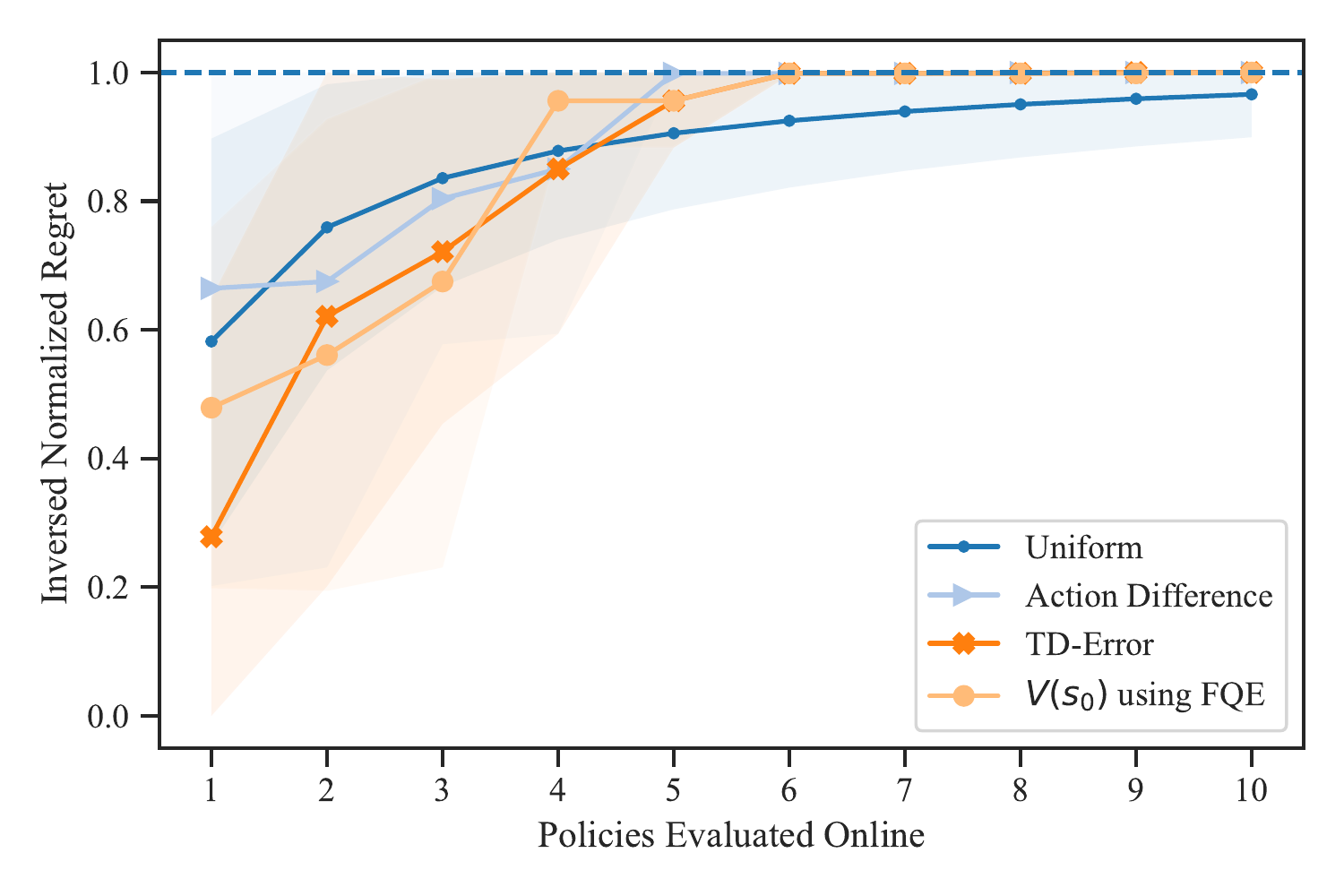}}
        \caption{CQL, CityLearn, Low-10000}
        \label{fig:eop_ops_a}
    \end{subfigure}
    \begin{subfigure}[b]{0.32\textwidth}
        \centering
        \centerline{\includegraphics[width=\columnwidth]{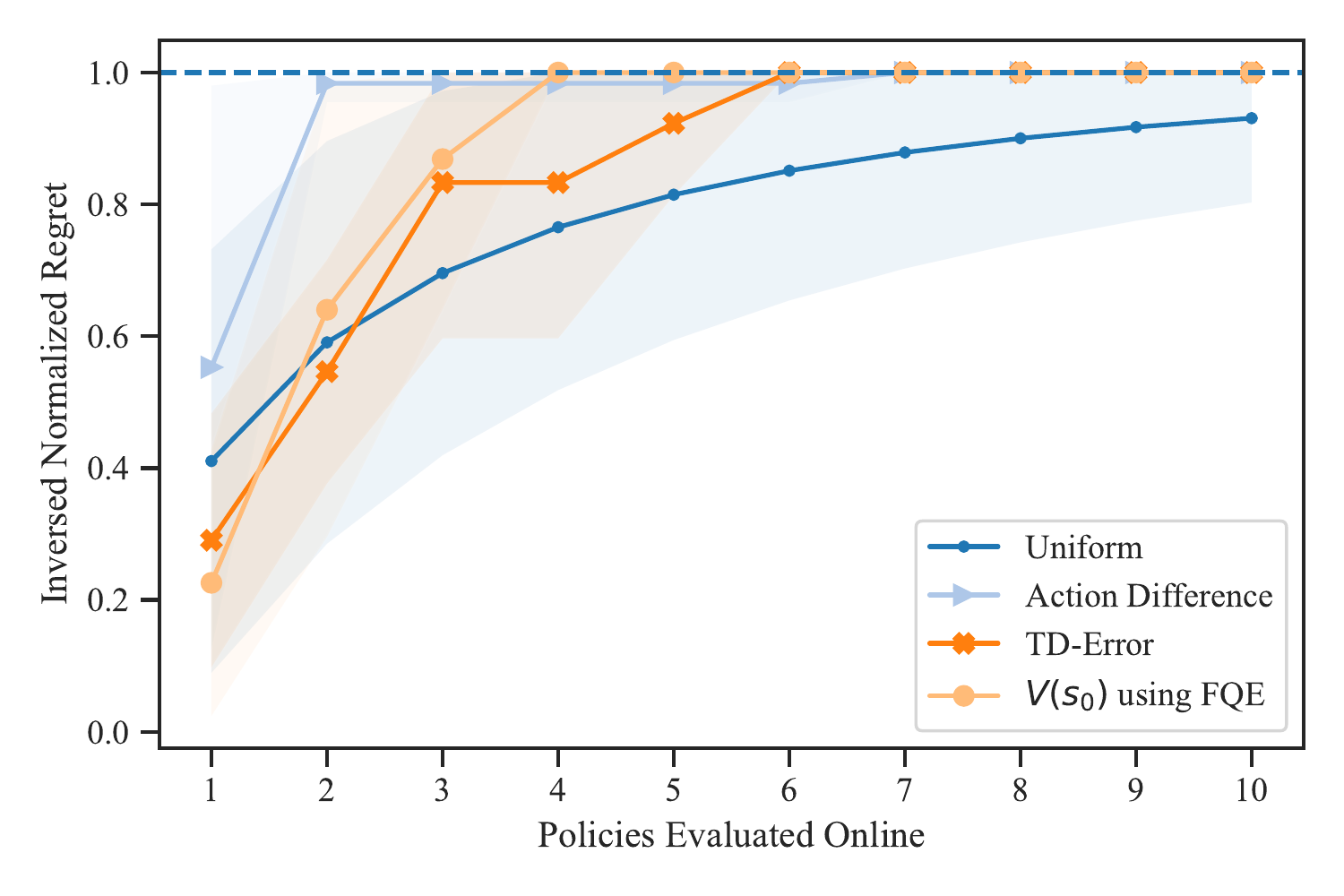}}
        \caption{TD3+BC, CityLearn, Low-10000}
        \label{fig:eop_ops_b}
    \end{subfigure}
    \begin{subfigure}[b]{0.32\textwidth}
        \centering
        \centerline{\includegraphics[width=\columnwidth]{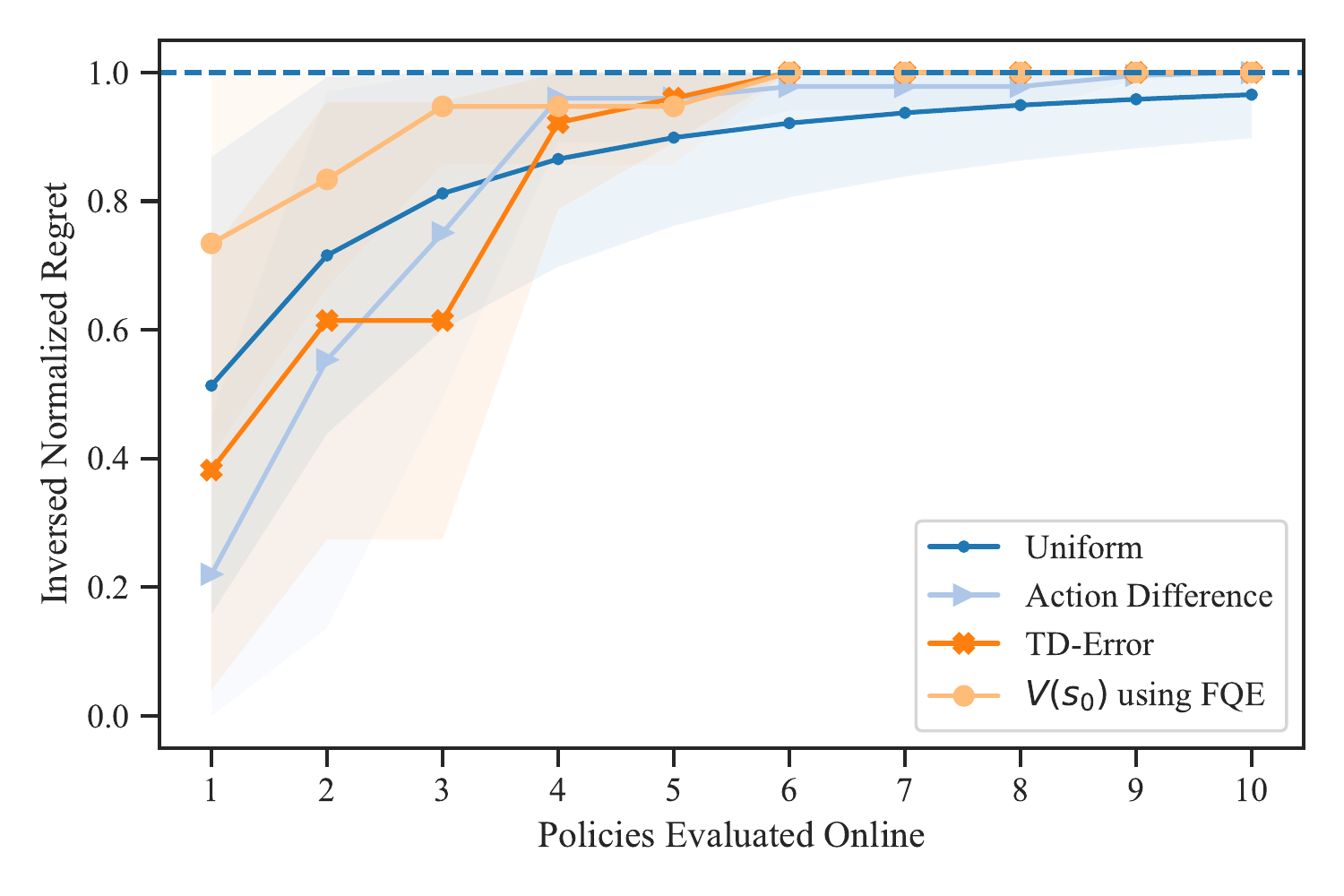}}
        \caption{CQL, FinRL, Low-1000}
        \label{fig:eop_ops_c}
    \end{subfigure}
    \caption{\textbf{Inverse of Normalized Regret@K over varied online evaluation budgets}. The suggested method can also be used for comparing offline policy selection methods. A similar pattern emerges -- preference for OPS methods is also budget-dependent. Furthermore, uniform policy selection can be competitive under low budgets, and action difference performs reasonably well across a range of environments, dataset sizes, and policy levels. Shadowed area represents one standard deviation.}
    \label{fig:eop_ops}
\end{figure*}

In Section \ref{sec:eop_math} we outlined an estimator of expected maximum performance under varied online evaluation budgets. Although, an assumption was made that besides randomly sampled hyperparameters, the policies are also selected uniformly. However, comparing deep ORL methods with different OPS methods is also of interest.

Since the policies selected under an arbitrary strategy (e.g. Fitted-Q Evaluation \cite{leBatchPolicyLearning2019}) are generally not i.i.d, the plug-in estimator derived above would be invalid. However, one can still estimate the parameters of interest using a vanilla average estimator.
\begin{equation}
    \hat{\theta_{b}} = \frac{1}{M}\sum^{M}_{r=1}max(V^{r}_{1}, ..., V^{r}_{b})
\end{equation}
where $M$ is a number of hyperparameter search rounds with offline policy selection, $V^{r}_{i}$ corresponds to an RV representing the result of online evaluation for the $i^{th}$ selected policy in round $r$. Note that this estimator loses the appealing computational side of EVP, and requires multiple runs of hyperparameter search with offline policy selection. This makes it more expensive in terms of computing time.

In one of our case studies (Section \ref{sec:case_ops}), we demonstrate how one could use EOP for comparing not only deep ORL algorithms, but OPS methods as well.

\section{Case Studies}
\label{sec:case_studies}

To further demonstrate the use of the proposed technique and to identify whether online evaluation budget changes the preference between deep ORL algorithms besides the environment analyzed in Section \ref{sec:online-eval-matters}, we consider several case studies covering a range of decision-making problems: robotics, finances, and energy management. 

\subsection{NeoRL, Robotic Tasks}
\label{sec:neorl_robotic_tasks}


Continuing the closer look at the results presented in \citet{qinNeoRLRealWorldBenchmark2021} from Section \ref{sec:online-eval-matters}, we build Expected Online Performance graphs for other robotics environments (Figure \ref{fig:eop_neorl}). This once again confirms that the preference is budget-dependent. Moreover, this is consistent across environments, dataset sizes, and policy levels (for a complete set of graphs, check the Appendix). 
There are many cases when the conclusion changes with a budget. For example, in the Walker-2d environment, CQL is clearly preferred to PLAS \cite{zhouPLASLatentAction2020a}, but only up to the 9-10 policies available for online evaluation. Another example can be seen in Figure \ref{fig:eop_neorl}b for the Hopper environment: Behavioral Cloning significantly outperforms its competitiors at low budgets, but loses to PLAS at higher ones. The same holds for the HalfCheetah environment, where CQL starts to prevail at higher budgets.

As the online budget is upper-bound by the total number of enumerated hyperparameters, it is clear (Figure \ref{fig:eop_neorl}) that different algorithms were tuned more or less excessively. This results in more optimistic results for one algorithm, and in more pessimistic for another. Consider BCQ \cite{leBatchPolicyLearning2019} in Figure \ref{fig:eop_neorl}a. It is tuned up to 13 hyperparameter assignments showing the best result, but as long as a competing algorithm, PLAS, is tuned for 14 and more assignments, it starts to outperform BCQ. An even more vivid example is depicted in Figure \ref{fig:eop_neorl}c, where MOPO \cite{yuMOPOModelbasedOffline2020} starts to outperform both BREMEN and CRR at 2x more hyperparameters tested. Note that reporting just one policy value (either using online policy selection or proxy tasks) hides this issue.


\subsection{NeoRL and Other Domains}
\label{sec:neorl_other_tasks}

To validate that our findings hold outside of the open-sourced experimental results provided by \citet{qinNeoRLRealWorldBenchmark2021}, and to cover a wider range of decision-making problems, we benchmark CQL \cite{kumarConservativeQLearningOffline2020}, TD3+BC \cite{fujimotoMinimalistApproachOffline2021}, and BC on the CityLearn \cite{vazquez-canteliCityLearnV1OpenAI2019}, FinRL \cite{liuFinRLDeepReinforcement2020}, and Industrial Benchmark \cite{heinBenchmarkEnvironmentMotivated2017} environments\footnote{Detailed descriptions of the environments and algorithms used can be found in the Appendix \ref{appendix:sec:hyperparams}, \ref{appendix:sec:envs_bases_datasets}.}. In addition, we make sure that the hyperparameter search budgets are equal for all the algorithms to avoid the issue described in the previous section. The hyperparameter grids were deferred to the Appendix \ref{appendix:sec:hyperparams}. We average mean returns over 100 evaluation trajectories and 3 seeds.

In Figure \ref{fig:eop_other_domains}, we see that Behavioral Cloning is quite competitive against both CQL and TD3+BC under limited online evaluation budgets. This akin to the results we observed in robotics environments (Figure \ref{fig:eop_neorl}), suggesting that BC is often more preferable in restricted settings to deep ORL algorithms.

\subsection{Offline Policy Selection}
\label{sec:case_ops}

As the EOP incorporates offline policy selection, we can also use it to compare how well OPS methods perform against each other.
To do so, we use inverse of normalized Regret@K (in our case at $B$) as a target metric. This allows us to answer the following question: "If we were able to run policies corresponding to $k$ hyperparameter settings in the actual environment and get reliable estimates for their values that way, how far would the best in the set we picked be from the best of all hyperparameter settings considered?" \cite{paineHyperparameterSelectionOffline2020}. But instead of reporting one value of $k$ as in \citet{paineHyperparameterSelectionOffline2020}, we can easily report on all the values of $B$. Moreover, the estimator outlined in Section \ref{sec:eop_math} can be used for presenting the results of uniform policy selection.

We do not aim to benchmark and compare the entire myriad of offline policy selection approaches, but rather to demonstrate how one can use the proposed tool for such purposes. To do so, we test several methods on the environments from the previous section, namely $V(s_0)$ using FQE, TD-Error, and Action Difference \cite{leBatchPolicyLearning2019,hussenotHyperparameterSelectionImitation2021}.

The results can be found in Figure \ref{fig:eop_ops}. First, we observe a pattern similar to the one described in Sections \ref{sec:online-eval-matters}, \ref{sec:neorl_robotic_tasks}, \ref{sec:neorl_other_tasks}: \textit{preference between offline policy selection methods is also budget-dependent}. Therefore, it is not enough to report the result of such methods under just one selected threshold.
Second, there is no clear winner among all setups, as even TD-Error may sometimes perform good (Figure \ref{fig:eop_ops_a}, for more -- check the Appendix). However, Action Difference often performs reasonably well across many dataset sizes and policy levels in Industrial Benchmark and CityLearn environments (Figures \ref{fig:eop_ops_a}, \ref{fig:eop_ops_b}). We hypothesize that such a selection method can serve as a post-training conservative regularizator (i.e. picking policies that are more similar to the behavioral ones), but that requires further investigation.
And the last notable observation is that uniform policy selection can be competitive to other considered methods, especially when the online evaluation budget is limited. Sometimes it can even perform the best among the entire set of methods (Figure \ref{fig:eop_ops_a}).

 
\section{Related Work}
To the best of our knowledge, the closest concept to our work is deployment-constrained RL \cite{matsushimaDeploymentEfficientReinforcementLearning2020, suMUSBOModelbasedUncertainty2021}. The core idea of this setting is to consider the number of policies deployed online and to reuse the data for iterative training from such deployments. \citet{matsushimaDeploymentEfficientReinforcementLearning2020,suMUSBOModelbasedUncertainty2021} propose new algorithms that are especially suited for this setting, claiming that they are more deployment-efficient. However, they also relied on an extensive hyperparameter search reporting for the best set of hyperparameters. This hides the actual number of policies evaluated online, while our approach prevents that. An interesting direction for future work would be to adapt the EOP for this iterative setting as well.

\citet{konyushkovaActiveOfflinePolicy2021} formulated a new problem, which is an extension to OPS, Active Offline Policy Selection (A-OPS). The major difference is to allow for an OPS method to have a feedback loop from newly trained policies deployed online, and re-adjust which policy should be run next. While this is an important step forward, EOP can actually subsume A-OPS as one of the OPS methods, since sequential policy testing is allowed. Furthermore, our paper aims to consolidate all the parts of a deep ORL pipeline, while \citet{konyushkovaActiveOfflinePolicy2021} focuses on a new problem.


Recently, \citet{agarwalDeepReinforcementLearning2022} scrutinized the evaluation methodology of deep RL algorithms, and advocated for a set of statistical tools to be employed for more reliable comparison. However, \citet{agarwalDeepReinforcementLearning2022} focuses on reliable evaluation \textit{after hyperparameter tuning}, while our work highlights its importance in offline deep RL setting, and argues that results should be reported under varied hyperparameter tuning capacity when comparing deep ORL algorithms.

\citet{brandfonbrenerQuantileFilteredImitation2021} noted the extensive online evaluation budgets used in recent works on deep ORL. To address this issue, comparison between algorithms was made under a small hyperparameter tuning budget ($B=4$). However, evaluating only under a limited budget may not be enough. Our paper demonstrates that the preference between algorithms can be budget-dependent, requiring evaluation under various budgets.

There is a sizeable body of work on Offline Policy Evaluation \cite{voloshinEmpiricalStudyOffPolicy2020, fuBenchmarksDeepOffPolicy2021} and, specifically, on Offline Policy Selection \cite{paineHyperparameterSelectionOffline2020, hussenotHyperparameterSelectionImitation2021, yangOfflinePolicySelection2020}. This work does not aim to compare or benchmark these types of methods, but to provide a procedure for comparing deep ORL algorithms with OPS (and promote the usage of simple uniform selection) for achieving reliable conclusions. In addition, we demonstrate that a similar pattern, online-budget dependence, is relevant for OPS methods as well.

Behavioral Cloning is typically reported in deep ORL papers as a baseline, and many papers claim to beat this baseline \cite{kumarConservativeQLearningOffline2020, fujimotoMinimalistApproachOffline2021, kumarStabilizingOffPolicyQLearning2019, leBatchPolicyLearning2019, wangCriticRegularizedRegression2020, wuBehaviorRegularizedOffline2019}. However, when considering learning from human demonstrations, \citet{mandlekarWhatMattersLearning2021} demonstrated the superiority of BC over deep ORL algorithms, especially when recurrence is employed in network architecture. Reinforcing the effectiveness of BC, this work suggests that BC is preferable not only in settings with human demonstrations, but across a diverse range of decision-making problems, given that the online evaluation budget is severely limited.




\section{Closing Remarks}








\textbf{Motivation}:
While a lot of RL community efforts are focused on offline RL datasets' general evaluation, this paper questions the methods' performance under the entire spectrum of their hyperparameters and available resources.
As different problems and contexts may allow for different online evaluation budgets, we argue that they can also make different optimal solutions possible within these constraints.
We hope that the proposed evaluation technique and our findings will encourage the ORL community to report performance results under different online evaluation budgets.

\textbf{Limitations}:
Although this work emphasizes the importance of several online evaluations, it does not investigate the possible evaluation risks. Many real-world applications have critical corner cases, especially in the autonomous driving, healthcare, and finance domains \cite{10.1007/s10664-020-09881-0}. We note that the EOP has limited applicability in such risk-sensitive scenarios due to its focus on maximum performance.


\textbf{Opportunities}:
EOP proposes a unified methodology for finding the best-performing setup under different online budget constraints. Unlike Deep Learning domains, the deep ORL evaluation is still in active development. The possibility of having a standard performance evaluation report opens avenues for adopting more precise methods for different tasks and contexts or creating online budget-dependent ORL algorithms.





\section{Conclusion}

A lot of community effort was recently devoted to developing new algorithms and datasets, while noting that the whole evaluation pipeline is still to be improved upon \cite{fuD4RLDatasetsDeep2021, gulcehreRLUnpluggedSuite2021}. In this work, we demonstrated (Section \ref{sec:online-eval-matters}) one of the problems with such pipelines, and proposed a technique named EOP (Section \ref{sec:eop_math}) to address it when presenting the results of deep ORL algorithms. 
Several empirical results were found (Section \ref{sec:case_studies}): (1) Behavioral Cloning is often more
favorable under a limited evaluation budget, (2) Online Policy Selection method preferences are also budget-dependent.


\bibliography{icml-bibliography}
\bibliographystyle{icml2022}

\clearpage
\appendix

\section{Target Metrics}
\subsection{Performance Normalized by the Best Policy in Training Data}
\begin{equation*}
    V_{normalized}(\pi) = \frac{V(\pi) - V(\pi_{best})}{V(\pi_{best})}
\end{equation*}
where $\pi_{best}$ is the best performing policy of all the policies that collected the training data. We make sure that the performance is bigger than zero for all the environments.

\subsection{Inversed Normalized Regret@K}
Given the true values of all $N$ policies,
\begin{equation*}
    Regret@K = \frac{V(\pi) - V(\pi_{worst})}{V(\pi_{best}) - V(\pi_{worst})}
\end{equation*}

where $\pi$ is the best performing policy among first $K$ ranked by a given offline policy selection method, $\pi_{best}$ is the best performing across all $N$ policies, and $\pi_{worst}$ is the worst performing across all $N$ policies.

\section{Experimental Setup in Other Domains}
\begin{figure*}[h]
 \centering
\begin{subfigure}[b]{.32\textwidth}
 \centering
  \centerline{\includegraphics[width=\textwidth]{figures/finrl_999_high_uniform.pdf}}
  \caption{High-Level Policy, 999 }
\end{subfigure}
\begin{subfigure}[b]{.32\textwidth}
 \centering
  \centerline{\includegraphics[width=\textwidth]{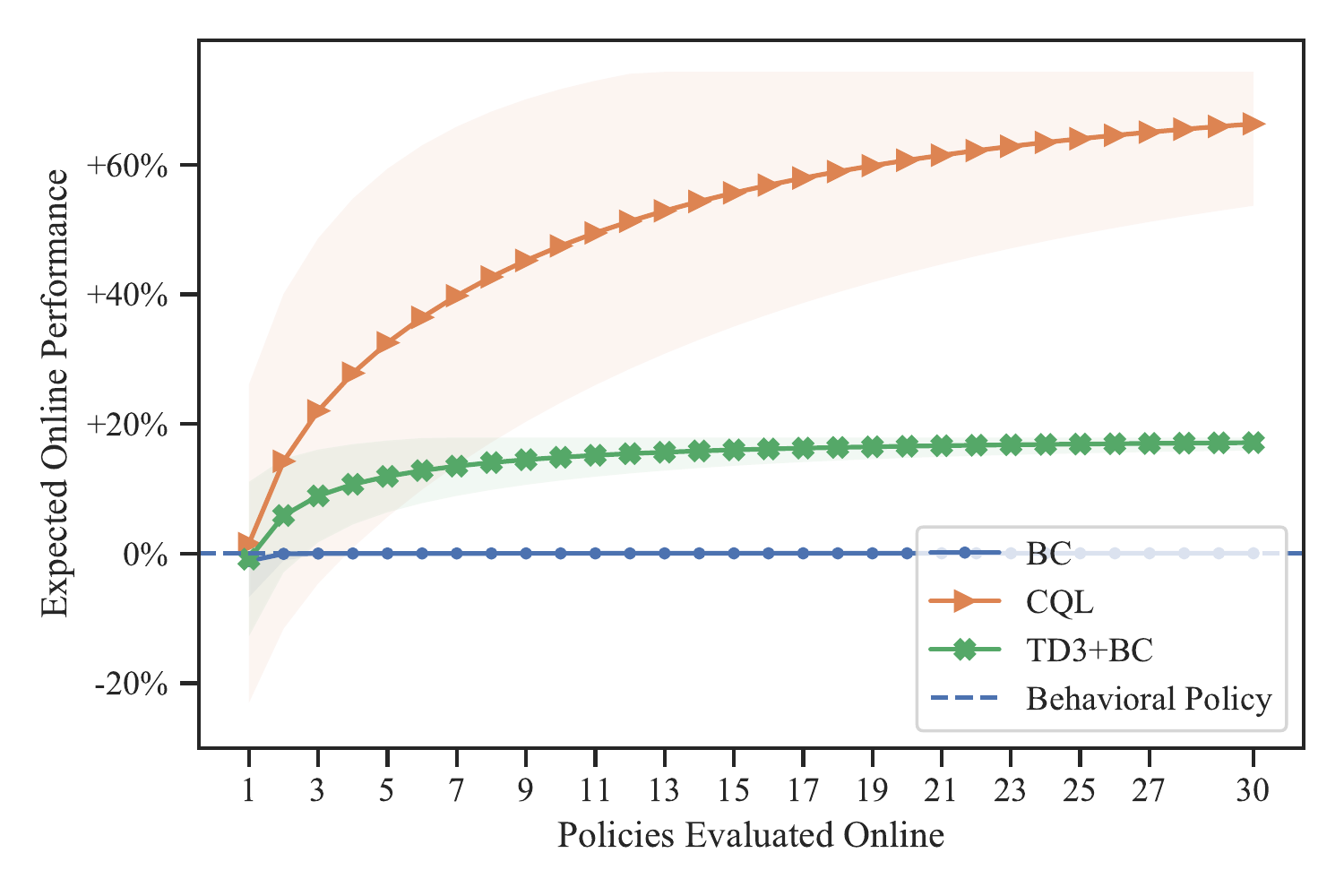}}
  \caption{Medium-Level Policy, 999 }
\end{subfigure}
\begin{subfigure}[b]{.32\textwidth}
 \centering
  \centerline{\includegraphics[width=\textwidth]{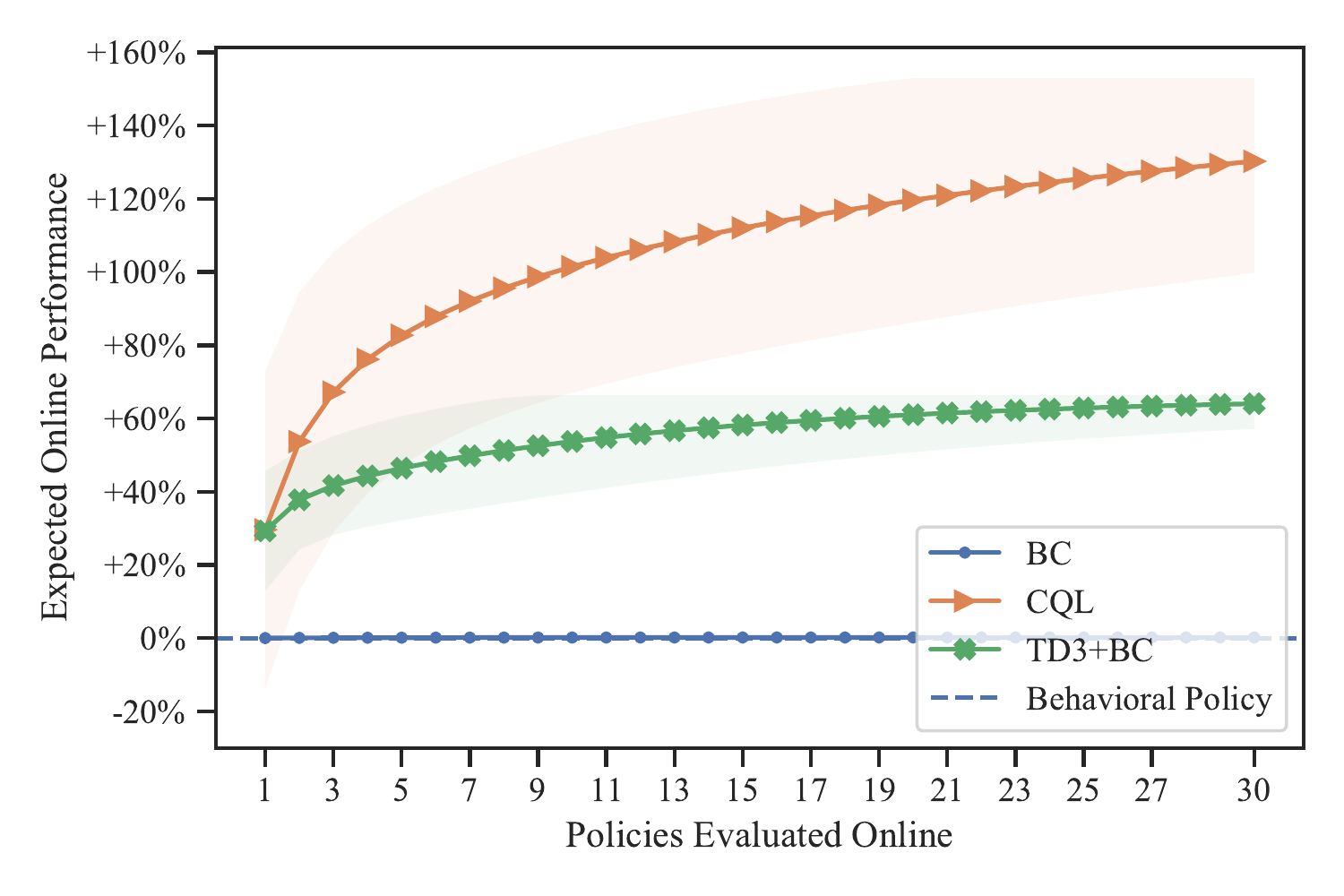}}
  \caption{Low-Level Policy, 999}
\end{subfigure}

\begin{subfigure}[b]{.32\textwidth}
 \centering
  \centerline{\includegraphics[width=\textwidth]{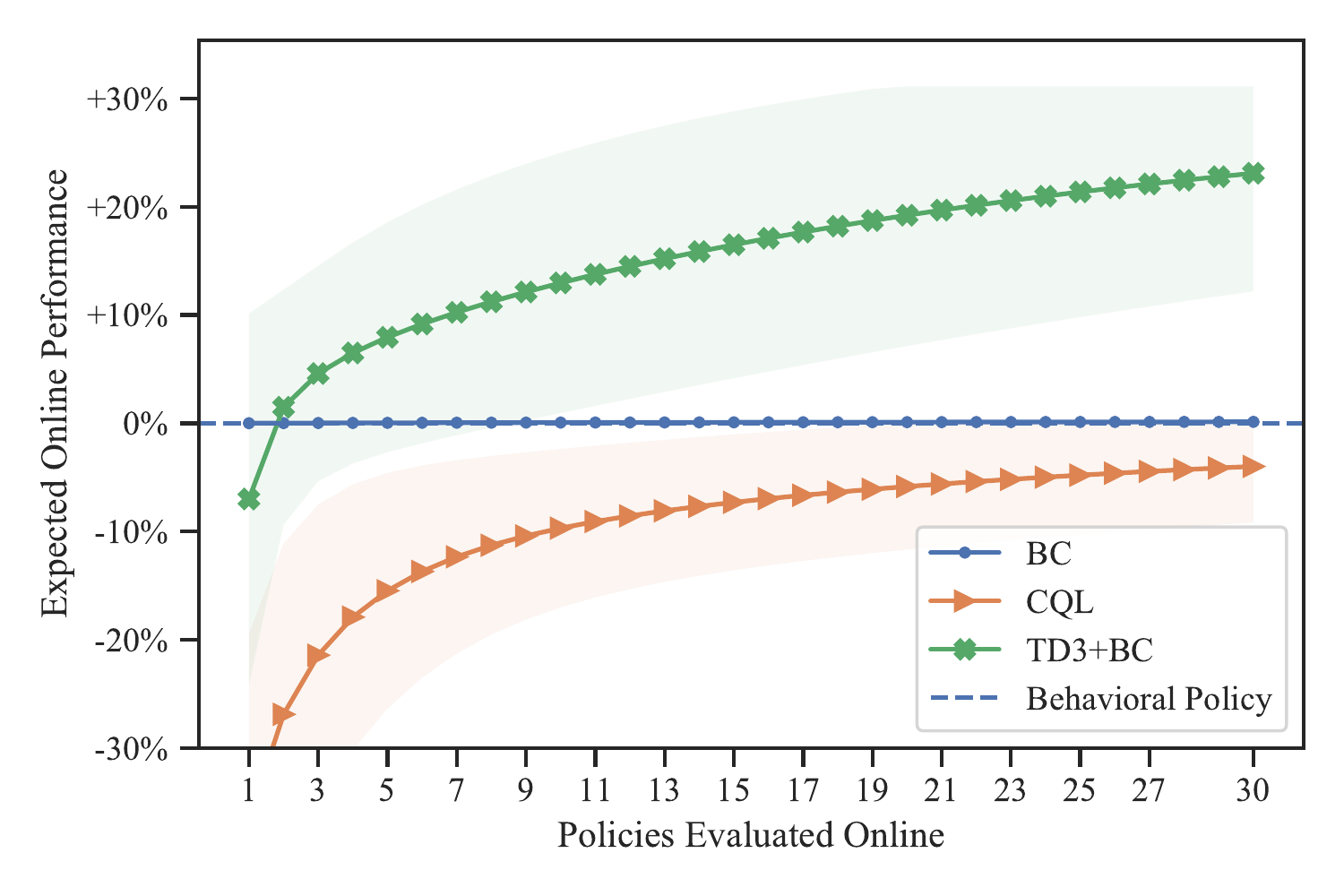}}
  \caption{High-Level Policy, 99 }
\end{subfigure}
\begin{subfigure}[b]{.32\textwidth}
 \centering
  \centerline{\includegraphics[width=\textwidth]{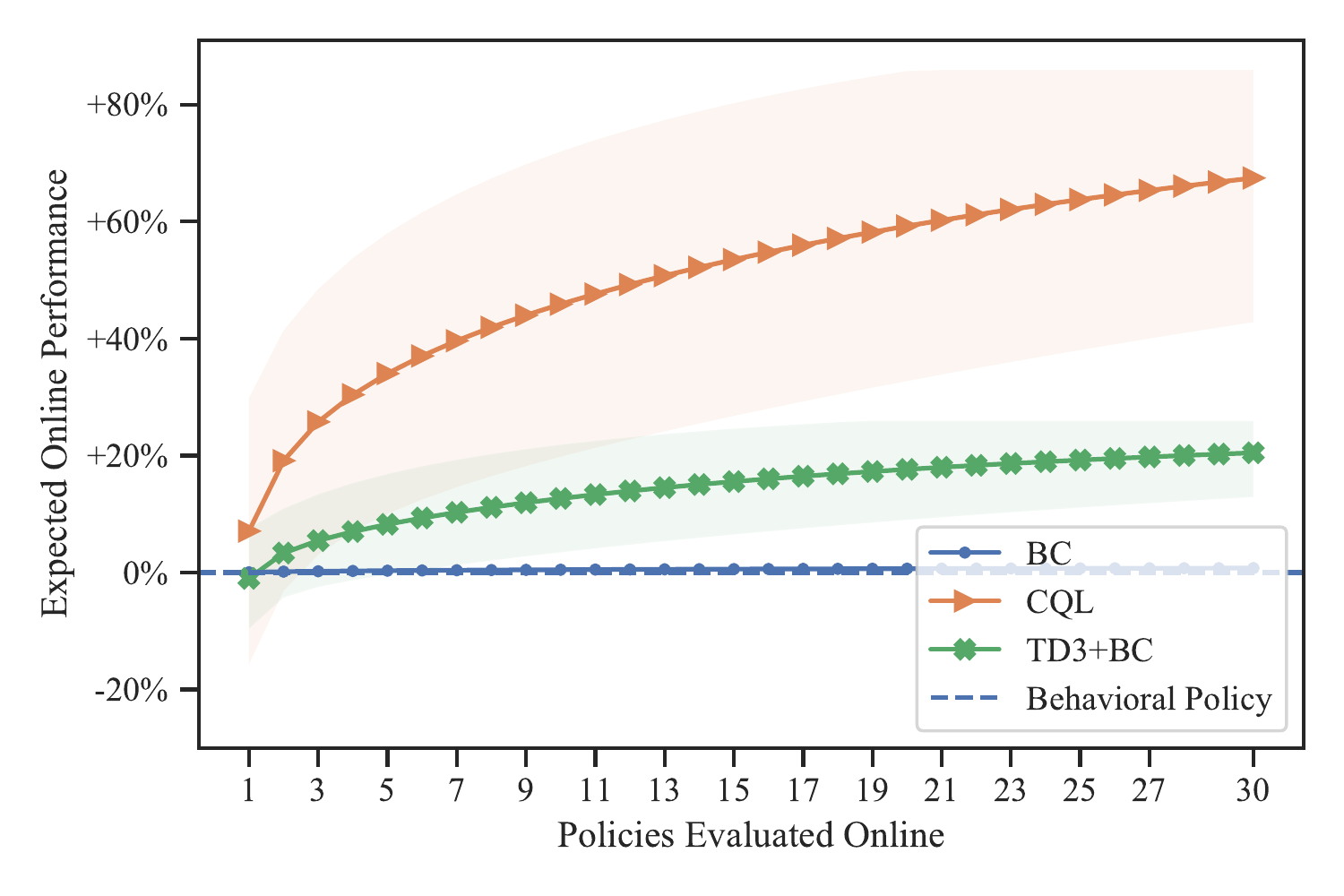}}
  \caption{Medium-Level Policy, 99 }
\end{subfigure}
\begin{subfigure}[b]{.32\textwidth}
 \centering
  \centerline{\includegraphics[width=\textwidth]{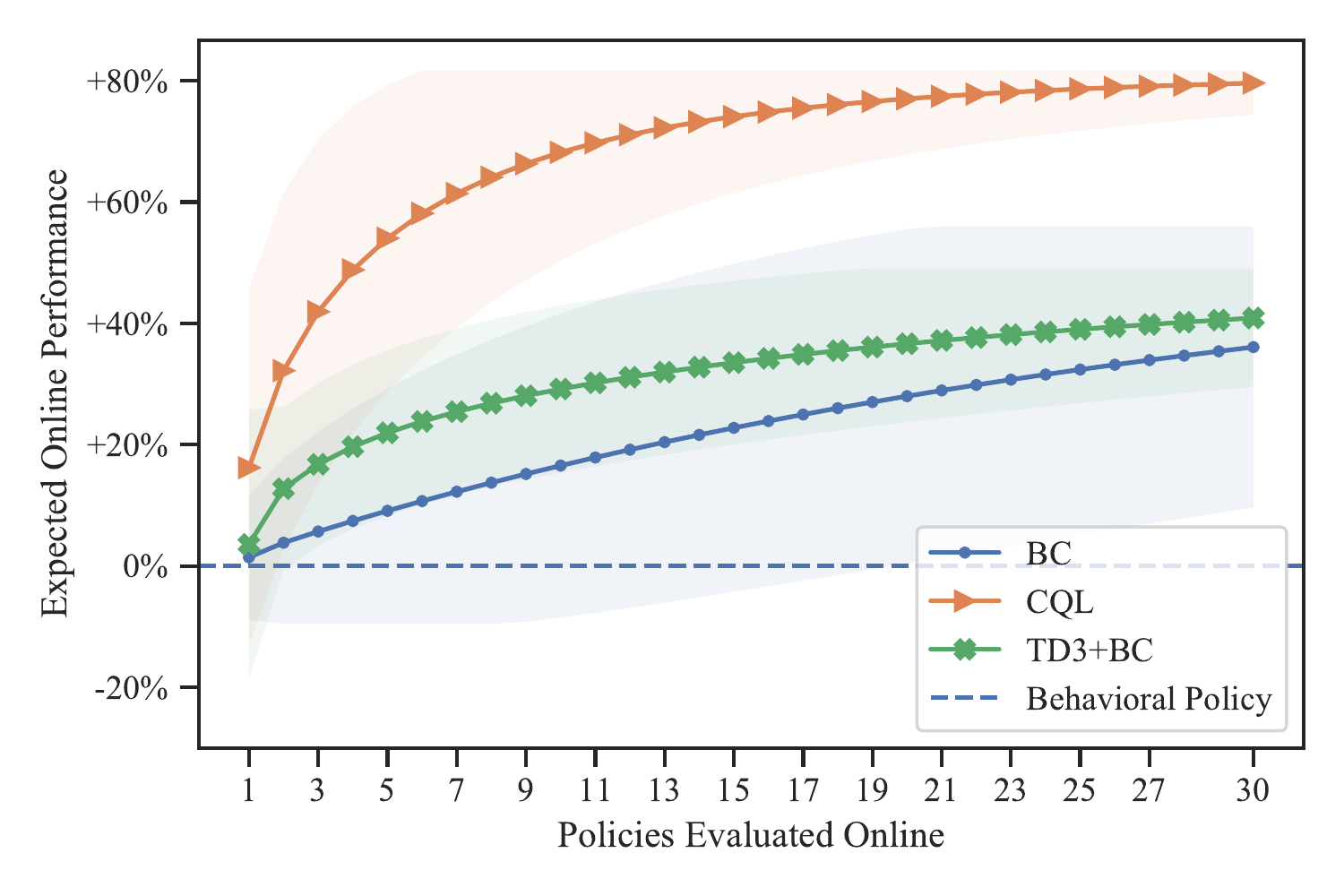}}
  \caption{Low-Level Policy, 99 }
\end{subfigure}

\caption{\textbf{Expected Online Performance under uniform policy selection. FinRL}, one of our experiments, comparing CQL, TD3+BC, and BC under equal amounts of online evaluation budgets.}
\label{fig:appendix:cql_finrl}
\end{figure*}

\begin{figure*}[h]
 \centering
\begin{subfigure}[b]{.32\textwidth}
 \centering
  \centerline{\includegraphics[width=\textwidth]{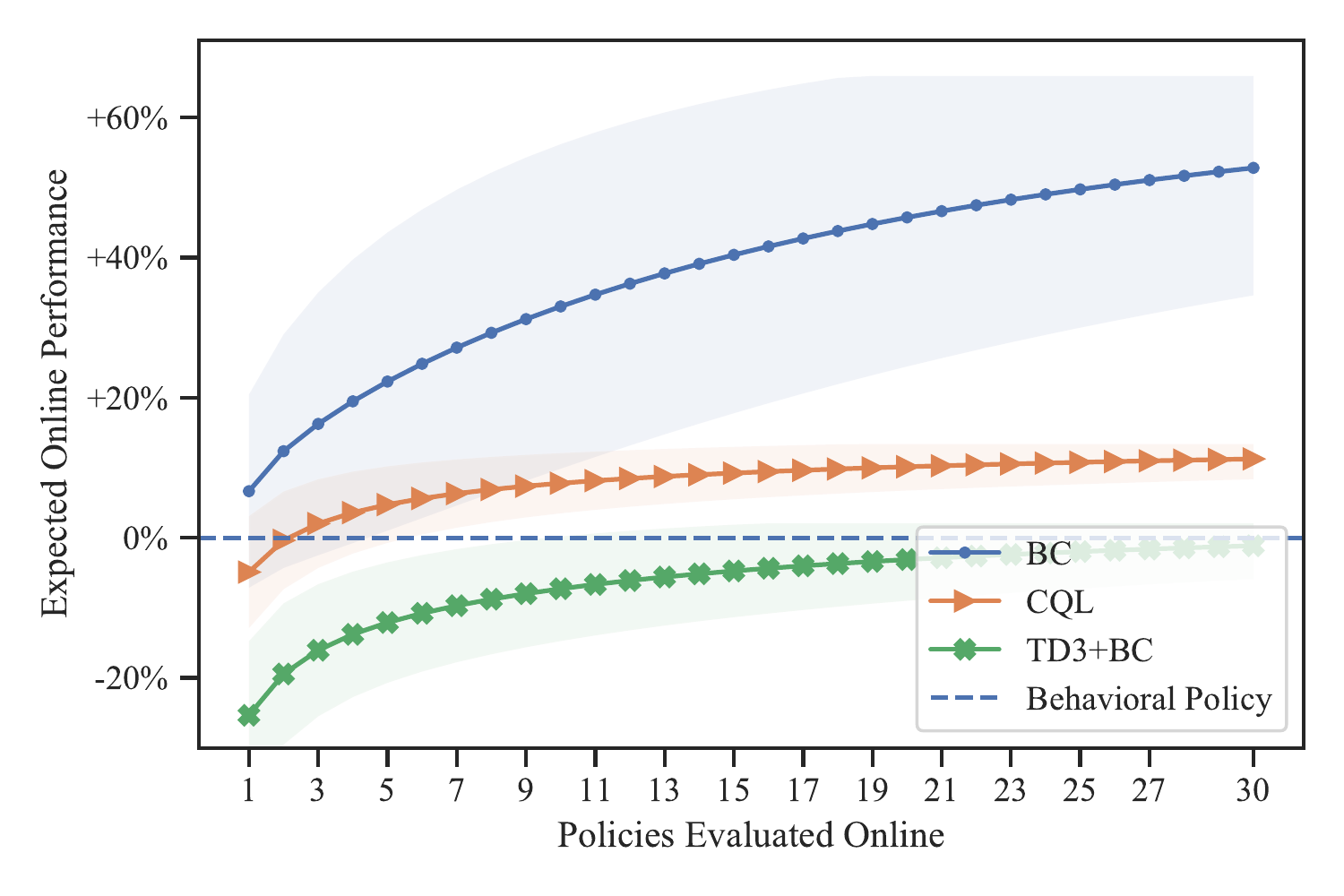}}
  \caption{Medium-Level Policy, 9999 }
\end{subfigure}
\begin{subfigure}[b]{.32\textwidth}
 \centering
  \centerline{\includegraphics[width=\textwidth]{figures/citylearn_999_medium_uniform.pdf}}
  \caption{Medium-Level Policy, 999 }
\end{subfigure}
\begin{subfigure}[b]{.32\textwidth}
 \centering
  \centerline{\includegraphics[width=\textwidth]{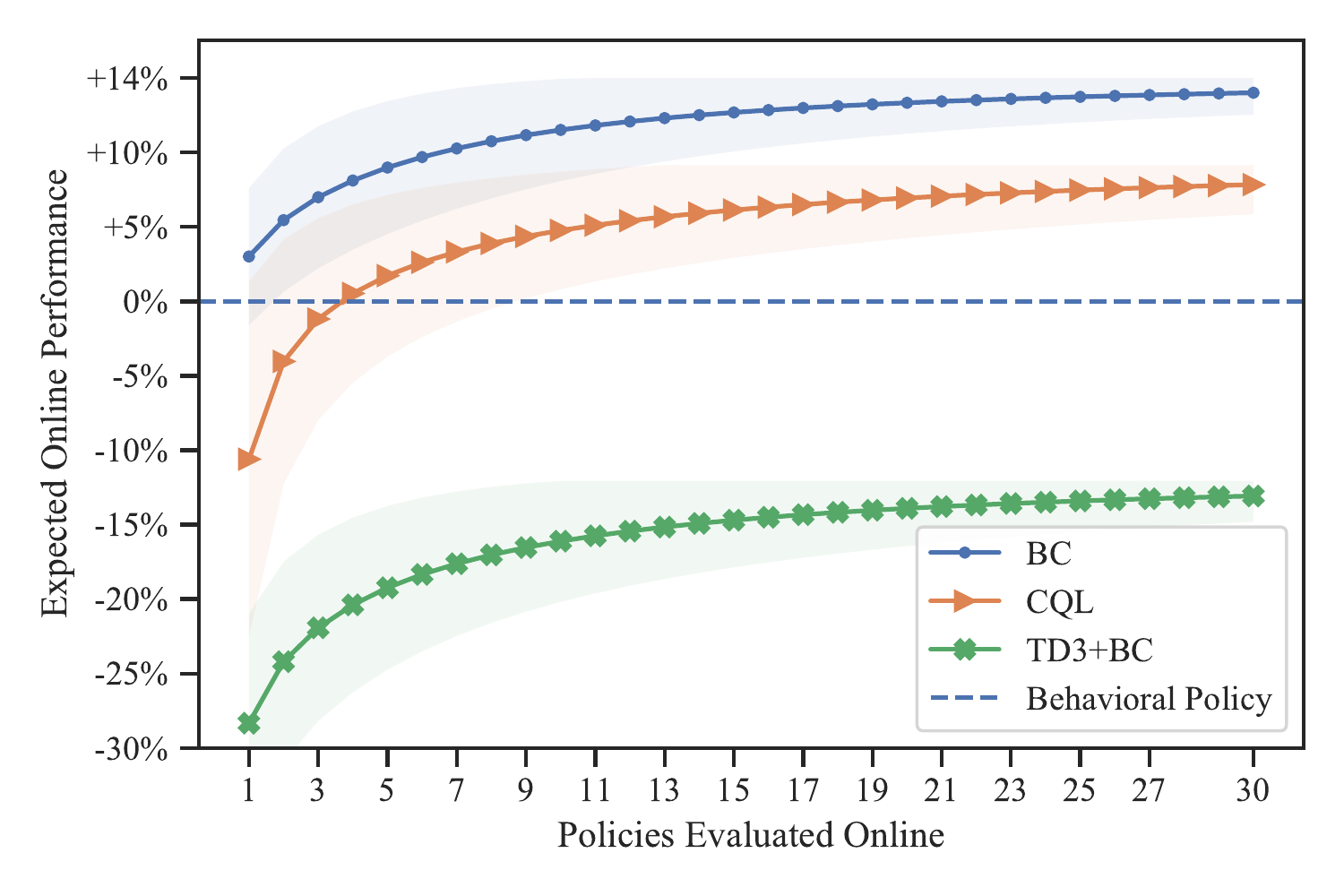}}
  \caption{Medium-Level Policy, 99 }
\end{subfigure}

\begin{subfigure}[b]{.32\textwidth}
 \centering
  \centerline{\includegraphics[width=\textwidth]{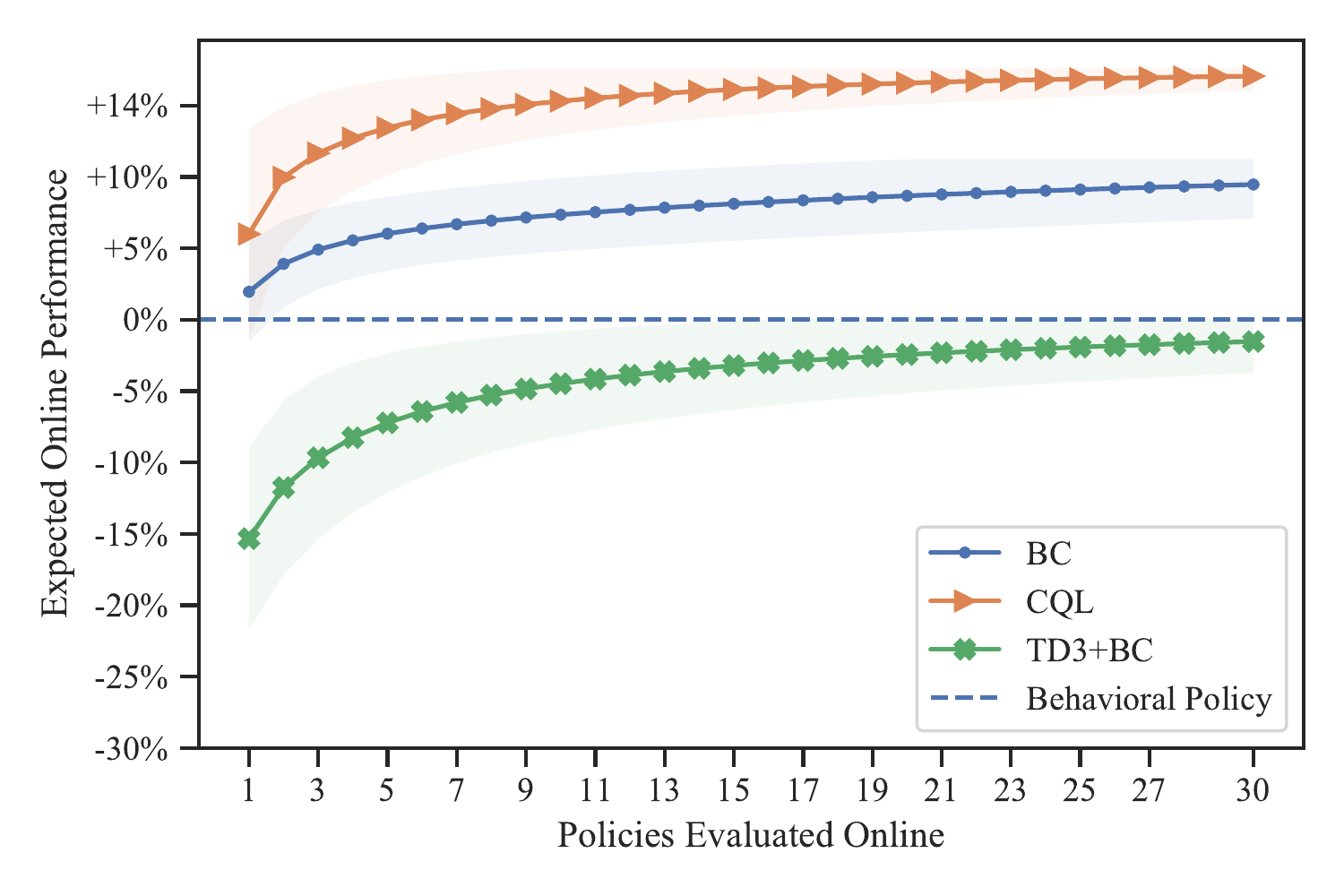}}
  \caption{Low-Level Policy, 9999 }
\end{subfigure}
\begin{subfigure}[b]{.32\textwidth}
 \centering
  \centerline{\includegraphics[width=\textwidth]{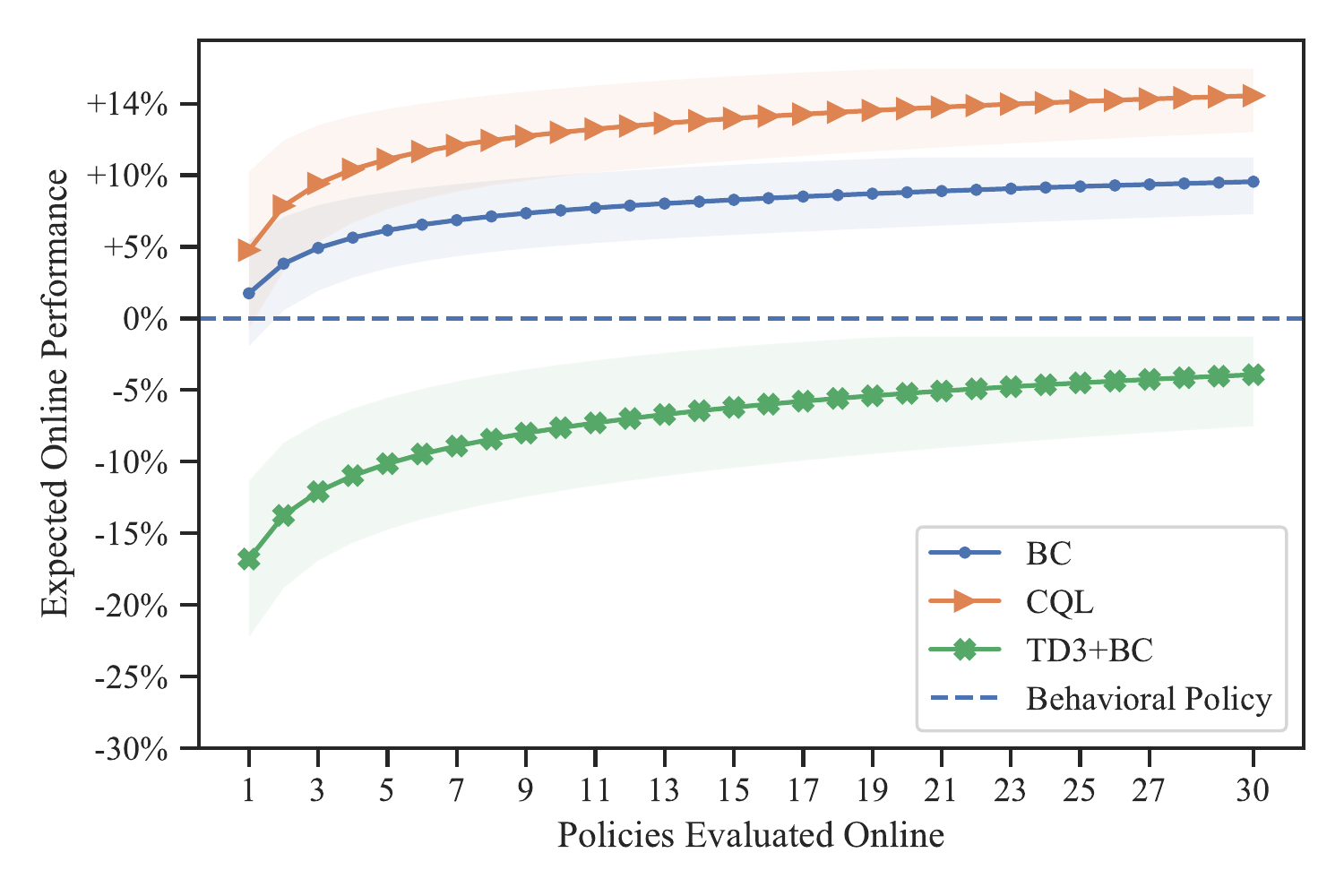}}
  \caption{Low-Level Policy, 999 }
\end{subfigure}
\begin{subfigure}[b]{.32\textwidth}
 \centering
  \centerline{\includegraphics[width=\textwidth]{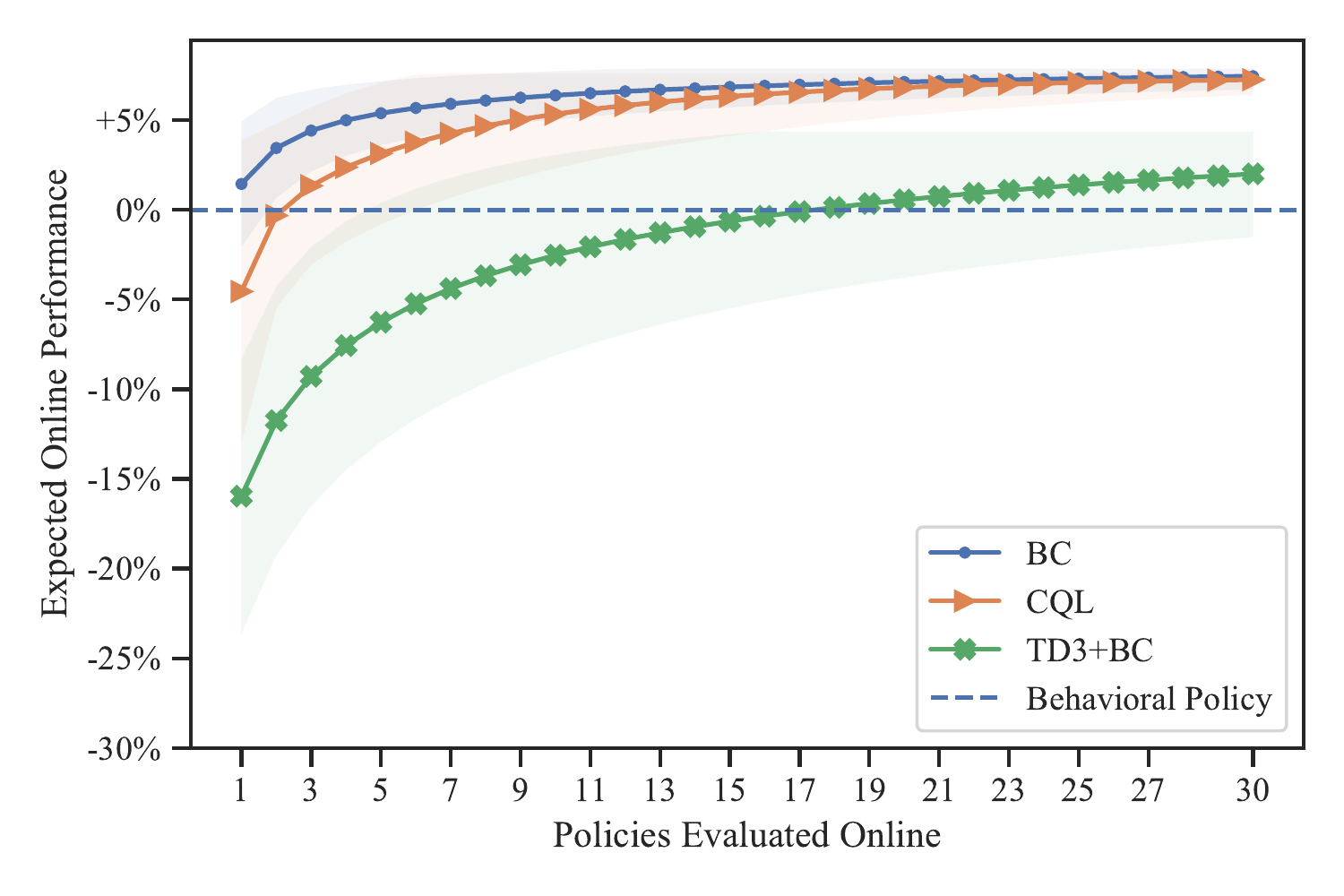}}
  \caption{Low-Level Policy, 99 }
\end{subfigure}

\caption{\textbf{Expected Online Performance under uniform policy selection. CityLearn}, one of our experiments, comparing CQL, TD3+BC, and BC under equal amounts of online evaluation budgets.}
\label{fig:appendix:cql_citylearn}
\end{figure*}

\begin{figure*}[h]
 \centering
\begin{subfigure}[b]{.32\textwidth}
 \centering
  \centerline{\includegraphics[width=\textwidth]{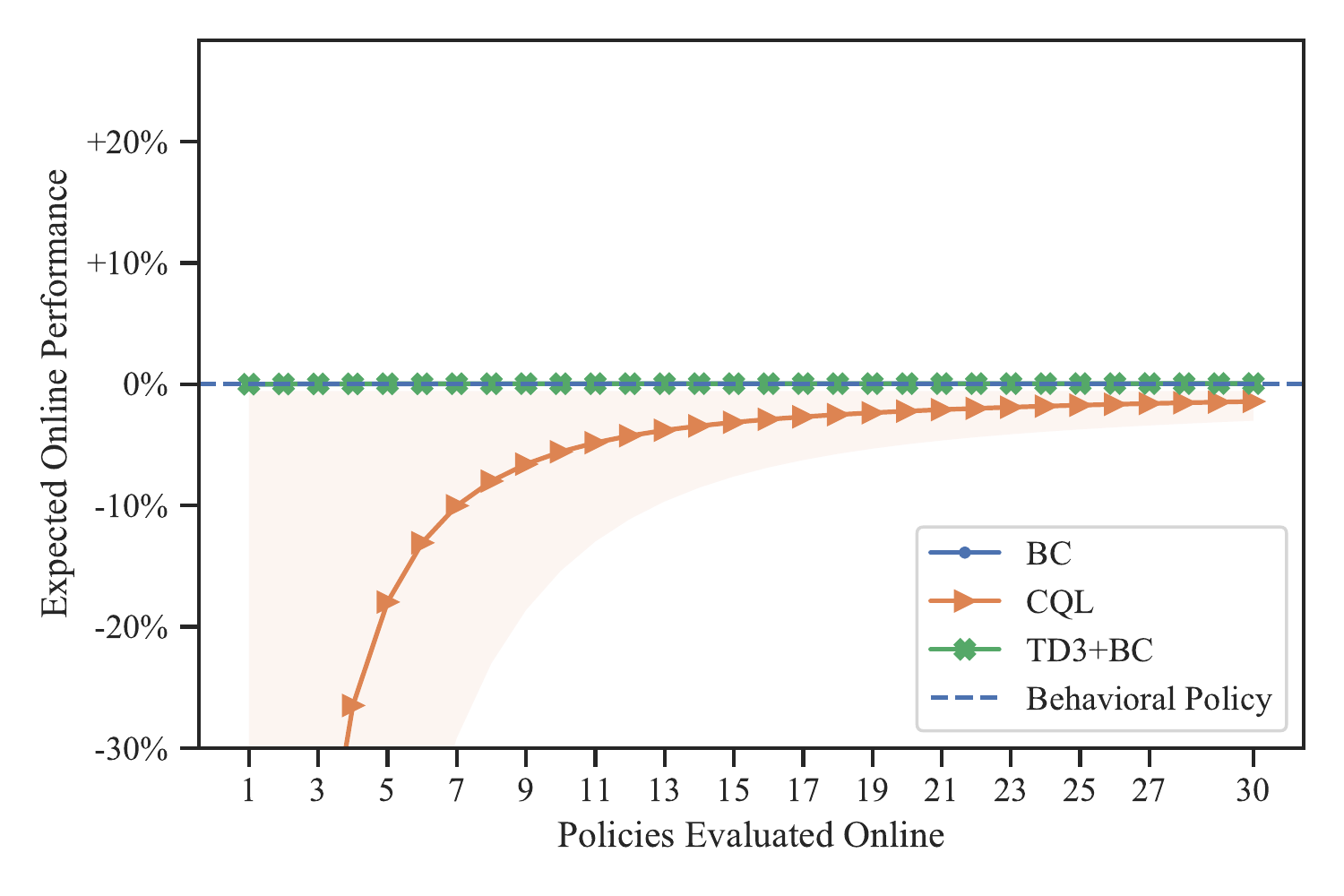}}
  \caption{High-Level Policy, 9999 }
\end{subfigure}
\begin{subfigure}[b]{.32\textwidth}
 \centering
  \centerline{\includegraphics[width=\textwidth]{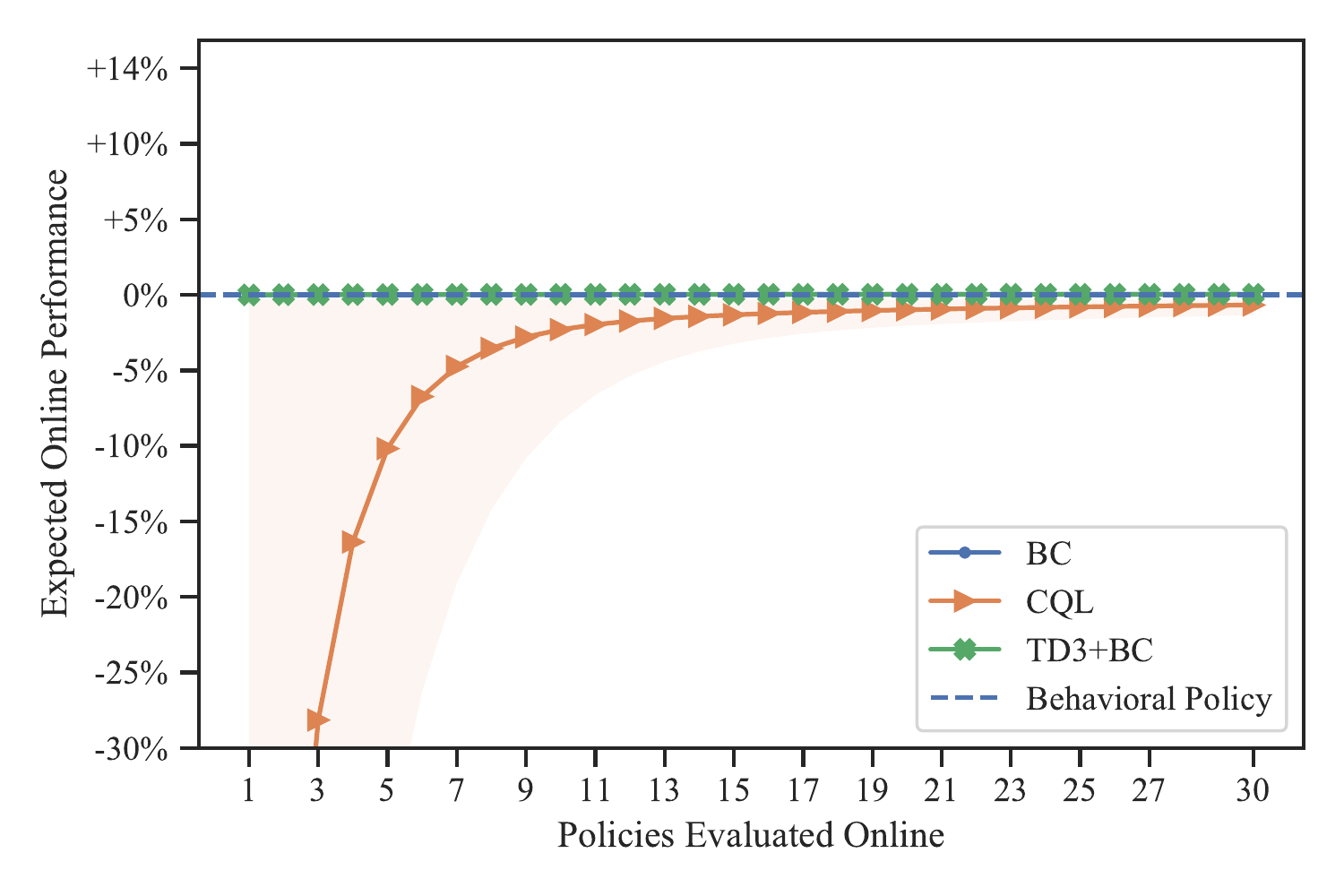}}
  \caption{High-Level Policy, 999 }
\end{subfigure}
\begin{subfigure}[b]{.32\textwidth}
 \centering
  \centerline{\includegraphics[width=\textwidth]{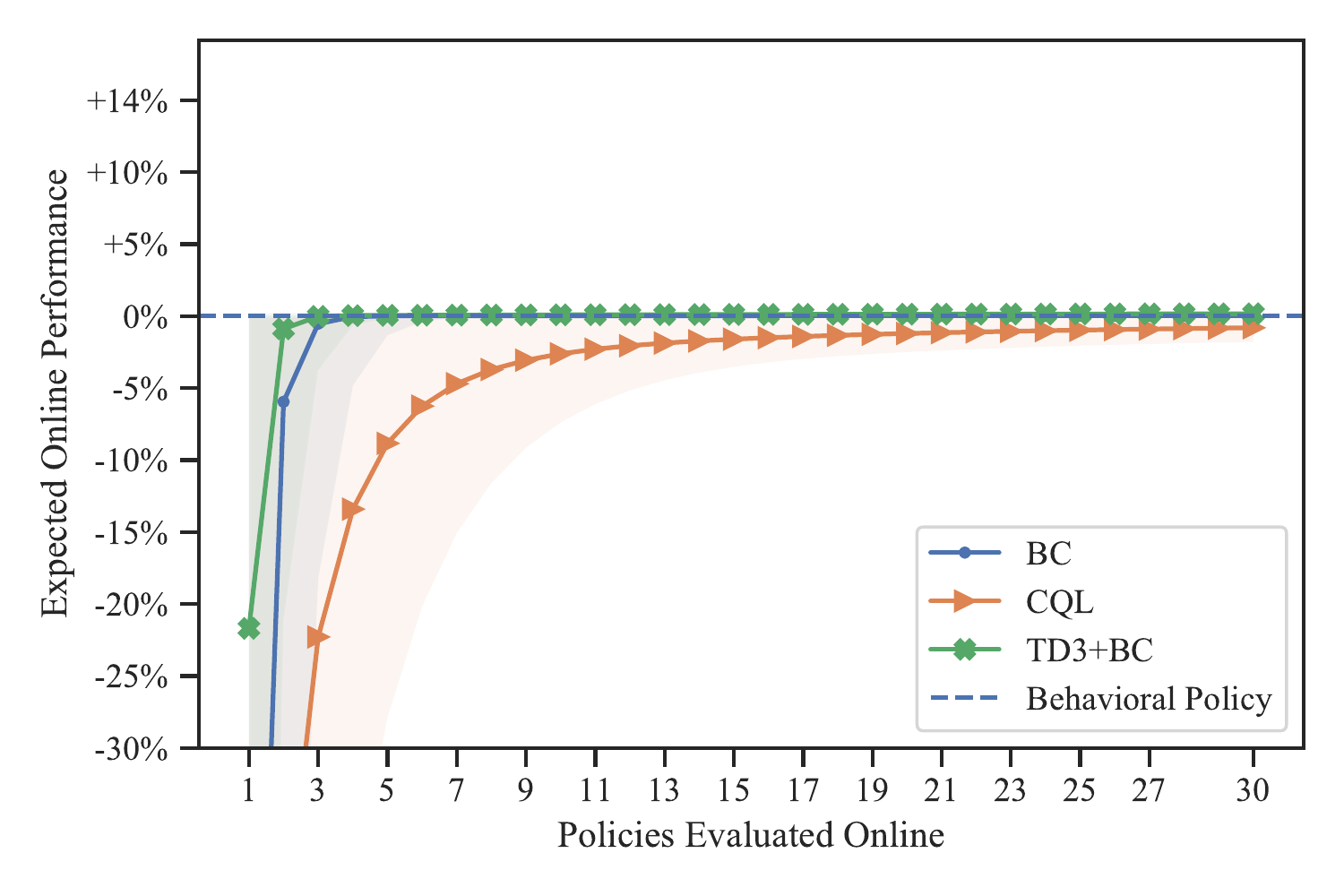}}
  \caption{High-Level Policy, 99 }
\end{subfigure}

\begin{subfigure}[b]{.32\textwidth}
 \centering
  \centerline{\includegraphics[width=\textwidth]{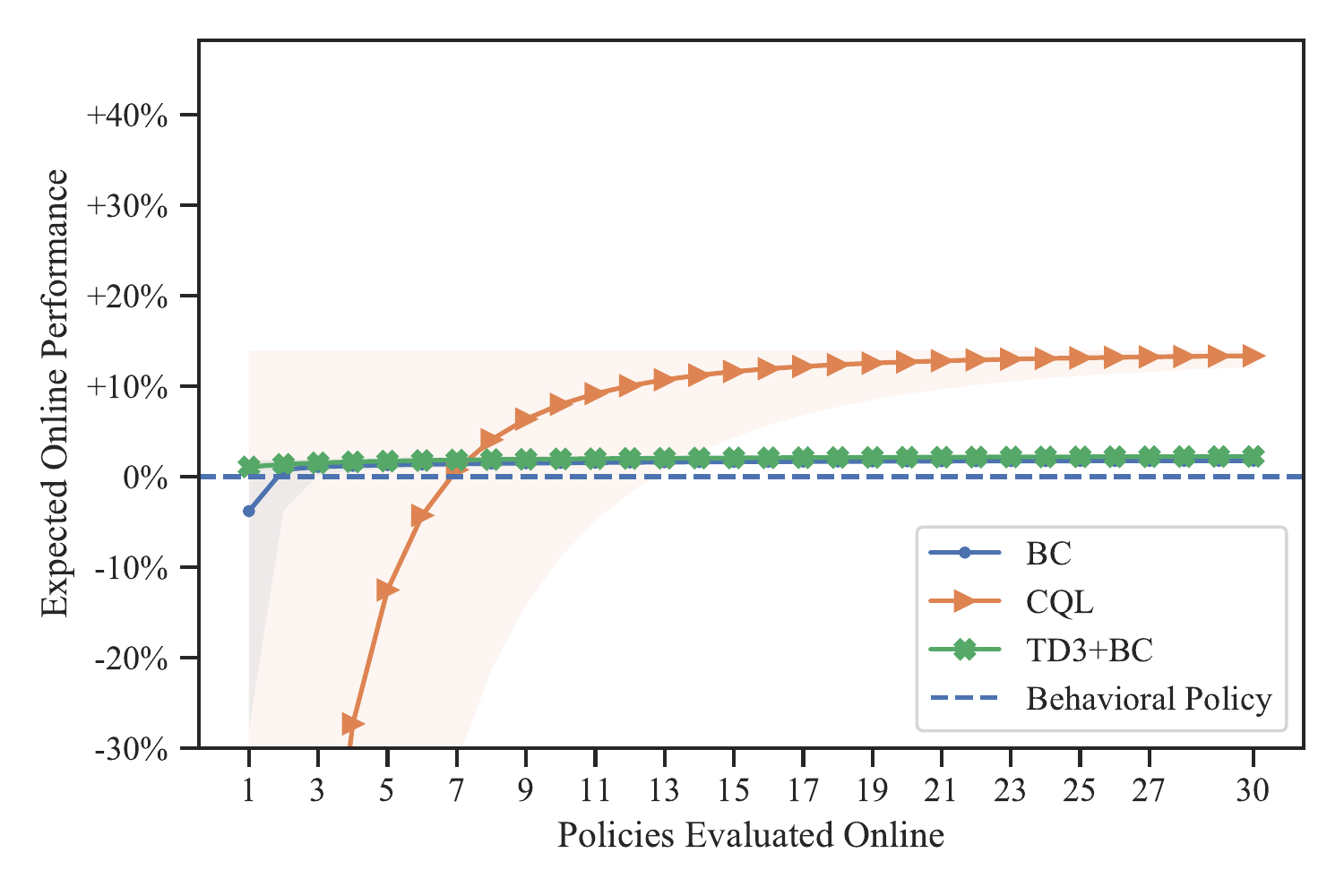}}
  \caption{Medium-Level Policy, 9999 }
\end{subfigure}
\begin{subfigure}[b]{.32\textwidth}
 \centering
  \centerline{\includegraphics[width=\textwidth]{figures/industrial_999_medium_uniform.pdf}}
  \caption{Medium-Level Policy, 999 }
\end{subfigure}
\begin{subfigure}[b]{.32\textwidth}
 \centering
  \centerline{\includegraphics[width=\textwidth]{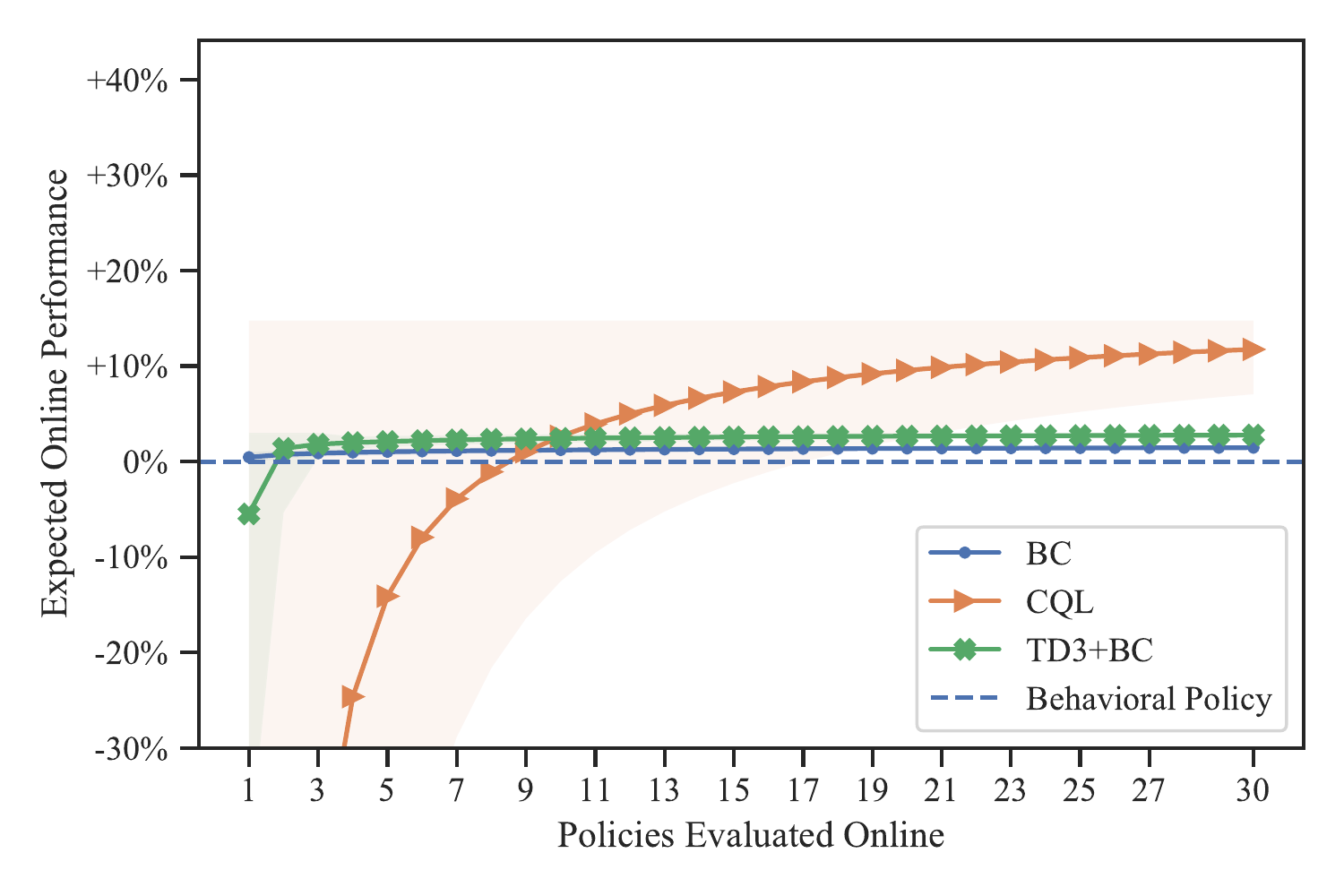}}
  \caption{Medium-Level Policy, 99 }
\end{subfigure}

\begin{subfigure}[b]{.32\textwidth}
 \centering
  \centerline{\includegraphics[width=\textwidth]{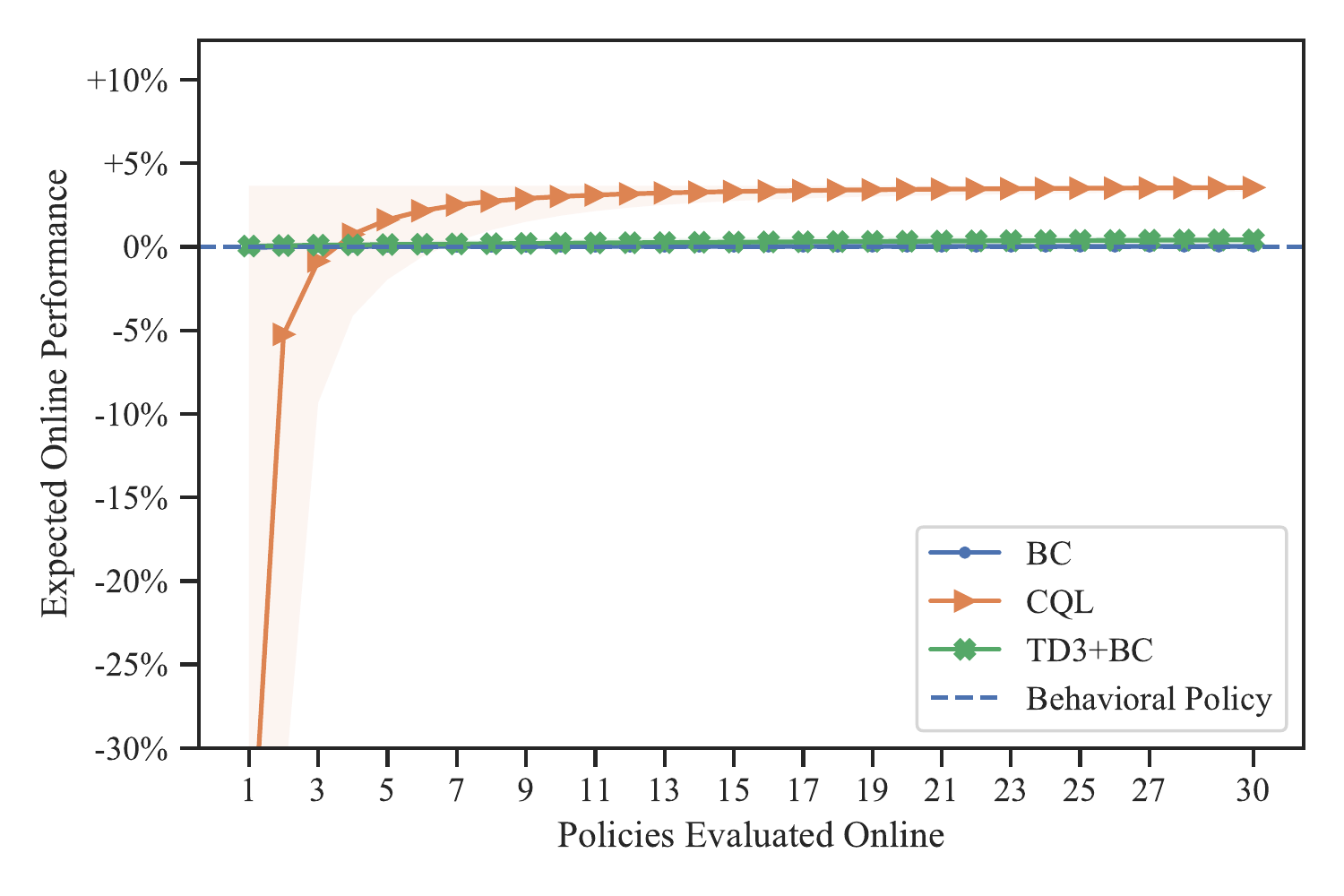}}
  \caption{Low-Level Policy, 9999 }
\end{subfigure}
\begin{subfigure}[b]{.32\textwidth}
 \centering
  \centerline{\includegraphics[width=\textwidth]{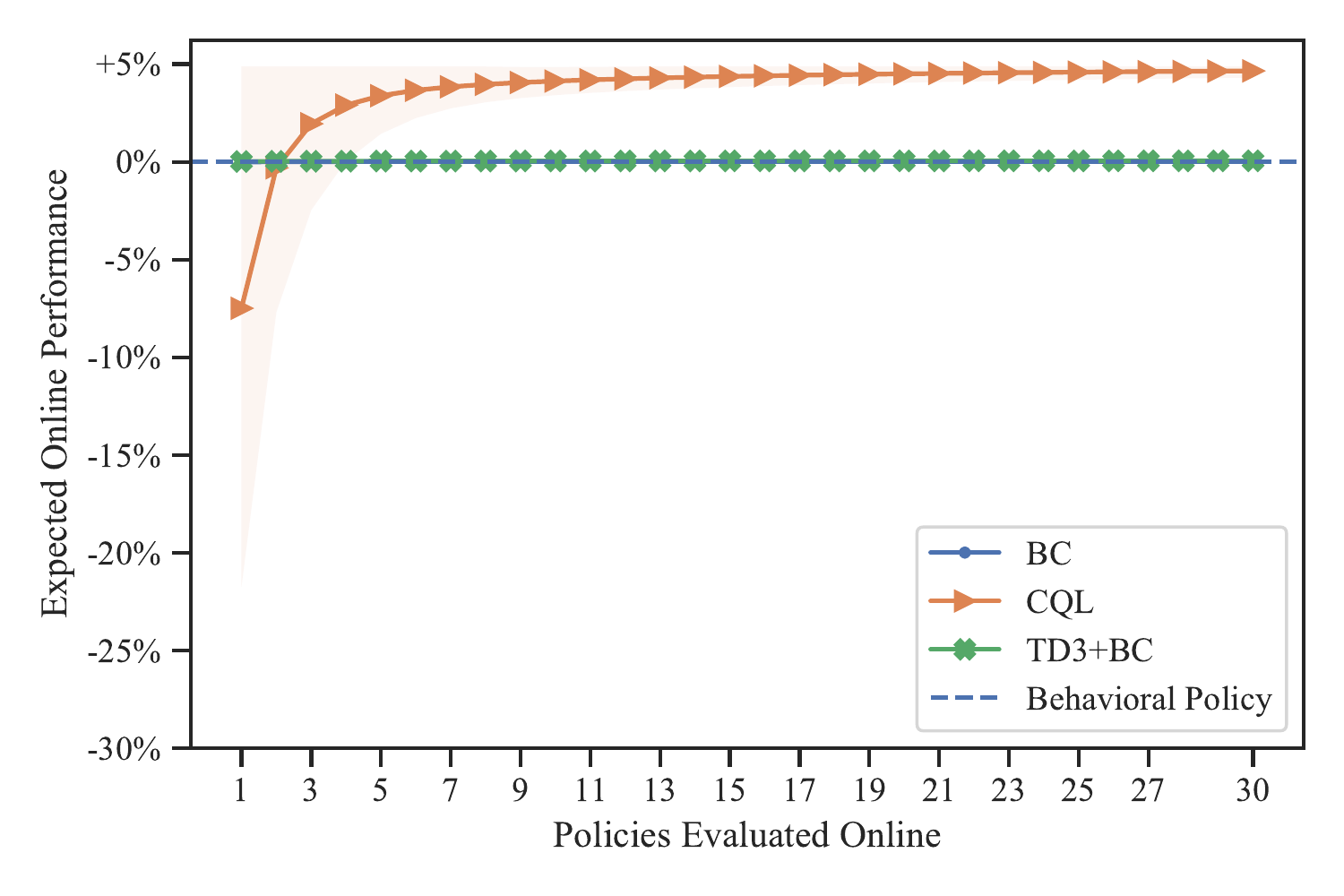}}
  \caption{Low-Level Policy, 999 }
\end{subfigure}
\begin{subfigure}[b]{.32\textwidth}
 \centering
  \centerline{\includegraphics[width=\textwidth]{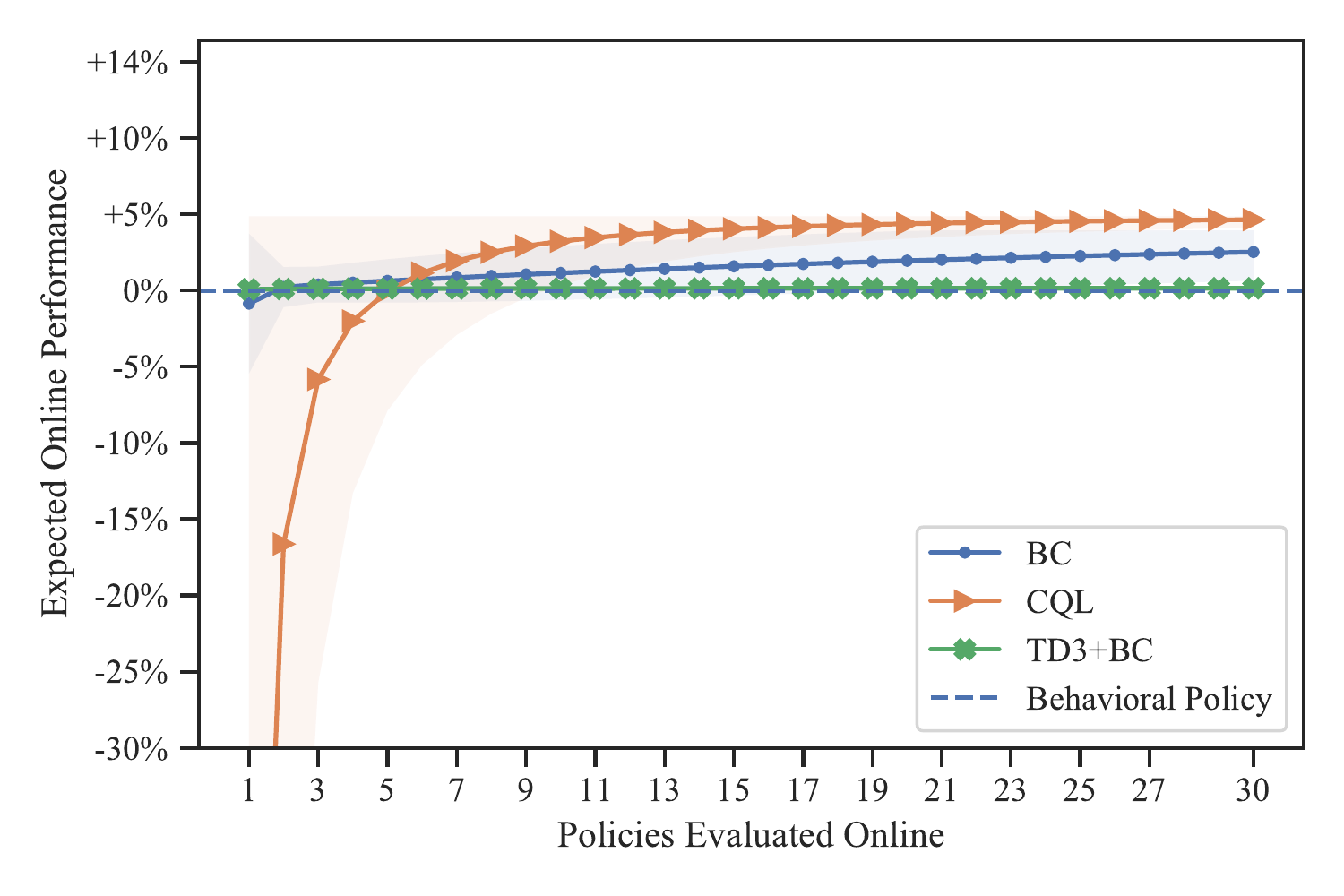}}
  \caption{Low-Level Policy, 99 }
\end{subfigure}

\caption{\textbf{Expected Online Performance under uniform policy selection. Industrial Benchmark}, one of our experiments, comparing CQL, TD3+BC, and BC under equal amounts of online evaluation budgets.}
\label{fig:appendix:cql_industrial}
\end{figure*}

In Sections \ref{sec:neorl_other_tasks} and \ref{sec:case_ops}, we run an additional set of experiments with deep ORL algorithms. Here, we describe them in more details. All the resulting EOP graphs can be found in Figures \ref{fig:appendix:cql_finrl}, \ref{fig:appendix:cql_citylearn}, \ref{fig:appendix:cql_industrial}.

\subsection{Multi-Level Policies and Dataset Sizes}

We replicate a procedure from \citet{qinNeoRLRealWorldBenchmark2021} for obtaining behavioral policies and datasets collection: we run Soft Actor-Critic \cite{haarnojaSoftActorCriticOffPolicy2018} until convergence and extract from 2 to 3 checkpointed policies with varying levels of expertise: low, medium, and high.

Afterwards, these policies are used to collect datasets of varying sizes. Similar to \citet{qinNeoRLRealWorldBenchmark2021}, we use the 99-999-9999 trajectories scheme. However, we do not inject any randomization while collecting the datasets, making the setting closer to more stringent real-life setups.

\subsection{On Hyperparameters}
\label{appendix:sec:hyperparams}

To define the space of hyperparameters, we mimicked a procedure common among deep learning practitioners. For a given algorithm, we scan through relevant papers and extract values specified there. However, instead of taking specific hyperparameter assignments, we randomize across the extracted values, as the optimal choice is largely problem-dependent. The exact hyperparameter space can be found in Table \ref{tab:hyperparams}.

\subsubsection{CQL}
We use hyperparameter values mentioned in \citet{kumarConservativeQLearningOffline2020} and \citet{ qinNeoRLRealWorldBenchmark2021}. Actor and critic networks are the same as in \citet{qinNeoRLRealWorldBenchmark2021}: two separate MLPs with 2 hidden layers and 256 units per layer. The implementation can be found in the attached source code.

\subsubsection{TD3+BC}
For this algorithm, we rely on hyperparameters specified in \citet{fujimotoMinimalistApproachOffline2021} and also extend it with learning rates from \citet{kumarConservativeQLearningOffline2020}. Actor and critic networks are the same as in \citet{fujimotoMinimalistApproachOffline2021}: two separate MLPs with 2 hidden layers and 256 units per layer. The implementation can be found in the attached source code.

\subsubsection{BC}
The search space for Behavioral Cloning is taken from \citet{mandlekarWhatMattersLearning2021} and \citet{qinNeoRLRealWorldBenchmark2021}. The network architecture is the same as in the algorithms above. Note that we do not use early stopping as in \citet{qinNeoRLRealWorldBenchmark2021}, but rather fix the number of gradient steps.

\subsection{Environments, Baselines, and Datasets}
\label{appendix:sec:envs_bases_datasets}

For a more precise picture of the high-dimensionality of the problems, we include the environment configurations table from \citet{qinNeoRLRealWorldBenchmark2021} below.

\begin{table}[!h]
	\centering
	\caption{Environment configurations. Taken from \citet{qinNeoRLRealWorldBenchmark2021}.}
	\vskip 0.15in
	\scalebox{0.99}{\begin{tabular}{ccccc}
			\hline
			Environment & \begin{tabular}[c]{@{}c@{}}Observation \\ Shape\end{tabular} & \begin{tabular}[c]{@{}c@{}}Action \\ Shape\end{tabular} & \begin{tabular}[c]{@{}c@{}}Have \\ Done\end{tabular} & \begin{tabular}[c]{@{}c@{}}Max \\ Timesteps\end{tabular}\\
			\hline
			IB & 182 & 3 & False & 1000\\
			FinRL & 181 & 30 & False & 2516\\
			CL & 74 & 14 & False & 1000\\
			\hline
	\end{tabular}}
	\label{tab:env_config}
\end{table}

As discussed in the main text, we reproduced the procedure from \citet{qinNeoRLRealWorldBenchmark2021} to train behavioral policies and generate datasets. We omitted the CityLearn high-level policy since we could not match the performance reported in \citet{qinNeoRLRealWorldBenchmark2021} in a reasonable amount of time. Online performance for the policies used in the paper is in Table \ref{tab:behavioral_performance}.

\begin{table}[!h]
	\centering
	\caption{Behavioral policies.}
	\vskip 0.15in
	\scalebox{0.99}{\begin{tabular}{ccccc}
			\hline
			Environment & \begin{tabular}[c]{@{}c@{}}Low \\ Level\end{tabular} & \begin{tabular}[c]{@{}c@{}}Medium \\ Level\end{tabular} & \begin{tabular}[c]{@{}c@{}}High \\ Level\end{tabular}\\
			\hline
			Industrial Benchmark & -323856 & -276379 & -234101 \\
			CityLearn & 27797 & 32498 & -\\
			FinRL & 221 & 294 & 446\\
			\hline
	\end{tabular}}
	\label{tab:behavioral_performance}
\end{table}

\subsection{On Offline Policy Selection methods}

After sampling $N$ hyperparameters, running an offline RL algorithm, and obtaining $N$ policies, $B$ of them are selected for online evaluation $\pi_{1}, ..., \pi_{B}$. Generally, the probability of being selected for each policy can be defined as $p(\pi_{i} | h_{i}, D_{v}, S)$, where $h_{i}$ is a tuple of hyperparameters used for training policy $\pi_{i}$, $D_{v}$ is a validation dataset, and $S$ is a offline policy selection strategy.

There is a plethora of approaches to offline policy evaluation and selection \cite{irpanOffPolicyEvaluationOffPolicy2019, fuBenchmarksDeepOffPolicy2021, voloshinEmpiricalStudyOffPolicy2020}. In this work, we utilize a few selection methods already utilized in the robotics domain: those based on value functions \cite{paineHyperparameterSelectionOffline2020} and action distributions \cite{mandlekarWhatMattersLearning2021}.
    
\subsubsection{Uniform}
Uniform Selection is an extremely simple baseline that corresponds to picking policies completely at random. It ignores all the information regarding the validation dataset. The probability of a policy $\pi_{i}$ being selected for online evaluation is simply proportional to the probability of its hyperparameters $p(h_{i})$.

\begin{equation} \label{eq:uniform_selection}
    p(\pi_{i} | h_{i}, D_{v}, S = uniform) \propto p(h_{i})
\end{equation}

As we demonstrated in the main text, this naive strategy serves as a competitive baseline to other selection methods, has good computational properties, and is strongly linked with a comparison tool utilized in natural language processing problems \cite{dodgeShowYourWork2019}.

\subsubsection{Expected Initial State Value}
These selection methods estimate an expected value of the target policy $\pi_{i}$ for the initial state distribution $E_{s_{0} \sim D_{V}}[Q(s_{0}, \pi_{i}(s_{0}))]$ and re-rank the policies accordingly. The $Q$ function can be reused from a training process (e.g. a critic from Conservative Q-Learning) -- \textit{$V(s_{0})$ using Critic}, or be refitted using Fitted-Q Evaluation \cite{leBatchPolicyLearning2019} -- we refer to it as \textit{$V(s_{0})$ using FQE}. These methods were found to be superior to others in simulated robotic tasks \cite{paineHyperparameterSelectionOffline2020}.

\subsubsection{Temporal Difference Error}

This approach uses an average of the temporal difference across all transition tuples $(s, a, r, s^{'})$ in the validation dataset $E_{(s, a, r, s^{'}) \sim D_{V}}[r + \gamma Q(s^{'}, \pi_{i}(s^{'})) - Q(s, a)]$. The lower the error for a policy, the higher its rank. Note that this method is rather indicative of the critics' quality and was shown to perform poorly \cite{paineHyperparameterSelectionOffline2020}. However, we still include it to obtain more evidence in novel environments (as far as we know, this approach was not tested for our set of environments) and to probe its ability to extract policies that improve over the baseline rather than try to find the best one.

\subsubsection{Action Difference}
This method measures how good a trained policy $\pi_{i}$ matches the behavioral policy $\pi_{\mu}$ by estimating an average expected difference in action $E_{(s, a) \sim D_{V}}[(\pi_{i}(s) - a)^2]$. The lower the discrepancy, the higher the trained policy is ranked. This method is expected to result in finding policies close to the behavioral, with no sudden drop in performance at the expense of lower expected improvement.

\subsection{Computational Resources}
The experiments were run on a computational cluster with 14x NVIDIA Tesla V100, 256GB RAM, and Intel(R) Xeon(R) Gold 6154 CPU @ 3.00GHz (72 cores) for 13 days.

\section{Why Spearman's Rank Correlation Can Be Misleading for Offline Policy Selection Methods}
In Section \ref{sec:case_ops}, we demonstrated how the proposed approach can be used for comparing OPS methods using Regret@K metric, highlighting that the preference between these methods is budget-dependent.

However, it may be pointed out that this is not really needed, as Spearman's rank correlation is usually representative of the ranking quality for such methods. Here, we show how it can fail by giving two examples in Table \ref{fig:appendix:spearman-fails}. The second ranking is what we usually want in offline RL setting -- getting the best policy with the least amount of online executions as possible. However, its rank correlation is significantly lower than for the first ranking.
\begin{table}[!ht]
\caption{An example where Spearman's rank correlation coefficient is not representative of the desiderata in deep ORL setting.}
\vskip 0.15in
\begin{center}
\begin{small}
\begin{sc}
\begin{tabular}{lcccr}
\toprule
True Ranking & Ranking\#1 & Ranking\#2 \\
\midrule
1& 5& 1\\
2& 4& 2\\
3& 3& 10\\
4& 2& 9\\
5& 1& 8\\
6& 6& 7\\
7& 7& 6\\
8& 8& 5\\
9& 9& 4\\
10& 10& 3\\
\midrule
Spearman's Rho& 0.76& -0.02\\
\bottomrule
\end{tabular}
\end{sc}
\end{small}
\end{center}
\label{fig:appendix:spearman-fails}
\end{table}

\begin{table*}[!h]
	\centering
	\caption{Hyperparameters search space.}
	\vskip 0.15in
	\scalebox{0.99}{\begin{tabular}{cc}
		\toprule
		\textbf{Algorithms} & \textbf{Search Space}\\
		\midrule
		\textbf{CQL} & \begin{tabular}[c]{@{}c@{}}
		variant $\in$ $\{\mathcal{H}, \rho\}$\\
		$\alpha \in \{5, 10\}$\\
		$\tau \in \{-1, 2, 5, 10\}$\\
		approximate-max backup $\in$ \{True, False\}\\
		actor alpha tuning $\in$ \{True, False\}\\
		actor learning rate $\in$ $\{3e-5, 3e-4, 1e-4, 1e-3\}$\\
		critic learning rate $\in$ $\{3e-4, 1e-4, 1e-3\}$\\ 
		tau $\in$ $\{5e-3, 1e-2\}$\\ 
		batch size $\in$ $\{256, 512\}$\\
		$\gamma \in [0.9, 1.0]$\\
		number of gradient steps $\in$ \{$3e5$\}
		    \end{tabular}
		\\
		\midrule
		\textbf{TD3+BC} & \begin{tabular}[c]{@{}c@{}}
		$\alpha \in \{1.0, 2.0, 2.5, 3.0, 4.0\}$\\
		actor learning rate $\in$ $\{3e-5, 3e-4, 1e-4, 1e-3\}$\\
		critic learning rate $\in$ $\{3e-4, 1e-4, 1e-3\}$\\ 
		tau $\in$ $\{5e-3, 1e-2\}$\\ 
		batch size $\in$ $\{256, 512\}$\\
		$\gamma \in [0.9, 1.0]$\\
		number of gradient steps $\in$ \{$3e5$\}
		    \end{tabular}
		\\
		\midrule
		\textbf{BC} & \begin{tabular}[c]{@{}c@{}}
		num modes in gaussian mixture model $\in \{1, 5, 10, 100\}$\\
		learning rate $\in$ $\{1e-3, 3e-4, 1e-4\}$\\ 
		batch size $\in$ $\{256, 512\}$\\
		number of gradient steps $\in$ \{$2e5$\}
	        \end{tabular}
		\\
		\bottomrule
	\end{tabular}}
	\label{tab:hyperparams}
\end{table*}

Spearman's Rho may punish what we typically care about more (the left side of the ranking), and appraise what we care about less (the right side of the ranking). Meanwhile, usage of Regret@K graphs helps to avoid this problem, capturing the left side.

\begin{figure*}[!h]
 \centering
\begin{subfigure}[b]{.32\textwidth}
 \centering
  \centerline{\includegraphics[width=\textwidth]{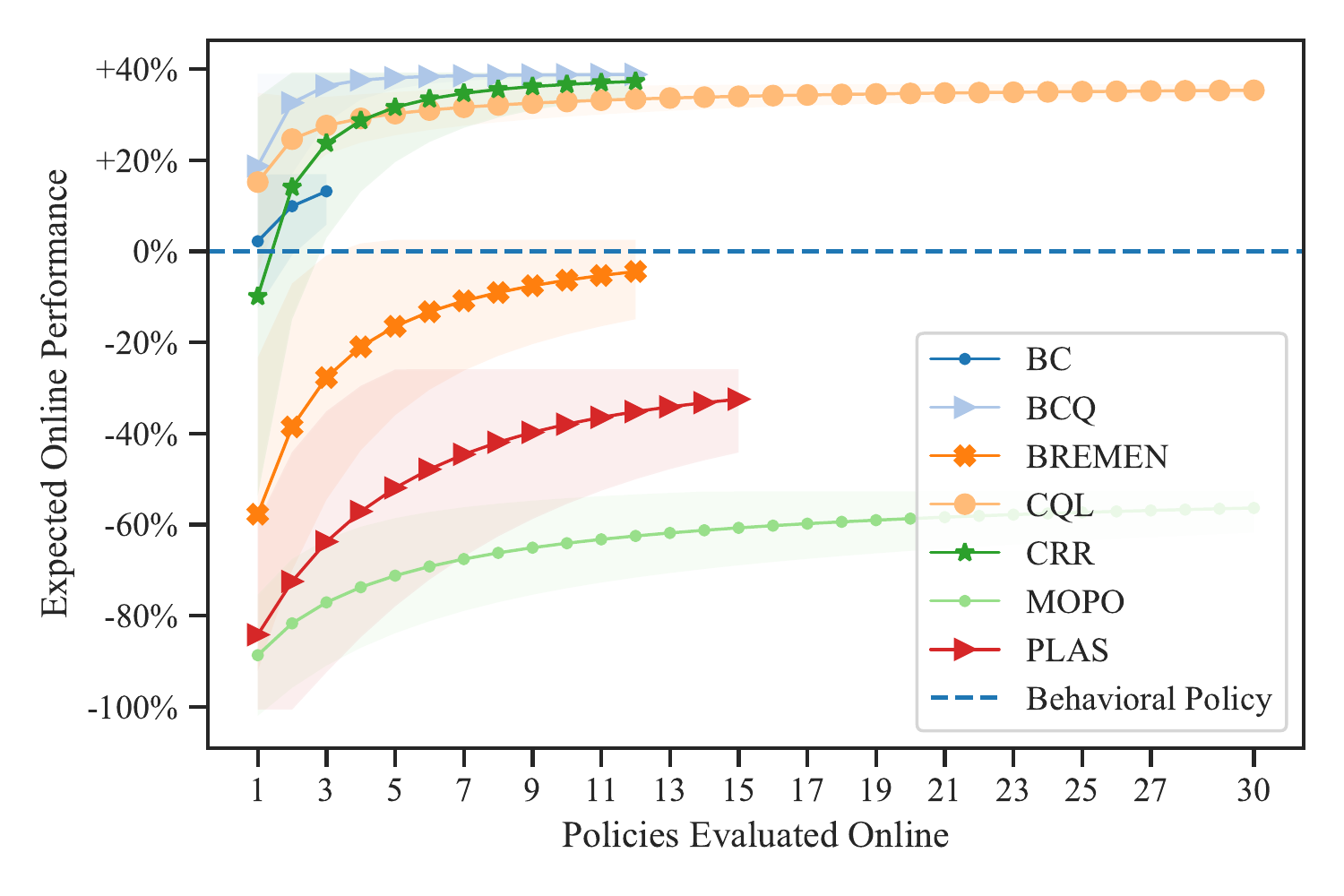}}
  \caption{High-Level Policy, 10000}
\end{subfigure}
\begin{subfigure}[b]{.32\textwidth}
 \centering
  \centerline{\includegraphics[width=\textwidth]{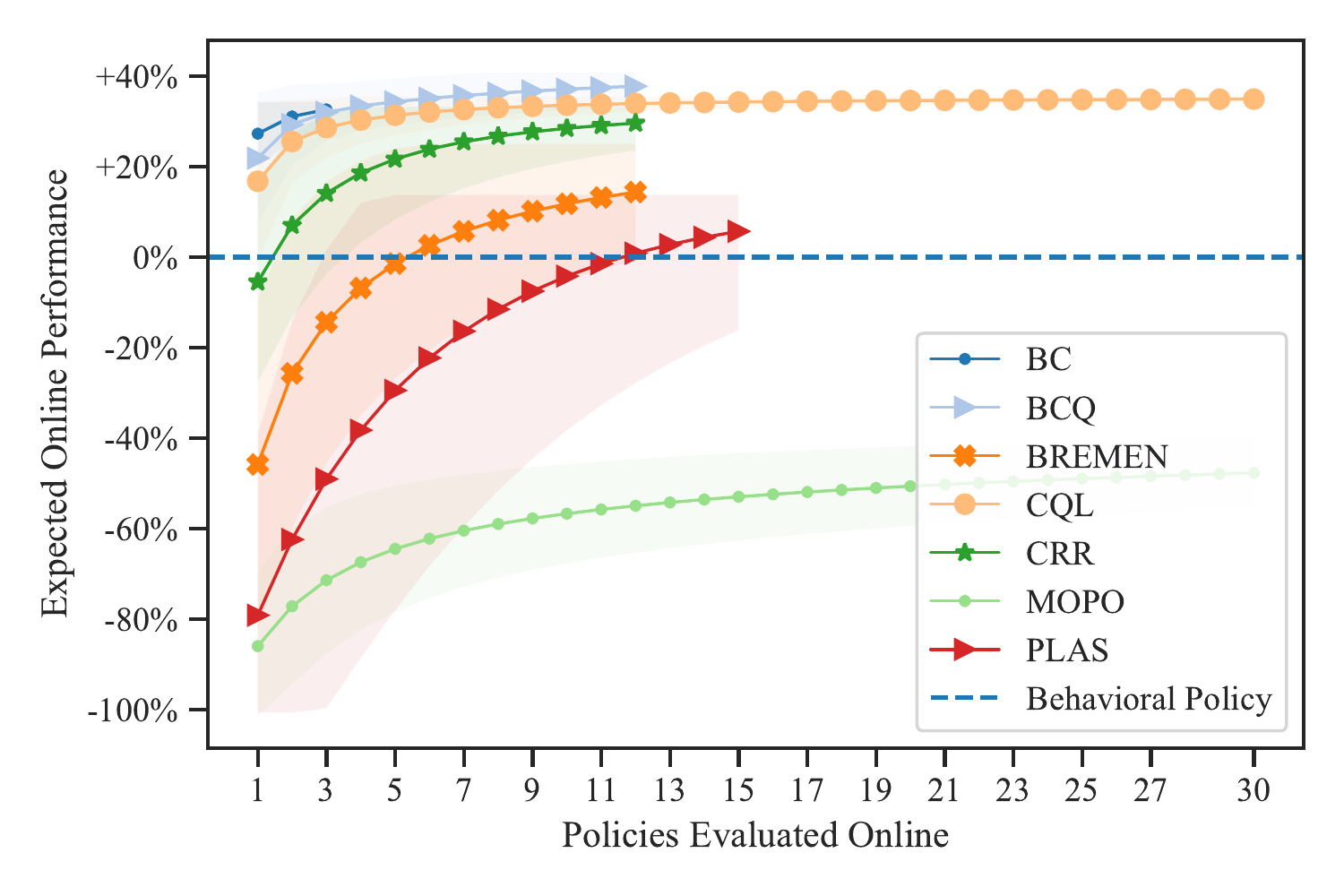}}
  \caption{High-Level Policy, 1000}
\end{subfigure}
\begin{subfigure}[b]{.32\textwidth}
 \centering
  \centerline{\includegraphics[width=\textwidth]{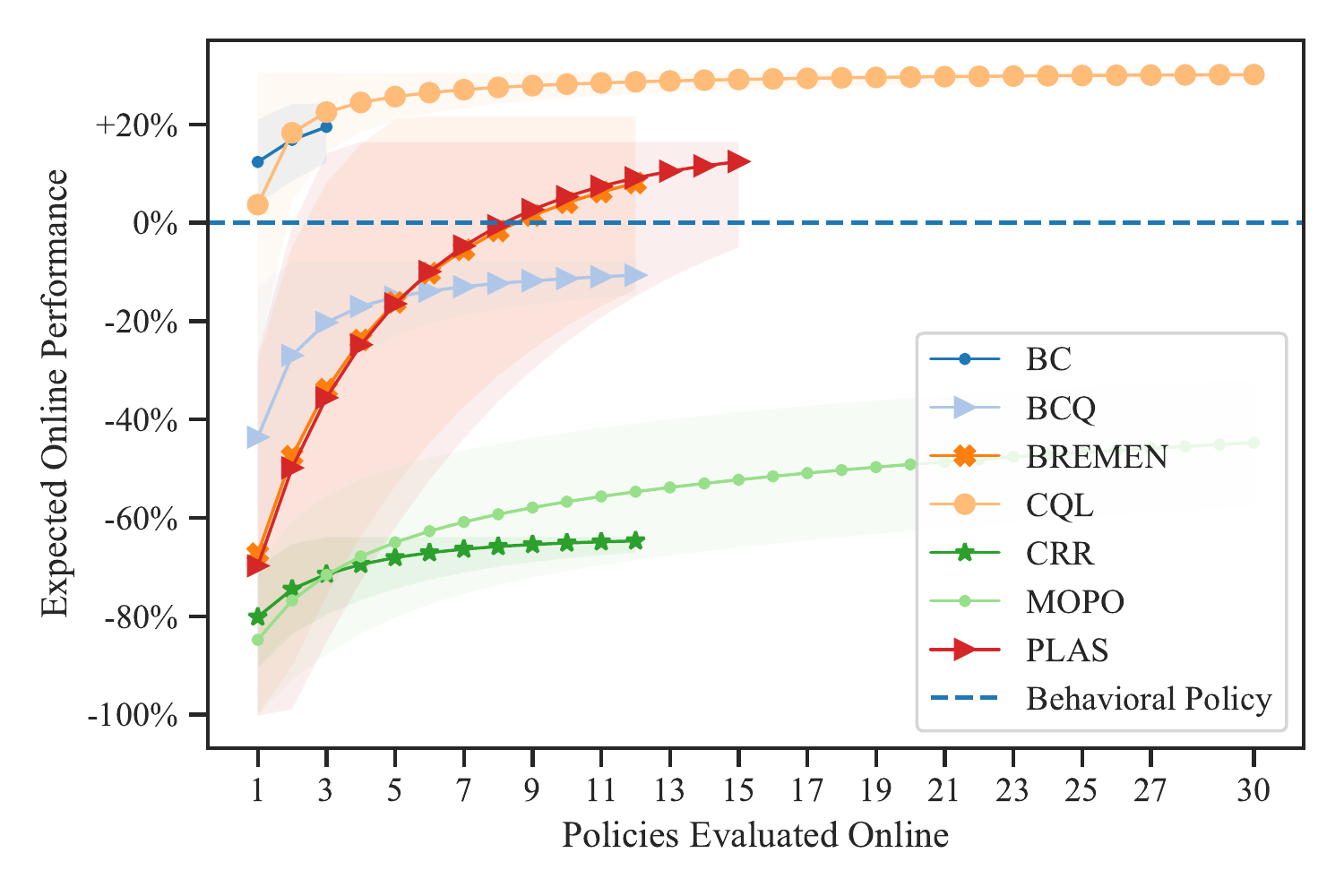}}
  \caption{High-Level Policy, 100}
\end{subfigure}

\begin{subfigure}[b]{.32\textwidth}
 \centering
  \centerline{\includegraphics[width=\textwidth]{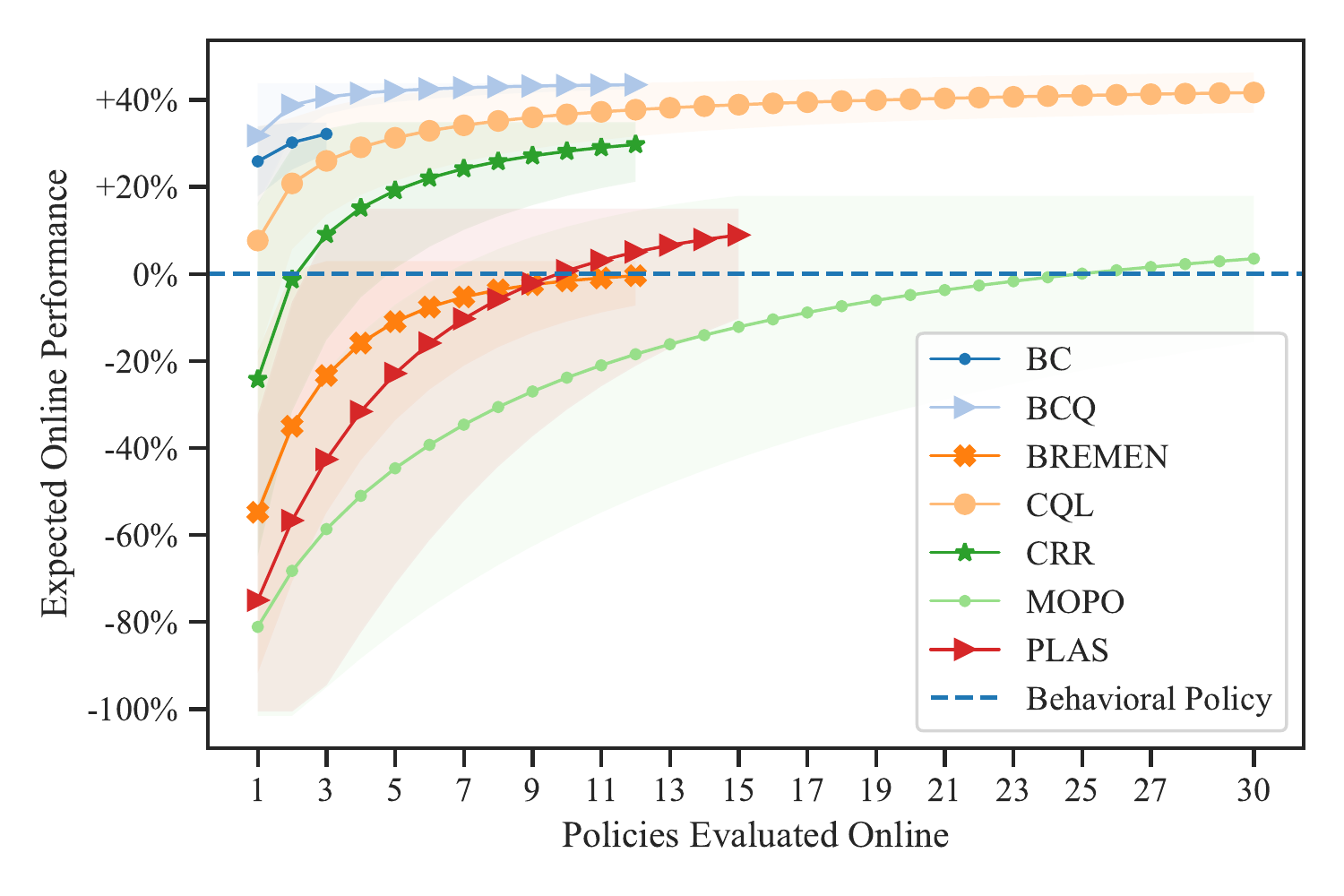}}
  \caption{Medium-Level Policy, 10000}
\end{subfigure}
\begin{subfigure}[b]{.32\textwidth}
 \centering
  \centerline{\includegraphics[width=\textwidth]{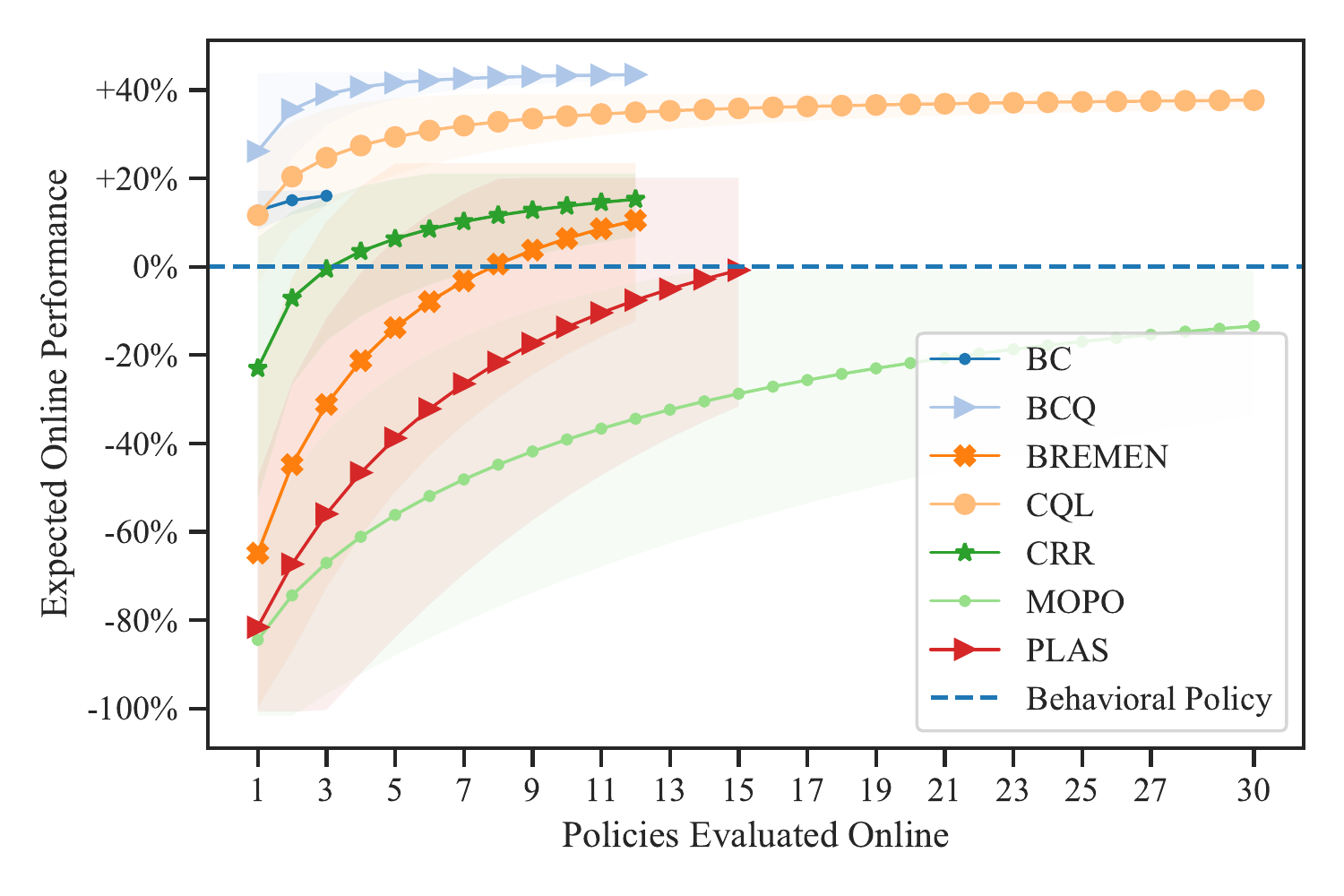}}
  \caption{Medium-Level Policy, 1000}
\end{subfigure}
\begin{subfigure}[b]{.32\textwidth}
 \centering
  \centerline{\includegraphics[width=\textwidth]{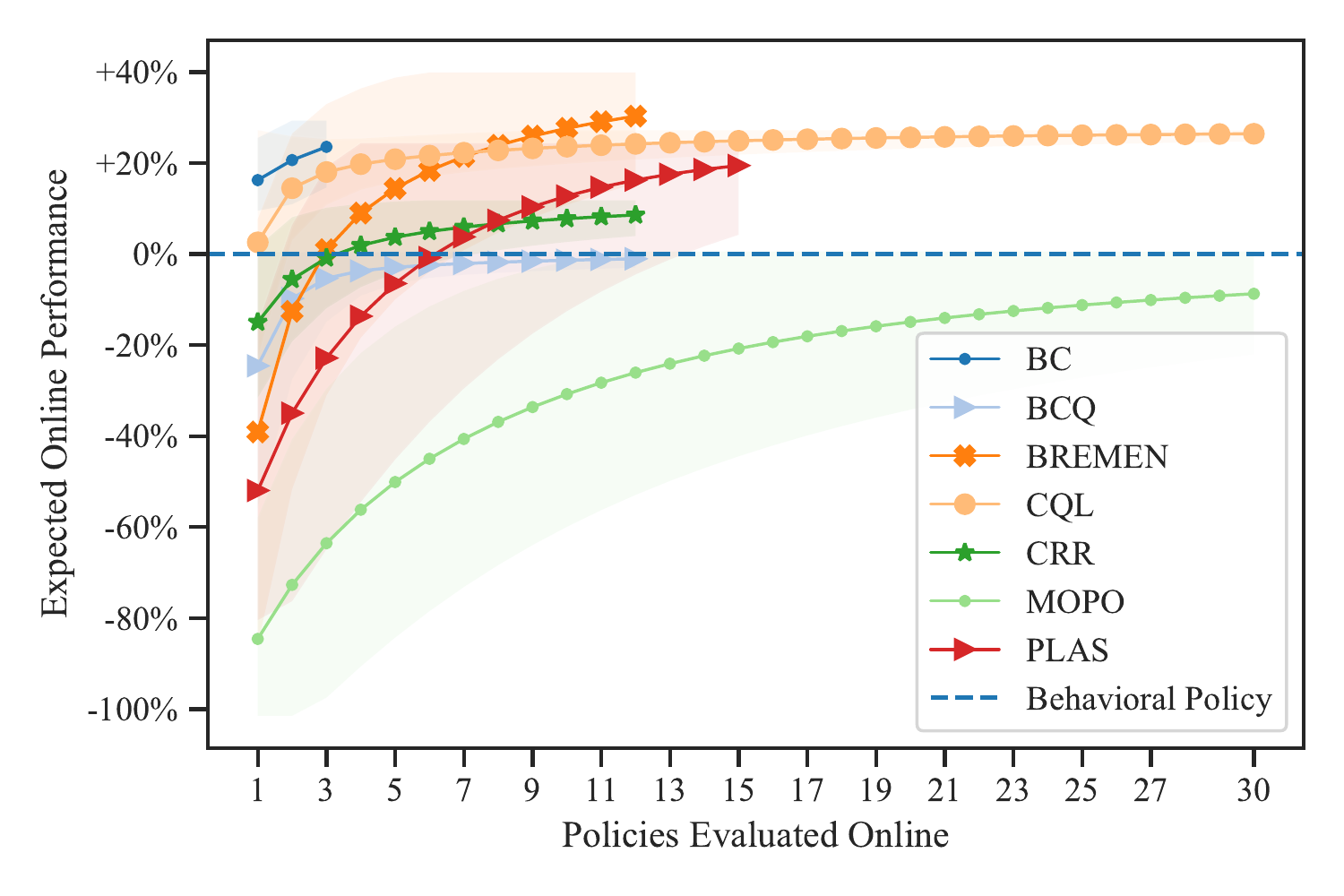}}
  \caption{Medium-Level Policy, 100 }
\end{subfigure}

\begin{subfigure}[b]{.32\textwidth}
 \centering
  \centerline{\includegraphics[width=\textwidth]{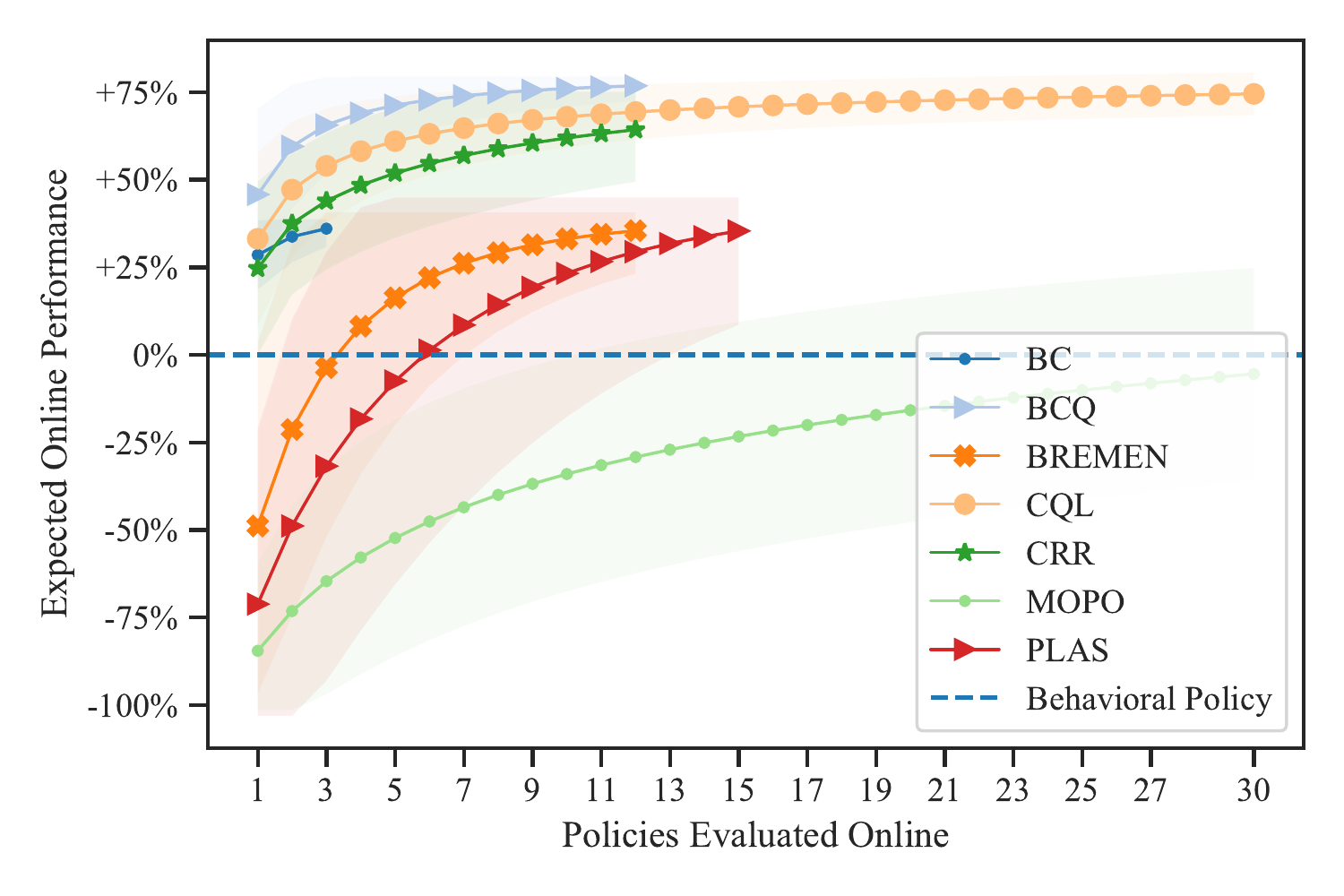}}
  \caption{Low-Level Policy, 10000 }
\end{subfigure}
\begin{subfigure}[b]{.32\textwidth}
 \centering
  \centerline{\includegraphics[width=\textwidth]{figures_appendix/Walker2d-v3_1000_low_uniform.pdf}}
  \caption{Low-Level Policy, 1000 }
\end{subfigure}
\begin{subfigure}[b]{.32\textwidth}
 \centering
  \centerline{\includegraphics[width=\textwidth]{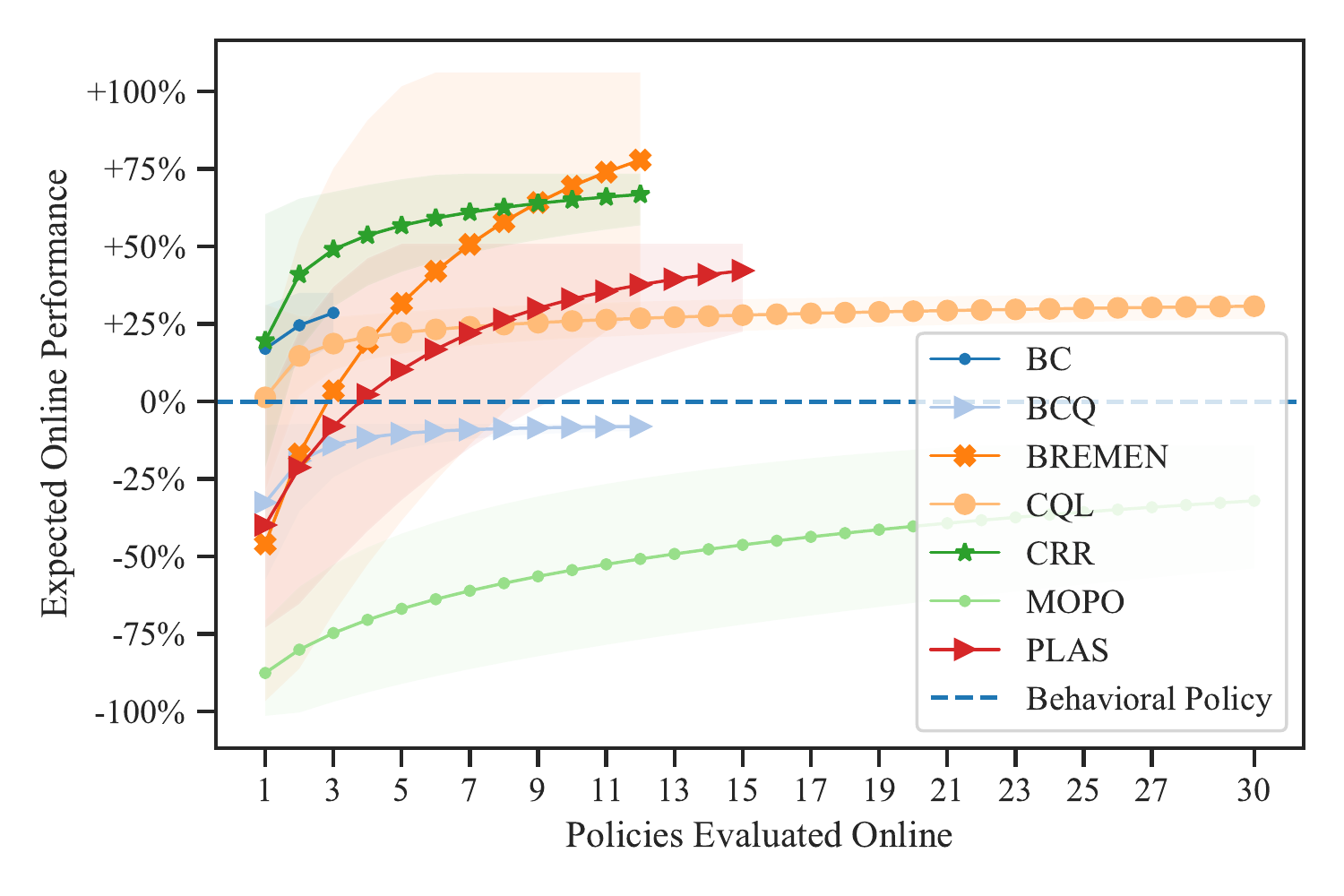}}
  \caption{Low-Level Policy, 100 }
\end{subfigure}
\caption{\textbf{Expected Online Performance under uniform policy selection. Walker2d-v3}, a robotic task from \citet{qinNeoRLRealWorldBenchmark2021}, the graphs are computed using their open-sourced online evaluations for different hyperparameter assignments.}
\label{fig:appendix:walker-2d}
\end{figure*}

\begin{figure*}[!h]
 \centering
\begin{subfigure}[b]{.32\textwidth}
 \centering
  \centerline{\includegraphics[width=\textwidth]{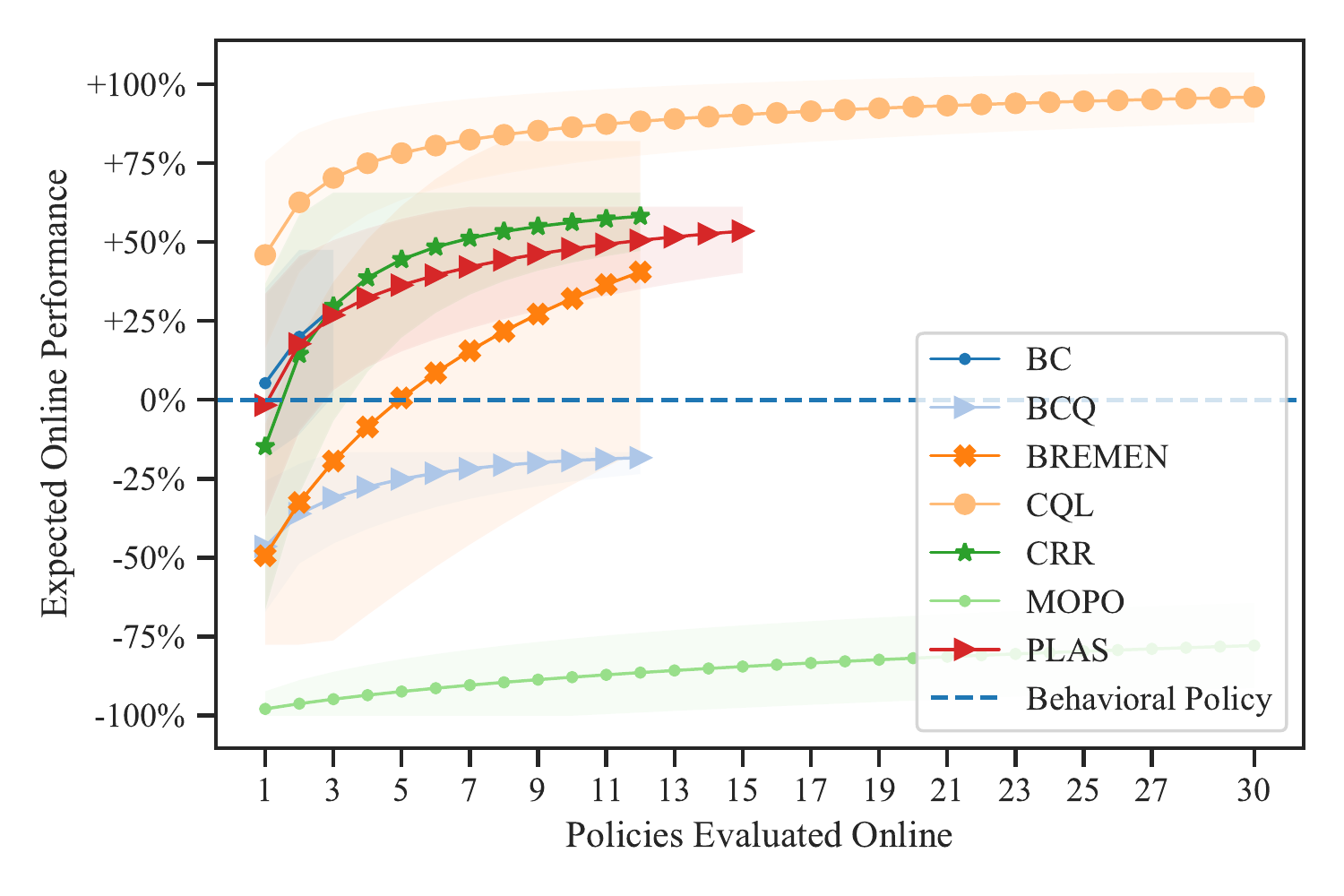}}
  \caption{High-Level Policy, 10000 }
\end{subfigure}
\begin{subfigure}[b]{.32\textwidth}
 \centering
  \centerline{\includegraphics[width=\textwidth]{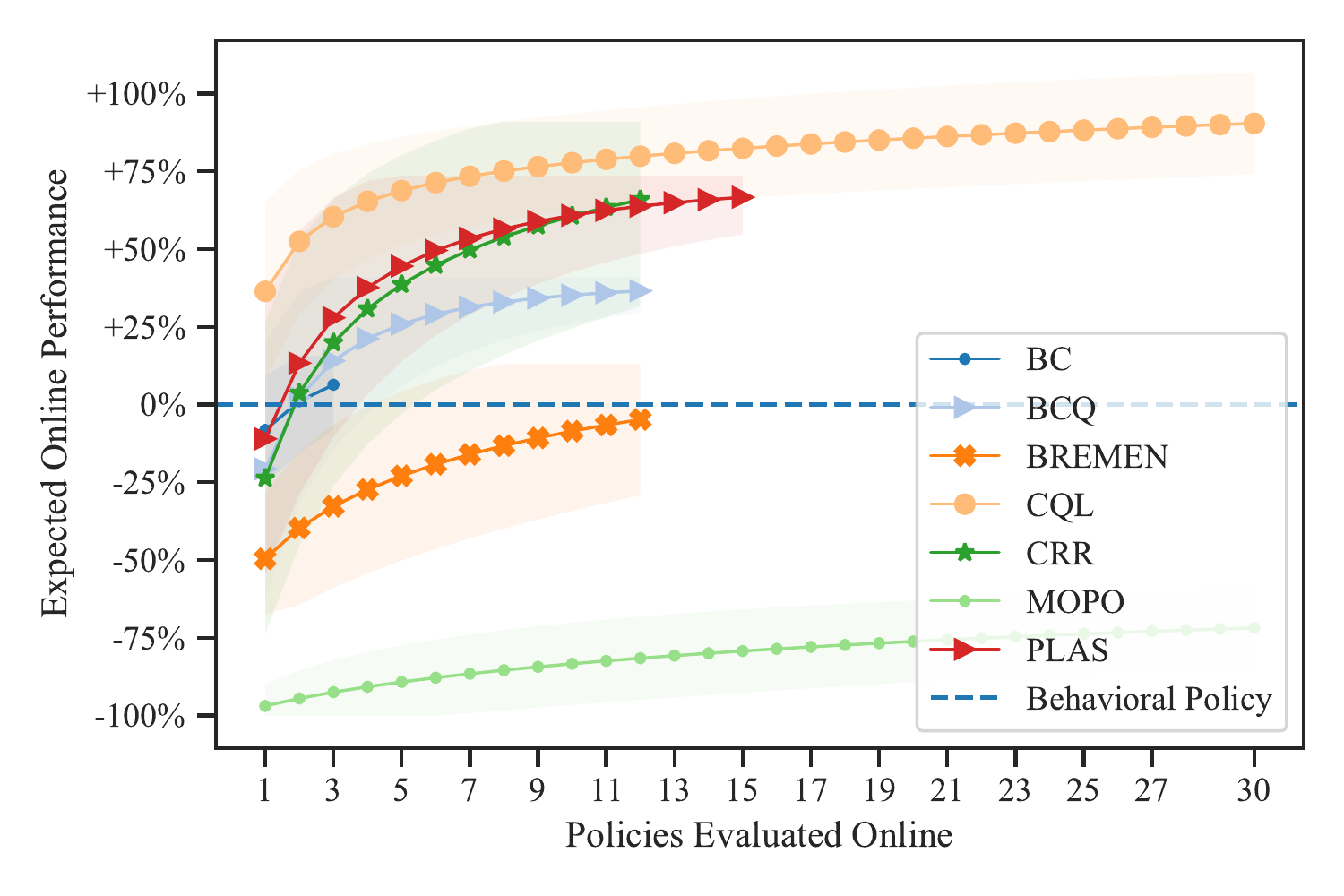}}
  \caption{High-Level Policy, 1000 }
\end{subfigure}
\begin{subfigure}[b]{.32\textwidth}
 \centering
  \centerline{\includegraphics[width=\textwidth]{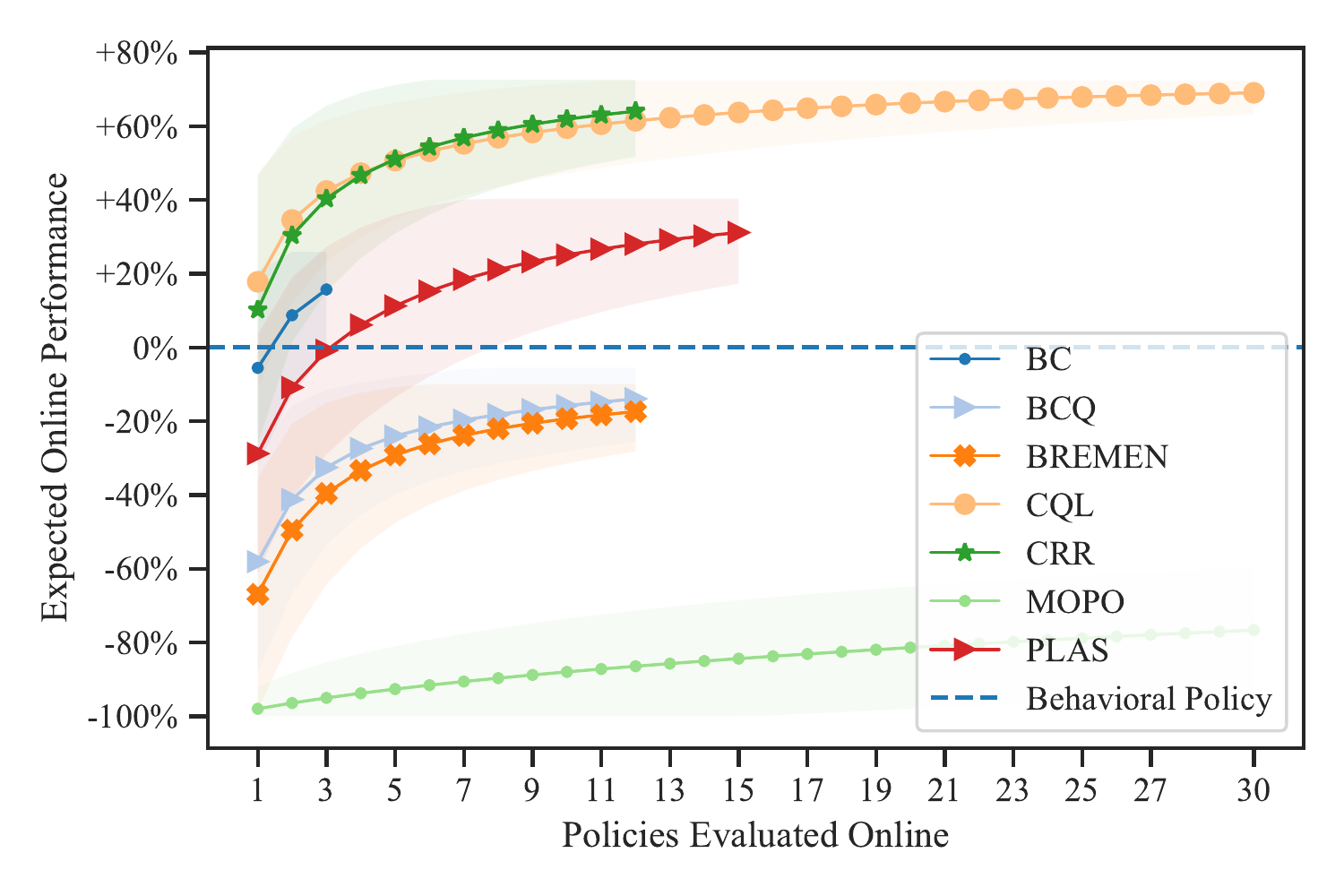}}
  \caption{High-Level Policy, 100 }
\end{subfigure}

\begin{subfigure}[b]{.32\textwidth}
 \centering
  \centerline{\includegraphics[width=\textwidth]{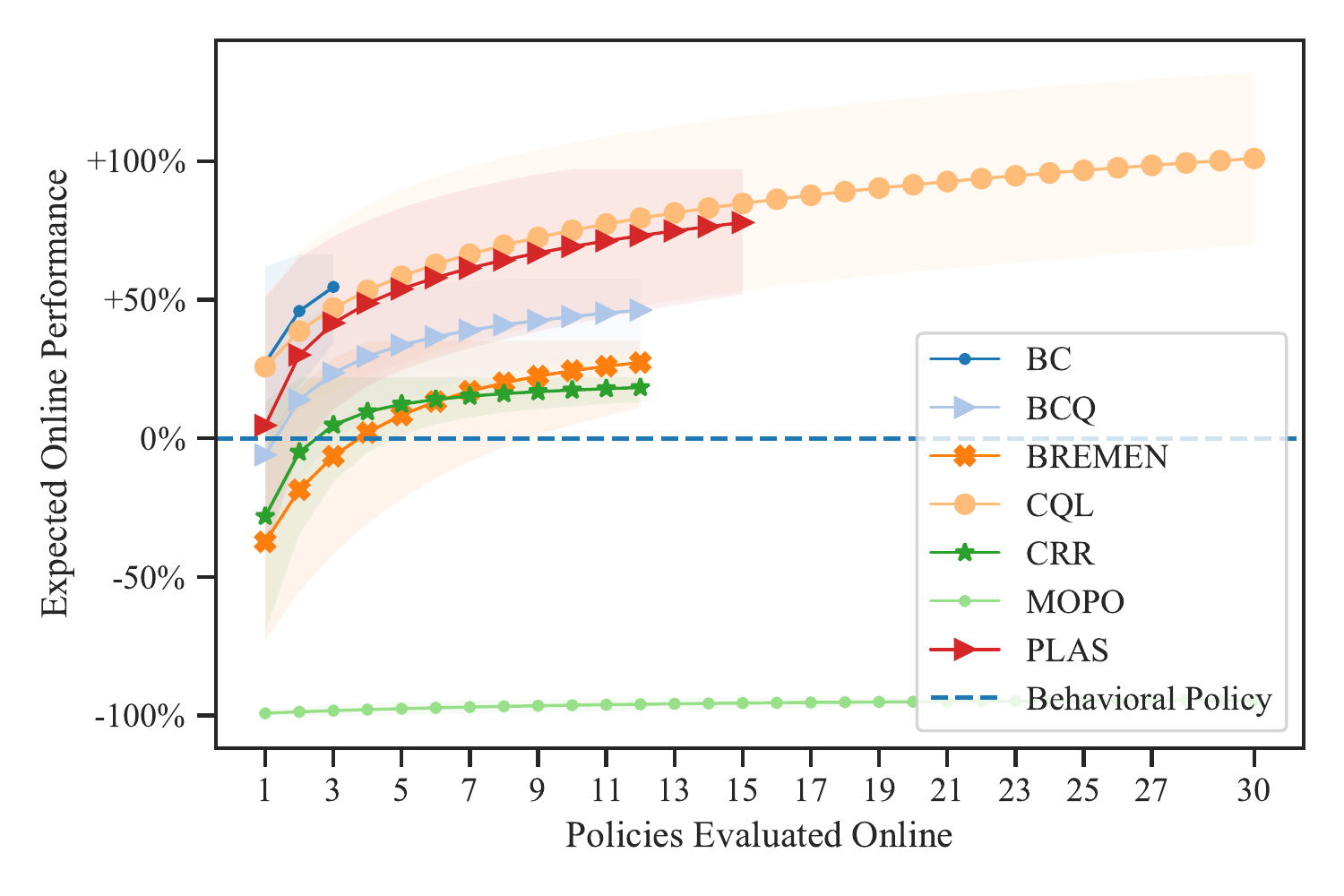}}
  \caption{Medium-Level Policy, 10000 }
\end{subfigure}
\begin{subfigure}[b]{.32\textwidth}
 \centering
  \centerline{\includegraphics[width=\textwidth]{figures_appendix/Hopper-v3_1000_medium_uniform.pdf}}
  \caption{Medium-Level Policy, 1000 }
\end{subfigure}
\begin{subfigure}[b]{.32\textwidth}
 \centering
  \centerline{\includegraphics[width=\textwidth]{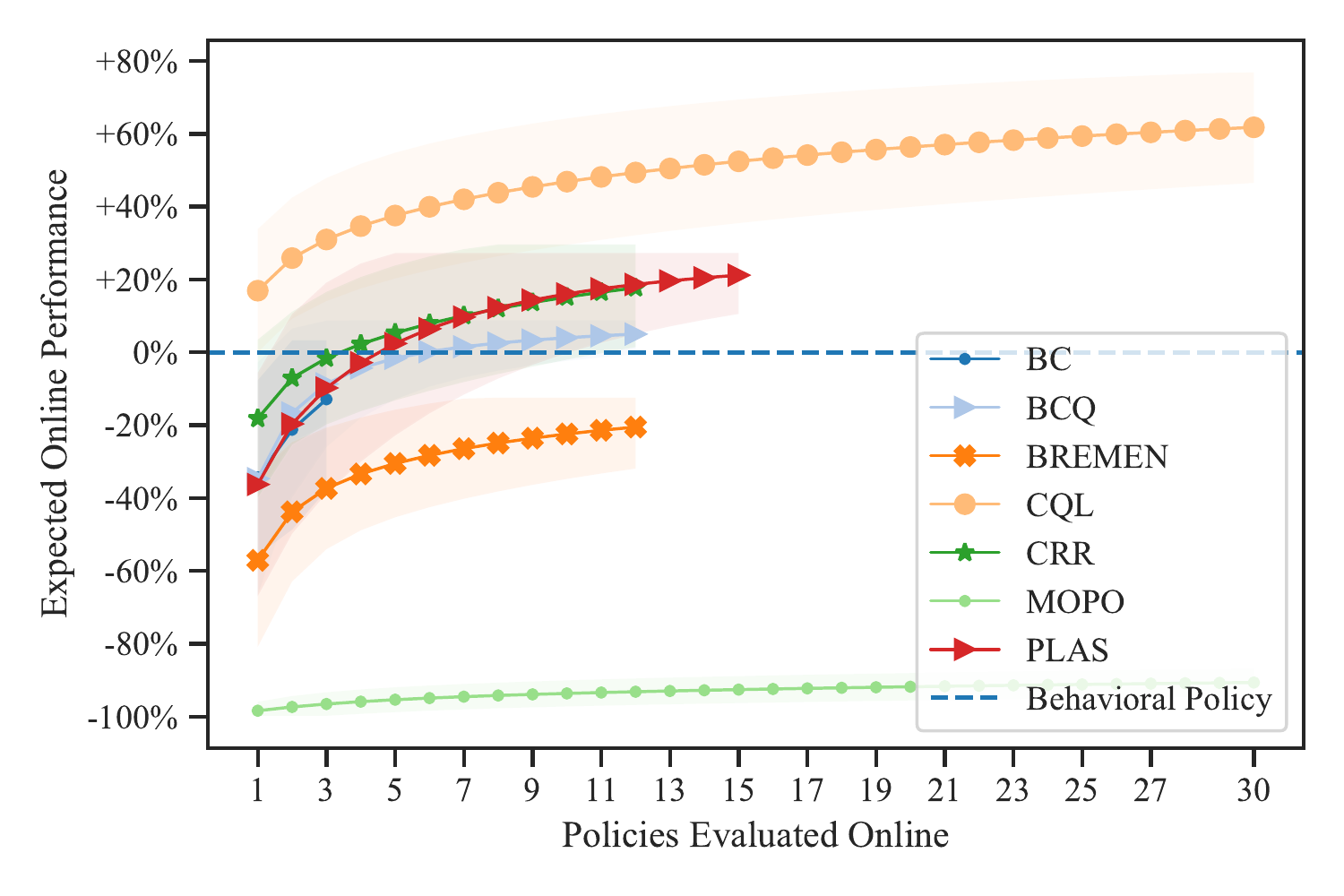}}
  \caption{Medium-Level Policy, 100 }
\end{subfigure}

\begin{subfigure}[b]{.32\textwidth}
 \centering
  \centerline{\includegraphics[width=\textwidth]{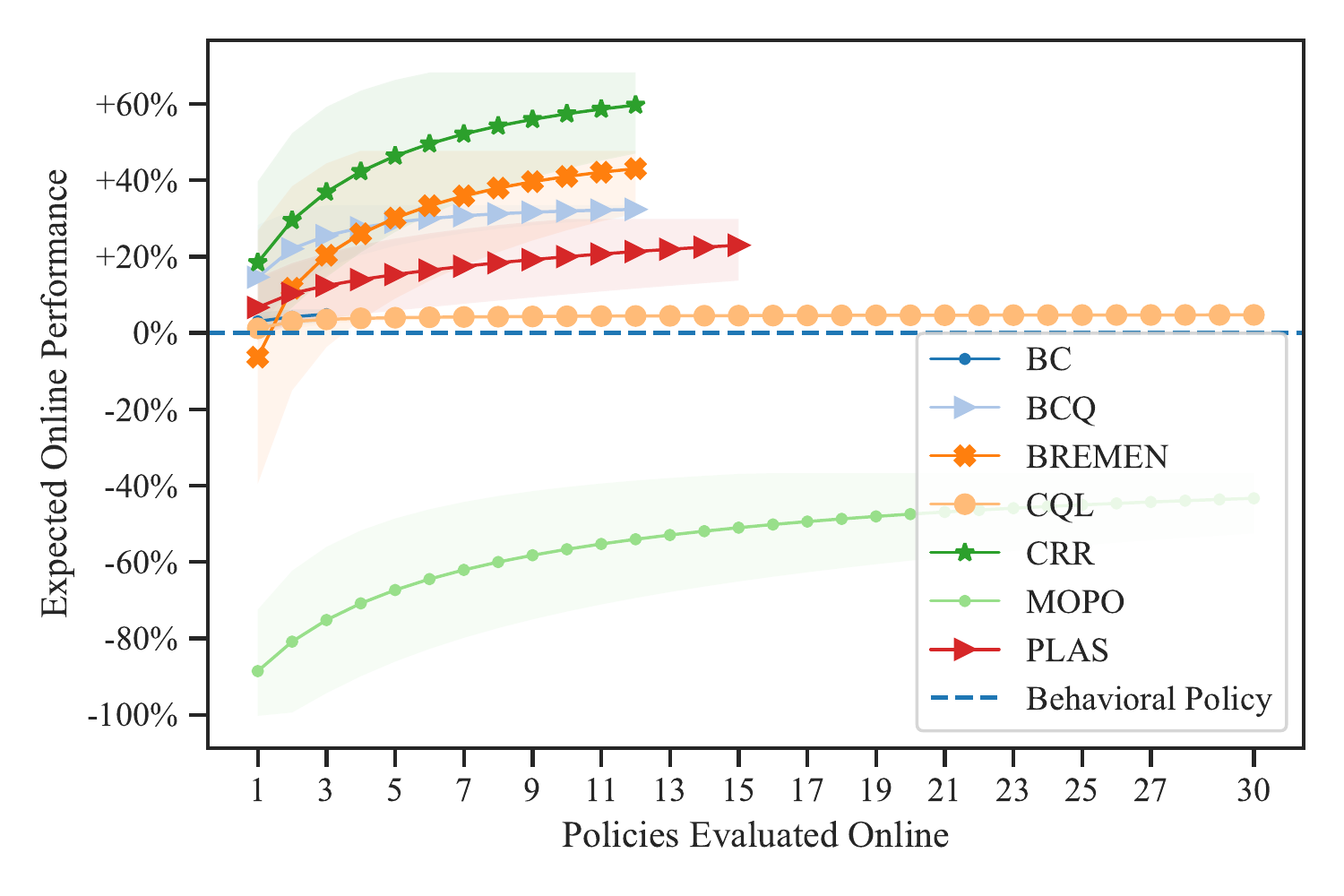}}
  \caption{Low-Level Policy, 10000 }
\end{subfigure}
\begin{subfigure}[b]{.32\textwidth}
 \centering
  \centerline{\includegraphics[width=\textwidth]{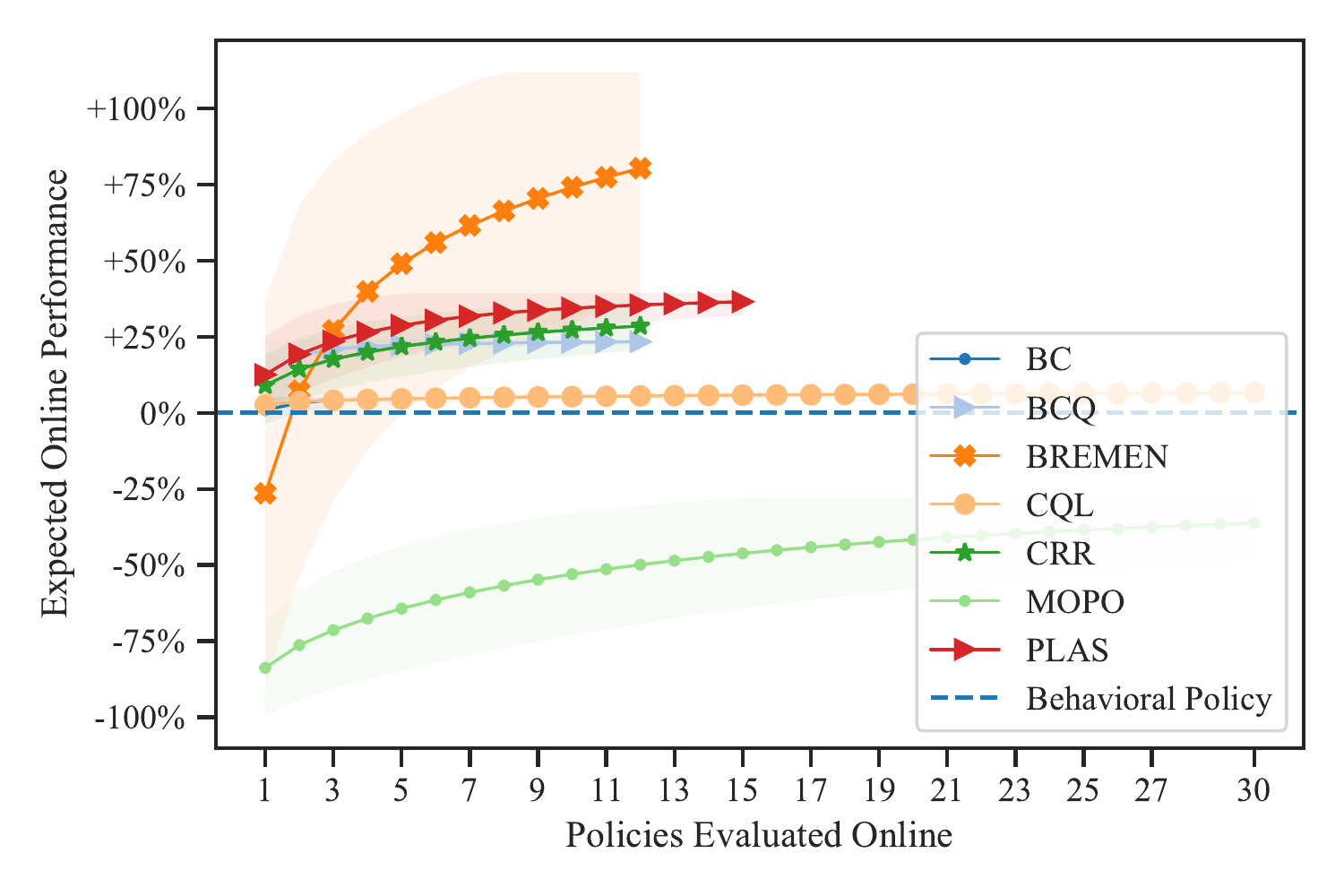}}
  \caption{Low-Level Policy, 1000 }
\end{subfigure}
\begin{subfigure}[b]{.32\textwidth}
 \centering
  \centerline{\includegraphics[width=\textwidth]{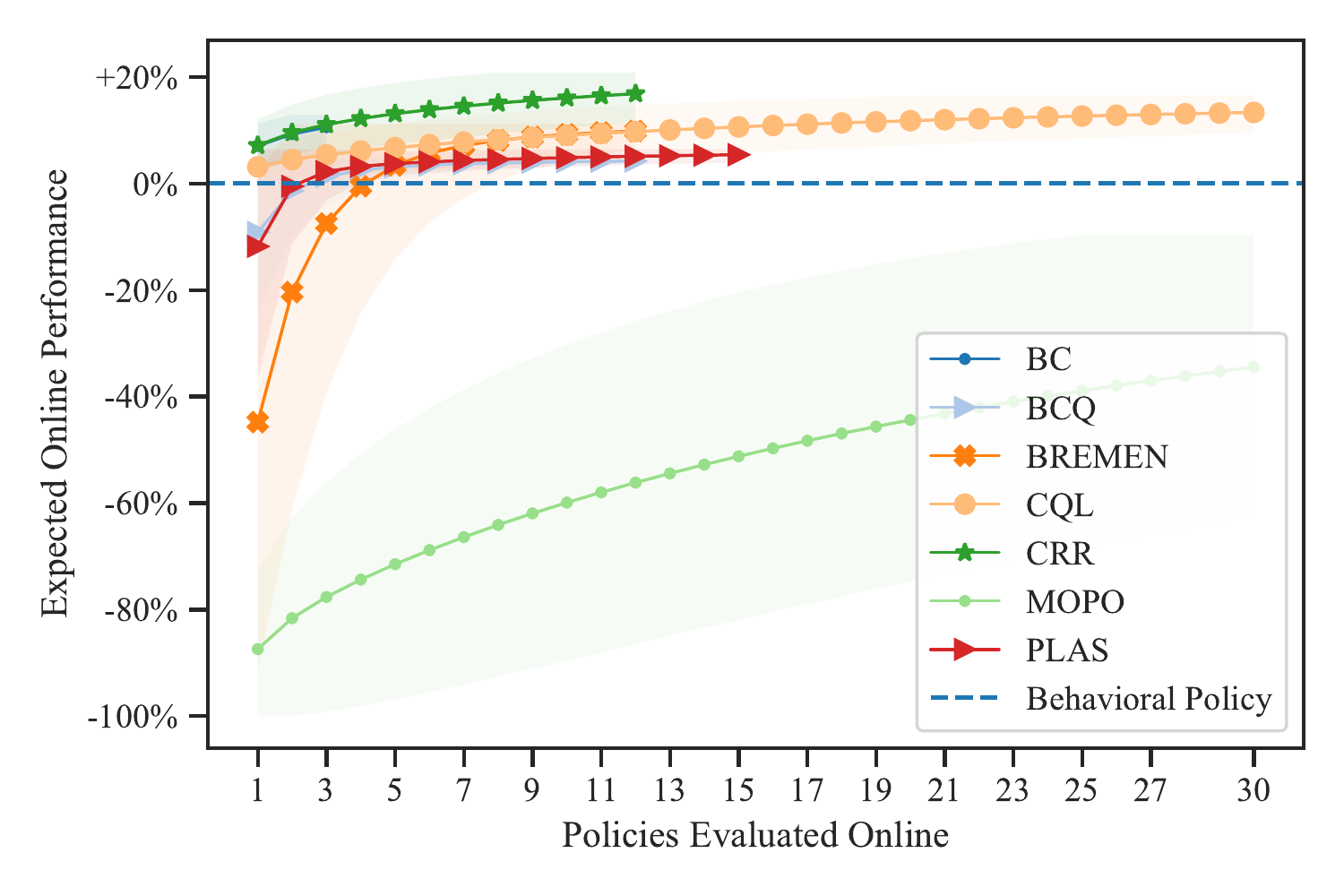}}
  \caption{Low-Level Policy, 100 }
\end{subfigure}
\caption{\textbf{Expected Online Performance under uniform policy selection. Hopper-v3}, a robotic task from \citet{qinNeoRLRealWorldBenchmark2021}, the graphs are computed using their open-sourced online evaluations for different hyperparameter assignments.}
\label{fig:appendix:hopper}
\end{figure*} 

\begin{figure*}[!h]
 \centering
\begin{subfigure}[b]{.32\textwidth}
 \centering
  \centerline{\includegraphics[width=\textwidth]{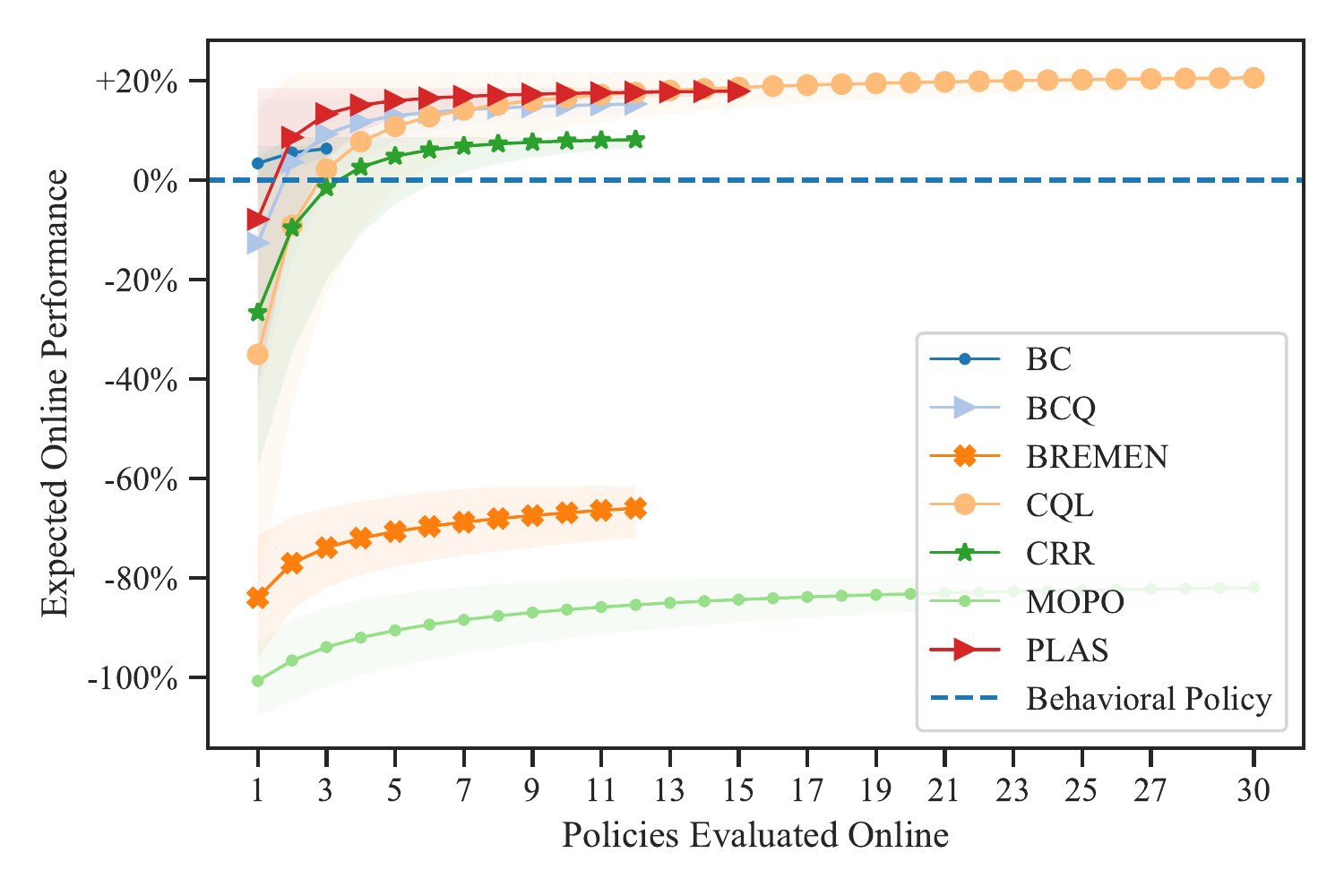}}
  \caption{High-Level Policy, 10000 }
\end{subfigure}
\begin{subfigure}[b]{.32\textwidth}
 \centering
  \centerline{\includegraphics[width=\textwidth]{figures_appendix/HalfCheetah-v3_1000_high_uniform.pdf}}
  \caption{High-Level Policy, 1000 }
\end{subfigure}
\begin{subfigure}[b]{.32\textwidth}
 \centering
  \centerline{\includegraphics[width=\textwidth]{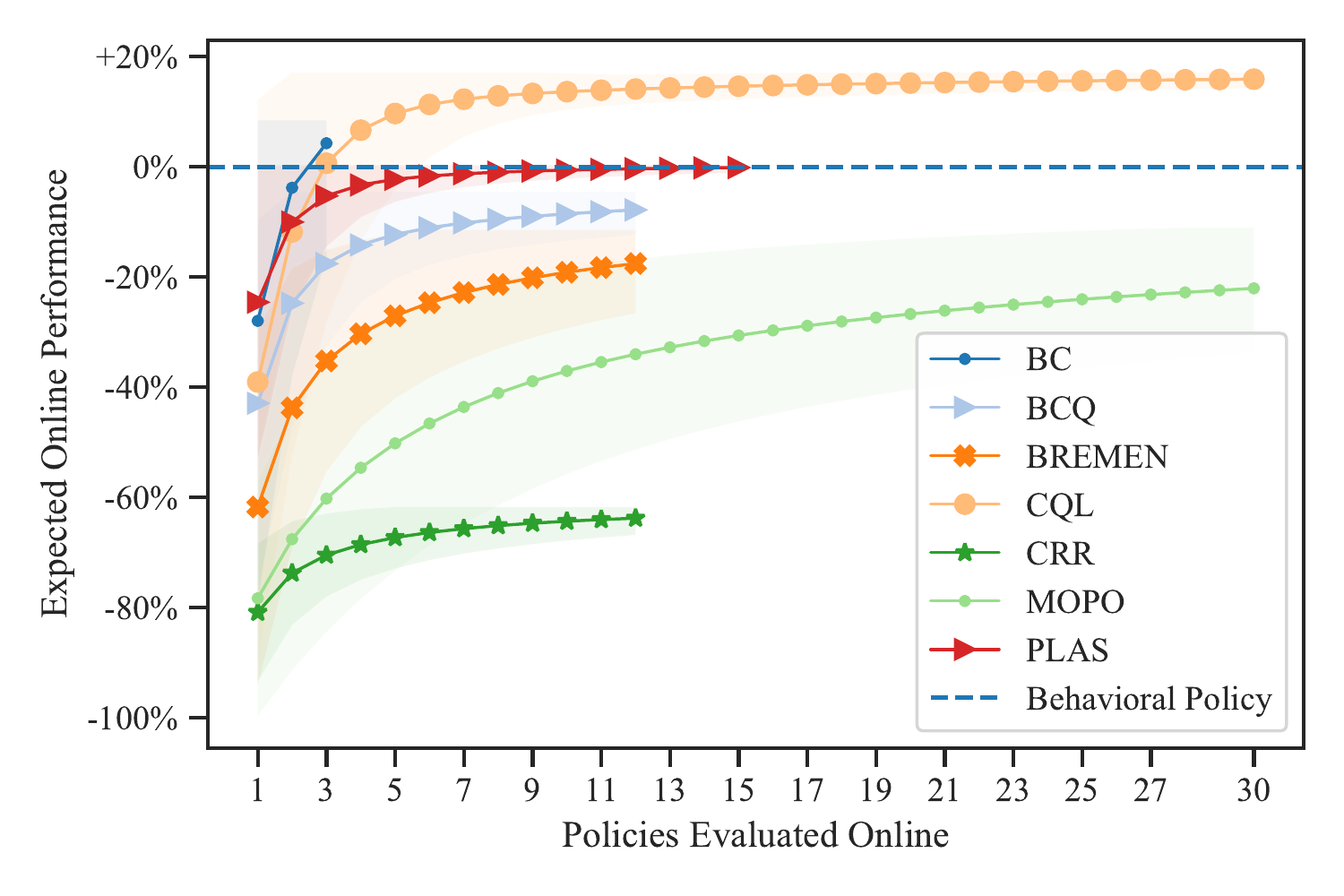}}
  \caption{High-Level Policy, 100 }
\end{subfigure}

\begin{subfigure}[b]{.32\textwidth}
 \centering
  \centerline{\includegraphics[width=\textwidth]{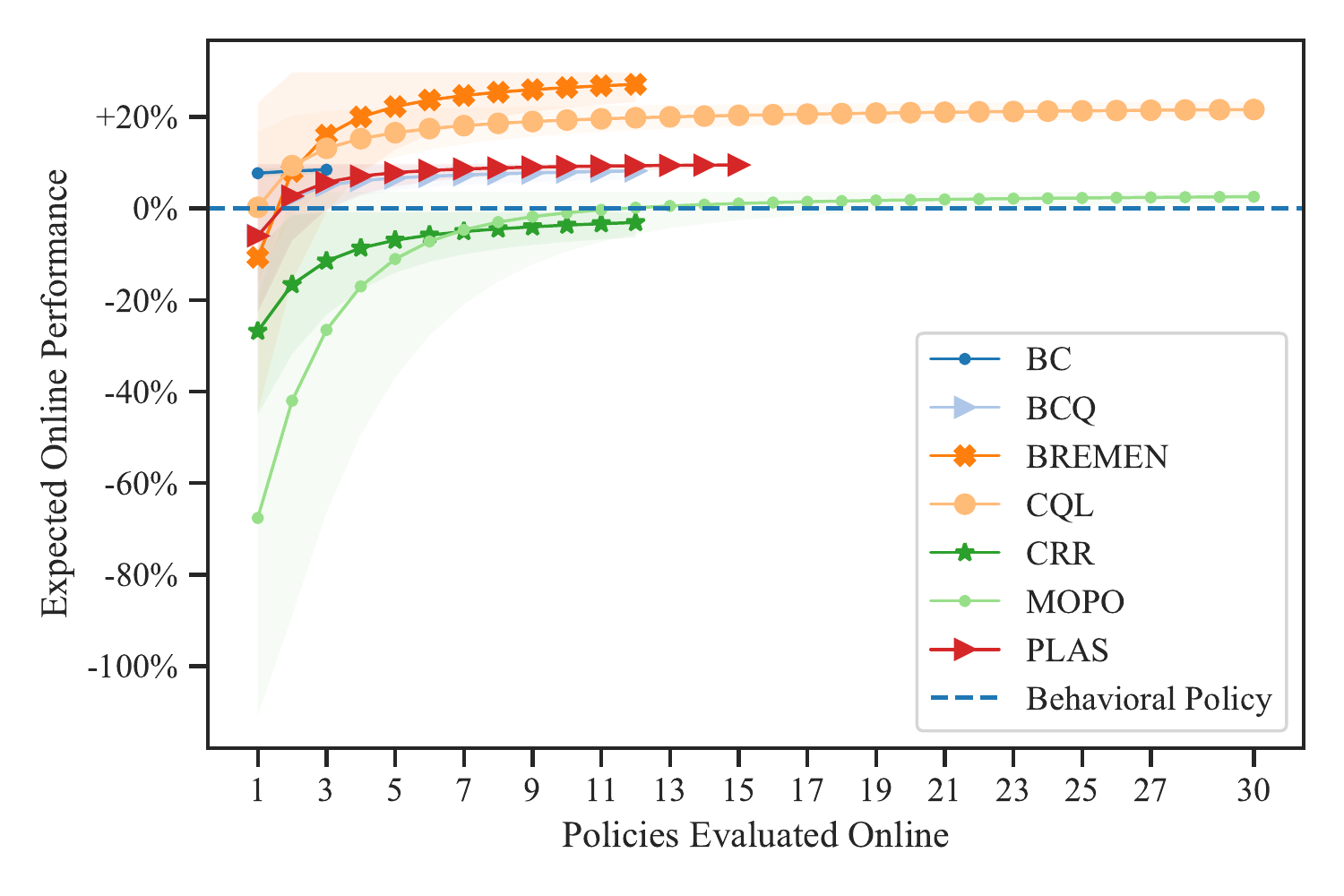}}
  \caption{Medium-Level Policy, 10000 }
\end{subfigure}
\begin{subfigure}[b]{.32\textwidth}
 \centering
  \centerline{\includegraphics[width=\textwidth]{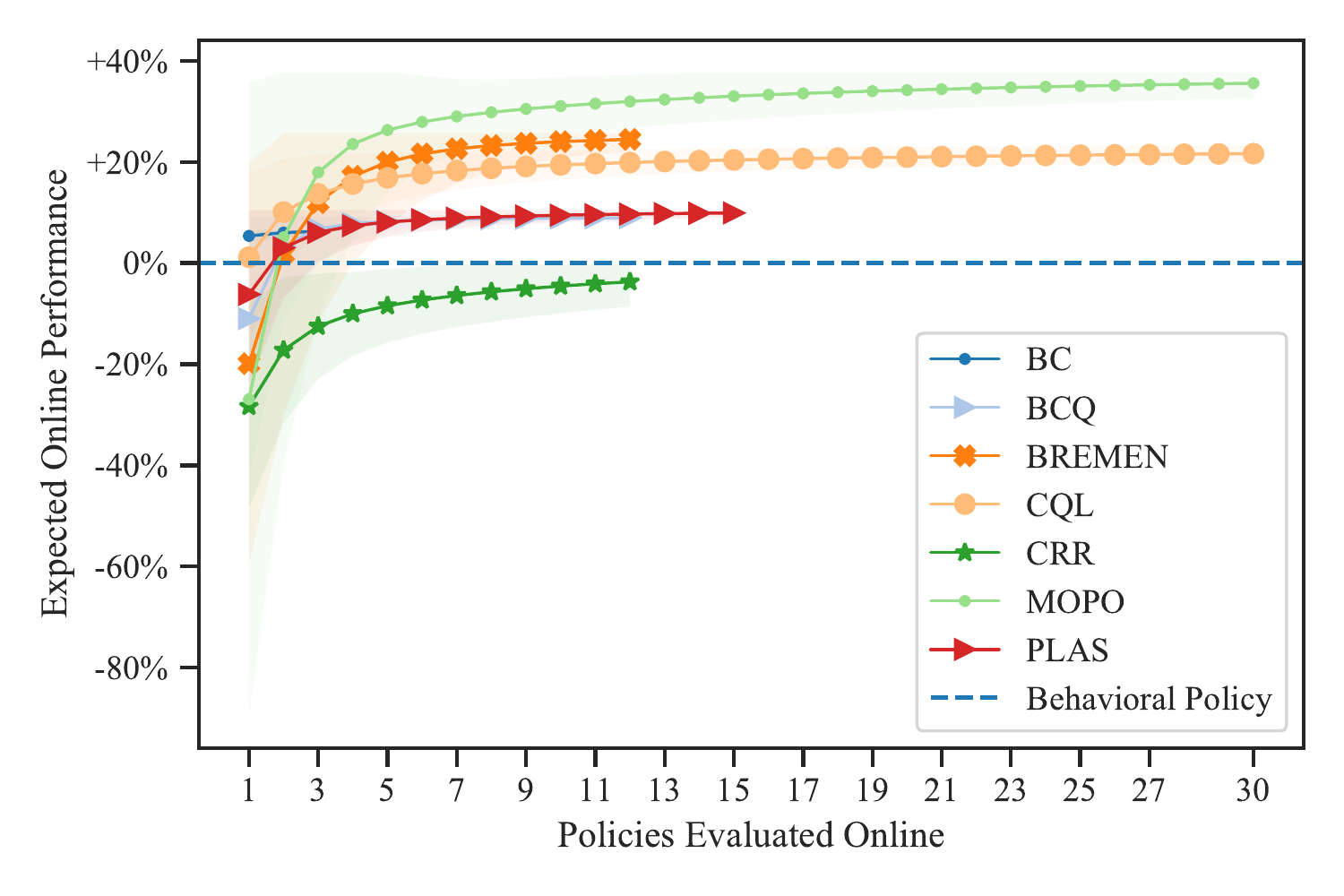}}
  \caption{Medium-Level Policy, 1000 }
\end{subfigure}
\begin{subfigure}[b]{.32\textwidth}
 \centering
  \centerline{\includegraphics[width=\textwidth]{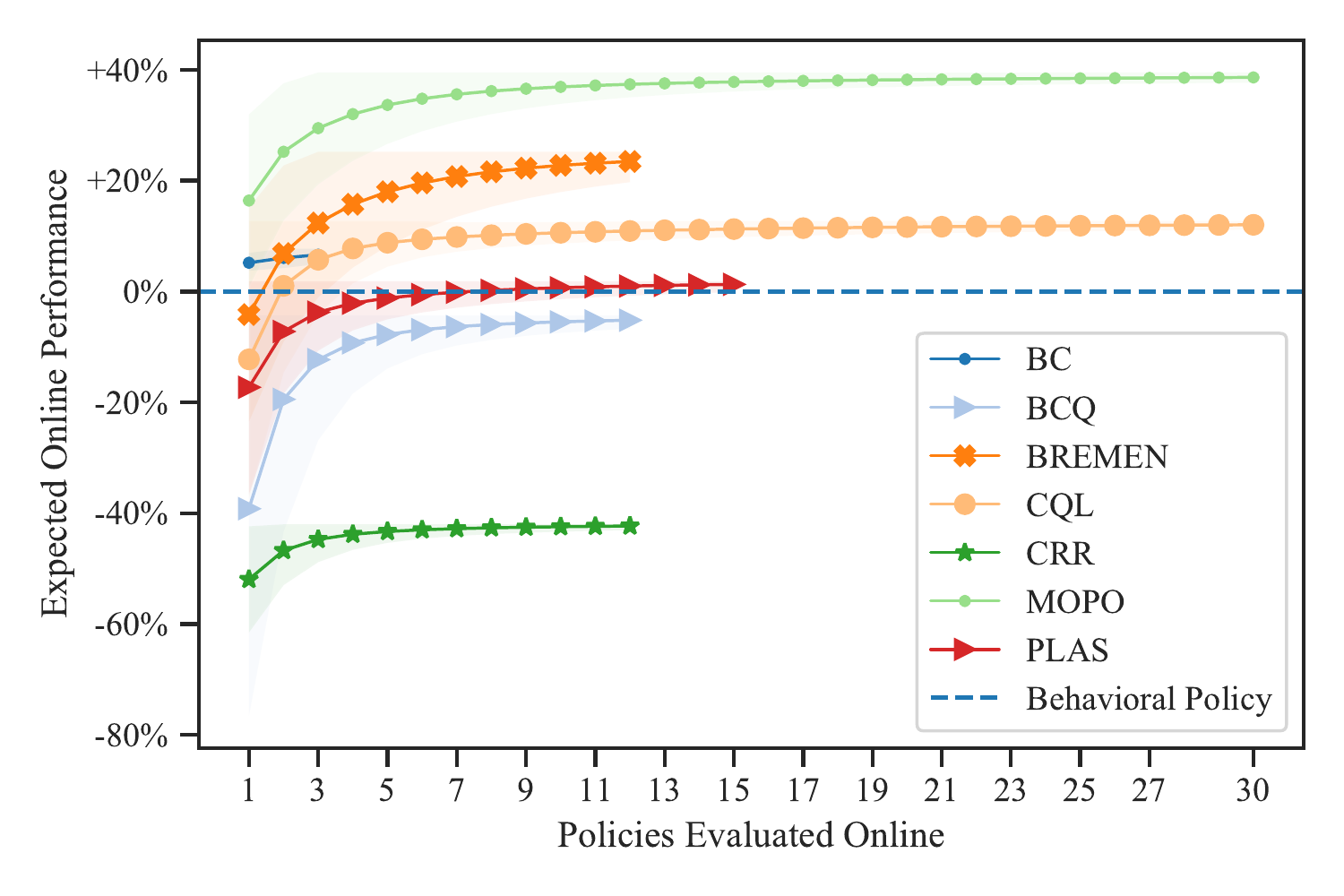}}
  \caption{Medium-Level Policy, 100 }
\end{subfigure}

\begin{subfigure}[b]{.32\textwidth}
 \centering
  \centerline{\includegraphics[width=\textwidth]{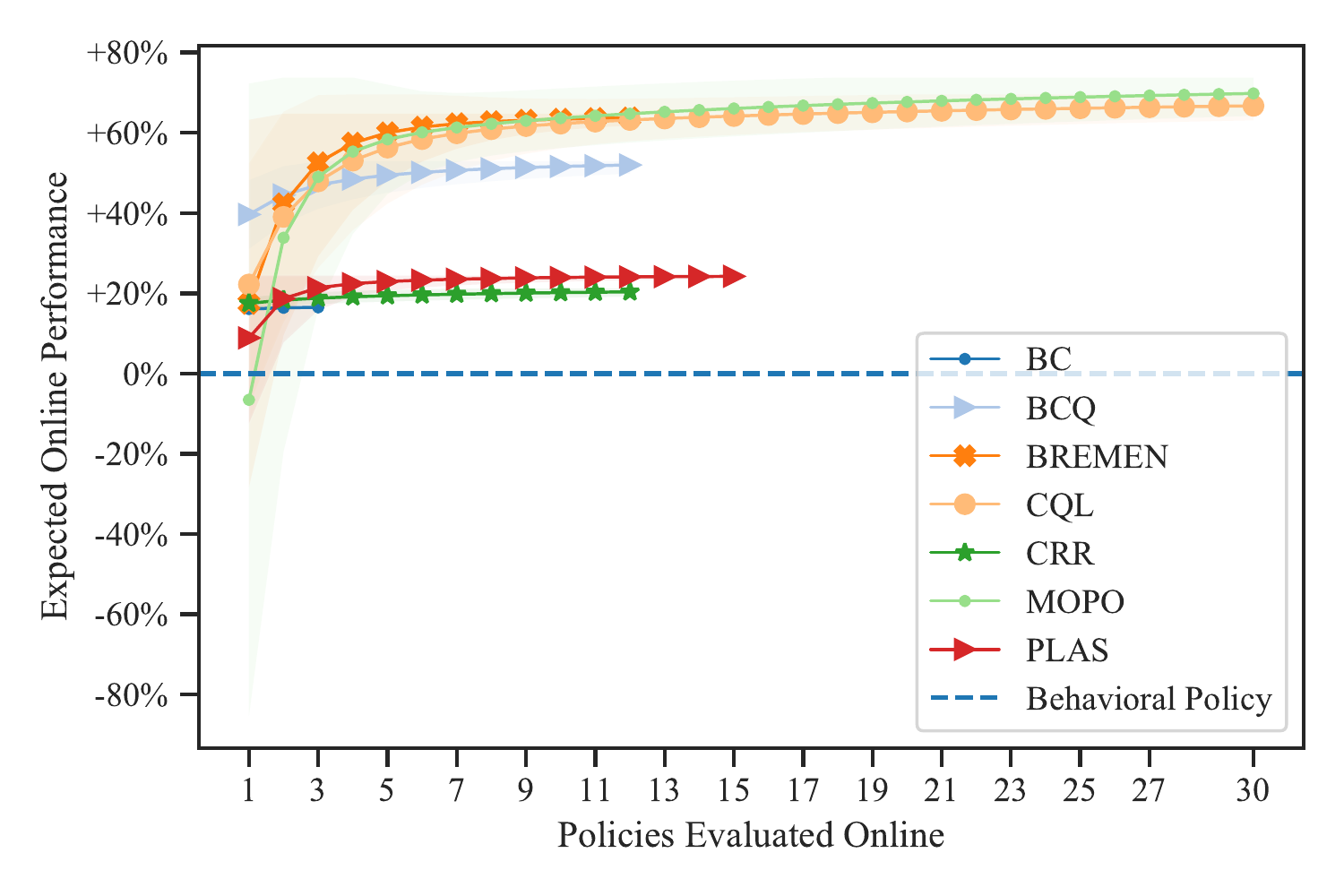}}
  \caption{Low-Level Policy, 10000 }
\end{subfigure}
\begin{subfigure}[b]{.32\textwidth}
 \centering
  \centerline{\includegraphics[width=\textwidth]{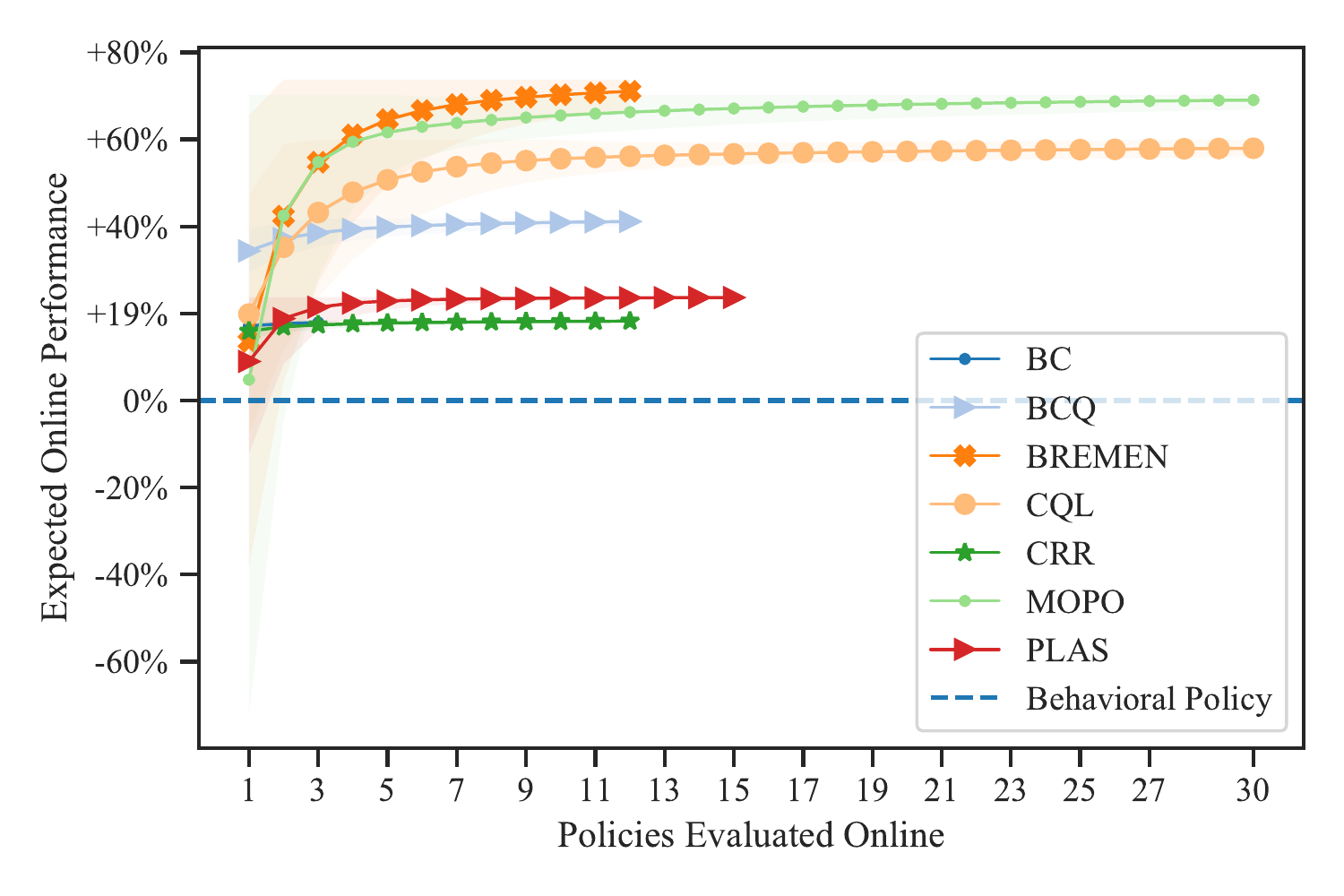}}
  \caption{Low-Level Policy, 1000 }
\end{subfigure}
\begin{subfigure}[b]{.32\textwidth}
 \centering
  \centerline{\includegraphics[width=\textwidth]{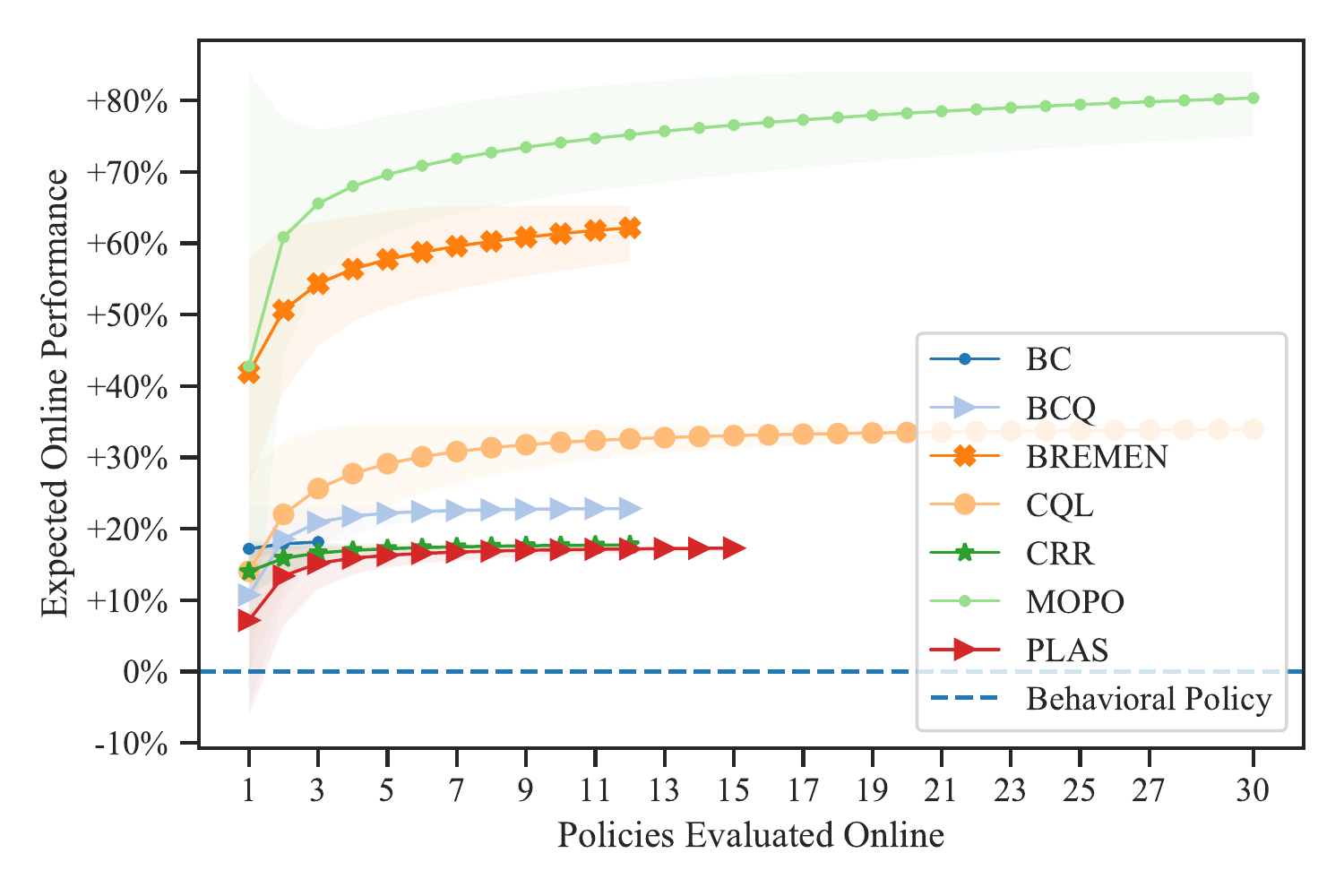}}
  \caption{Low-Level Policy, 100 }
\end{subfigure}
\caption{\textbf{Expected Online Performance under uniform policy selection. HalfCheetah-v3}, a robotic task from \citet{qinNeoRLRealWorldBenchmark2021}, the graphs are computed using their open-sourced online evaluations for different hyperparameter assignments.}
\label{fig:appendix:halfcheetah}
\end{figure*} 

\newpage
\begin{figure*}[h]
 \centering
\begin{subfigure}[b]{.32\textwidth}
 \centering
  \centerline{\includegraphics[width=\textwidth]{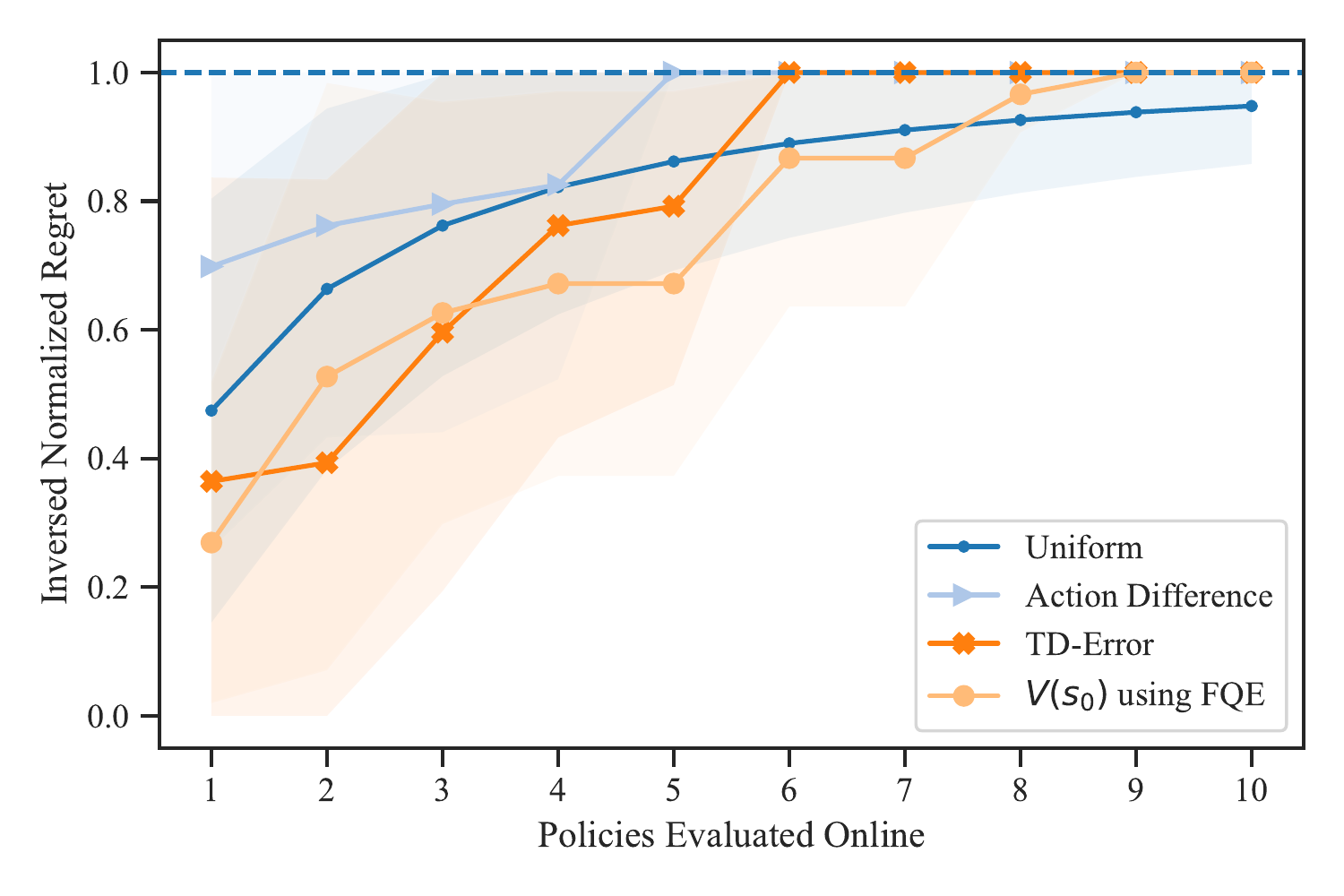}}
  \caption{High-Level Policy, 999 }
\end{subfigure}
\begin{subfigure}[b]{.32\textwidth}
 \centering
  \centerline{\includegraphics[width=\textwidth]{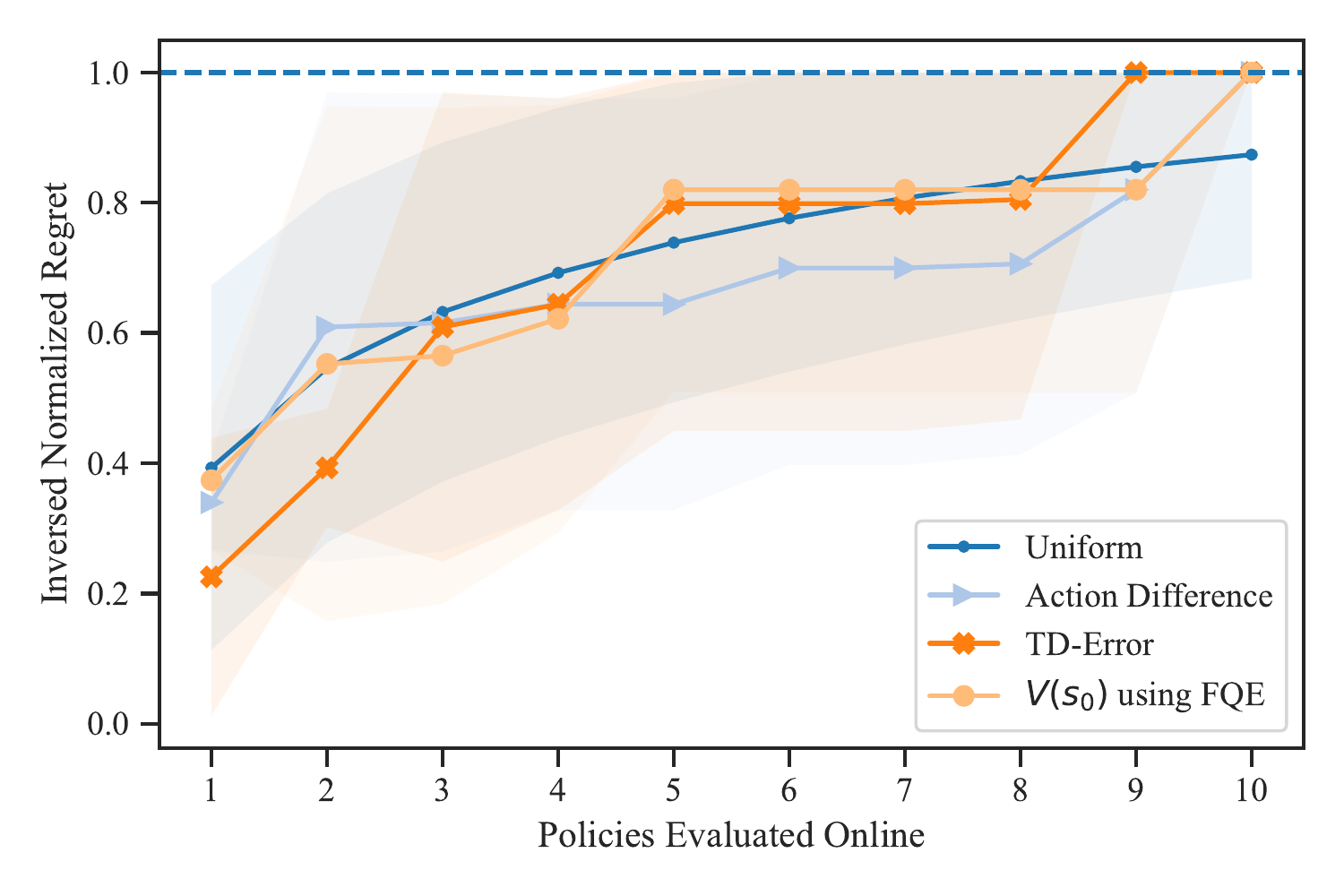}}
  \caption{Medium-Level Policy, 999 }
\end{subfigure}
\begin{subfigure}[b]{.32\textwidth}
 \centering
  \centerline{\includegraphics[width=\textwidth]{figures_appendix/ops_cql_finrl_999_low.pdf}}
  \caption{Low-Level Policy, 999 }
\end{subfigure}

\begin{subfigure}[b]{.32\textwidth}
 \centering
  \centerline{\includegraphics[width=\textwidth]{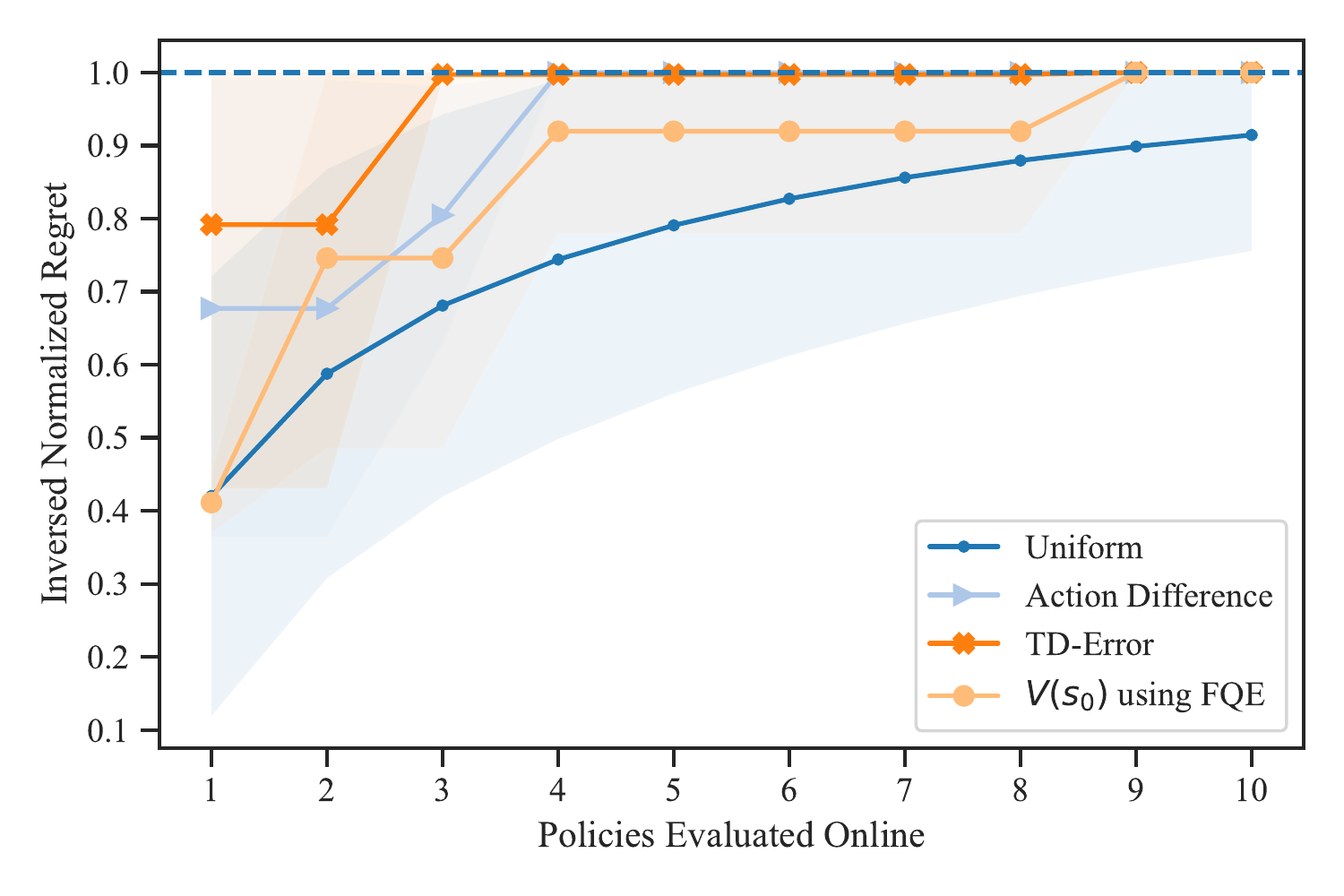}}
  \caption{High-Level Policy, 99 }
\end{subfigure}
\begin{subfigure}[b]{.32\textwidth}
 \centering
  \centerline{\includegraphics[width=\textwidth]{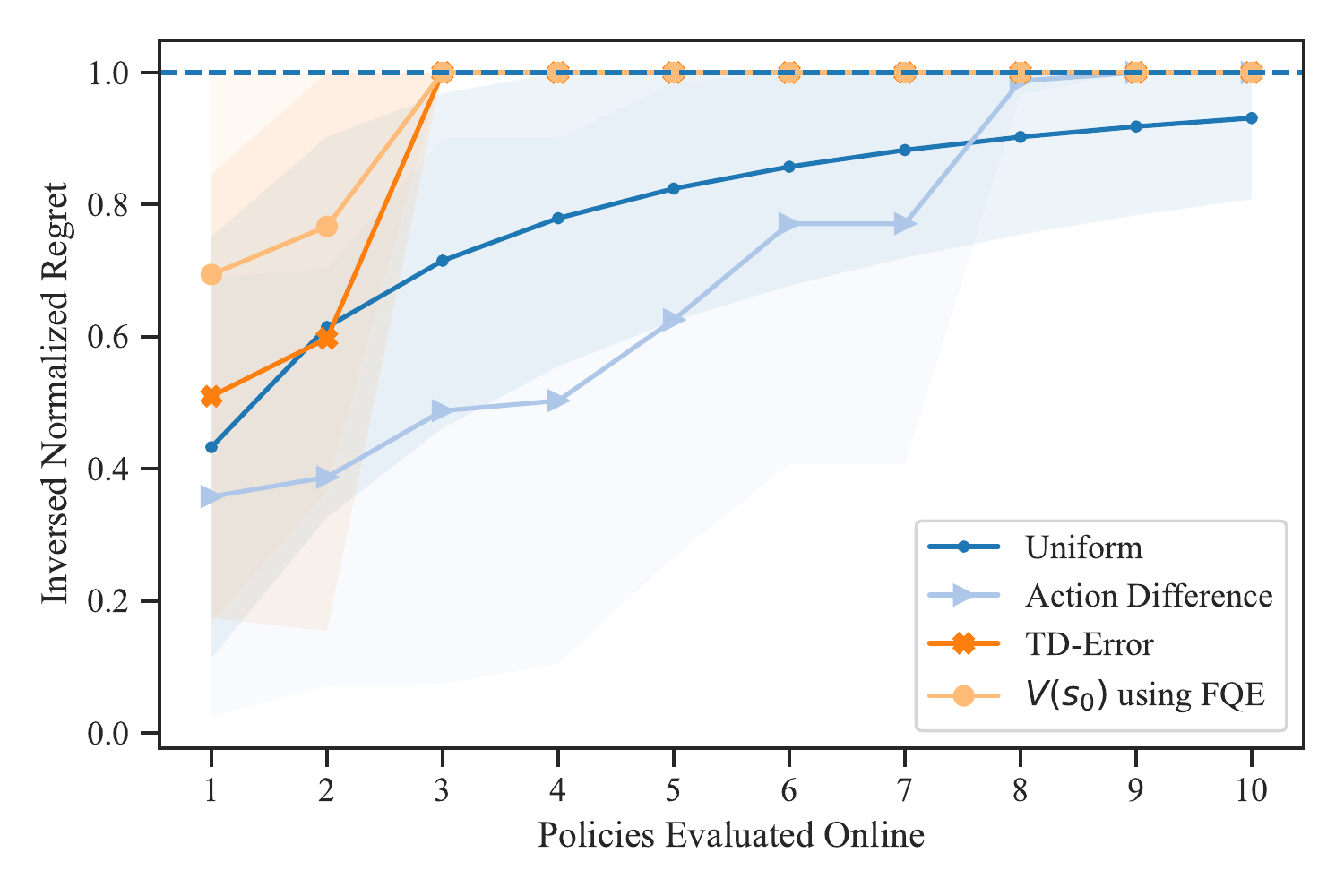}}
  \caption{Medium-Level Policy, 99 }
\end{subfigure}
\begin{subfigure}[b]{.32\textwidth}
 \centering
  \centerline{\includegraphics[width=\textwidth]{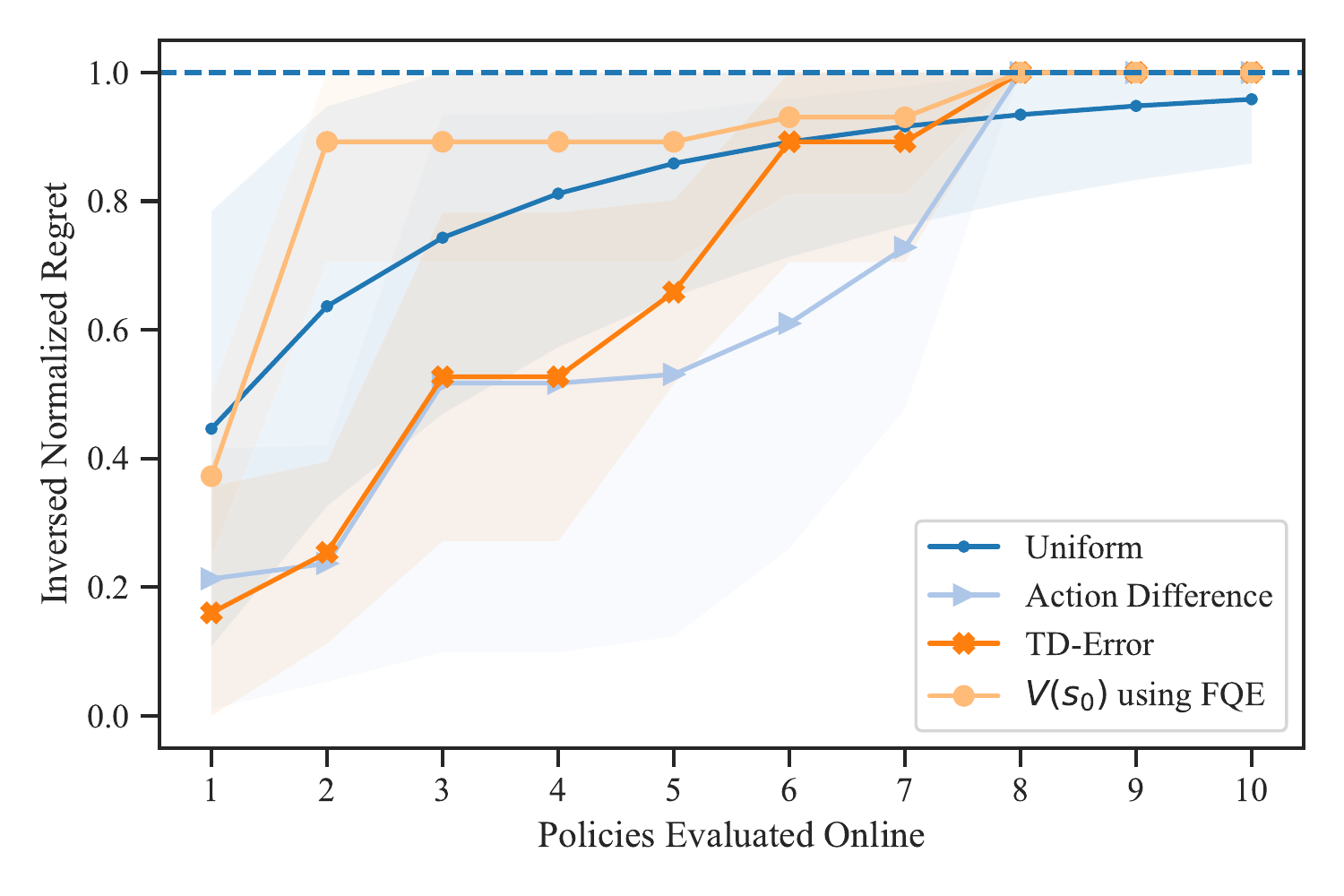}}
  \caption{Low-Level Policy, 99 }
\end{subfigure}

\caption{\textbf{CQL, FinRL. Inversed Normalized Regret under different offline policy selection methods using EOP graph.} The shaded area represents one standard deviation }
\label{fig:appendix:ops_cql_finrl}
\end{figure*}

\begin{figure*}[h]
 \centering
\begin{subfigure}[b]{.32\textwidth}
 \centering
  \centerline{\includegraphics[width=\textwidth]{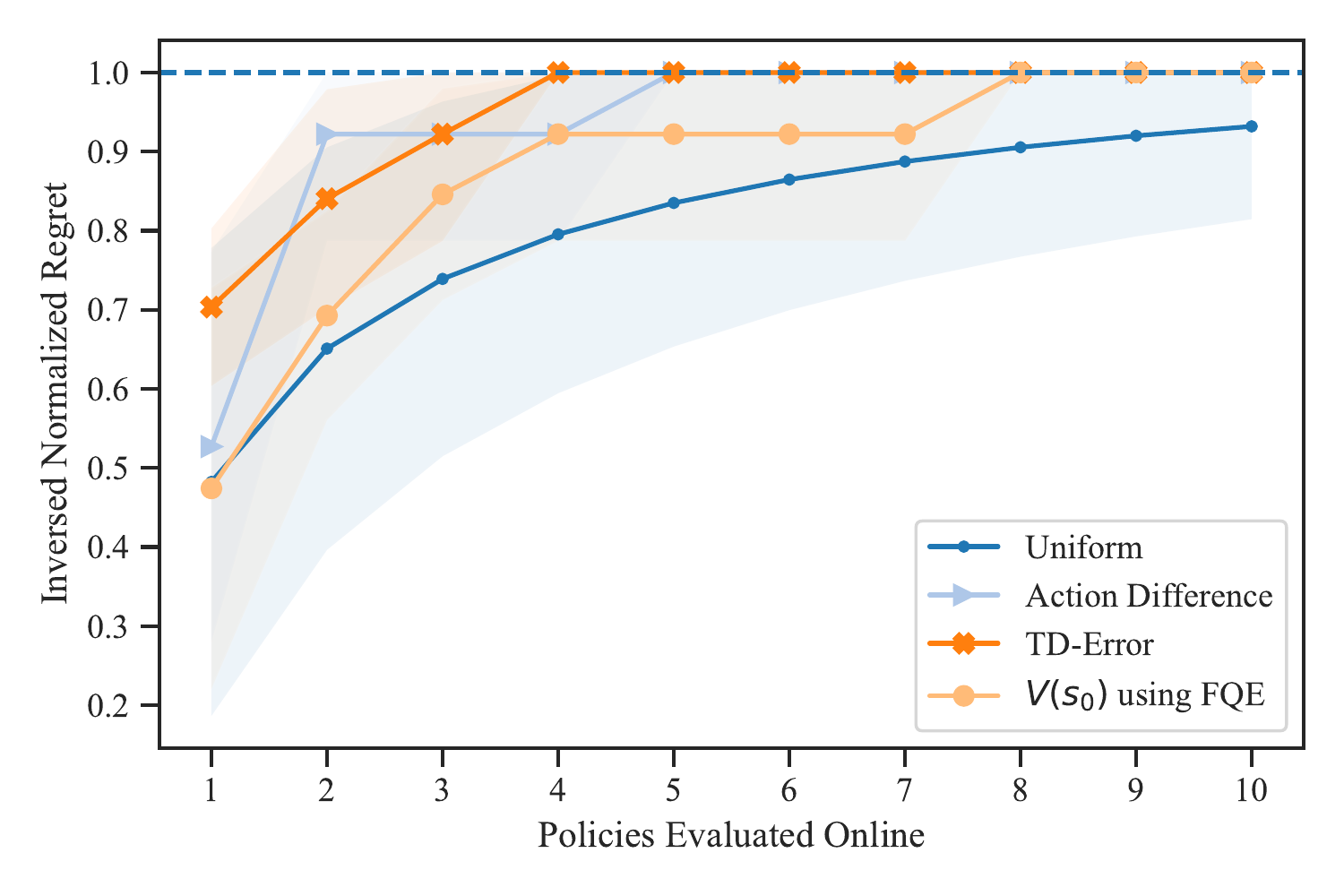}}
  \caption{Medium-Level Policy, 9999 }
\end{subfigure}
\begin{subfigure}[b]{.32\textwidth}
 \centering
  \centerline{\includegraphics[width=\textwidth]{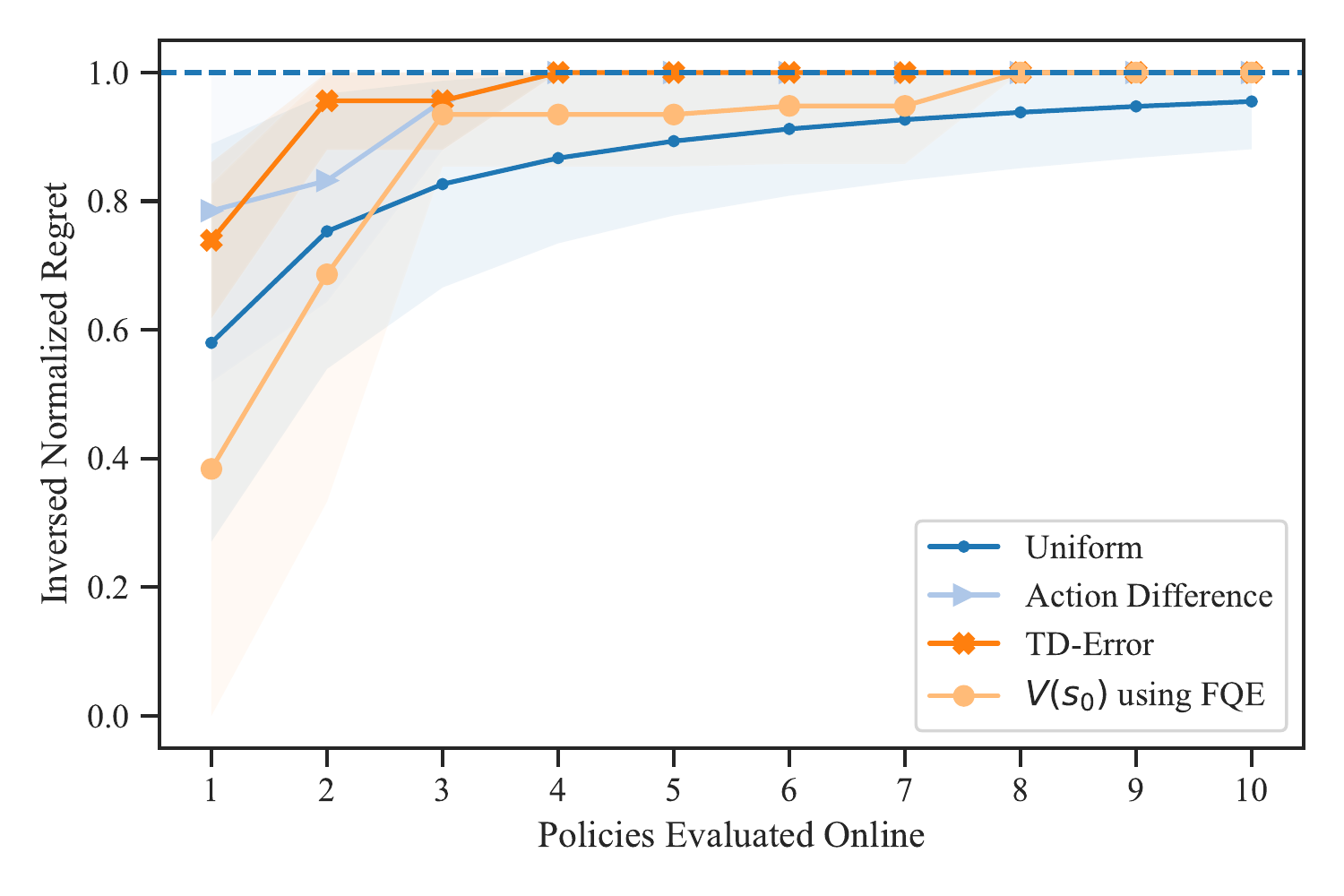}}
  \caption{Medium-Level Policy, 999 }
\end{subfigure}
\begin{subfigure}[b]{.32\textwidth}
 \centering
  \centerline{\includegraphics[width=\textwidth]{figures_appendix/ops_cql_citylearn_999_medium.pdf}}
  \caption{Medium-Level Policy, 99 }
\end{subfigure}

\begin{subfigure}[b]{.32\textwidth}
 \centering
  \centerline{\includegraphics[width=\textwidth]{figures_appendix/ops_cql_citylearn_9999_low.pdf}}
  \caption{Low-Level Policy, 9999 }
\end{subfigure}
\begin{subfigure}[b]{.32\textwidth}
 \centering
  \centerline{\includegraphics[width=\textwidth]{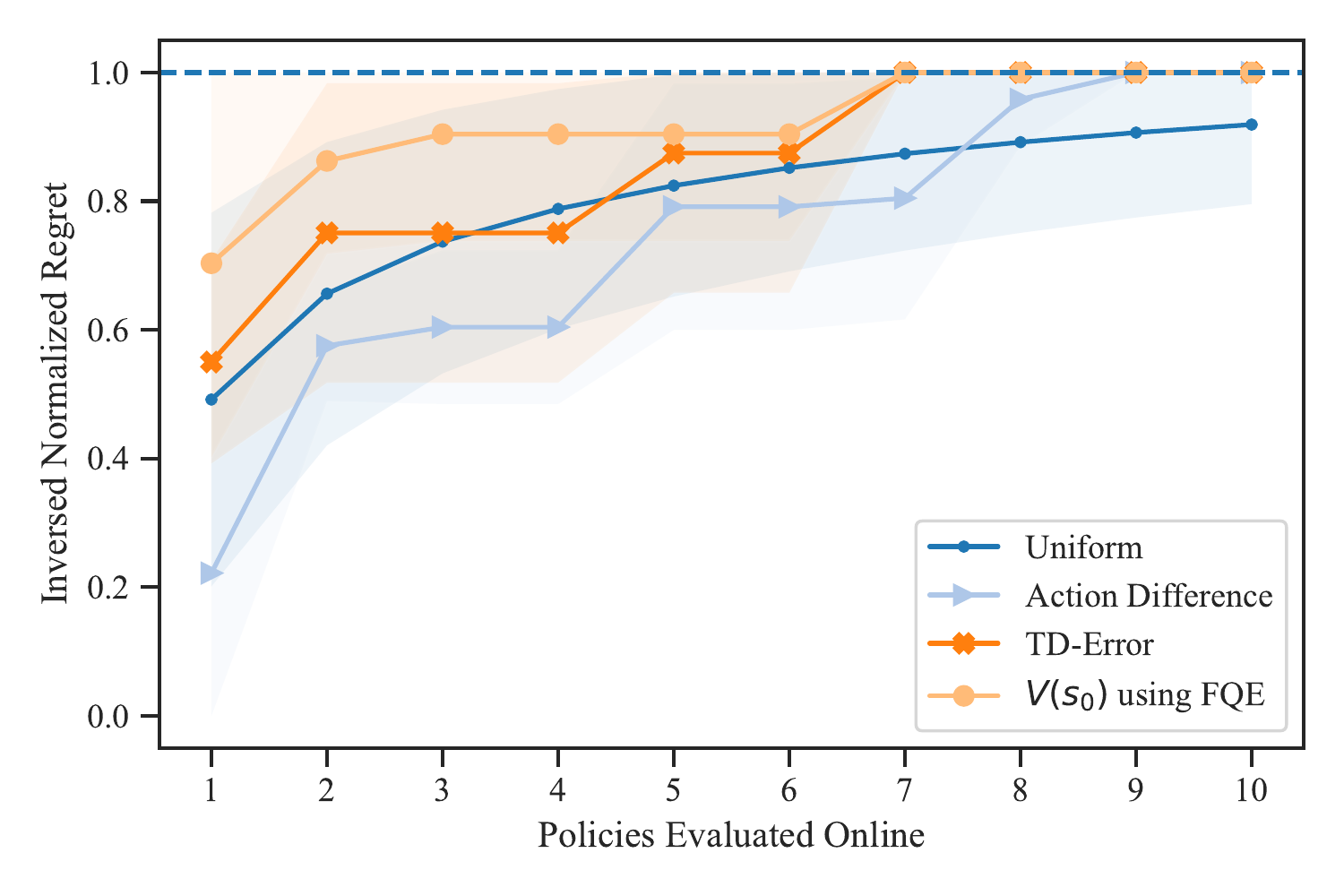}}
  \caption{Low-Level Policy, 999 }
\end{subfigure}
\begin{subfigure}[b]{.32\textwidth}
 \centering
  \centerline{\includegraphics[width=\textwidth]{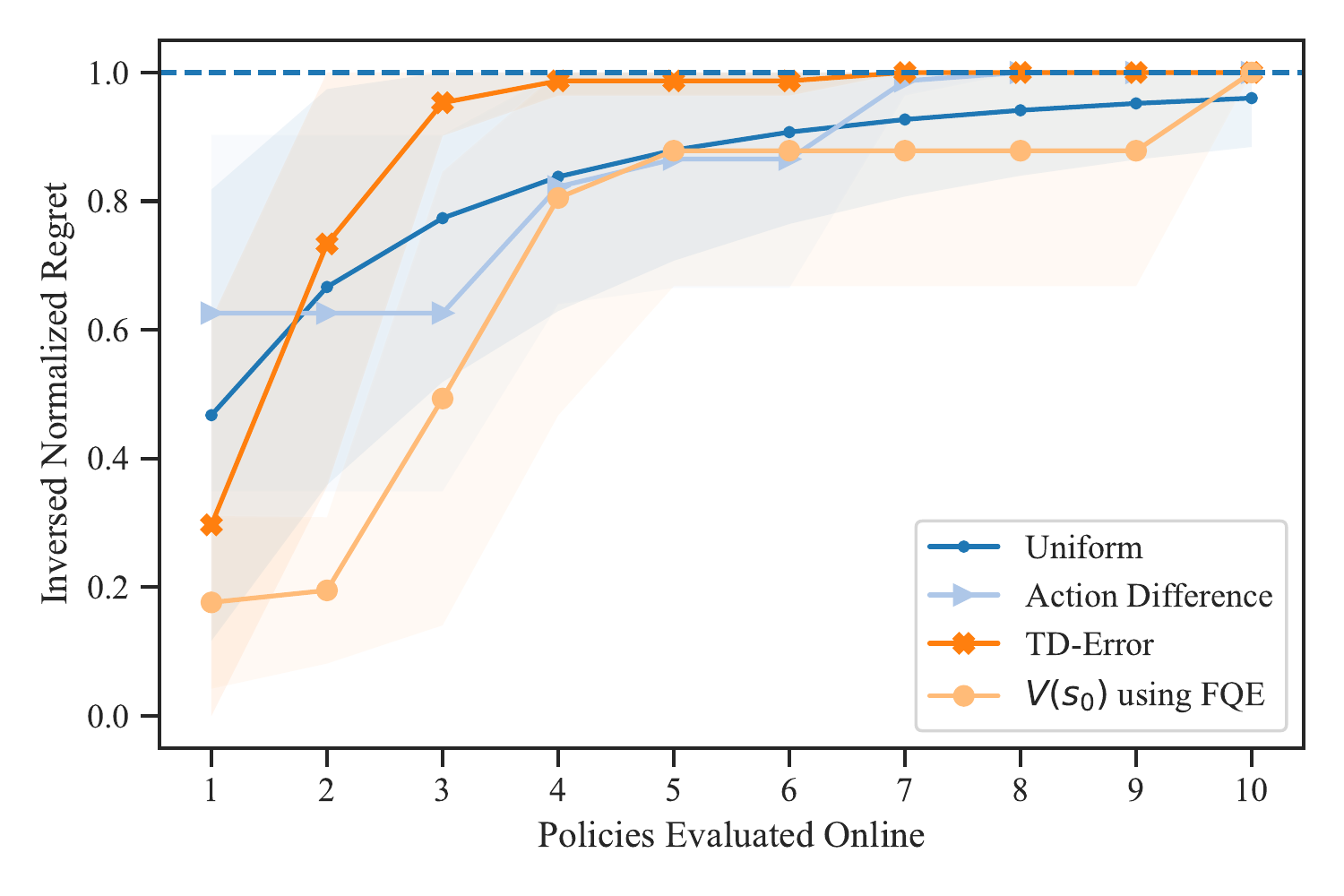}}
  \caption{Low-Level Policy, 99 }
\end{subfigure}

\caption{\textbf{CQL, CityLearn. Inversed Normalized Regret under different offline policy selection methods using EOP graph.} The shaded area represents one standard deviation }
\label{fig:appendix:ops_cql_citylearn}
\end{figure*}

\begin{figure*}[h]
 \centering
\begin{subfigure}[b]{.32\textwidth}
 \centering
  \centerline{\includegraphics[width=\textwidth]{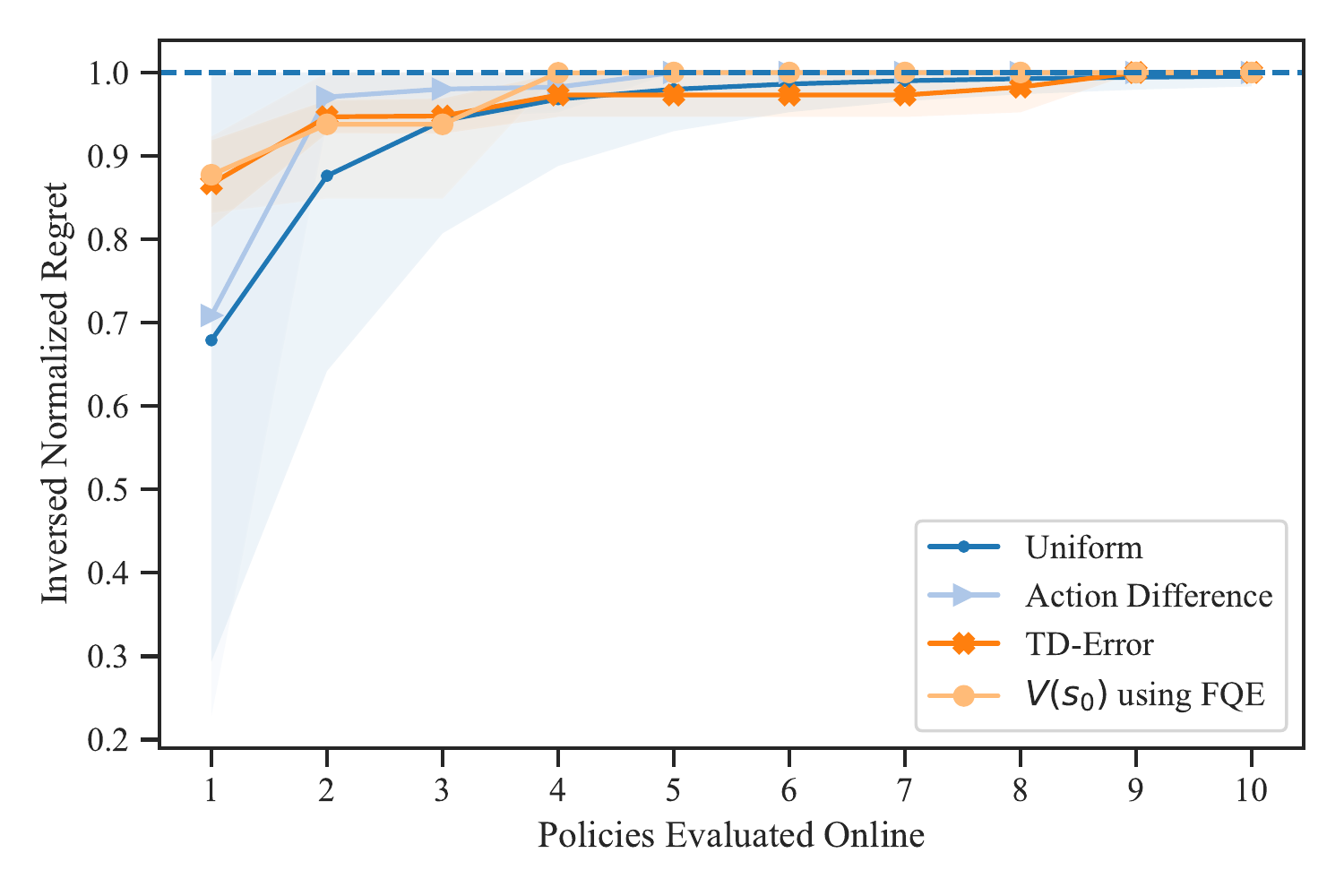}}
  \caption{High-Level Policy, 9999 }
\end{subfigure}
\begin{subfigure}[b]{.32\textwidth}
 \centering
  \centerline{\includegraphics[width=\textwidth]{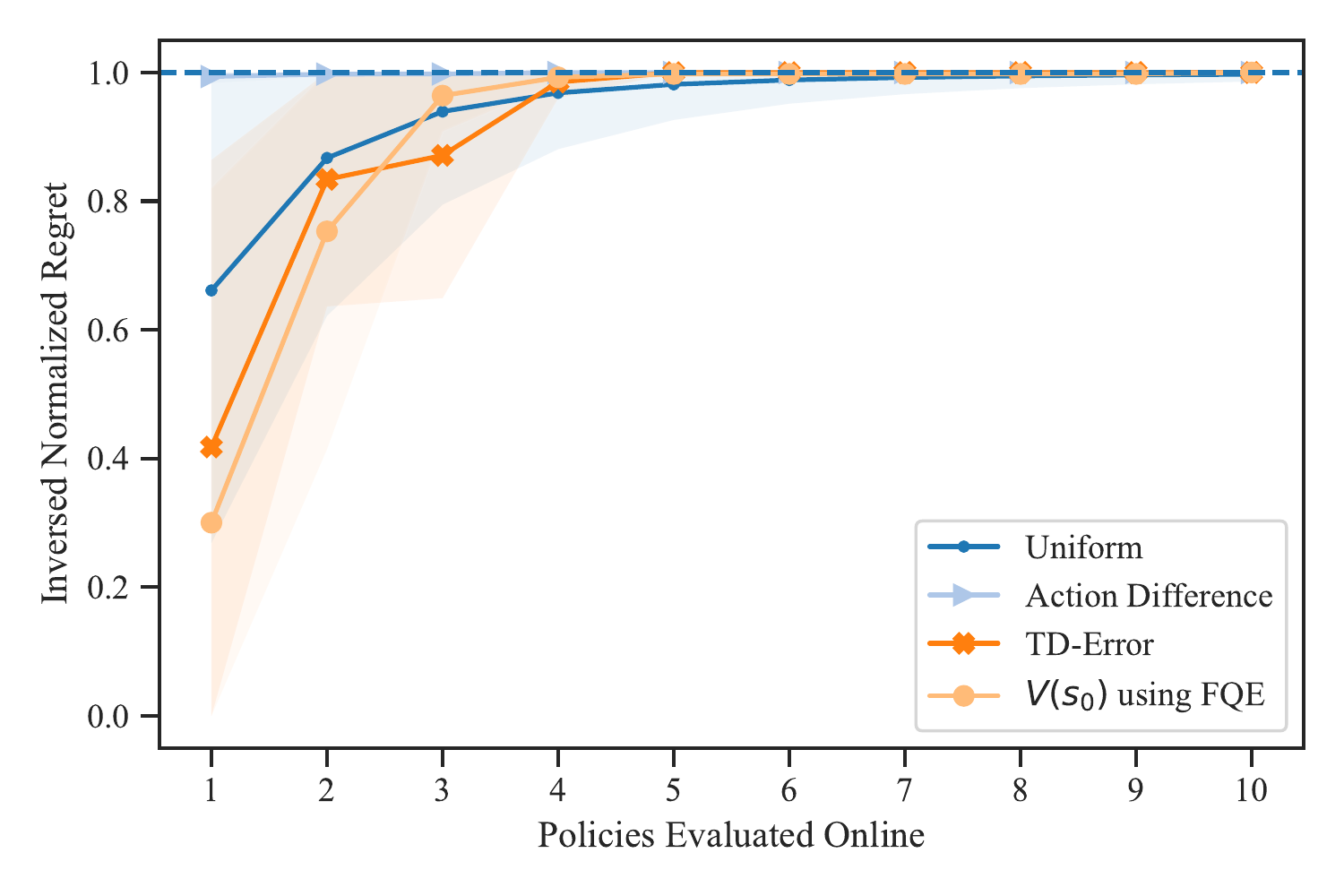}}
  \caption{High-Level Policy, 999 }
\end{subfigure}
\begin{subfigure}[b]{.32\textwidth}
 \centering
  \centerline{\includegraphics[width=\textwidth]{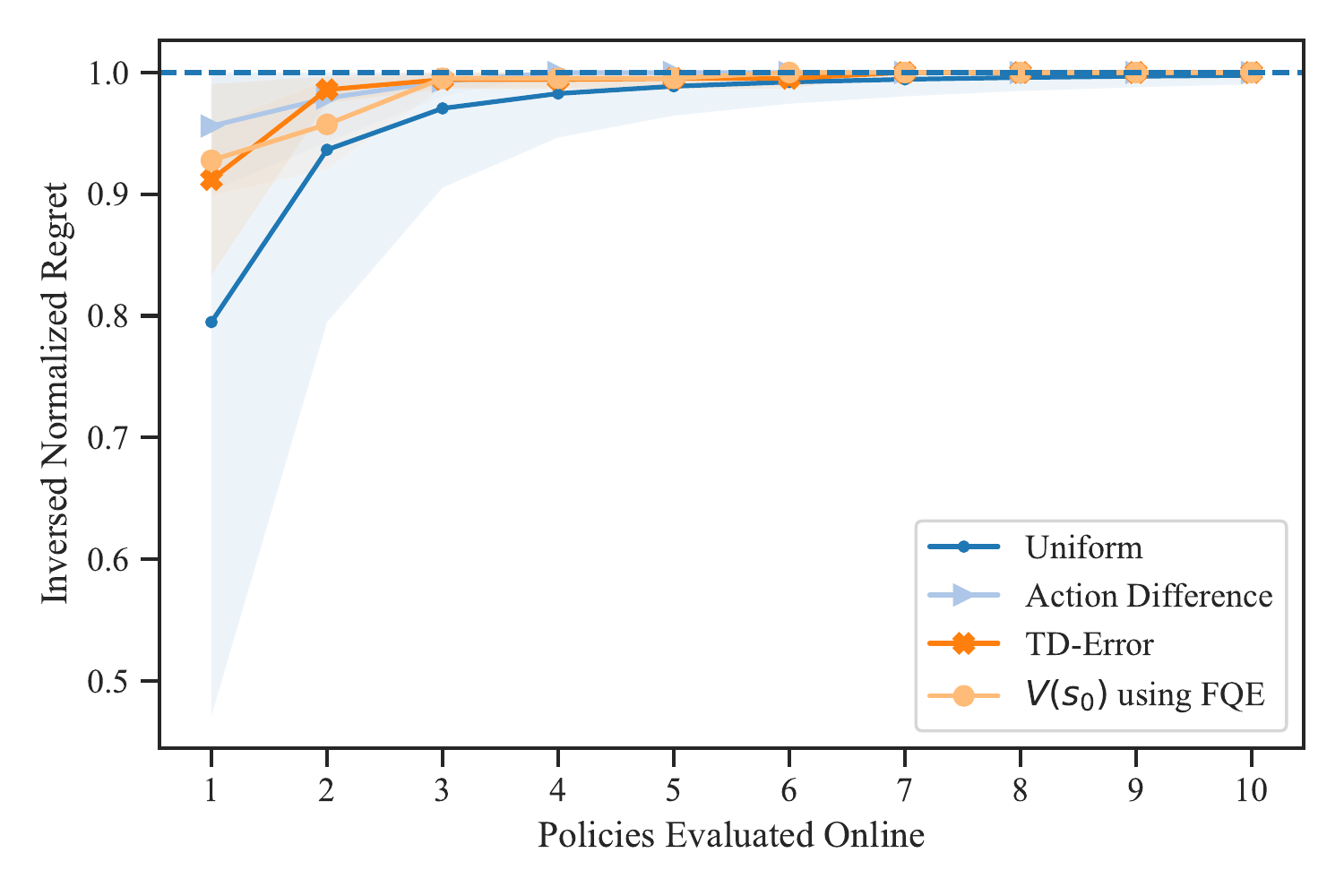}}
  \caption{High-Level Policy, 99 }
\end{subfigure}

\begin{subfigure}[b]{.32\textwidth}
 \centering
  \centerline{\includegraphics[width=\textwidth]{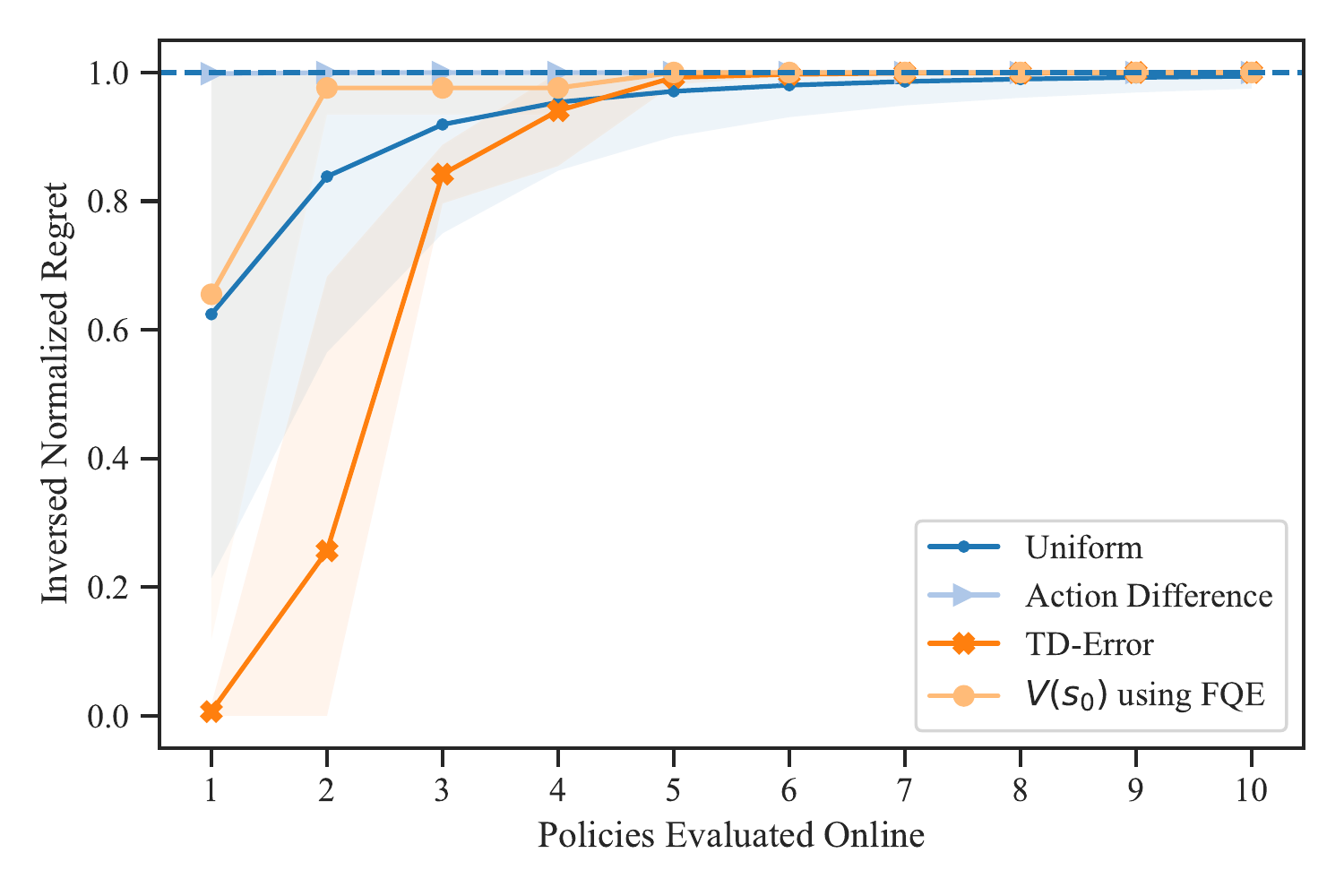}}
  \caption{Medium-Level Policy, 9999 }
\end{subfigure}
\begin{subfigure}[b]{.32\textwidth}
 \centering
  \centerline{\includegraphics[width=\textwidth]{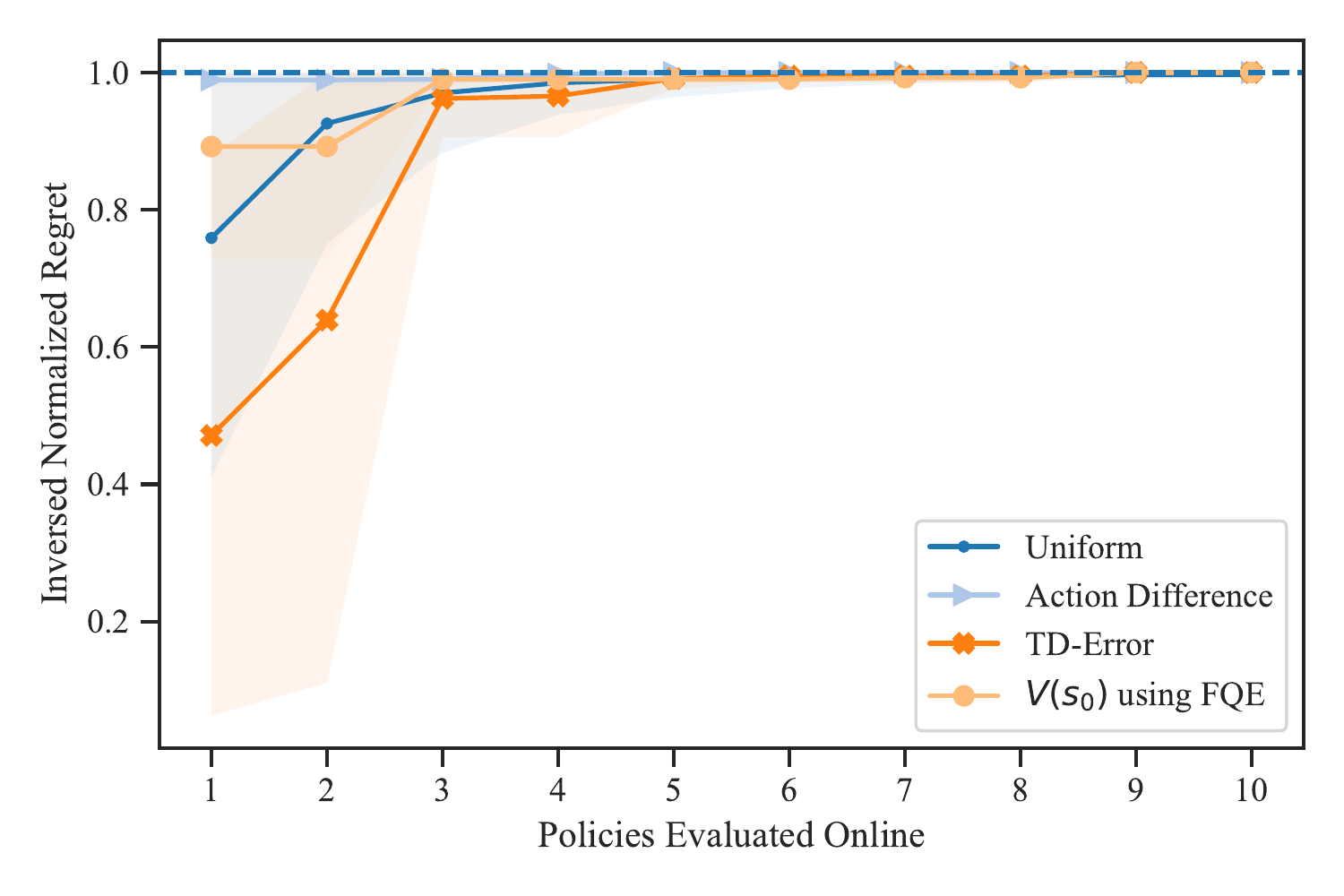}}
  \caption{Medium-Level Policy, 999 }
\end{subfigure}
\begin{subfigure}[b]{.32\textwidth}
 \centering
  \centerline{\includegraphics[width=\textwidth]{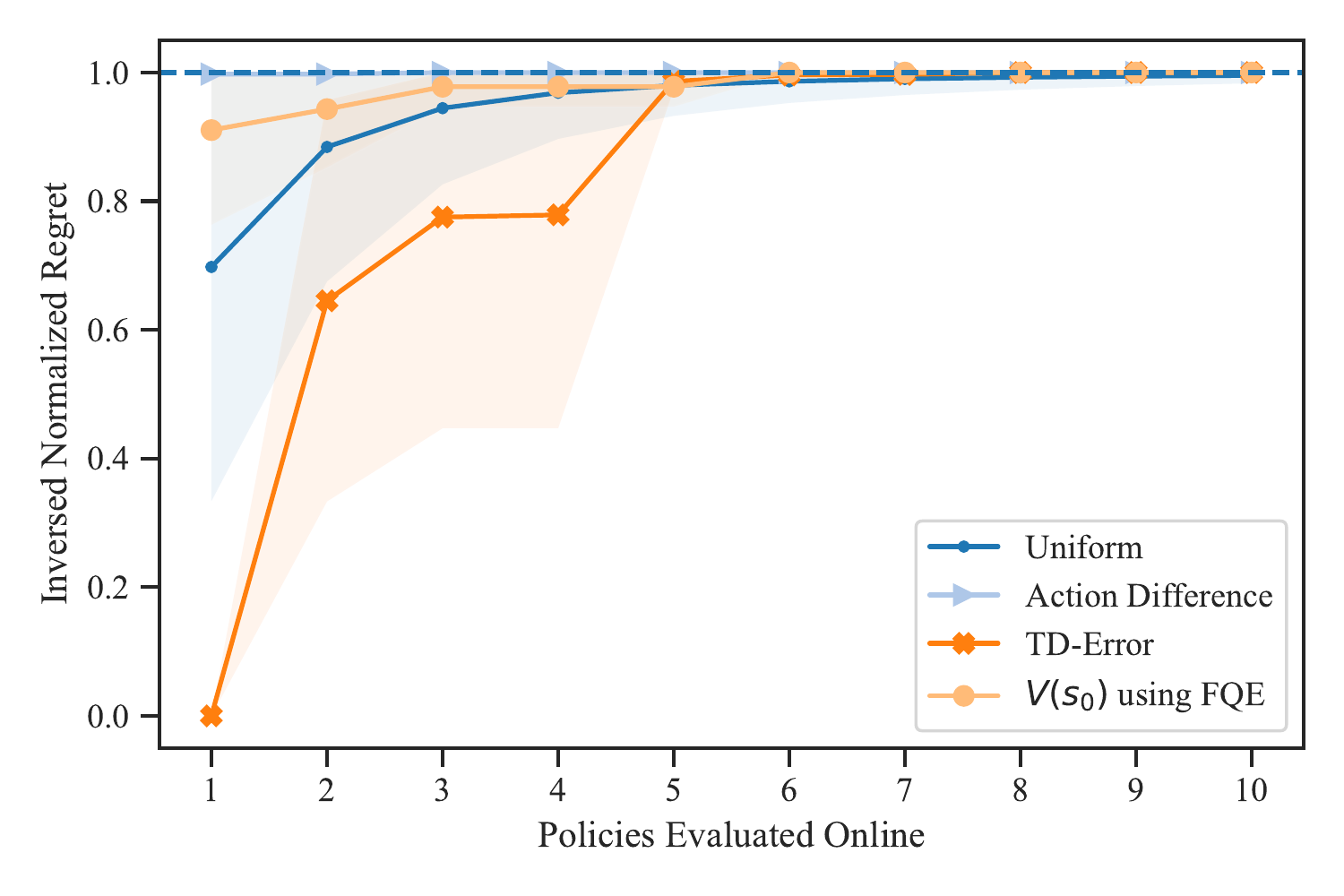}}
  \caption{Medium-Level Policy, 99 }
\end{subfigure}

\begin{subfigure}[b]{.32\textwidth}
 \centering
  \centerline{\includegraphics[width=\textwidth]{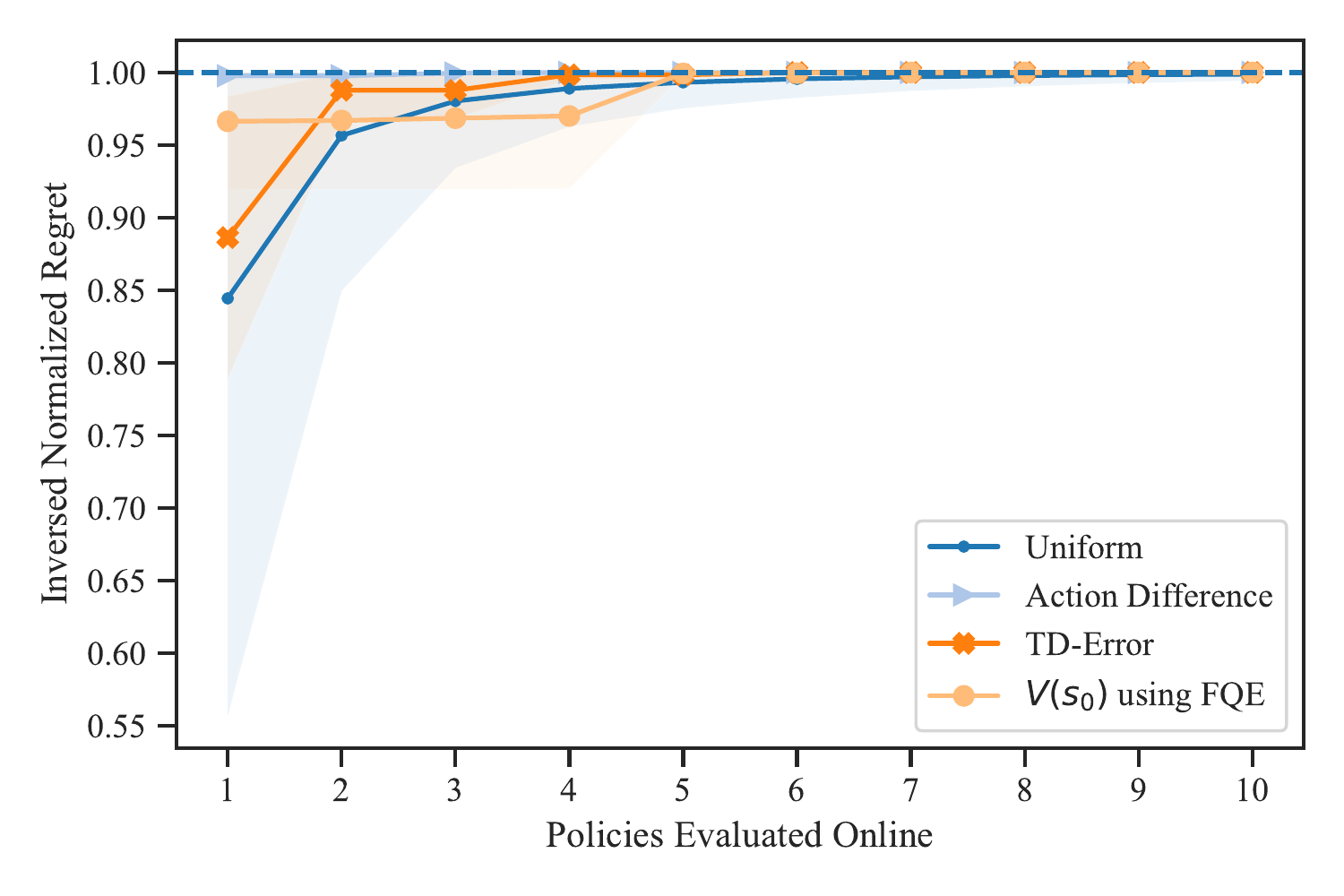}}
  \caption{Low-Level Policy, 9999 }
\end{subfigure}
\begin{subfigure}[b]{.32\textwidth}
 \centering
  \centerline{\includegraphics[width=\textwidth]{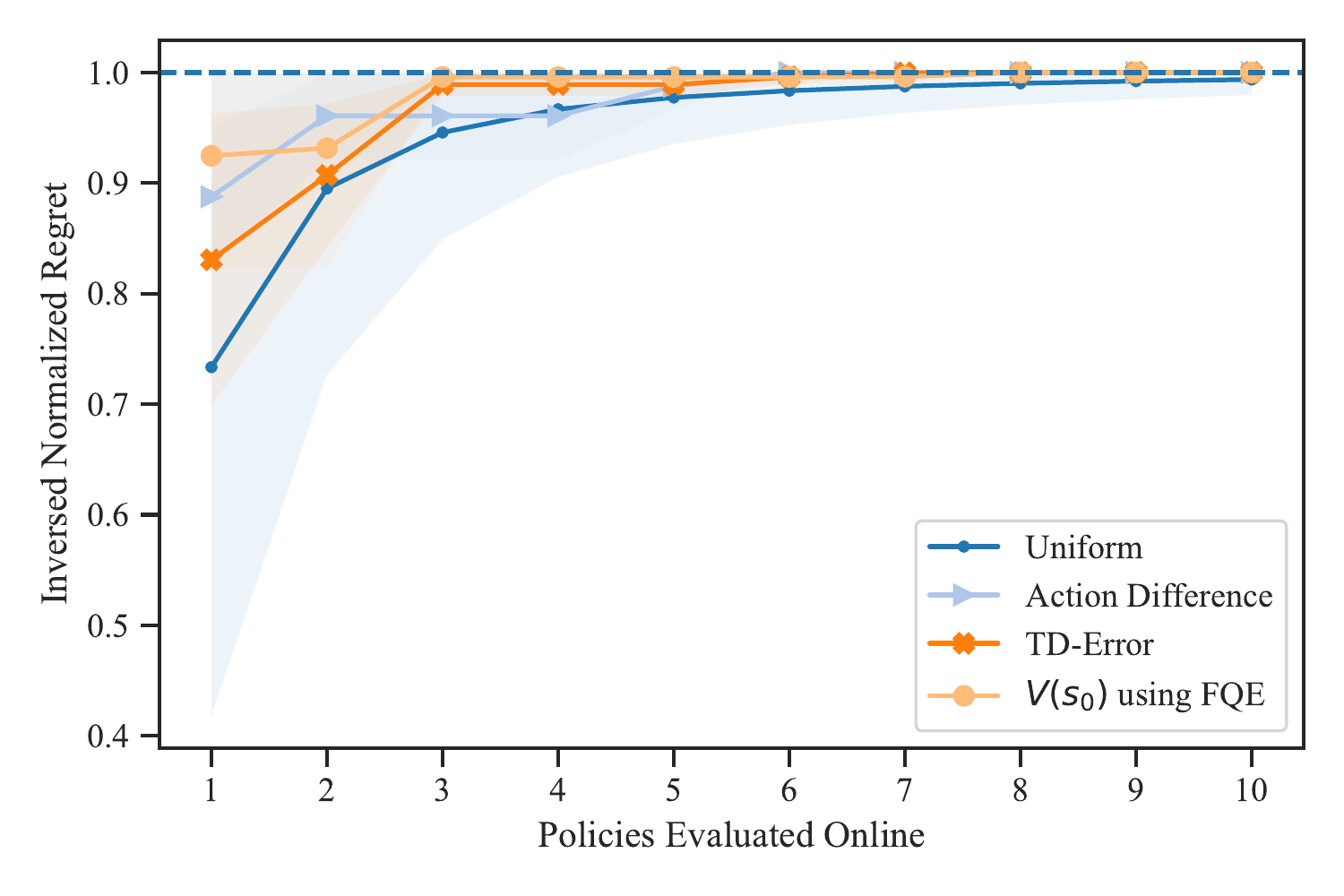}}
  \caption{Low-Level Policy, 999 }
\end{subfigure}
\begin{subfigure}[b]{.32\textwidth}
 \centering
  \centerline{\includegraphics[width=\textwidth]{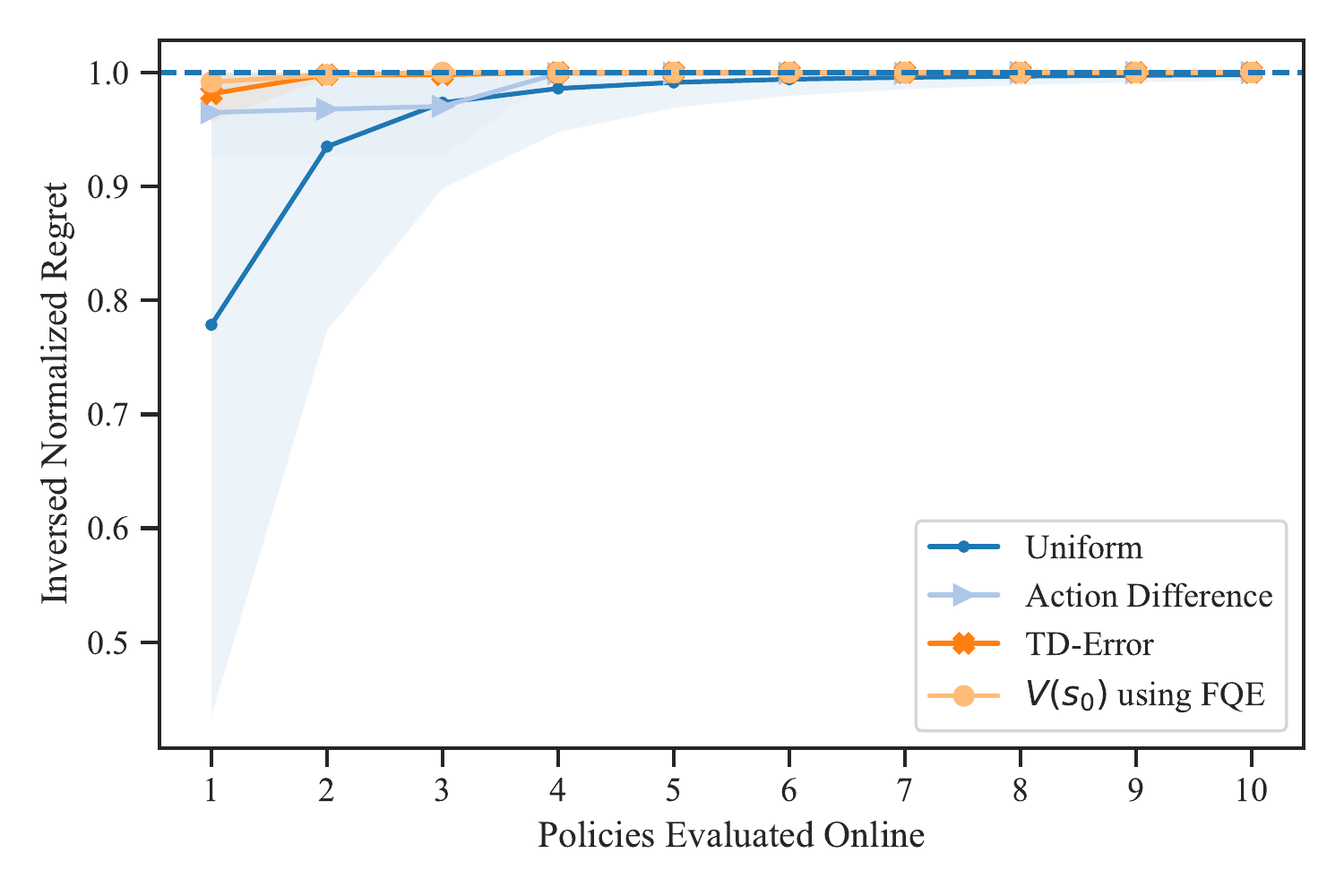}}
  \caption{Low-Level Policy, 99 }
\end{subfigure}

\caption{\textbf{CQL, Industrial Benchmark. Inversed Normalized Regret under different offline policy selection methods using EOP graph.} The shaded area represents one standard deviation }
\label{fig:appendix:ops_cql_industrial}
\end{figure*}

\begin{figure*}[h]
 \centering
\begin{subfigure}[b]{.32\textwidth}
 \centering
  \centerline{\includegraphics[width=\textwidth]{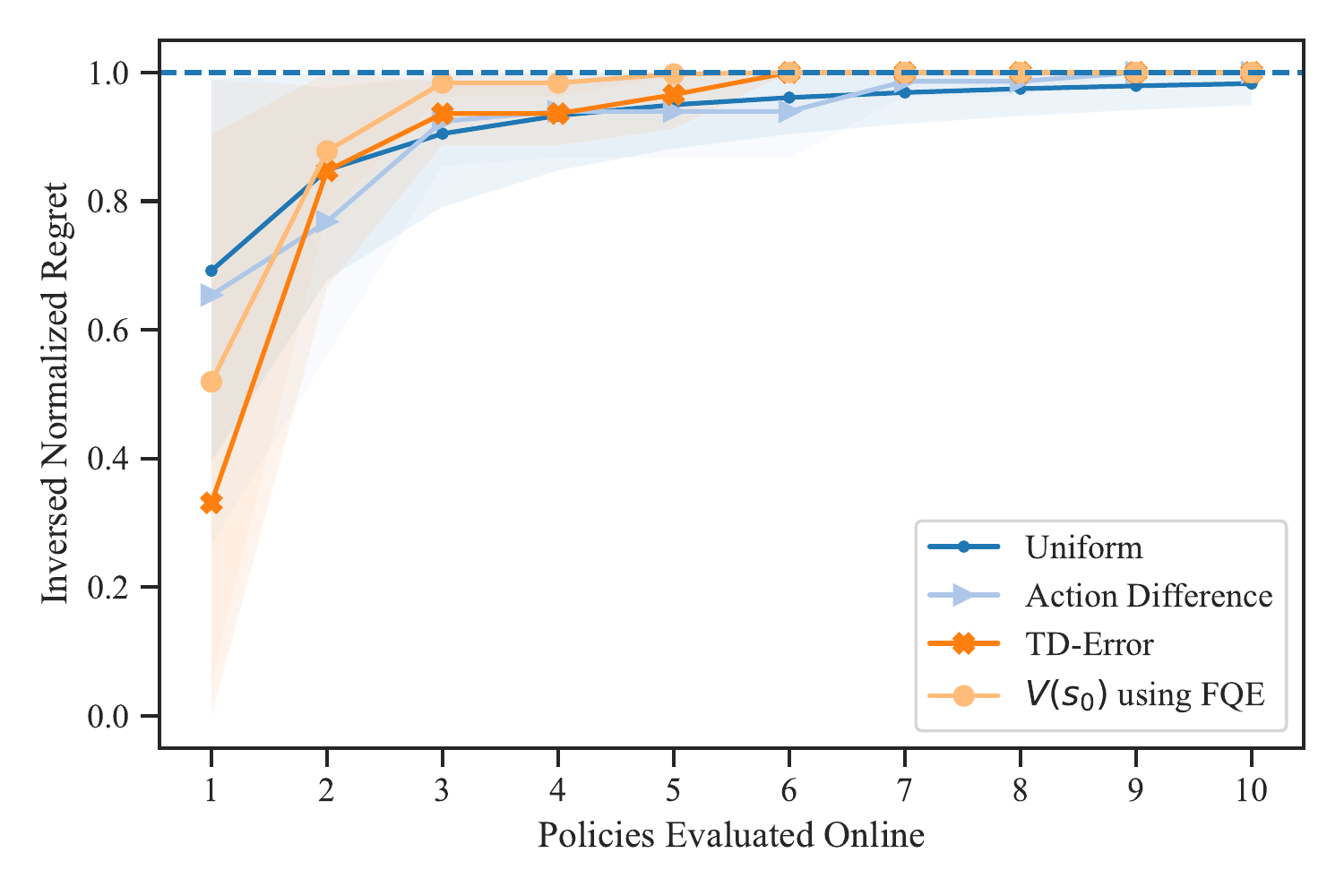}}
  \caption{High-Level Policy, 999 }
\end{subfigure}
\begin{subfigure}[b]{.32\textwidth}
 \centering
  \centerline{\includegraphics[width=\textwidth]{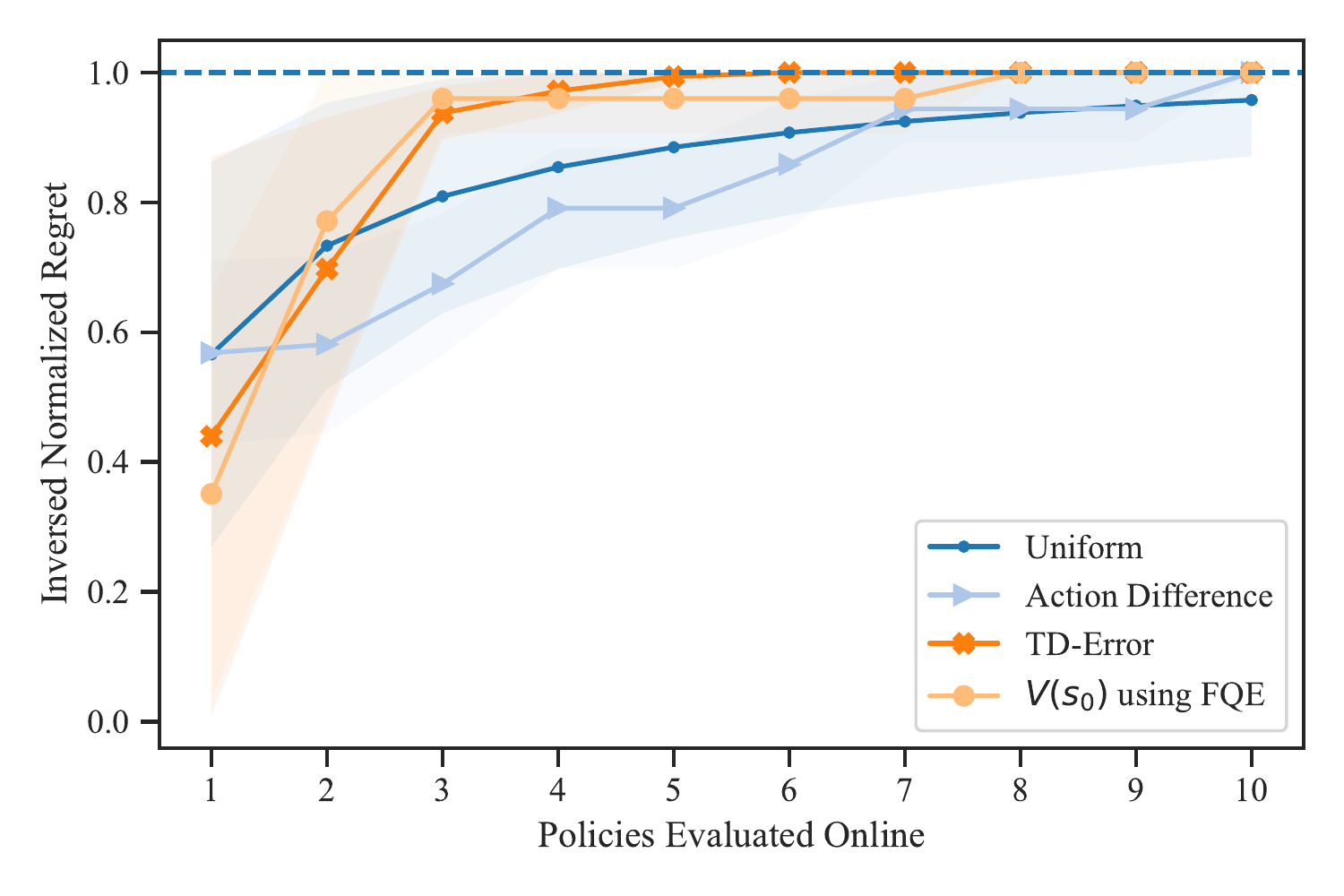}}
  \caption{Medium-Level Policy, 999 }
\end{subfigure}
\begin{subfigure}[b]{.32\textwidth}
 \centering
  \centerline{\includegraphics[width=\textwidth]{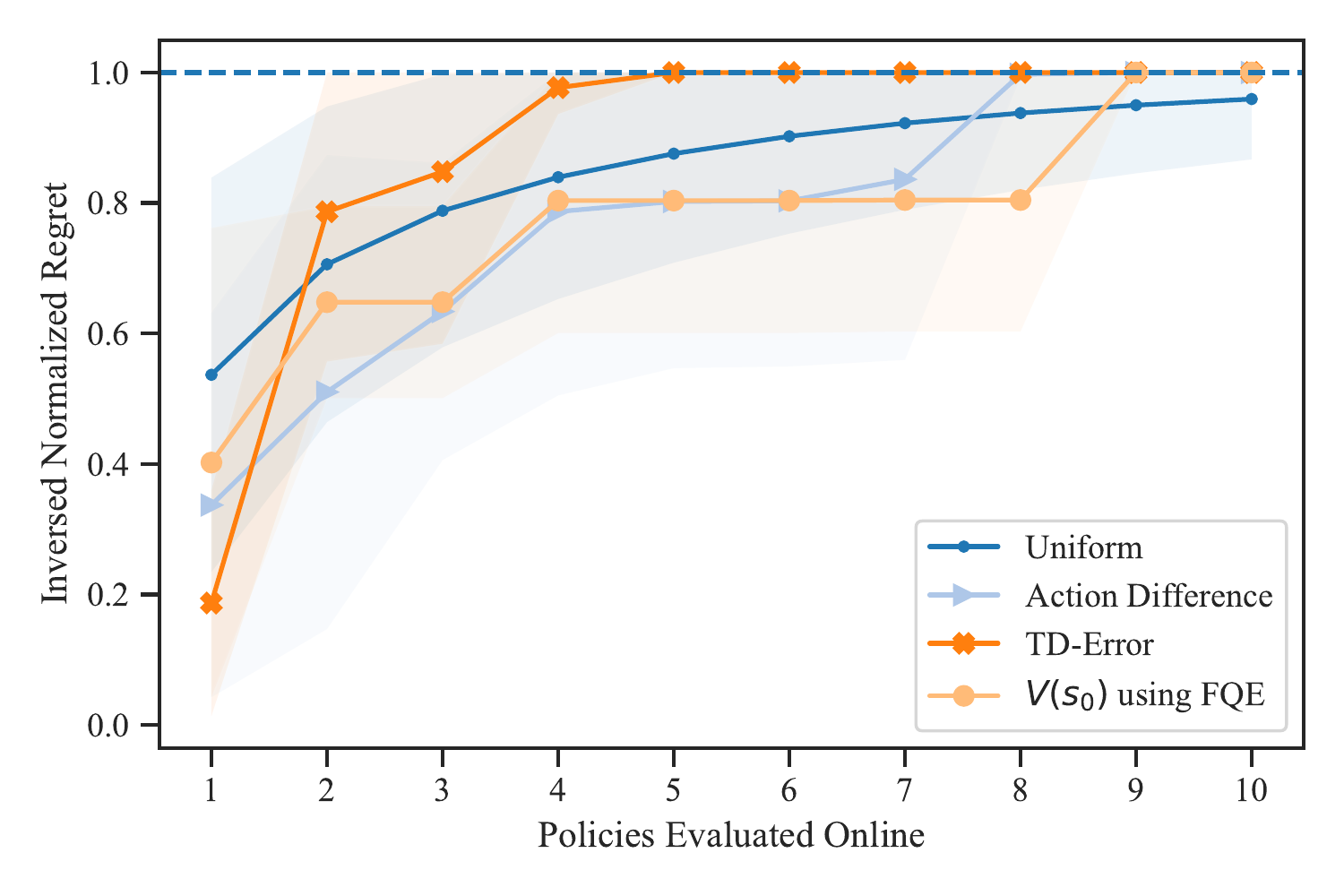}}
  \caption{Low-Level Policy, 999 }
\end{subfigure}

\begin{subfigure}[b]{.32\textwidth}
 \centering
  \centerline{\includegraphics[width=\textwidth]{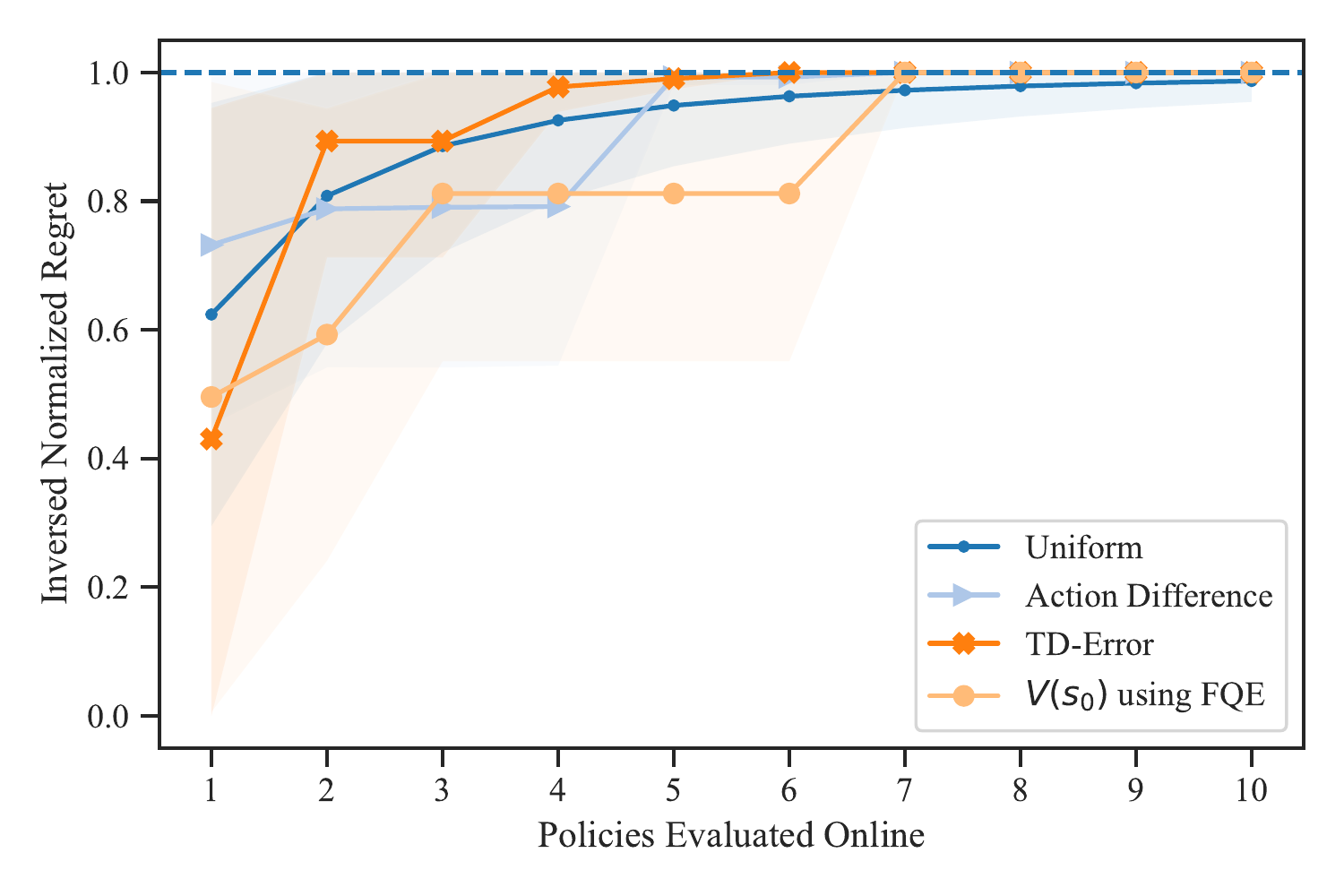}}
  \caption{High-Level Policy, 99 }
\end{subfigure}
\begin{subfigure}[b]{.32\textwidth}
 \centering
  \centerline{\includegraphics[width=\textwidth]{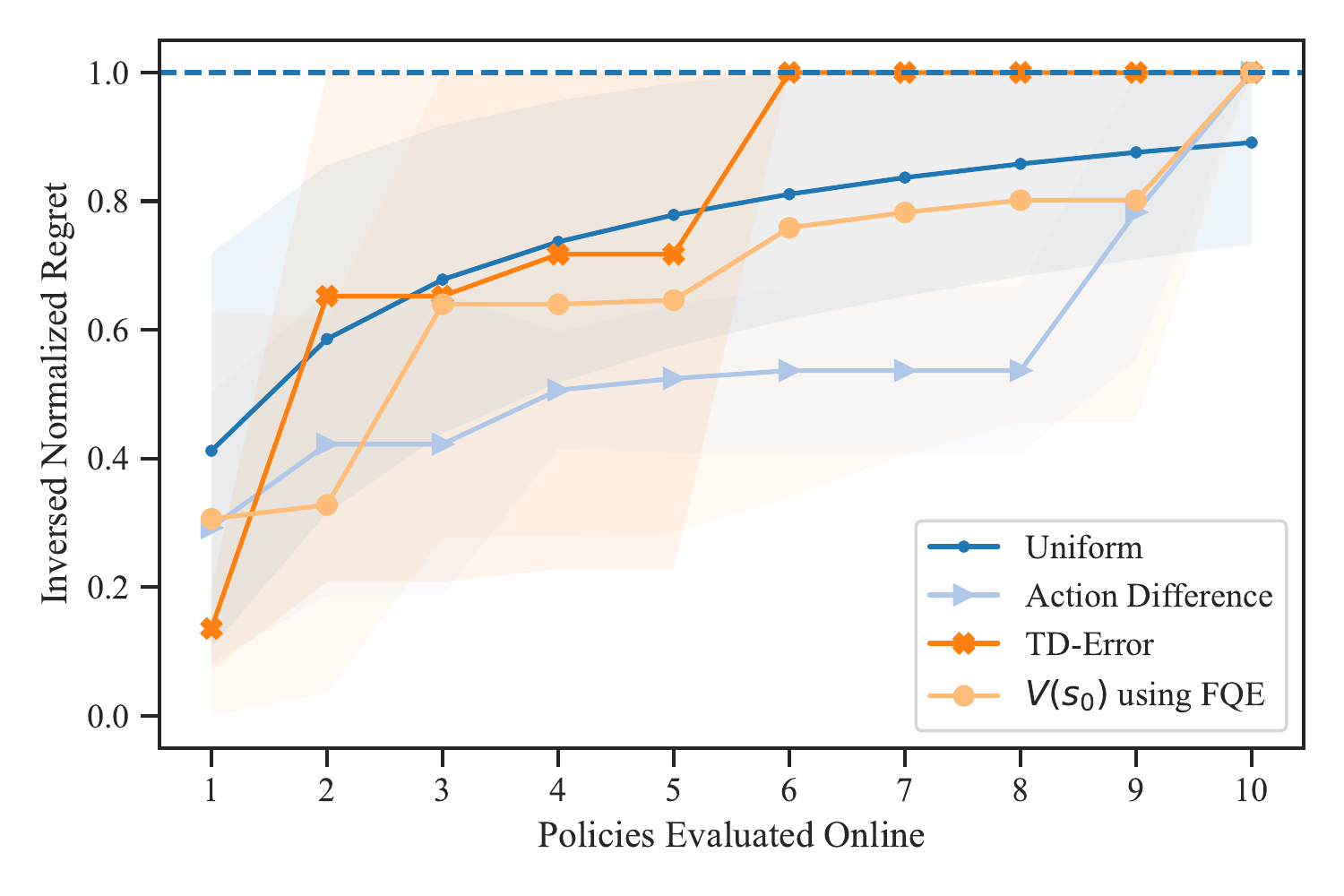}}
  \caption{Medium-Level Policy, 99 }
\end{subfigure}
\begin{subfigure}[b]{.32\textwidth}
 \centering
  \centerline{\includegraphics[width=\textwidth]{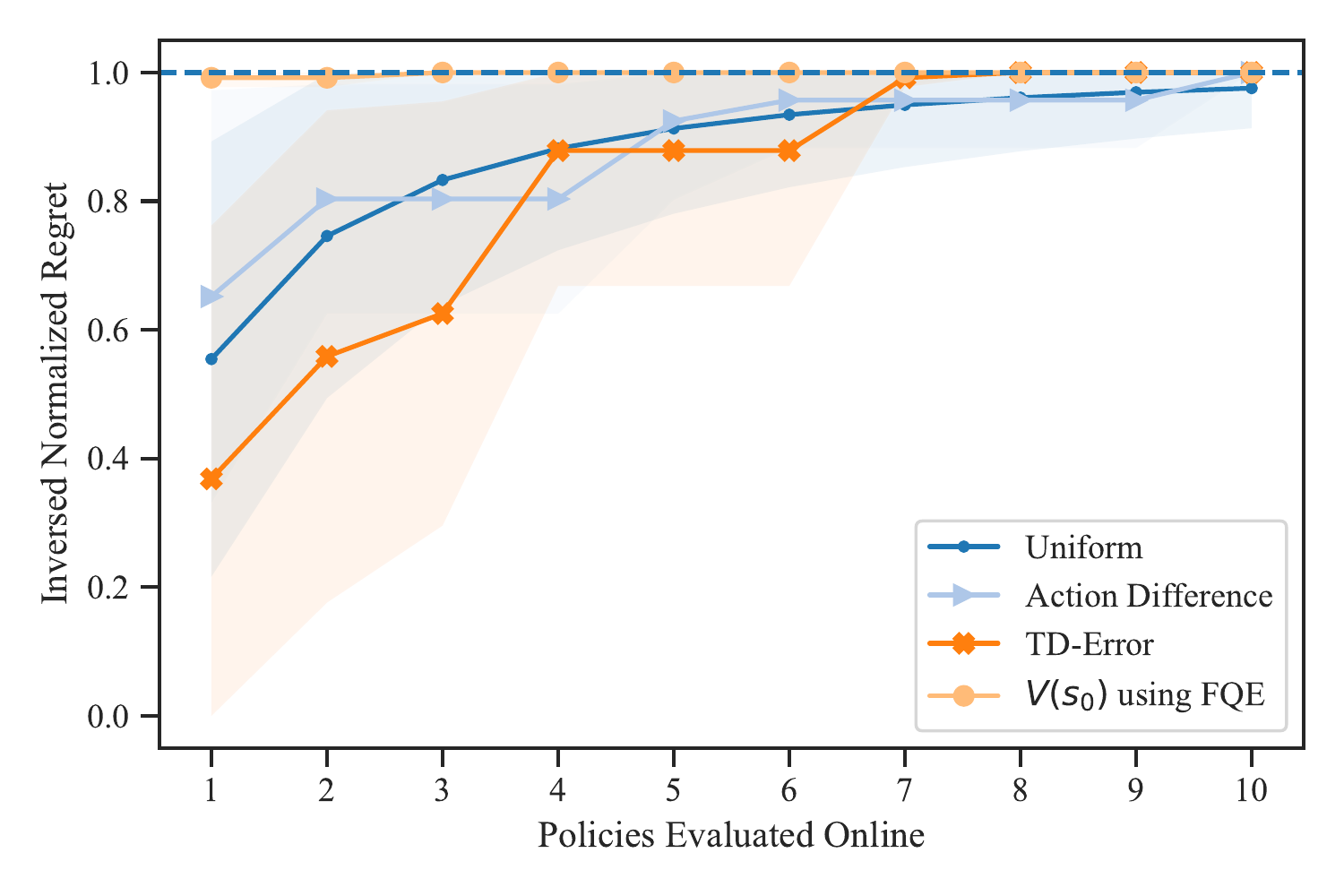}}
  \caption{Low-Level Policy, 99 }
\end{subfigure}

\caption{\textbf{TD3+BC, FinRL. Inversed Normalized Regret under different offline policy selection methods using EOP graph.} The shaded area represents one standard deviation }
\label{fig:appendix:ops_td3bc_finrl}
\end{figure*}

\begin{figure*}[h]
 \centering
\begin{subfigure}[b]{.32\textwidth}
 \centering
  \centerline{\includegraphics[width=\textwidth]{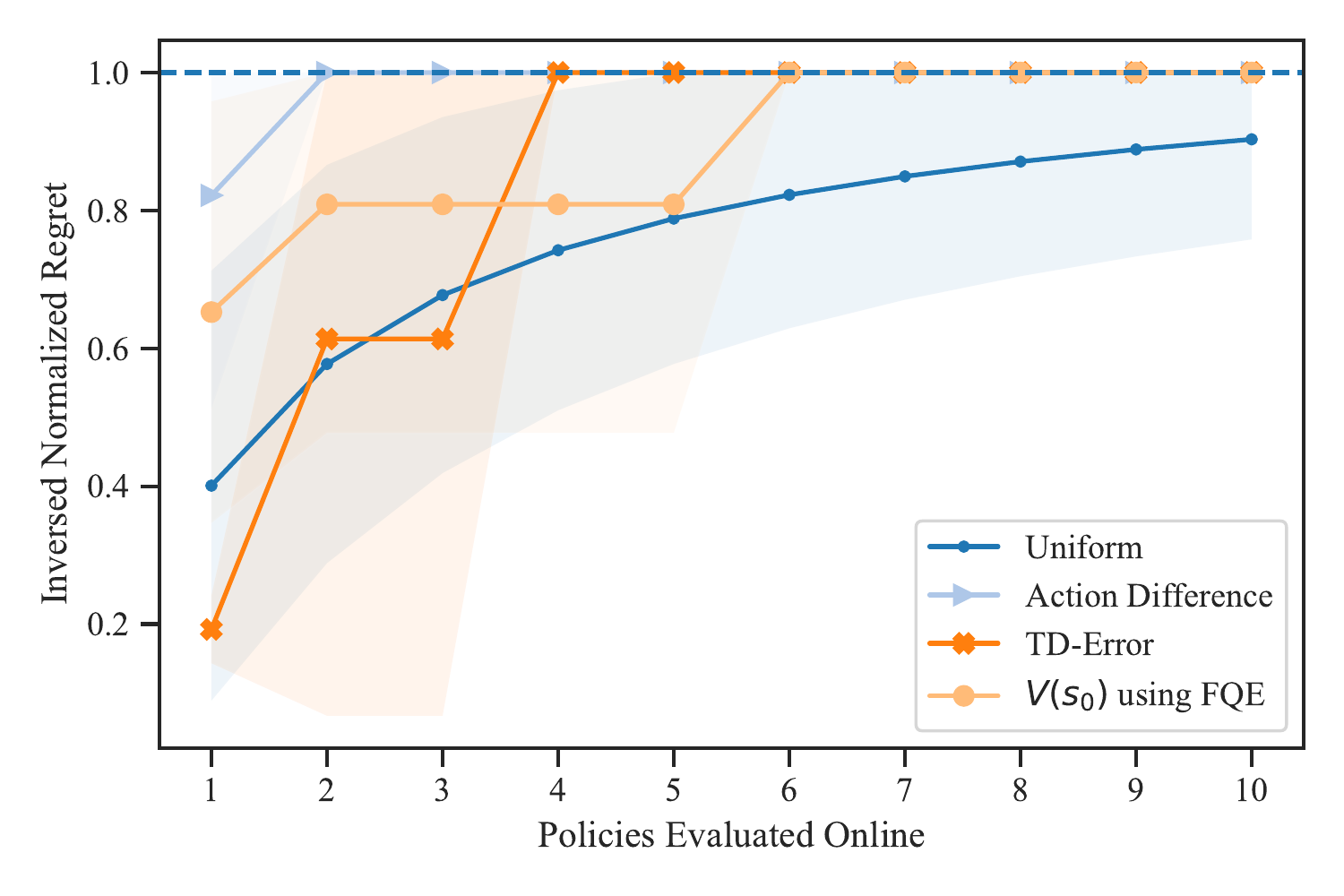}}
  \caption{Medium-Level Policy, 9999 }
\end{subfigure}
\begin{subfigure}[b]{.32\textwidth}
 \centering
  \centerline{\includegraphics[width=\textwidth]{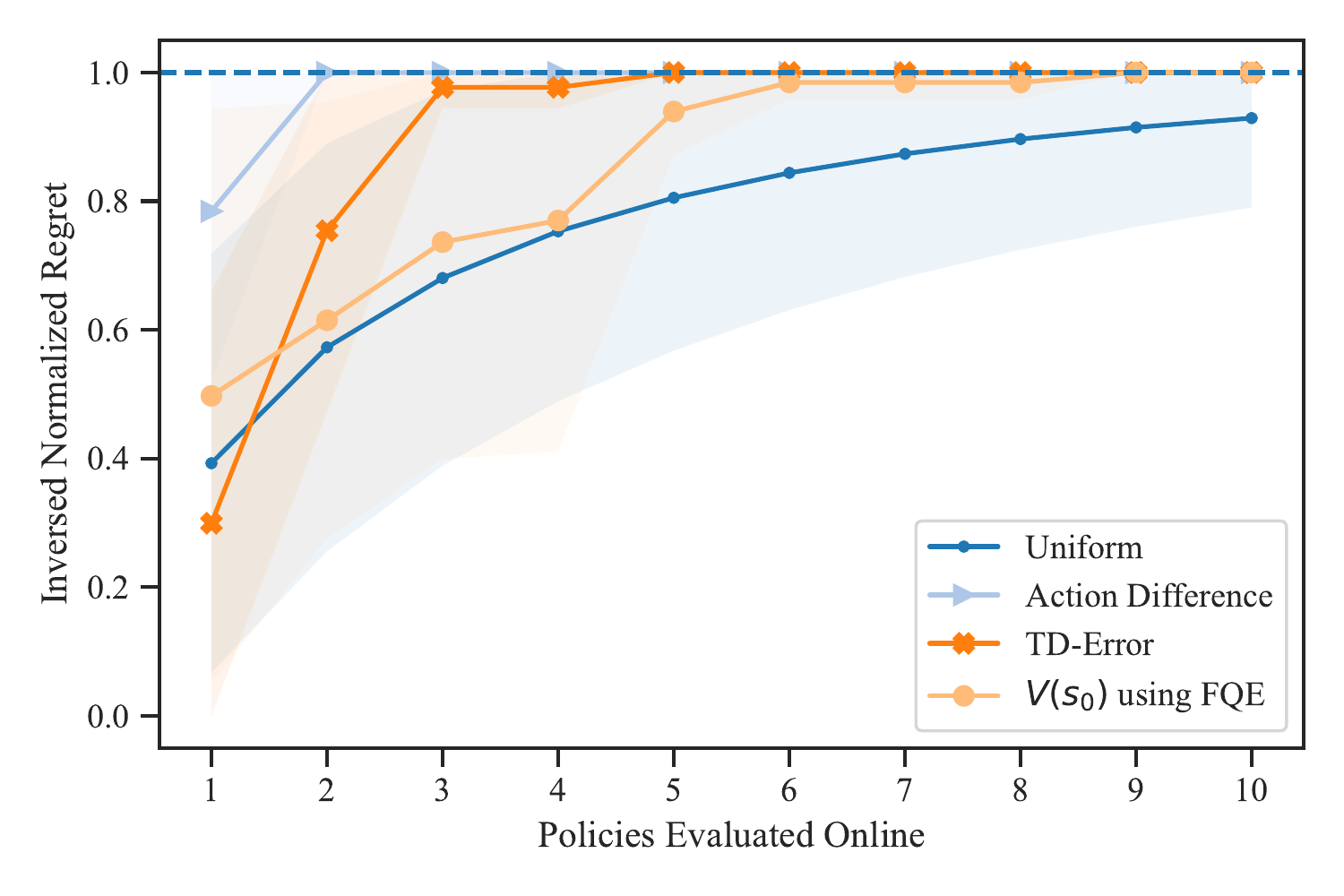}}
  \caption{Medium-Level Policy, 999 }
\end{subfigure}
\begin{subfigure}[b]{.32\textwidth}
 \centering
  \centerline{\includegraphics[width=\textwidth]{figures_appendix/ops_td3+bc_citylearn_999_medium.pdf}}
  \caption{Medium-Level Policy, 99 }
\end{subfigure}

\begin{subfigure}[b]{.32\textwidth}
 \centering
  \centerline{\includegraphics[width=\textwidth]{figures_appendix/ops_td3+bc_citylearn_9999_low.pdf}}
  \caption{Low-Level Policy, 9999 }
\end{subfigure}
\begin{subfigure}[b]{.32\textwidth}
 \centering
  \centerline{\includegraphics[width=\textwidth]{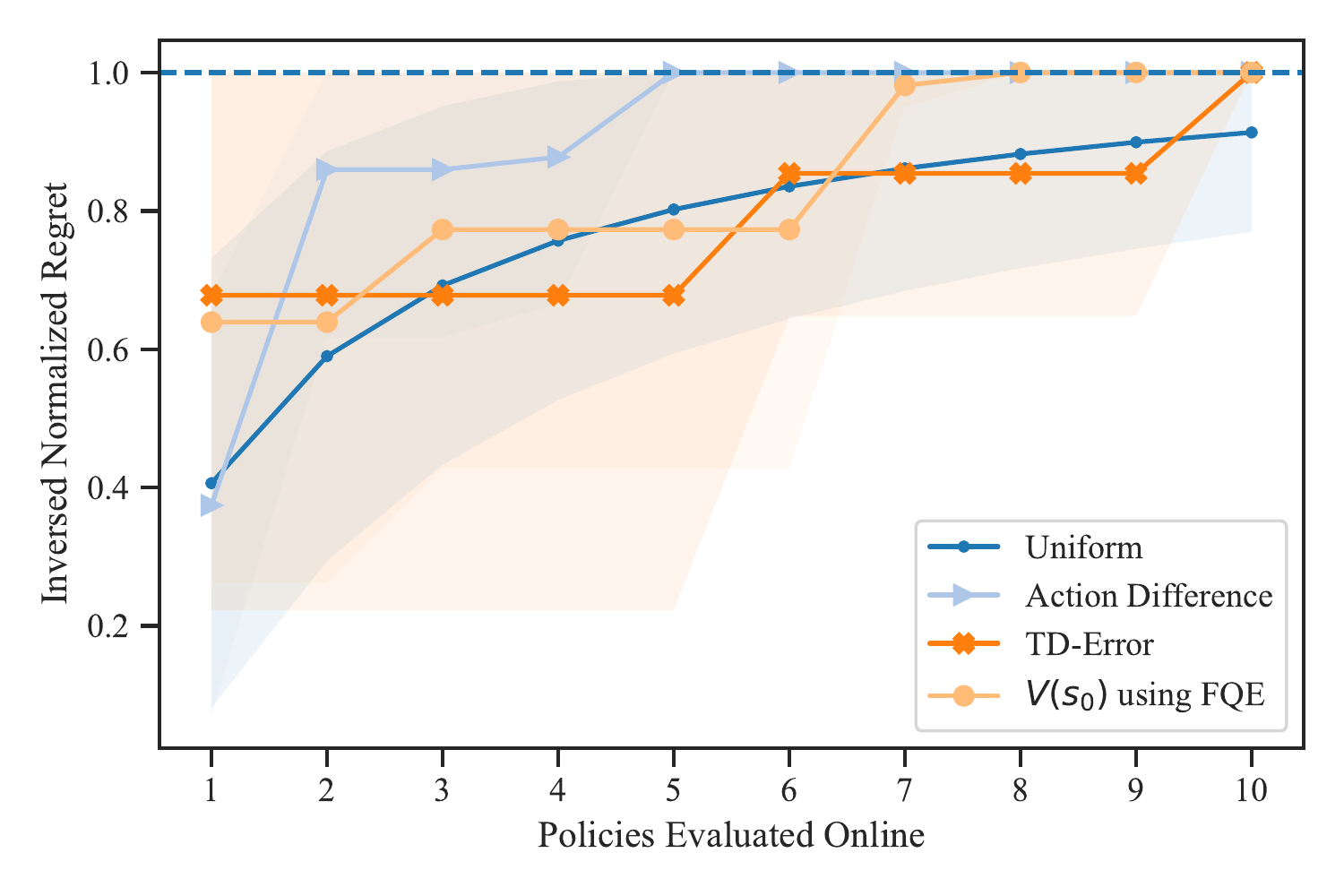}}
  \caption{Low-Level Policy, 999 }
\end{subfigure}
\begin{subfigure}[b]{.32\textwidth}
 \centering
  \centerline{\includegraphics[width=\textwidth]{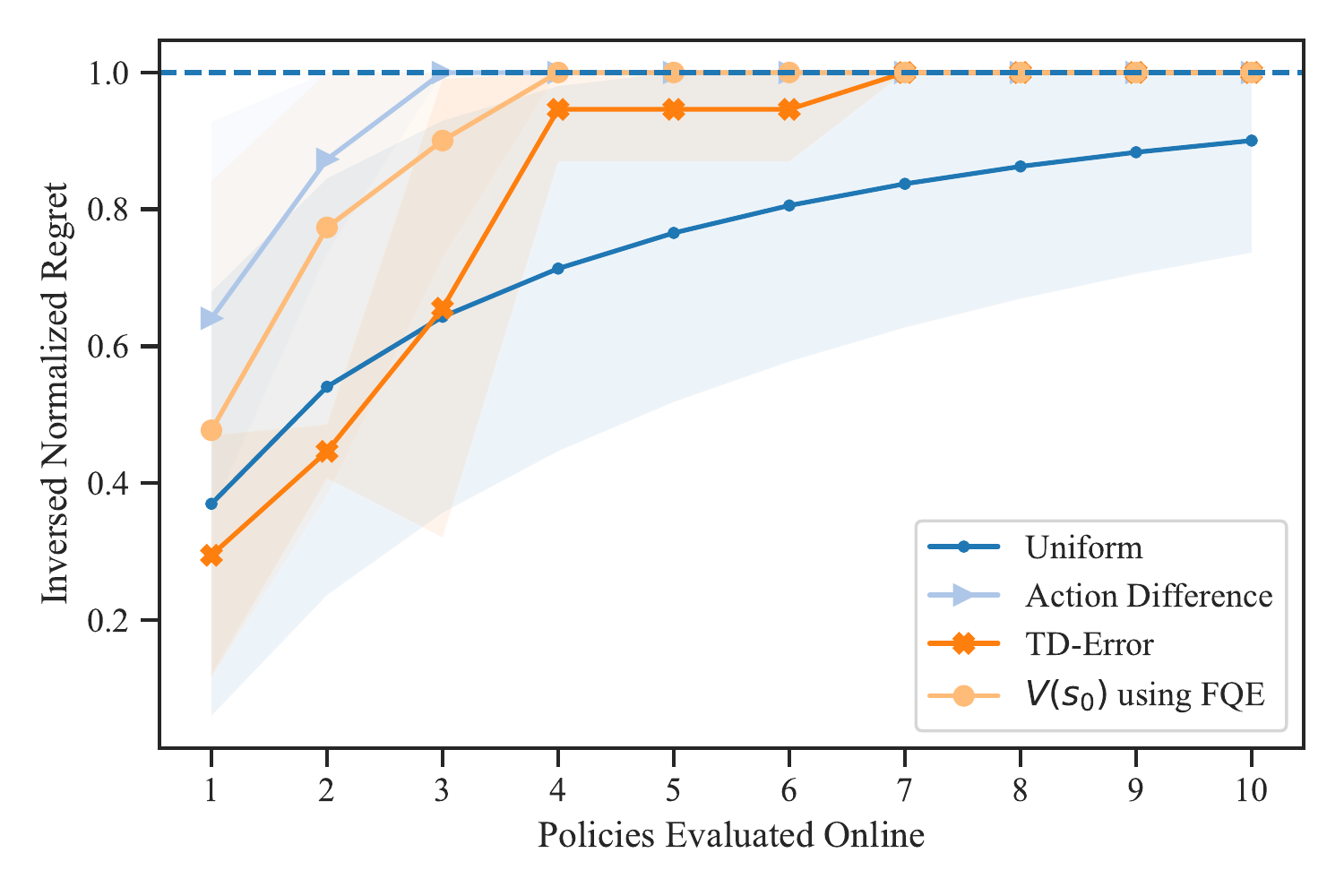}}
  \caption{Low-Level Policy, 99 }
\end{subfigure}

\caption{\textbf{TD3+BC, CityLearn. Inversed Normalized Regret under different offline policy selection methods using EOP graph.} The shaded area represents one standard deviation }
\label{fig:appendix:ops_td3bc_citylearn}
\end{figure*}

\begin{figure*}[h]
 \centering
\begin{subfigure}[b]{.32\textwidth}
 \centering
  \centerline{\includegraphics[width=\textwidth]{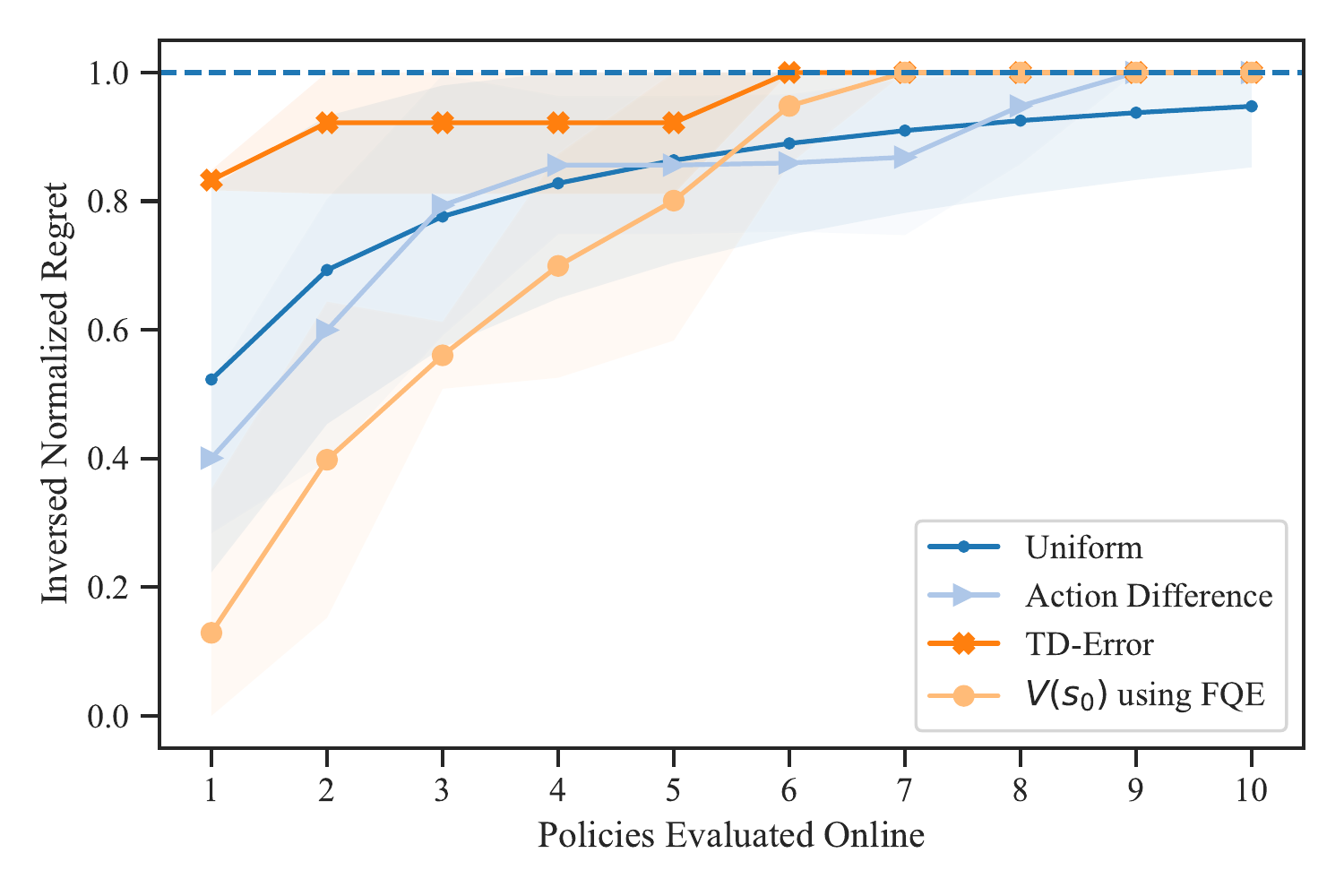}}
  \caption{High-Level Policy, 9999 }
\end{subfigure}
\begin{subfigure}[b]{.32\textwidth}
 \centering
  \centerline{\includegraphics[width=\textwidth]{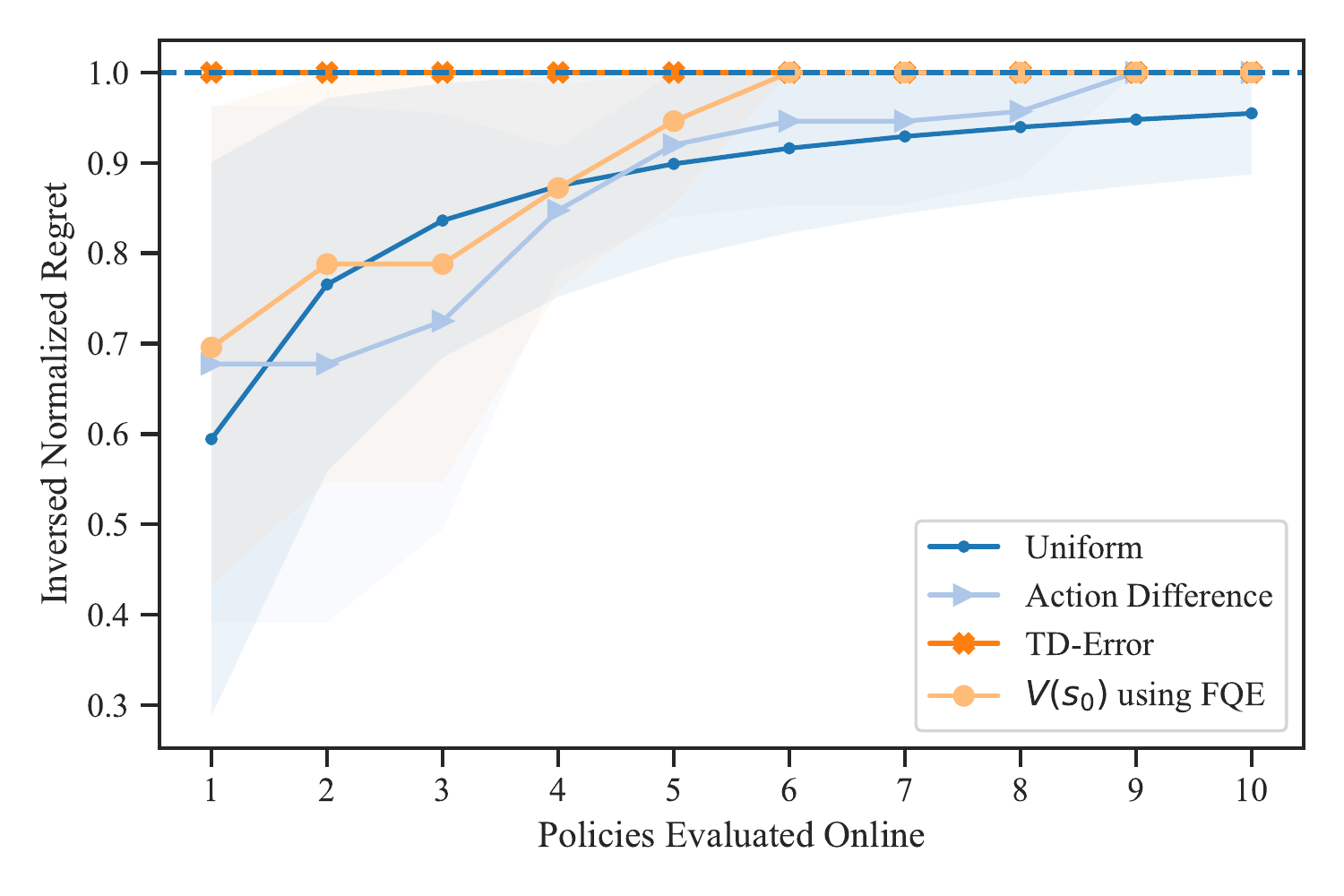}}
  \caption{High-Level Policy, 999 }
\end{subfigure}
\begin{subfigure}[b]{.32\textwidth}
 \centering
  \centerline{\includegraphics[width=\textwidth]{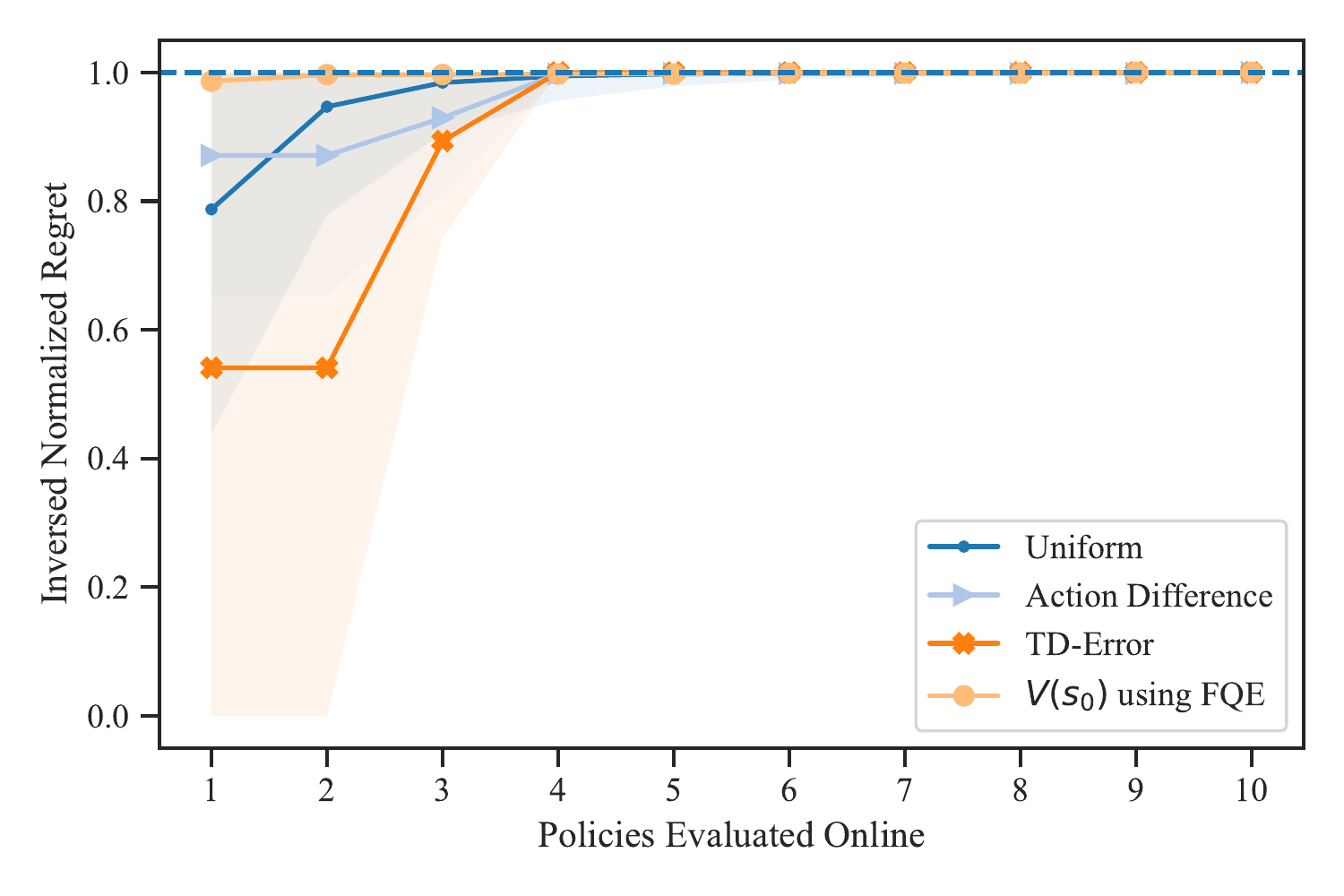}}
  \caption{High-Level Policy, 99 }
\end{subfigure}

\begin{subfigure}[b]{.32\textwidth}
 \centering
  \centerline{\includegraphics[width=\textwidth]{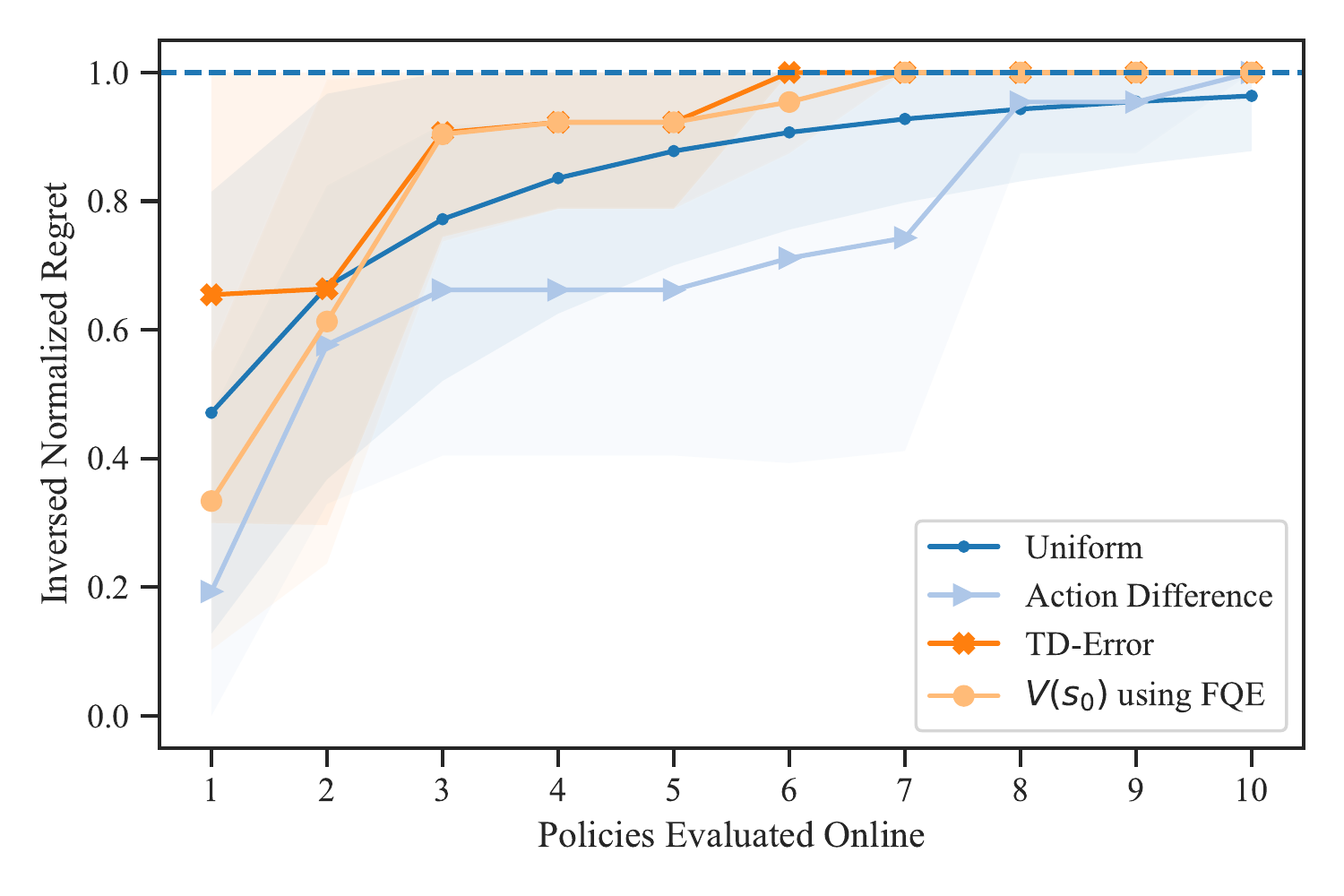}}
  \caption{Medium-Level Policy, 9999 }
\end{subfigure}
\begin{subfigure}[b]{.32\textwidth}
 \centering
  \centerline{\includegraphics[width=\textwidth]{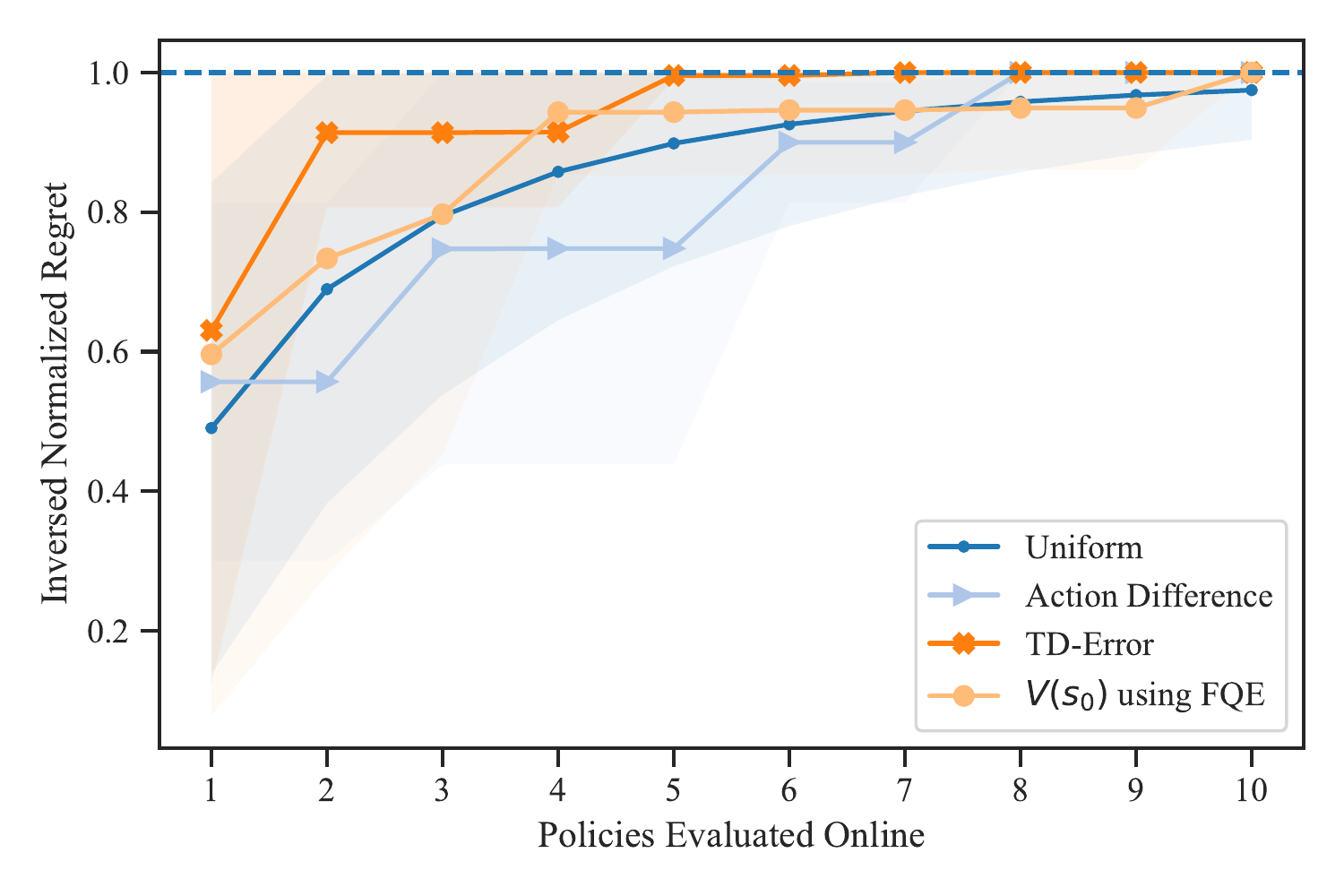}}
  \caption{Medium-Level Policy, 999 }
\end{subfigure}
\begin{subfigure}[b]{.32\textwidth}
 \centering
  \centerline{\includegraphics[width=\textwidth]{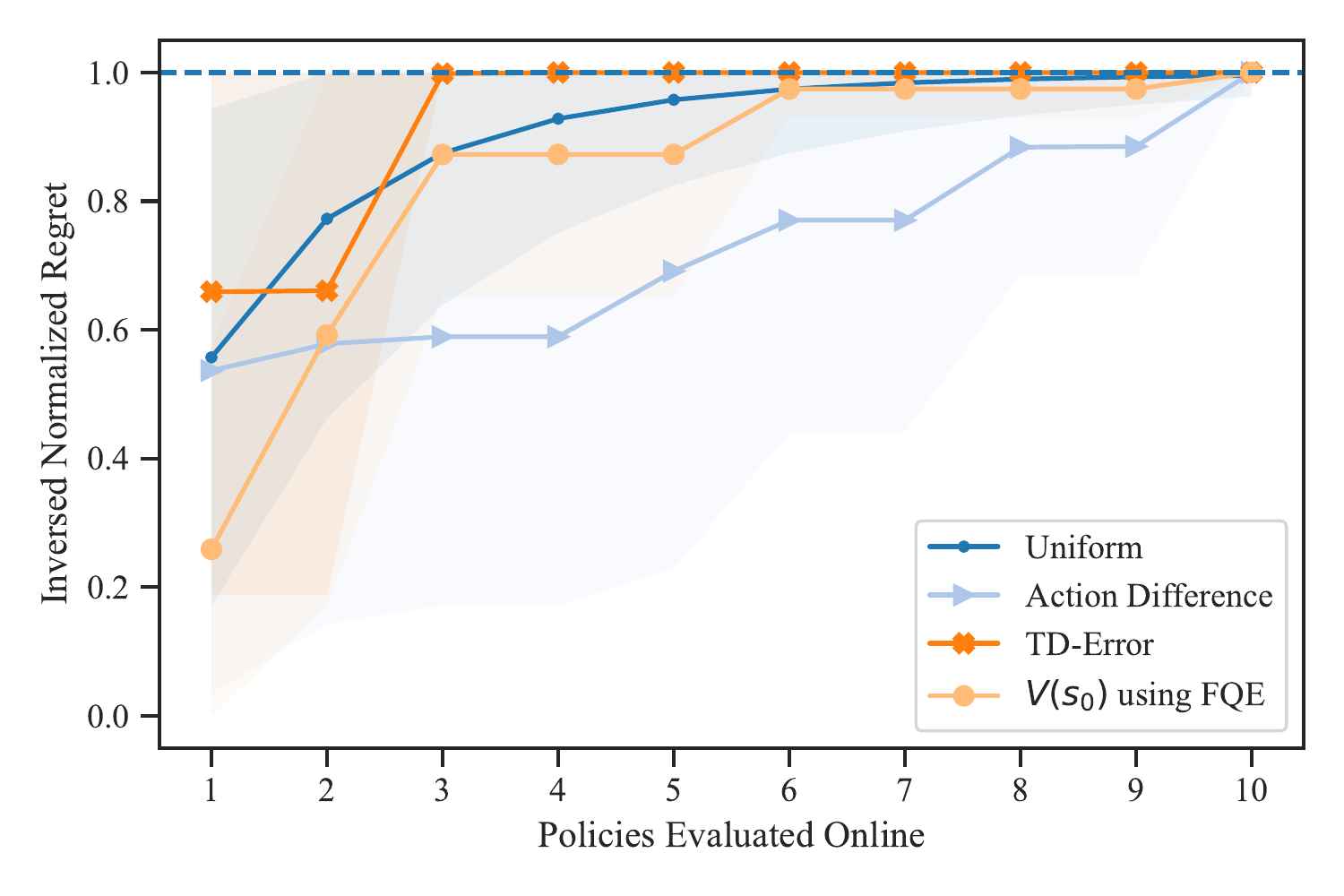}}
  \caption{Medium-Level Policy, 99 }
\end{subfigure}

\begin{subfigure}[b]{.32\textwidth}
 \centering
  \centerline{\includegraphics[width=\textwidth]{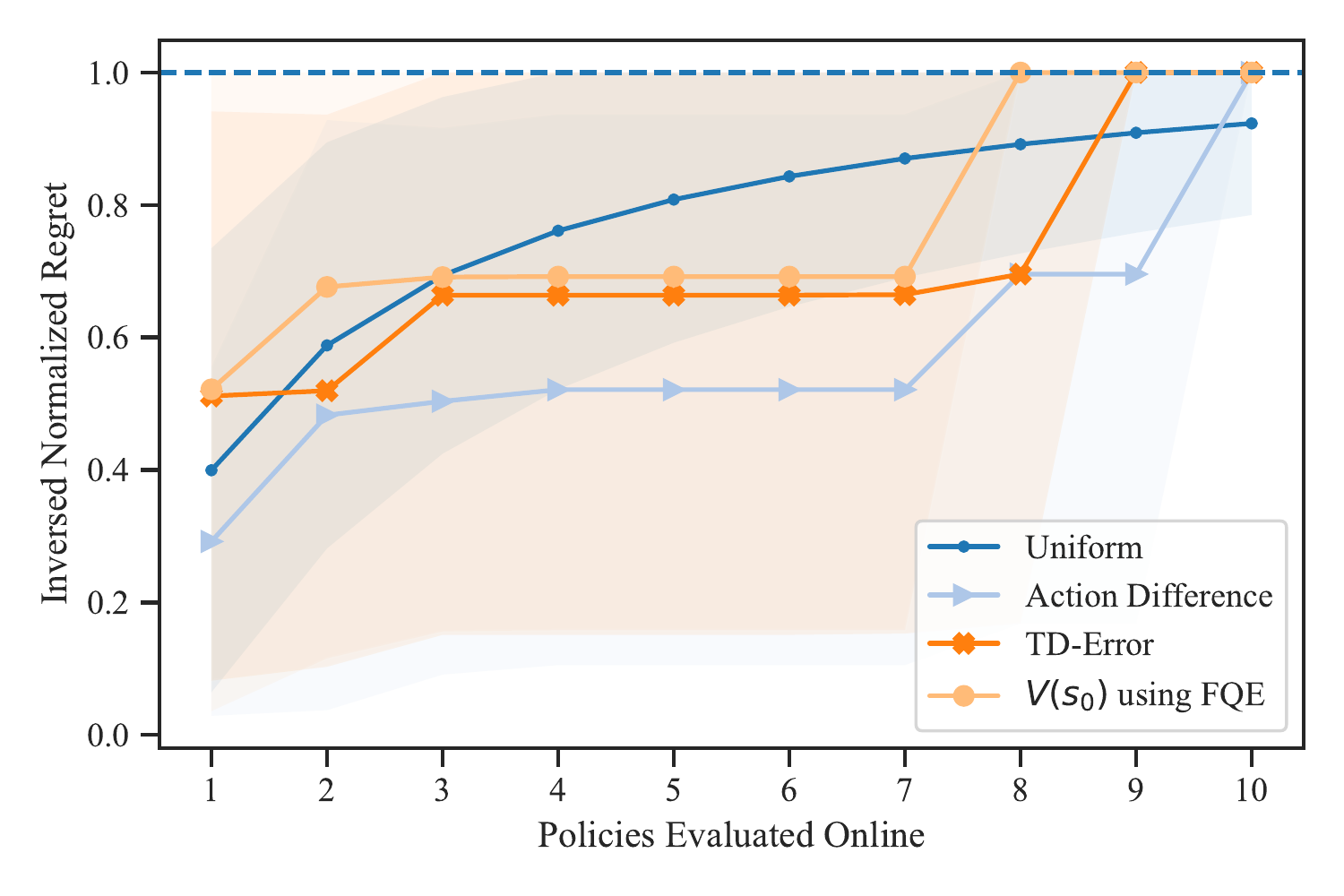}}
  \caption{Low-Level Policy, 9999 }
\end{subfigure}
\begin{subfigure}[b]{.32\textwidth}
 \centering
  \centerline{\includegraphics[width=\textwidth]{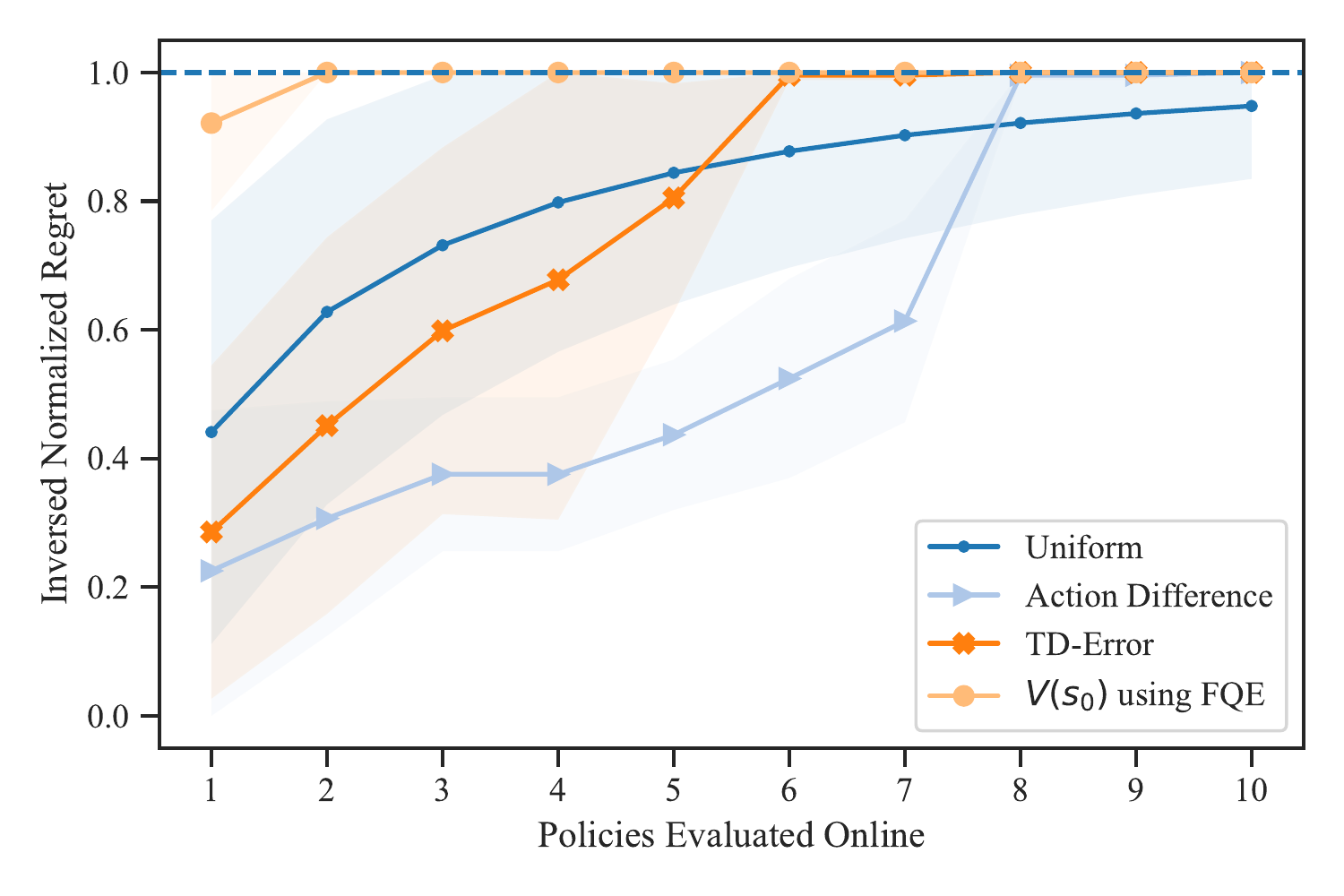}}
  \caption{Low-Level Policy, 999 }
\end{subfigure}
\begin{subfigure}[b]{.32\textwidth}
 \centering
  \centerline{\includegraphics[width=\textwidth]{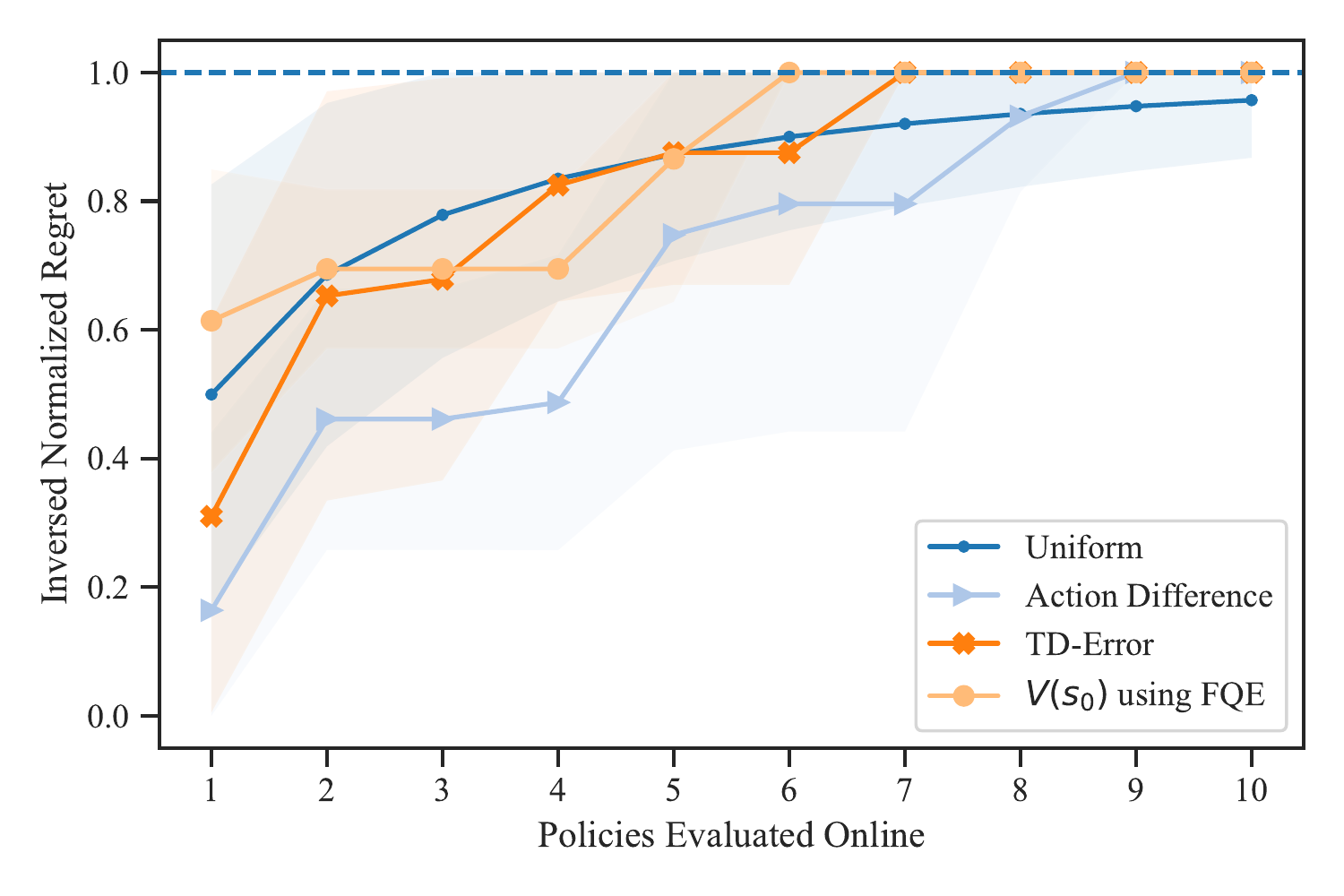}}
  \caption{Low-Level Policy, 99 }
\end{subfigure}

\caption{\textbf{TD3+BC, Industrial Benchmark. Inversed Normalized Regret under different offline policy selection methods using EOP graph.} The shaded area represents one standard deviation }
\label{fig:appendix:ops_td3bc_industrial}
\end{figure*}

\begin{figure*}[h]
 \centering
\begin{subfigure}[b]{.32\textwidth}
 \centering
  \centerline{\includegraphics[width=\textwidth]{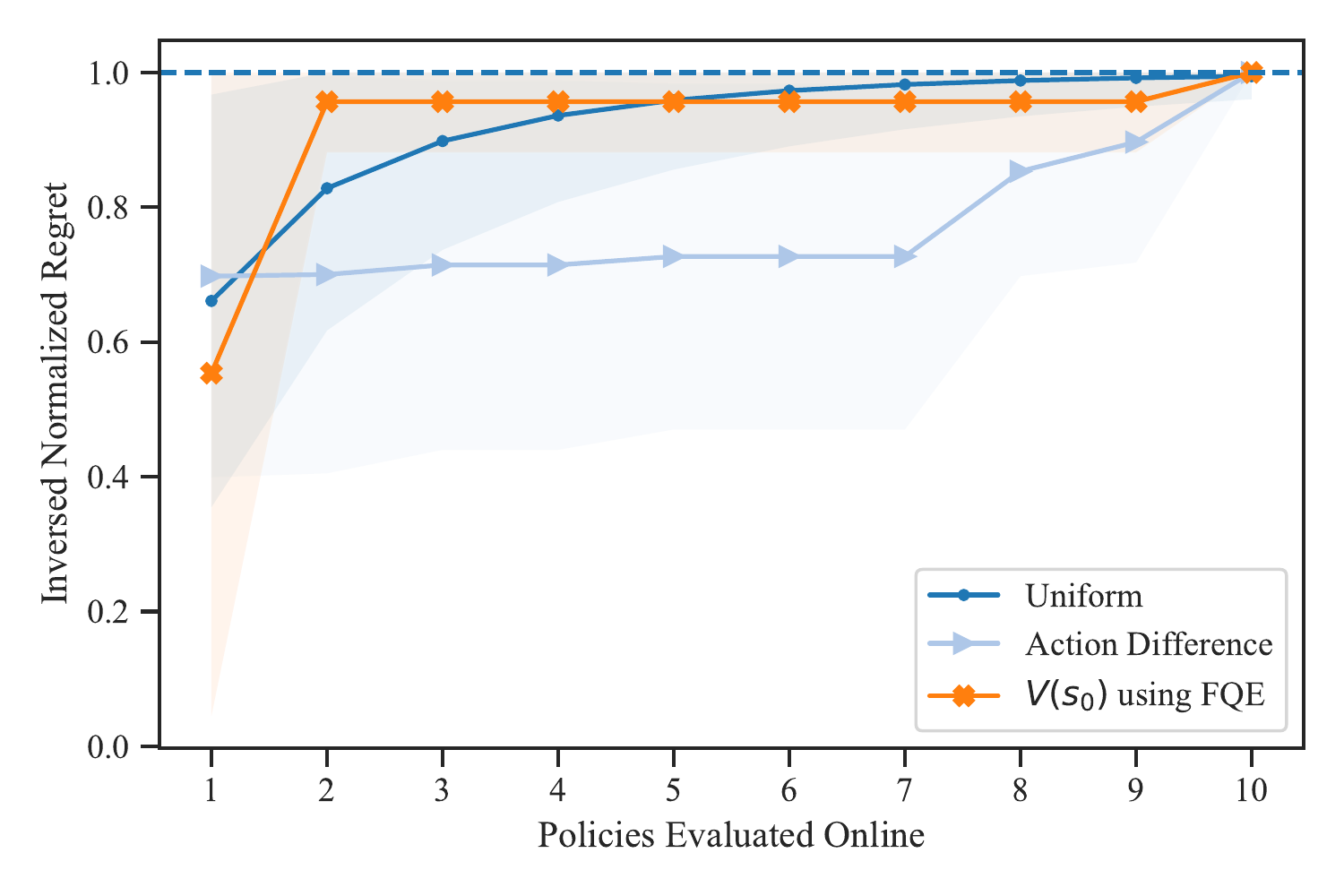}}
  \caption{High-Level Policy, 999 }
\end{subfigure}
\begin{subfigure}[b]{.32\textwidth}
 \centering
  \centerline{\includegraphics[width=\textwidth]{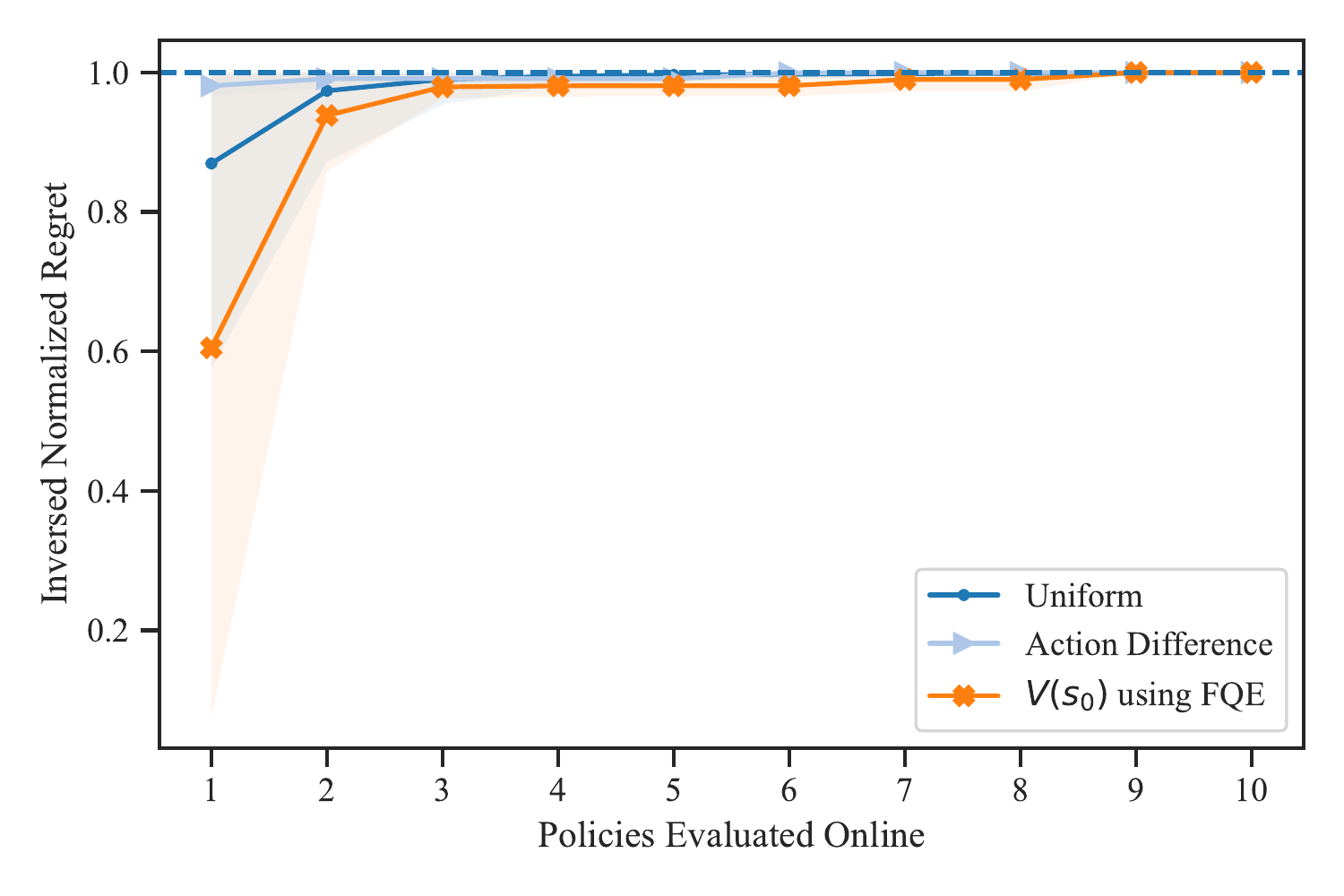}}
  \caption{Medium-Level Policy, 999 }
\end{subfigure}
\begin{subfigure}[b]{.32\textwidth}
 \centering
  \centerline{\includegraphics[width=\textwidth]{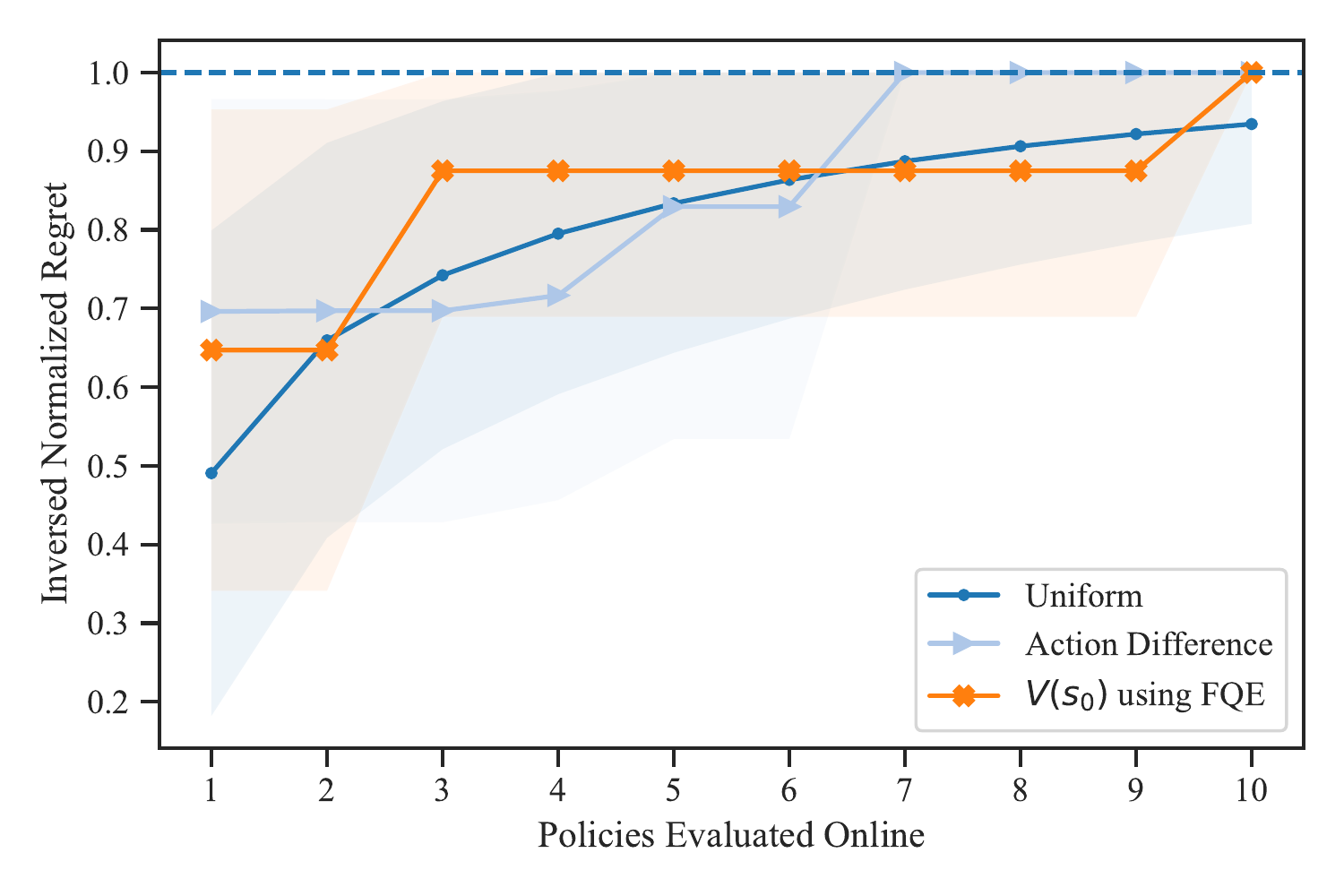}}
  \caption{Low-Level Policy, 999 }
\end{subfigure}

\begin{subfigure}[b]{.32\textwidth}
 \centering
  \centerline{\includegraphics[width=\textwidth]{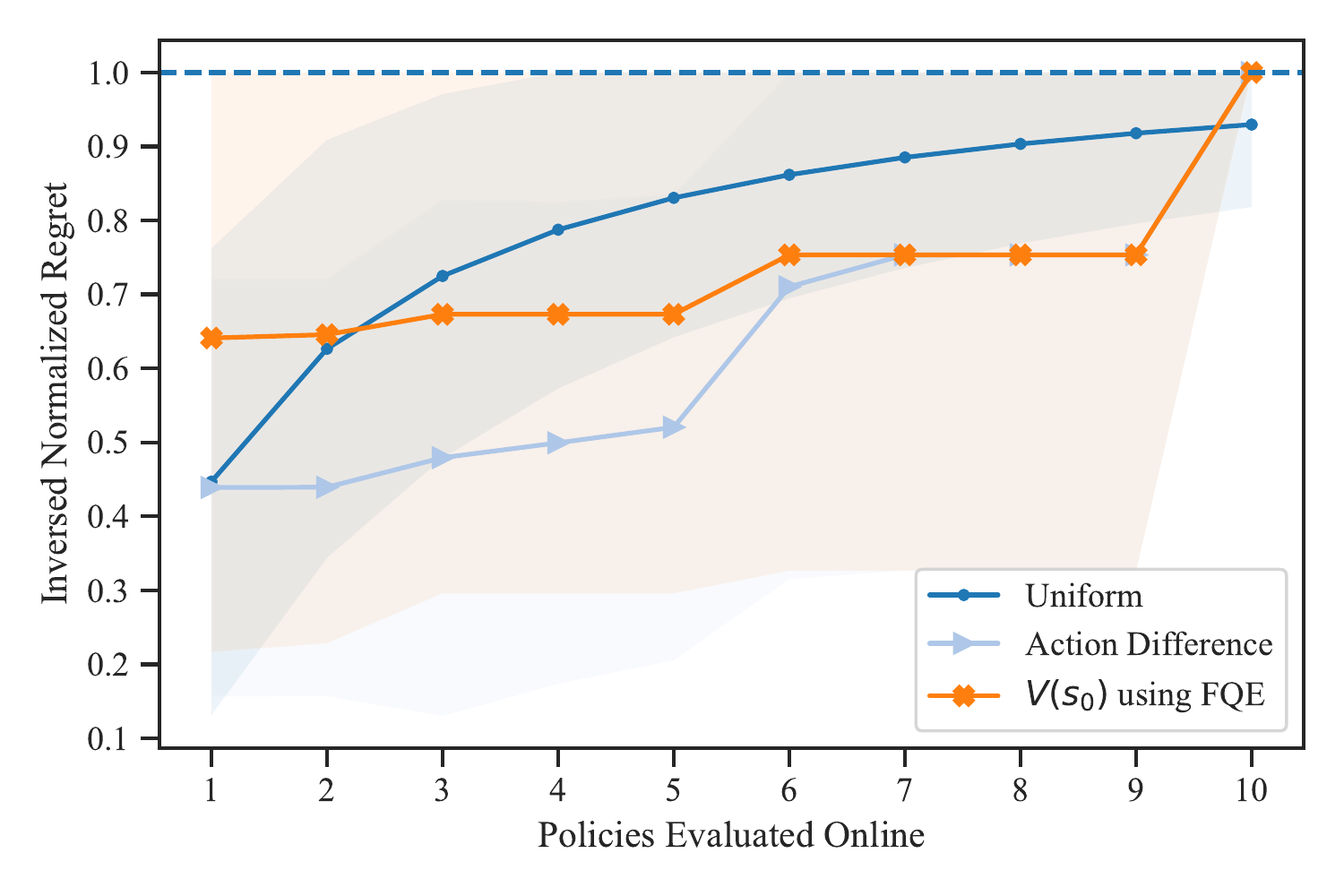}}
  \caption{High-Level Policy, 99 }
\end{subfigure}
\begin{subfigure}[b]{.32\textwidth}
 \centering
  \centerline{\includegraphics[width=\textwidth]{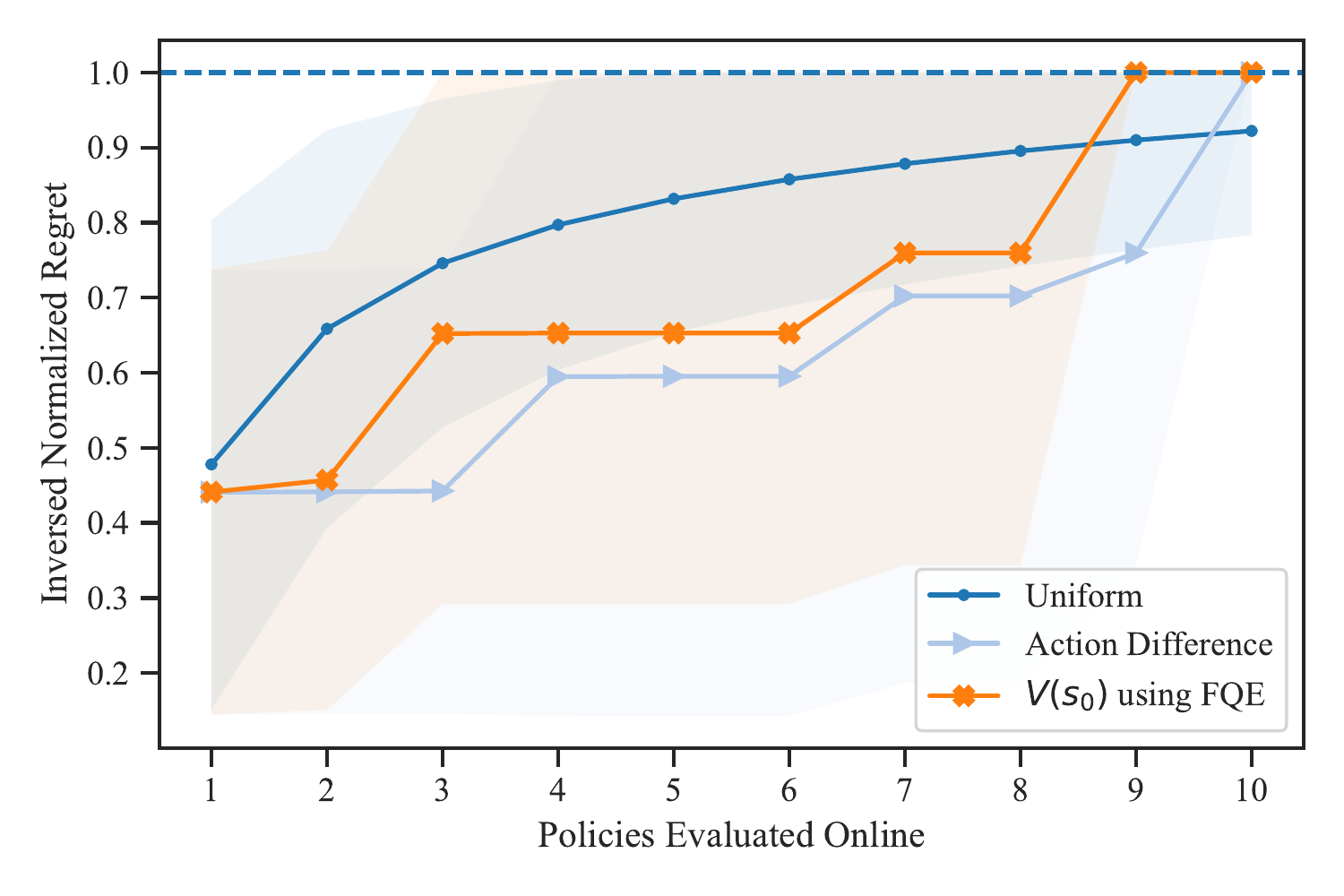}}
  \caption{Medium-Level Policy, 99 }
\end{subfigure}
\begin{subfigure}[b]{.32\textwidth}
 \centering
  \centerline{\includegraphics[width=\textwidth]{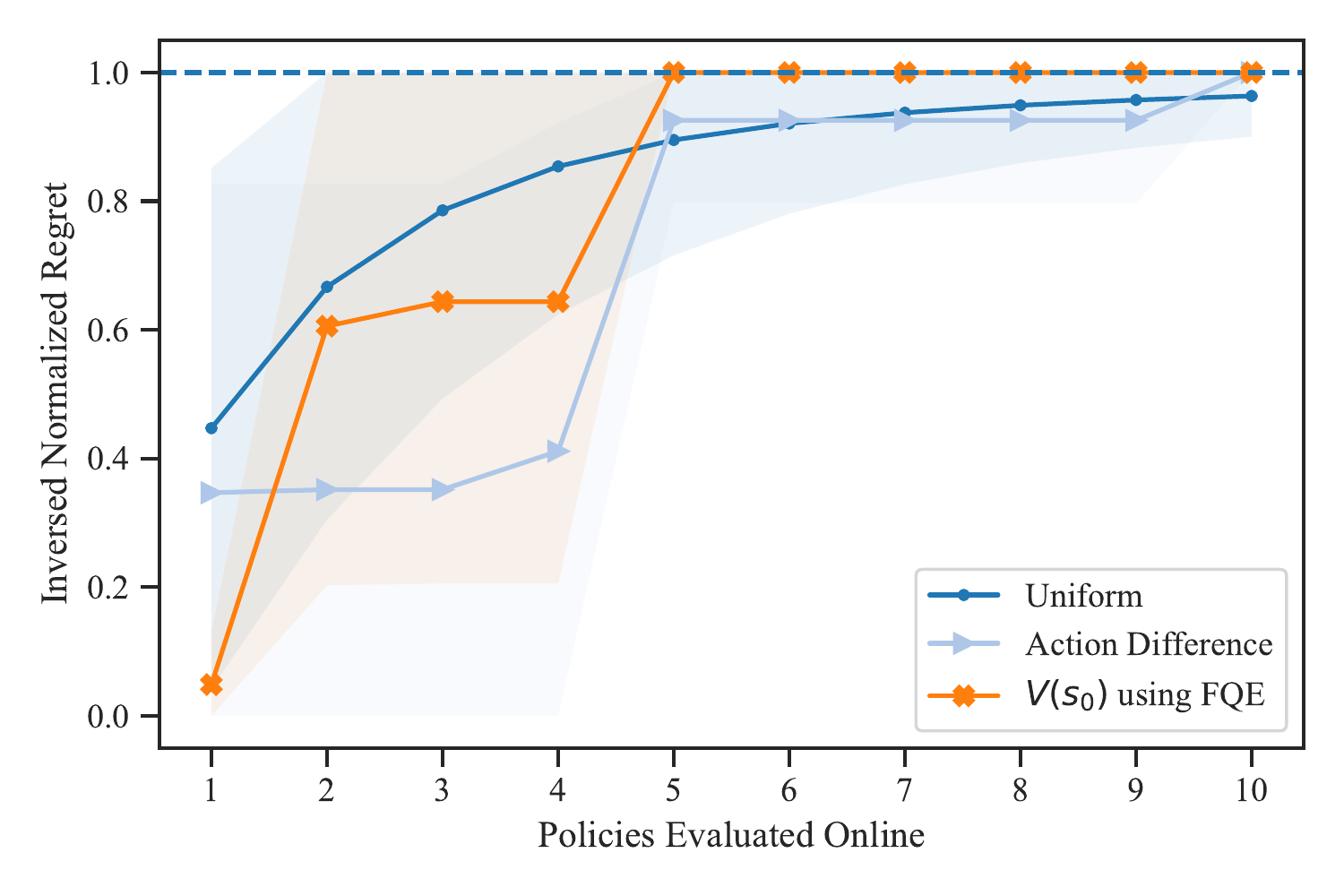}}
  \caption{Low-Level Policy, 99 }
\end{subfigure}

\caption{\textbf{BC, FinRL. Inversed Normalized Regret under different offline policy selection methods using EOP graph.} The shaded area represents one standard deviation }
\label{fig:appendix:ops_bc_finrl}
\end{figure*}

\begin{figure*}[h]
 \centering
\begin{subfigure}[b]{.32\textwidth}
 \centering
  \centerline{\includegraphics[width=\textwidth]{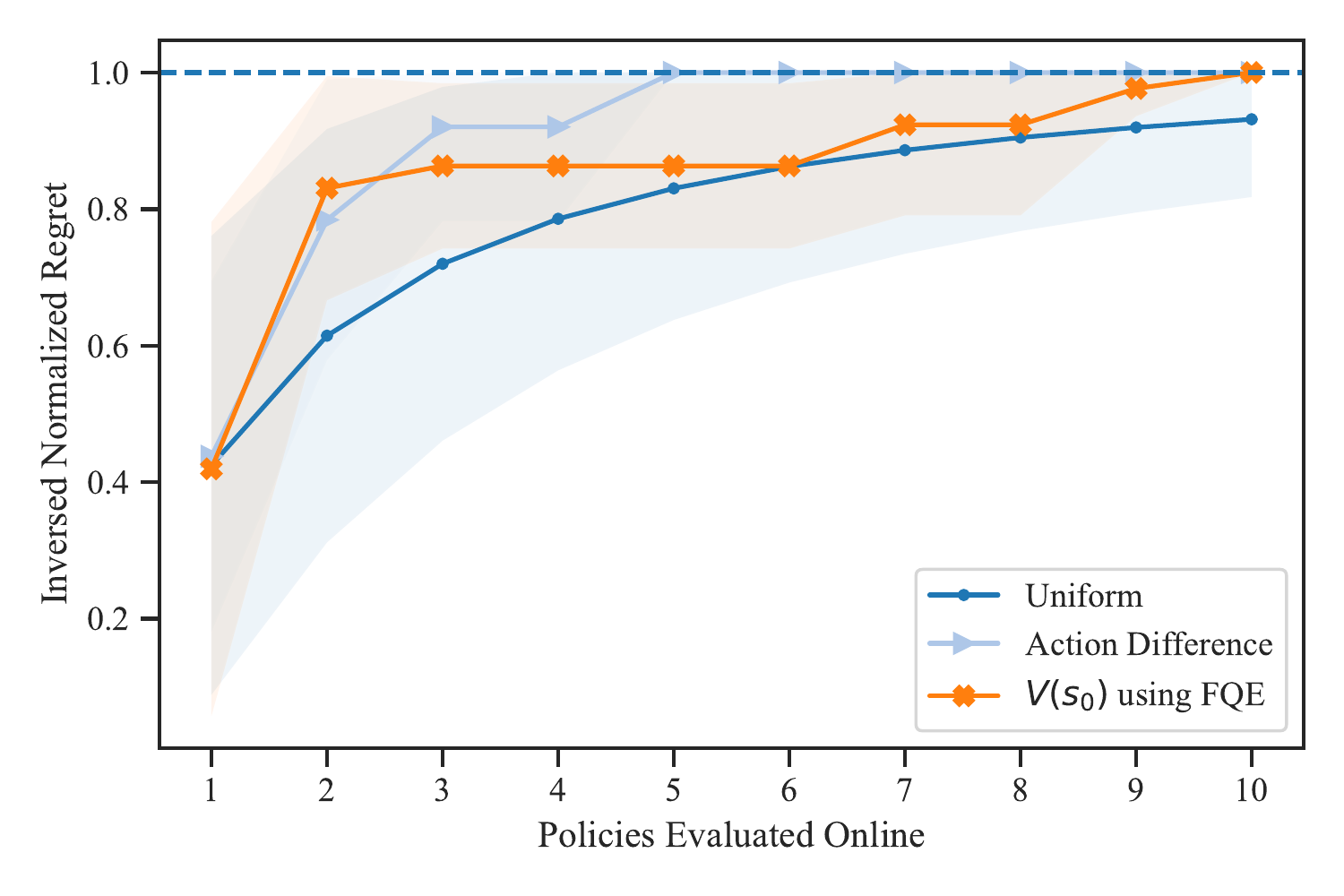}}
  \caption{Medium-Level Policy, 9999 }
\end{subfigure}
\begin{subfigure}[b]{.32\textwidth}
 \centering
  \centerline{\includegraphics[width=\textwidth]{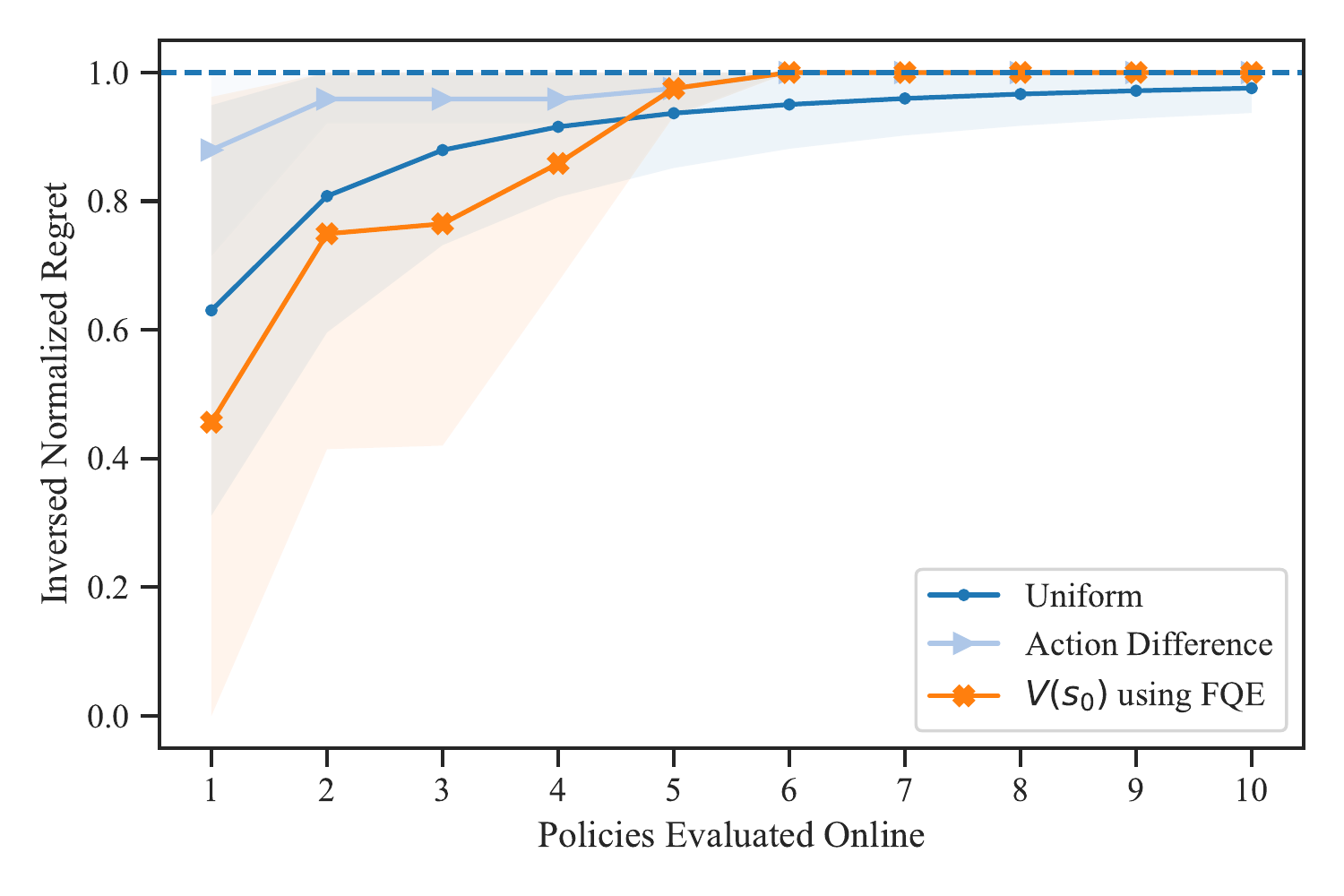}}
  \caption{Medium-Level Policy, 999 }
\end{subfigure}
\begin{subfigure}[b]{.32\textwidth}
 \centering
  \centerline{\includegraphics[width=\textwidth]{figures_appendix/ops_bc_citylearn_999_medium.pdf}}
  \caption{Medium-Level Policy, 99 }
\end{subfigure}

\begin{subfigure}[b]{.32\textwidth}
 \centering
  \centerline{\includegraphics[width=\textwidth]{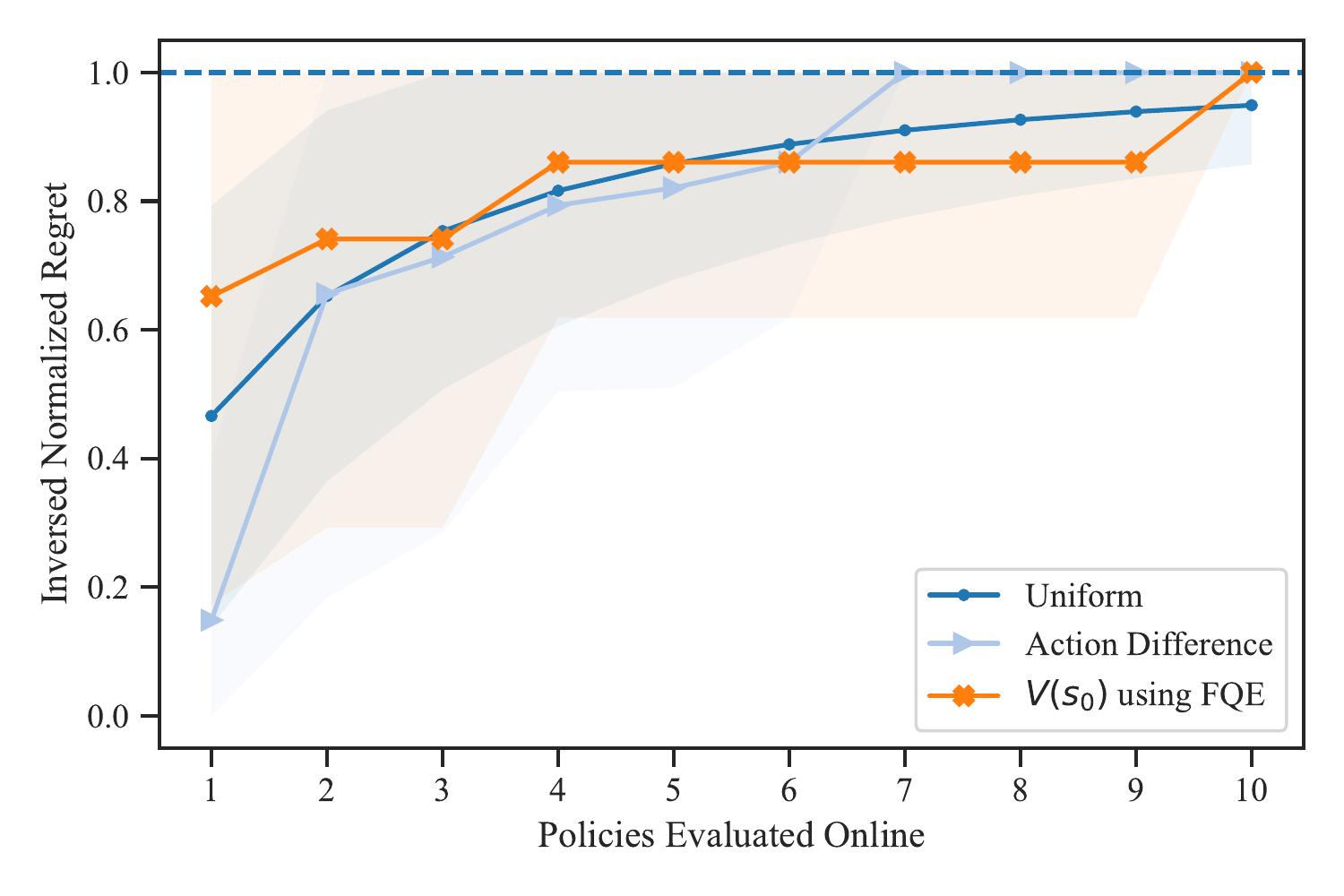}}
  \caption{Low-Level Policy, 9999 }
\end{subfigure}
\begin{subfigure}[b]{.32\textwidth}
 \centering
  \centerline{\includegraphics[width=\textwidth]{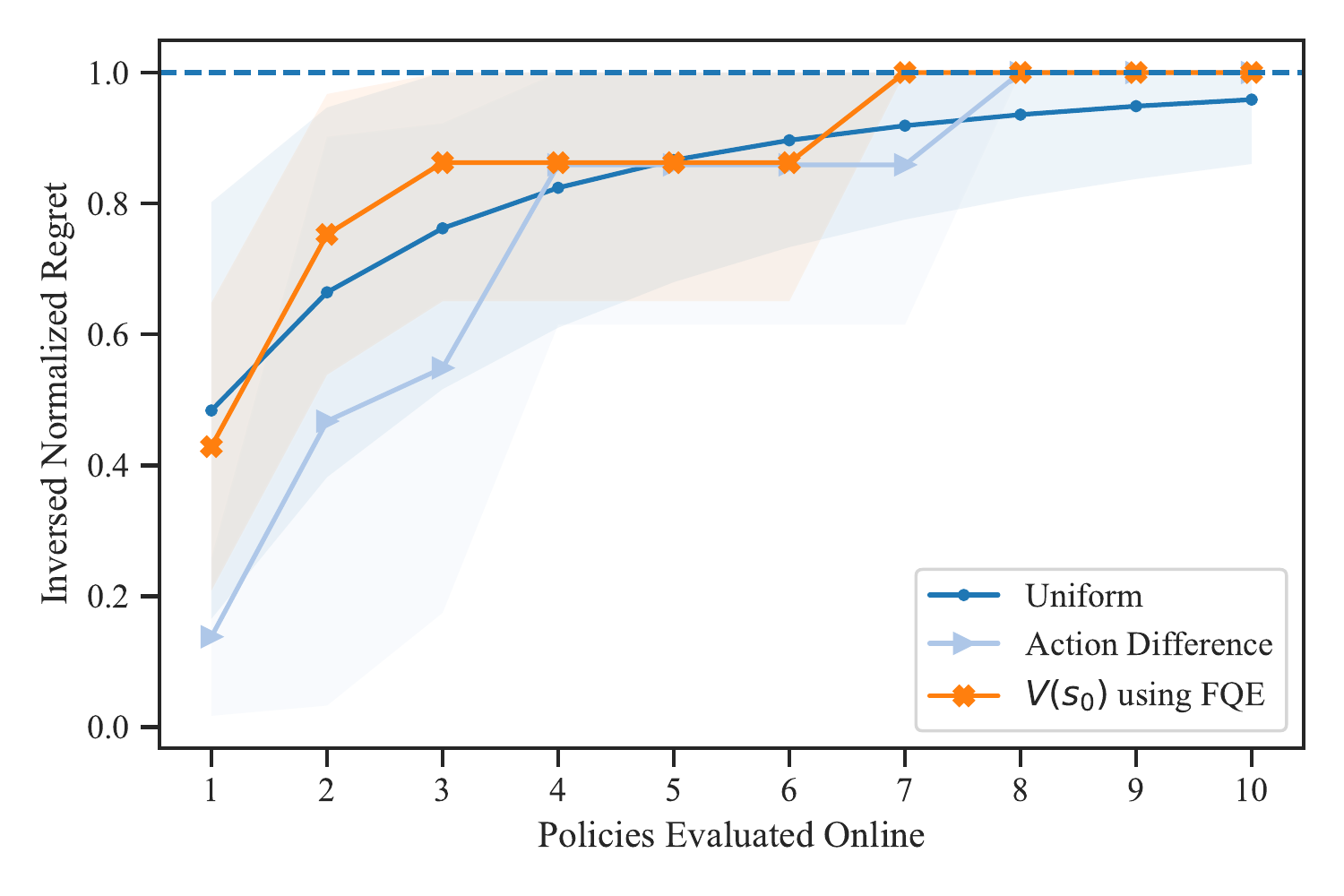}}
  \caption{Low-Level Policy, 999 }
\end{subfigure}
\begin{subfigure}[b]{.32\textwidth}
 \centering
  \centerline{\includegraphics[width=\textwidth]{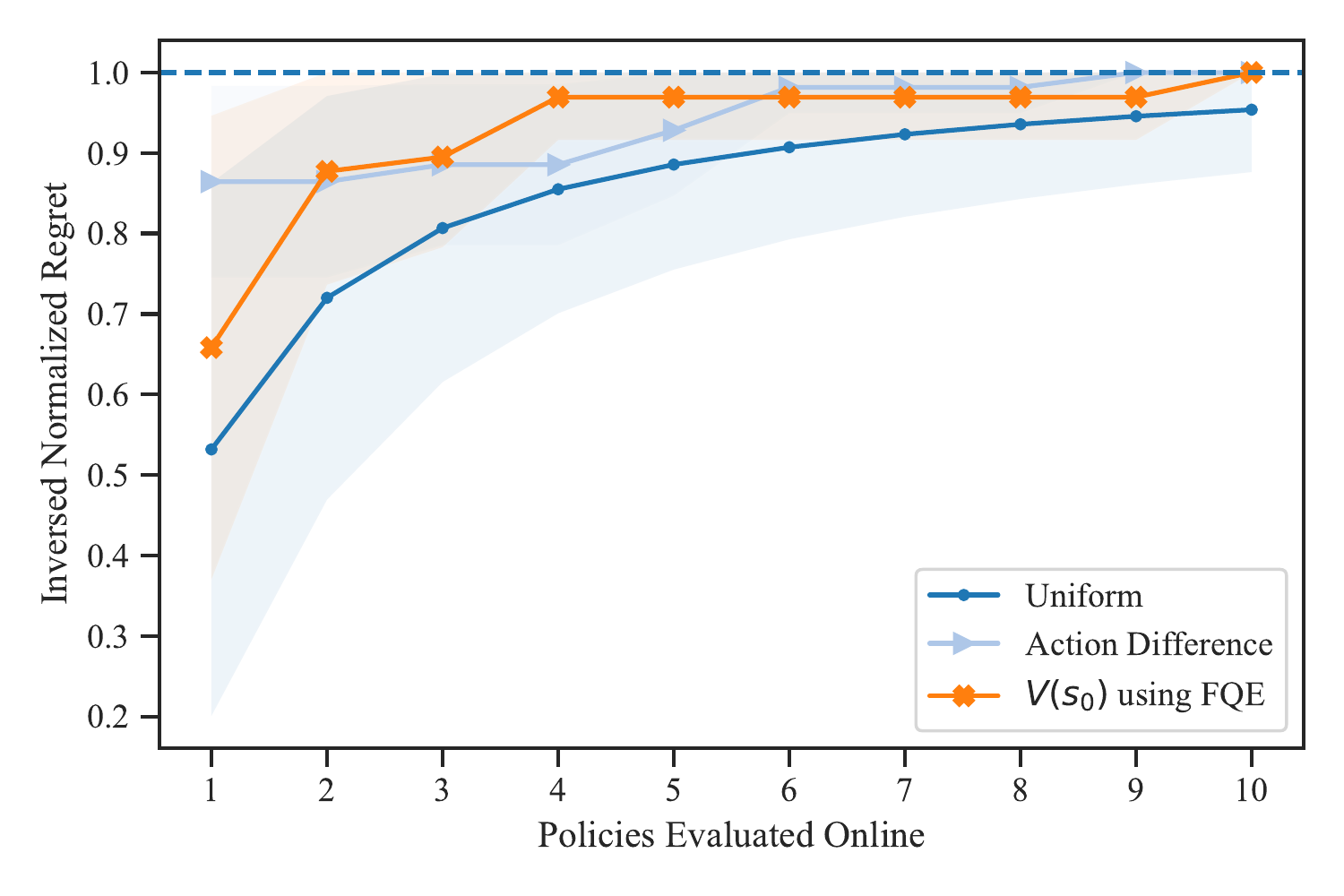}}
  \caption{Low-Level Policy, 99 }
\end{subfigure}

\caption{\textbf{BC, CityLearn. Inversed Normalized Regret under different offline policy selection methods using EOP graph.} The shaded area represents one standard deviation }
\label{fig:appendix:ops_bc_citylearn}
\end{figure*}

\begin{figure*}[h]
 \centering
\begin{subfigure}[b]{.32\textwidth}
 \centering
  \centerline{\includegraphics[width=\textwidth]{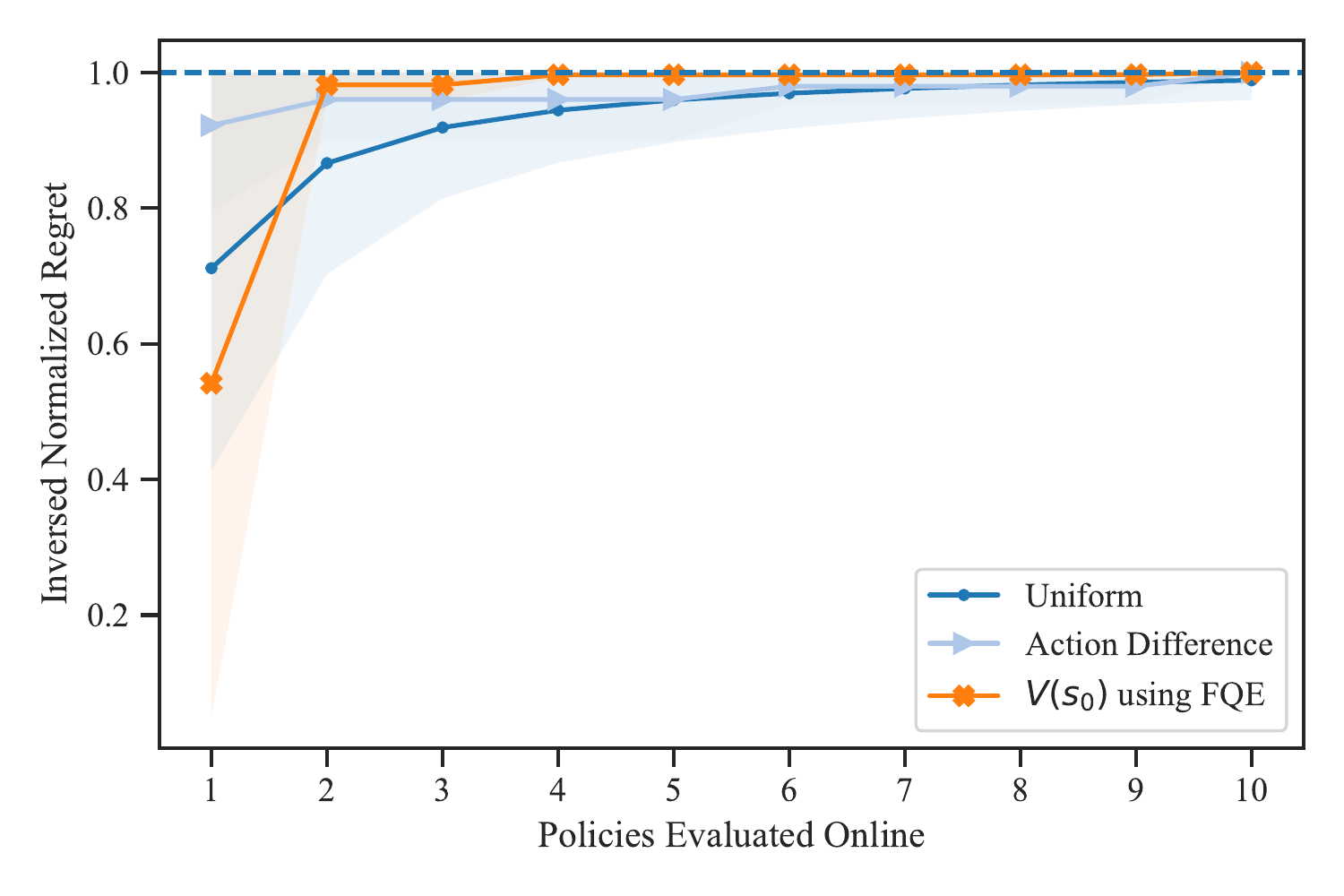}}
  \caption{High-Level Policy, 9999 }
\end{subfigure}
\begin{subfigure}[b]{.32\textwidth}
 \centering
  \centerline{\includegraphics[width=\textwidth]{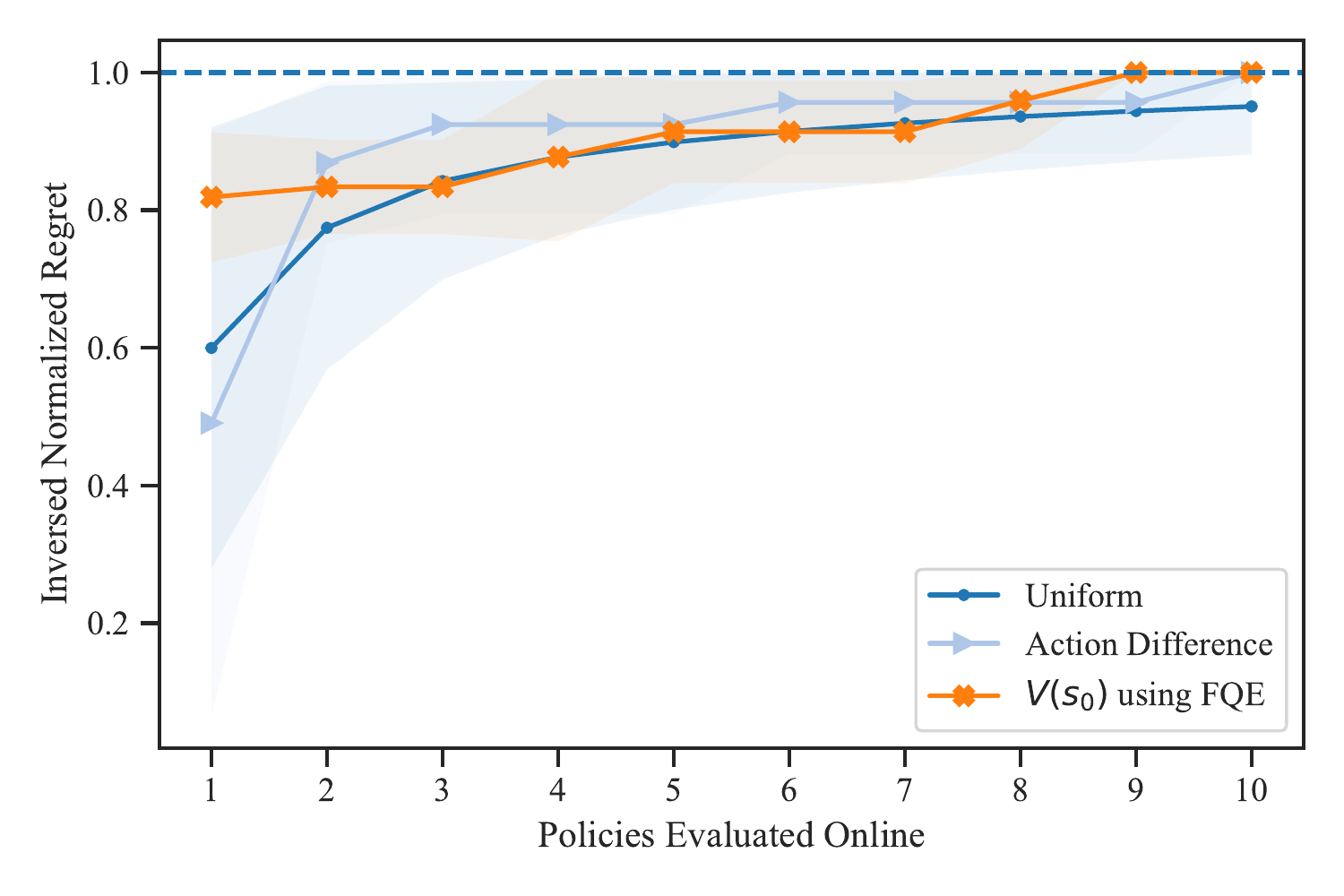}}
  \caption{High-Level Policy, 999 }
\end{subfigure}
\begin{subfigure}[b]{.32\textwidth}
 \centering
  \centerline{\includegraphics[width=\textwidth]{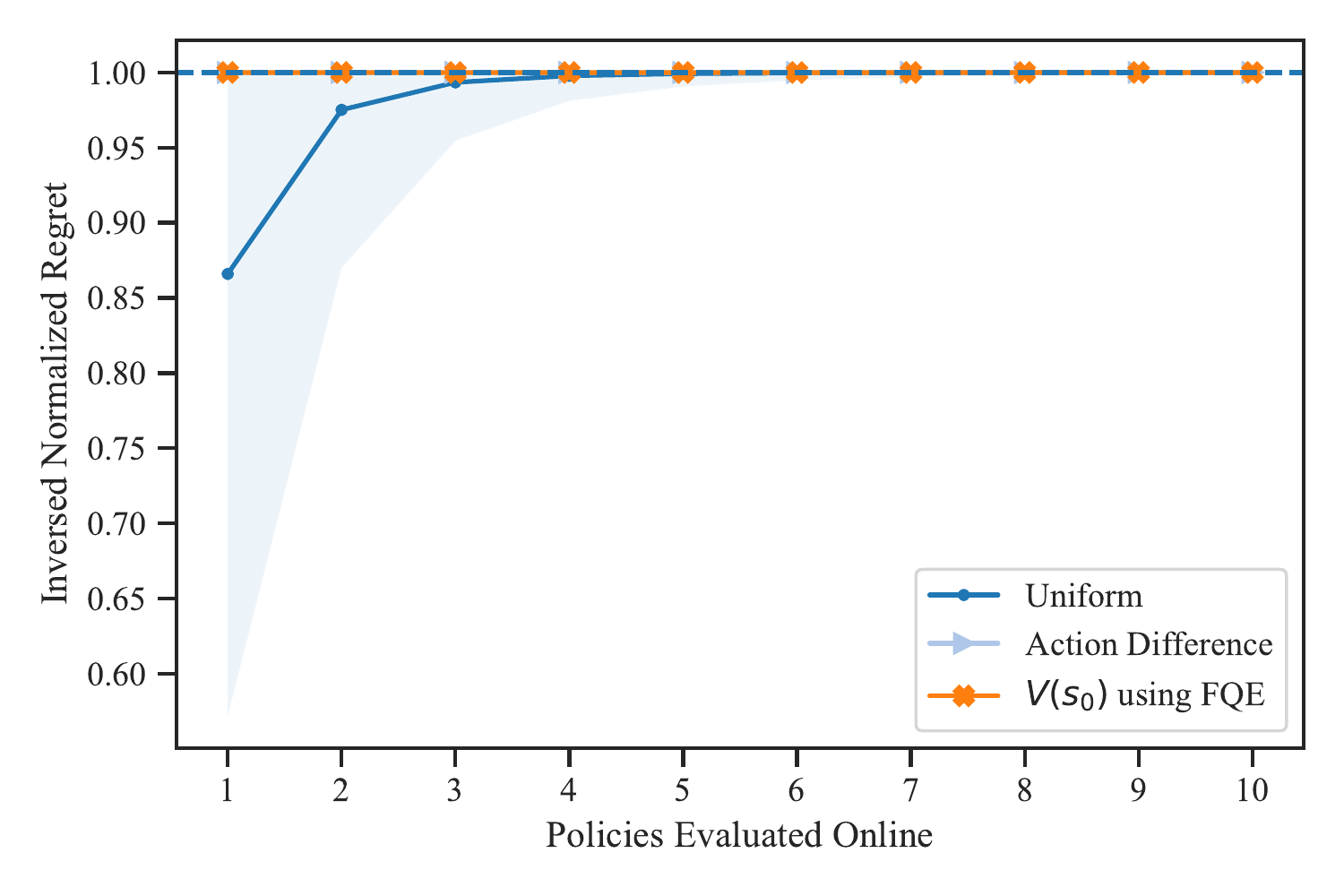}}
  \caption{High-Level Policy, 99 }
\end{subfigure}

\begin{subfigure}[b]{.32\textwidth}
 \centering
  \centerline{\includegraphics[width=\textwidth]{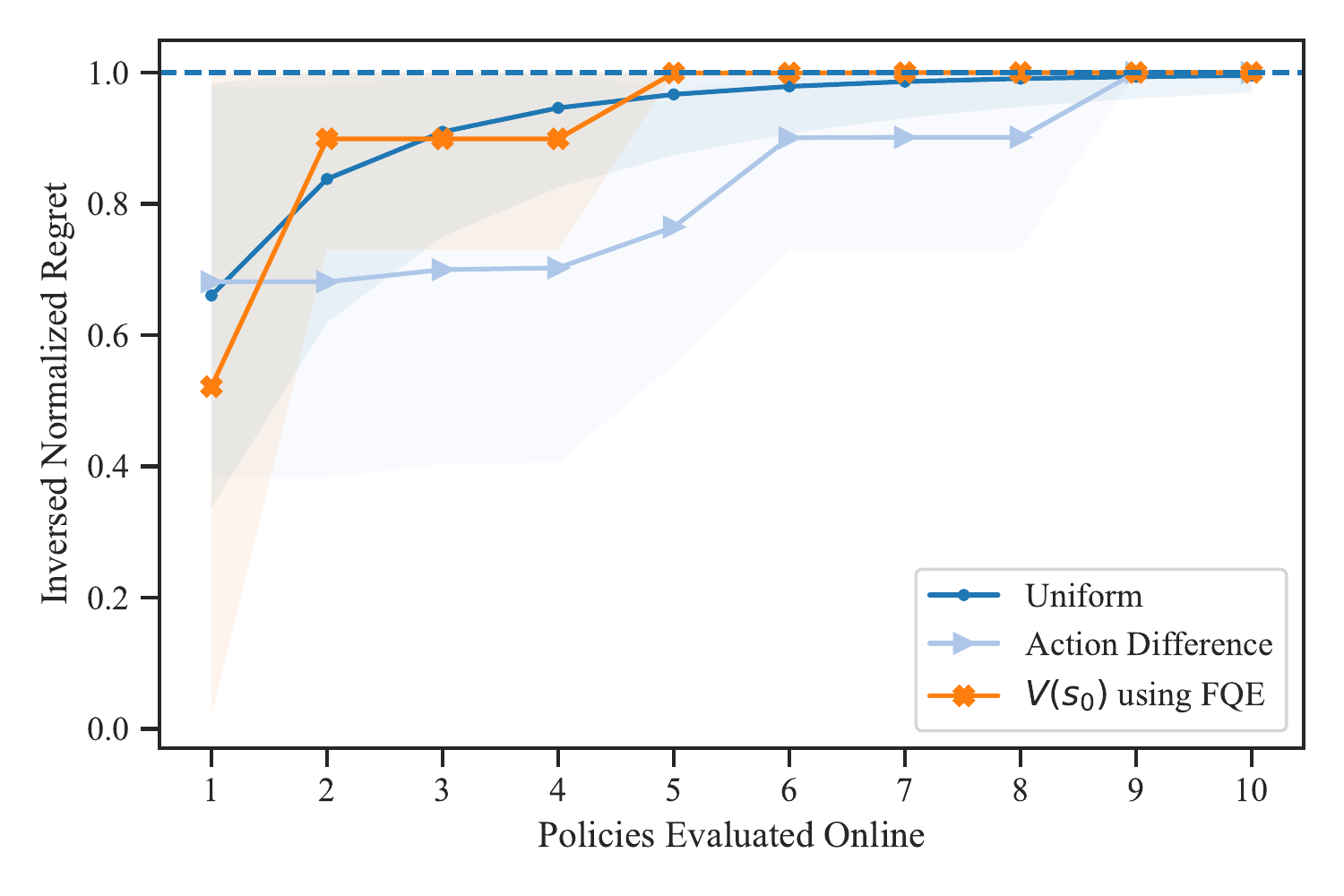}}
  \caption{Medium-Level Policy, 9999 }
\end{subfigure}
\begin{subfigure}[b]{.32\textwidth}
 \centering
  \centerline{\includegraphics[width=\textwidth]{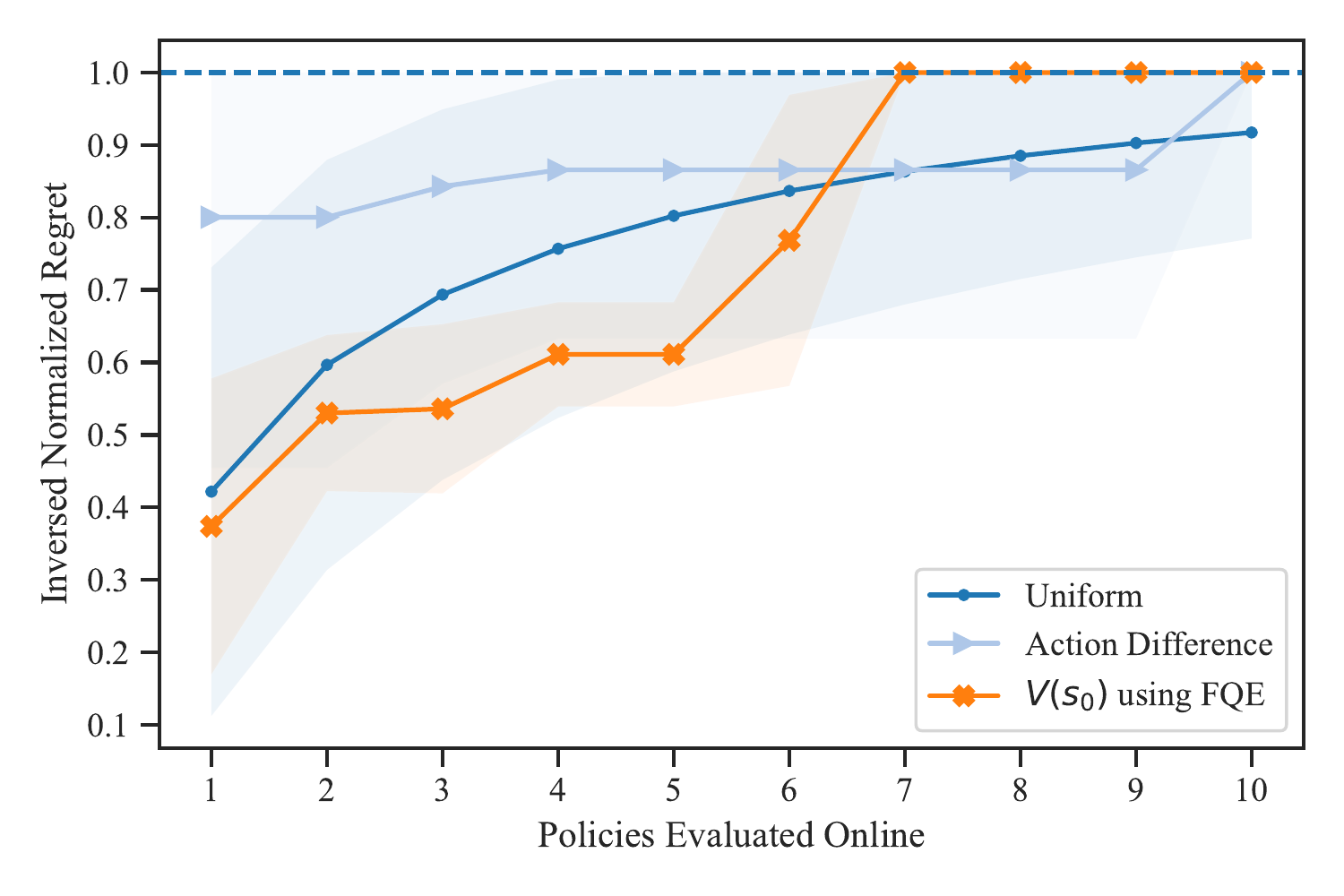}}
  \caption{Medium-Level Policy, 999 }
\end{subfigure}
\begin{subfigure}[b]{.32\textwidth}
 \centering
  \centerline{\includegraphics[width=\textwidth]{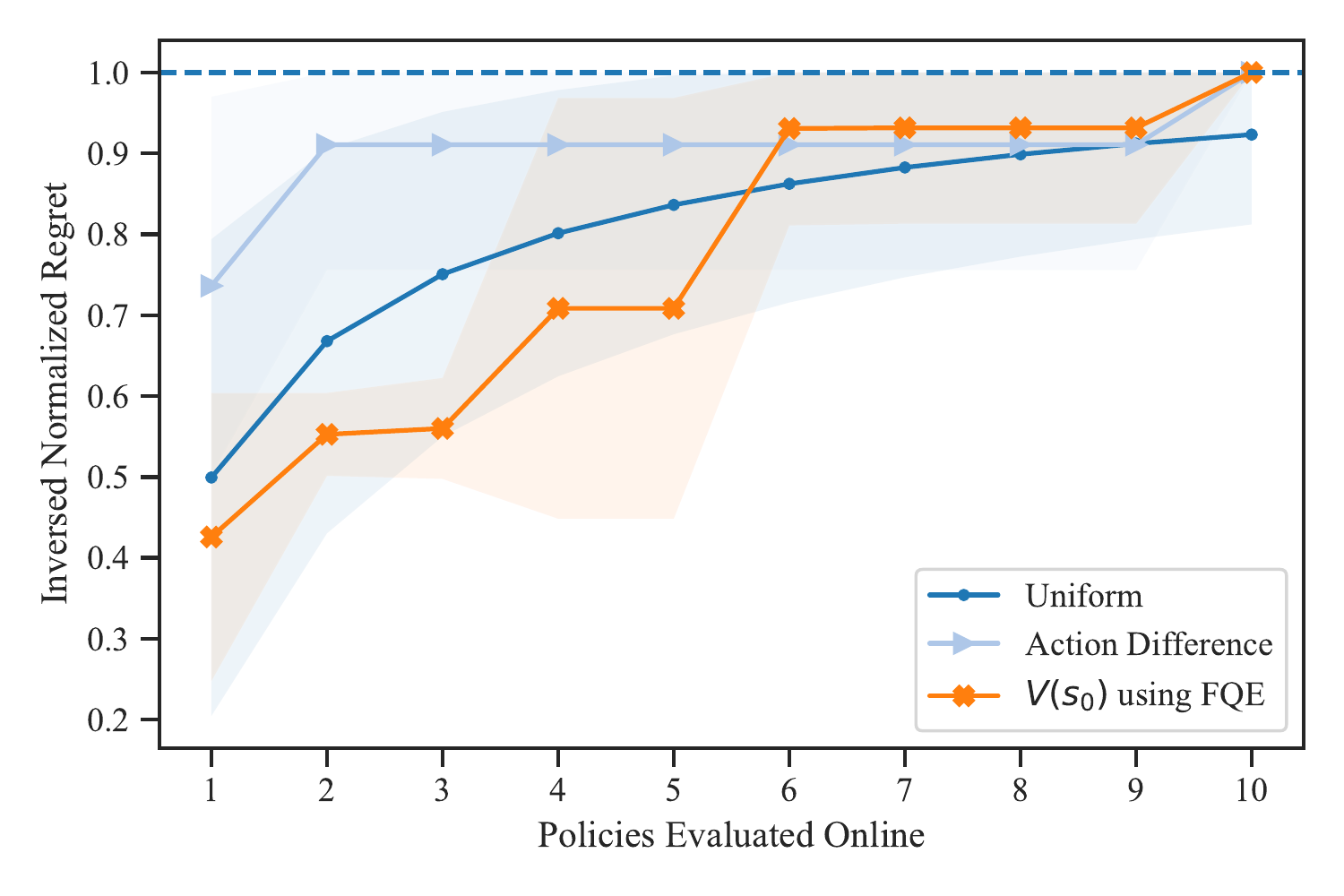}}
  \caption{Medium-Level Policy, 99 }
\end{subfigure}

\begin{subfigure}[b]{.32\textwidth}
 \centering
  \centerline{\includegraphics[width=\textwidth]{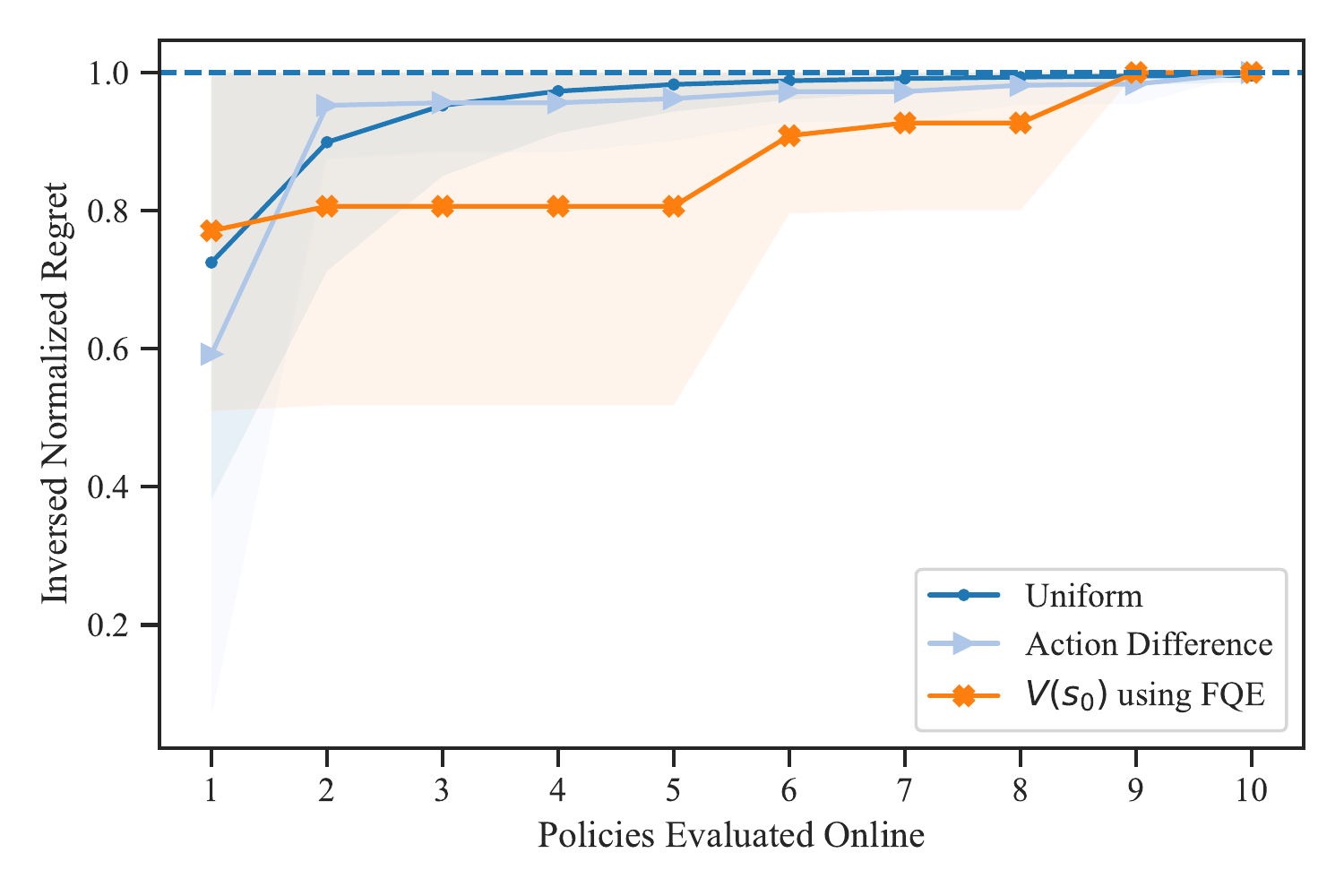}}
  \caption{Low-Level Policy, 9999 }
\end{subfigure}
\begin{subfigure}[b]{.32\textwidth}
 \centering
  \centerline{\includegraphics[width=\textwidth]{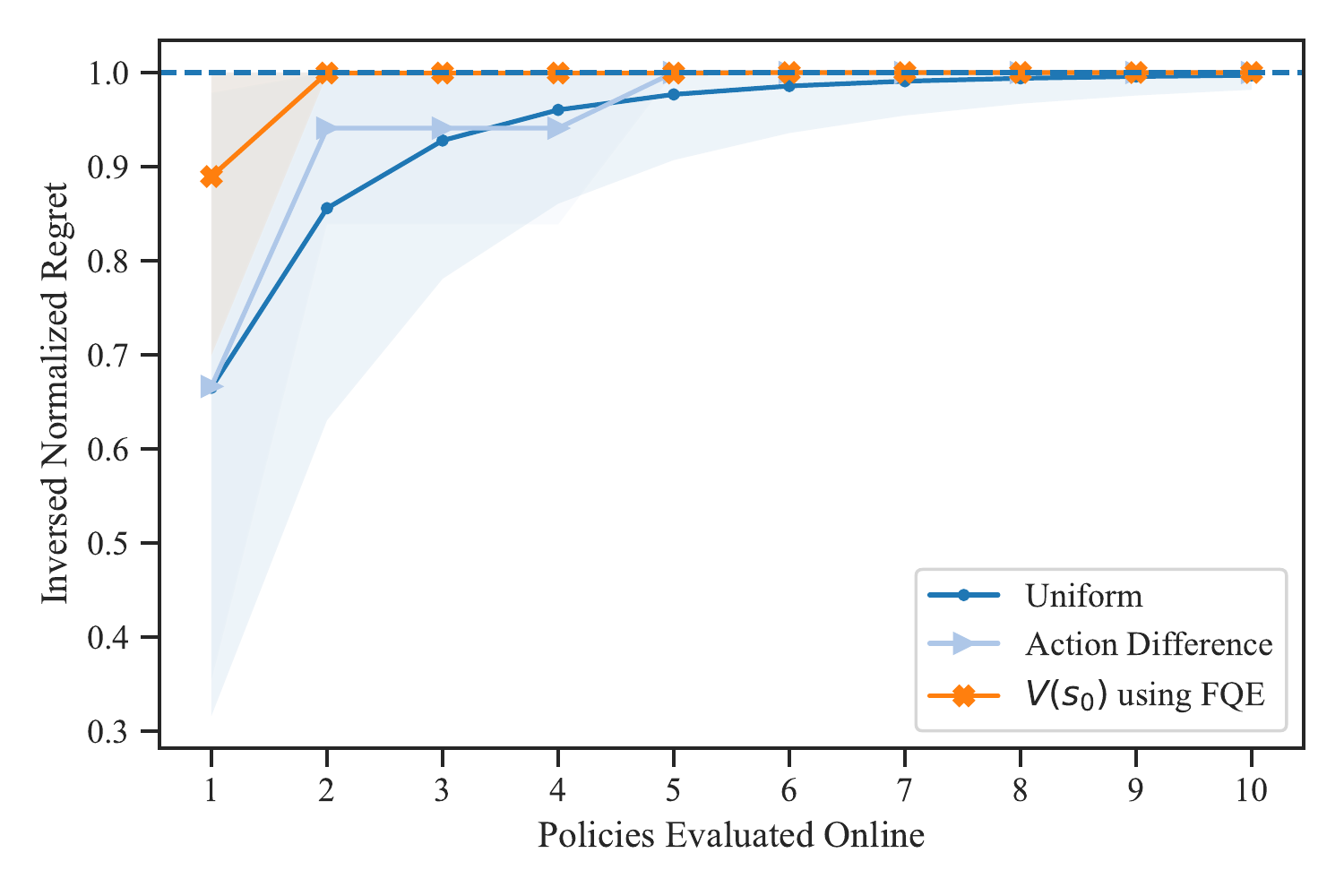}}
  \caption{Low-Level Policy, 999 }
\end{subfigure}
\begin{subfigure}[b]{.32\textwidth}
 \centering
  \centerline{\includegraphics[width=\textwidth]{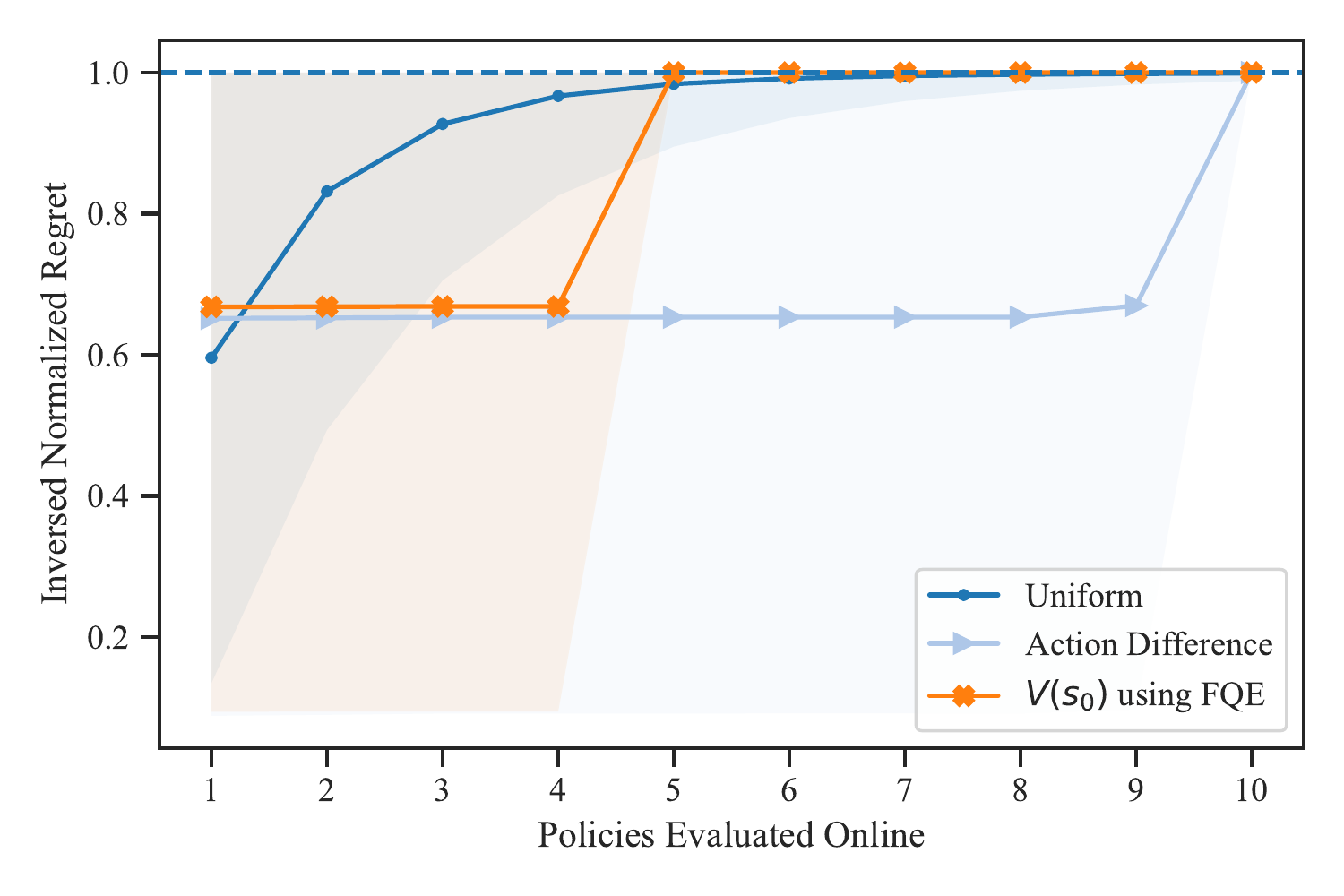}}
  \caption{Low-Level Policy, 99 }
\end{subfigure}

\caption{\textbf{BC, Industrial Benchmark. Inversed Normalized Regret under different offline policy selection methods using EOP graph.} The shaded area represents one standard deviation }
\label{fig:appendix:ops_bc_industrial}
\end{figure*}

\end{document}